%% file: robust_arxiv2020.tex
\documentclass[11pt,table]{article}
\pdfoutput=1

\usepackage[utf8]{inputenc} %
\usepackage[T1]{fontenc}    %
\usepackage[in]{fullpage}
\usepackage{xurl}
\usepackage{booktabs}       %
\usepackage{amsfonts}       %
\usepackage{nicefrac}       %
\usepackage{microtype}      %
\usepackage{amsmath,amssymb,amsthm}
\usepackage{mathtools}
\usepackage{mathrsfs}
\usepackage{thmtools}
\usepackage{thm-restate}
\usepackage{comment}
\usepackage{parskip}
\usepackage{siunitx}
\usepackage{etoolbox}
\usepackage{enumitem}
\usepackage{graphicx}
\usepackage[margin=1cm]{caption}
\usepackage{subcaption}
\usepackage{tabularx}
\usepackage{xcolor} 
\usepackage[numbers,sort]{natbib}
\usepackage{hyperref}
\usepackage{longtable}
\usepackage{afterpage}
\usepackage{algorithm}
\usepackage{algpseudocode}
\usepackage{bm}
\usepackage{etoc}
\usepackage{cleveref}

\extrafloats{100}

\input{macros}
\togglefalse{isneurips}

\usepackage[title]{appendix}

\title{Measuring Robustness to Natural Distribution Shifts\\in Image Classification}
\author{\begin{tabular}[t]{c@{\hskip 1.5cm}c@{\hskip 1.5cm}c}
    \vspace{0.1cm}Rohan Taori & Achal Dave & Vaishaal Shankar \\
    \vspace{0.5cm}UC Berkeley & CMU & UC Berkeley\\
    \vspace{0.1cm}Nicholas Carlini & Benjamin Recht & Ludwig Schmidt \\
    Google Brain & UC Berkeley & UC Berkeley
    \end{tabular}
    }

\begin{document}
\date{}
\maketitle

\begin{abstract}
\input{abstract}
\end{abstract}

\section{Introduction}
\label{sec:intro}
\input{sec1-intro.tex}

\section{Measuring robustness}
\label{sec:robustness-defn}
\input{sec2-robustness-defn.tex}

\section{Experimental setup}
\label{sec:setup}
\input{sec3-testbed.tex}

\section{Main results}
\label{sec:results}
\input{sec4-results.tex}

\iftoggle{isneurips}{}{
  \input{secX-more-data}
}

\section{Related work}
\input{sec5-related-work.tex}

\section{Conclusion}
\input{sec6-conclusion.tex}

\section*{Broader Impact}
\input{broader_impact.tex}

\section*{Acknowledgements}
\input{acknowledgements.tex}

\clearpage
\bibliography{references}
\bibliographystyle{icml2020}

\newpage
\captionsetup{margin=0cm}
\appendix
\input{appendix.tex}

\end{document}

%% file: macros.tex
\newcommand{\acca}{\ensuremath{\text{acc}_1}}
\newcommand{\accb}{\ensuremath{\text{acc}_2}}

\newcommand{\E}{\mathbb{E}}
\newcommand{\R}{\mathbb{R}}
\newcommand{\eps}{\varepsilon}

\newtoggle{isneurips}

\newcommand{\nummodels}{204}
\newcommand{\numevalsettings}{213}
\newcommand{\numstandardmodels}{78}
\newcommand{\nummoredatamodels}{30}
\newcommand{\numrobustnessmodels}{86}
\newcommand{\numcorruptions}{38}

\newcommand{\imagenet}{ImageNet}
\newcommand{\imagenetnew}{ImageNetV2}
\newcommand{\imagenetvid}{ImageNetVid}
\newcommand{\imagenetvidrobust}{ImageNet-Vid-Robust}
\newcommand{\objectnet}{ObjectNet}
\newcommand{\ytbb}{YTBB}
\newcommand{\ytbbrobust}{YTBB-Robust}
\newcommand{\imageneta}{ImageNet-A}
\newcommand{\imagenetr}{ImageNet-R}
\newcommand{\imagenetsketch}{ImageNet-Sketch}

\newcommand{\jft}{JFT-300M}

\newcommand{\model}[1]{\footnotesize #1}

\newlength{\postsecvspace}
\makeatletter
\def\foo#1{%
\setbox0\vbox{%
\global\let\foonext\foohyph
\hbadness\maxdimen
\hfuzz\maxdimen
#1\par
  \loop
  \unskip\unpenalty\unskip\unpenalty
  \setbox\z@\lastbox
  \ifvoid\z@
  \else
   \setbox\z@\hbox{\unhbox\z@}%
   \ifdim\wd\z@>\dimexpr\hsize+4pt\relax%
    \global\let\foonext\foowrap
   \fi
  \repeat}%
\foonext{#1}%
}
\def\foohyph#1{#1}
\def\foowrap#1{\foowrap@#1\foowrap@}

\def\foowrap@{\futurelet\tmp\foowrap@@}

\def\foowrap@@{%
\let\next\relax
\ifx\tmp\foowrap@
\let\foowrap@\relax
\else\ifx\tmp\lbrace
\else\ifx\tmp\@sptoken
\let\next\foowrap@sp
\else\ifcat\noexpand\tmp\active%
\let\next\foowrap@aa
\else
\let\next\foowrap@a
\fi\fi\fi\fi
\next
}

\def\foowrap@sp{\afterassignment\foowrap@\let\tmp= }

\def\foowrap@a#1{%
\leavevmode\penalty\z@#1\foowrap@}

\def\foowrap@aa#1#2{%
\leavevmode\penalty\z@#1#2%
\foowrap@}

\newcolumntype{R}{>{\raggedleft\arraybackslash}p{1.7cm}}
\iftoggle{isneurips}{
  \newcolumntype{W}{p{4cm}}
  \newcolumntype{T}{>{\raggedleft\arraybackslash}p{1.25cm}}
  \newcolumntype{E}{p{1.9cm}}
}{
  \newcolumntype{W}{p{5cm}}
  \newcolumntype{T}{>{\raggedleft\arraybackslash}p{1.5cm}}  
  \newcolumntype{E}{p{2.8cm}}
}

%% file: abstract.tex
We study how robust current ImageNet models are to distribution shifts arising from natural variations in datasets. 
Most research on robustness focuses on synthetic image perturbations (noise, simulated weather artifacts, adversarial examples, etc.), which leaves open how robustness on synthetic distribution shift relates to distribution shift arising in real data. 
Informed by an evaluation of \nummodels{} ImageNet models in \numevalsettings{} different test conditions, we find that there is often little to no transfer of robustness from current synthetic to natural distribution shift.
Moreover, most current techniques provide no robustness to the natural distribution shifts in our testbed.
The main exception is training on larger and more diverse datasets, which in multiple cases increases robustness, but is still far from closing the performance gaps.
Our results indicate that distribution shifts arising in real data are currently an open research problem.
We provide our testbed and data as a resource for future work at \href{https://modestyachts.github.io/imagenet-testbed/}{\color{blue}https://modestyachts.github.io/imagenet-testbed/}. 

%% file: sec1-intro.tex
Reliable classification under distribution shift is still out of reach for current machine learning \citep{datasetshiftbook,torralba2011unbiased,recht2019imagenet}. 
As a result, the research community has proposed a wide range of evaluation protocols that go beyond a single, static test set.
Common examples include noise corruptions \citep{Geirhos2018,hendrycks2019benchmarking}, spatial transformations \citep{fawzi2015manitest,engstrom2019exploring}, and adversarial examples \citep{szegedy2013intriguing,biggio2013}.
Encouragingly, the past few years have seen substantial progress in robustness to these distribution shifts, e.g., see \citep{madry2017towards,trades,geirhos2019imagenettrained,zhang2019making,engstrom2019exploring,yang2019invarianceinducing,TB19b,wong2019wasserstein,maini2019adversarial,cohen2019certified,raghunathan2018certified} among many others.
However, this progress comes with an important limitation: all of the aforementioned distribution shifts are \emph{synthetic}: the test examples are derived from well-characterized image modifications at the pixel level.

Synthetic distribution shifts are a good starting point for experiments since they are precisely defined and easy to apply to arbitrary images.
However, classifiers ultimately must be robust to distribution shifts arising naturally in the real world.
These distribution shifts may include subtle changes in scene compositions, object types, lighting conditions, and many others.
Importantly, these variations are \emph{not} precisely defined because they have not been created artificially.
The hope is that an ideal robust classifier is still robust to such natural distribution shifts.

In this paper, we investigate how robust current machine learning techniques are to distribution shift arising naturally from real image data without synthetic modifications.
To this end, we conduct a comprehensive experimental study in the context of ImageNet \cite{imagenet,ILSVRC15}.
ImageNet is a natural starting point since it has been the focus of intense research efforts over the past decade and a large number of pre-trained classification models, some with robustness interventions, are available for this task.
The core of our experimental study is a testbed of \nummodels{} pre-trained ImageNet models that we evaluate in \numevalsettings{} different settings, covering both the most popular models and distribution shifts.
Our testbed consists of $10^9$ model predictions and is 100 times larger than prior work \cite{kang2019transfer,recht2019imagenet,robustness,Geirhos2018}.
This allows us to draw several new conclusions about current robustness interventions:
\iftoggle{isneurips}{\vspace{.2cm}}{}

\textbf{Robustness measurements should control for accuracy.}
Existing work typically argues that an intervention improves robustness by showing that the accuracy on a robustness test set has improved (e.g., see \cite{zhang2019making,geirhos2019imagenettrained,xie2019selftraining,orhan2019robustness,hendrycks2020faces}).
We find that in many cases, this improved robustness can be explained by the model performing better on the standard, unperturbed test set.
For instance, using different model architectures does not substantially improve the robustness of a model beyond what would be expected from having a higher standard accuracy.
While training more accurate models is clearly useful, it is important to separate accuracy improvements from robustness improvements when interpreting the results.

\begin{figure*}[htb]
  	\centering
	\begin{subfigure}[t]{0.48\textwidth}
	  \includegraphics[width=\textwidth]{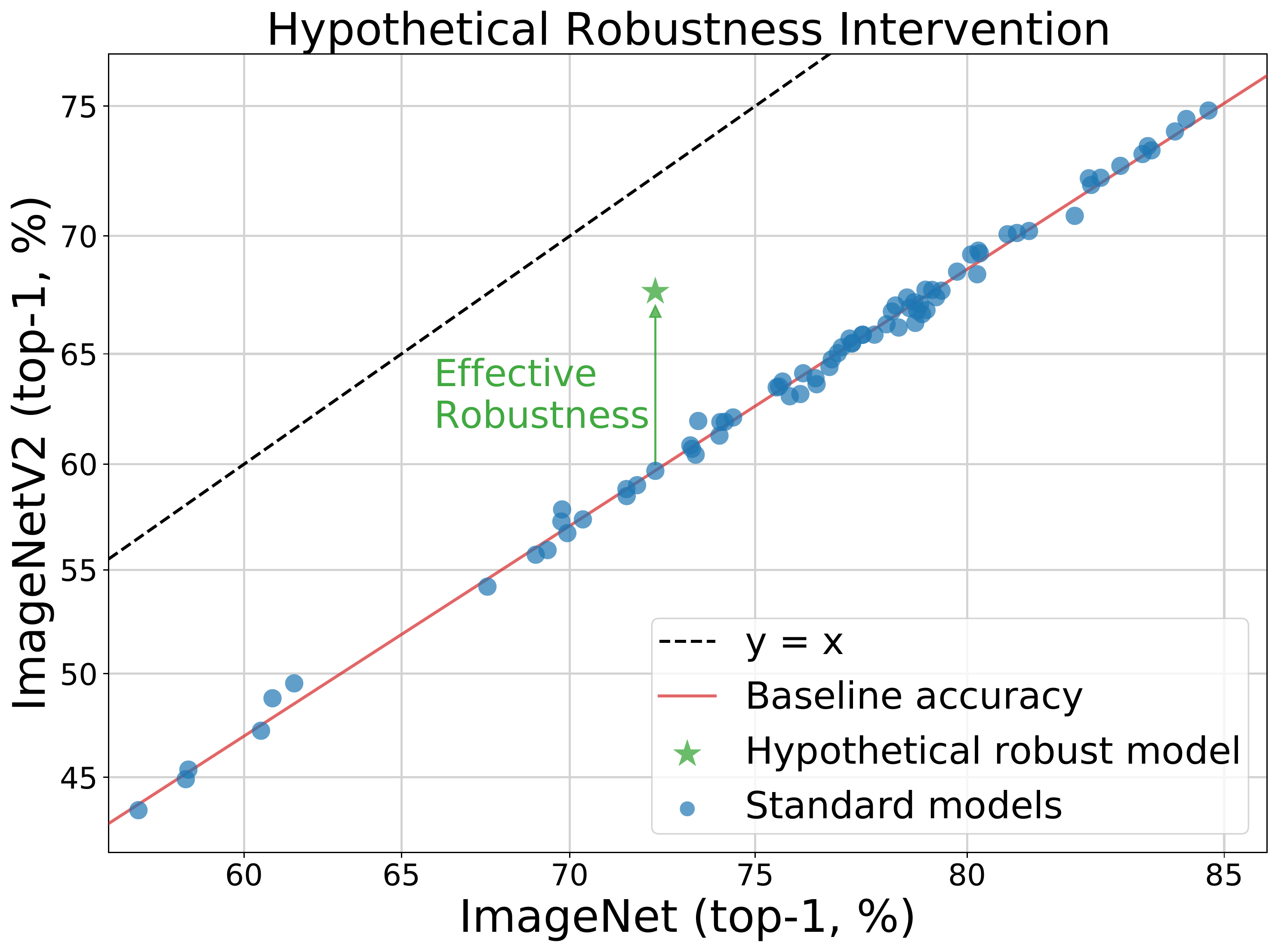}
	\end{subfigure}
	\hfill
	\begin{subfigure}[t]{0.48\textwidth}
	  \includegraphics[width=\textwidth]{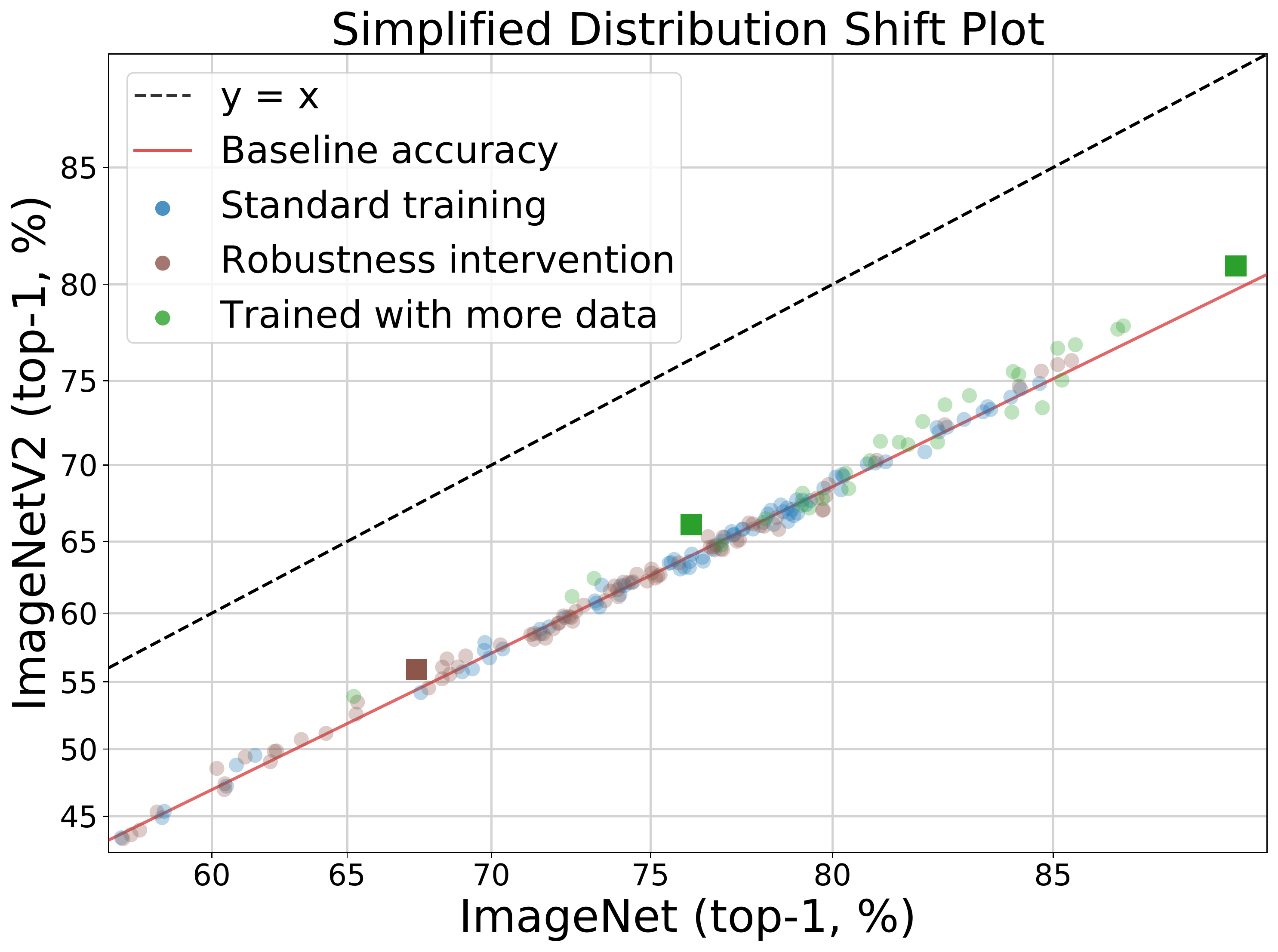}
	\end{subfigure}

  \caption{
    (Left)
    We plot \numstandardmodels{} standard models trained on ImageNet without any robustness interventions, showing both their accuracy on the standard test set (ImageNet, x-axis) and on a test set with distribution shift (ImageNetV2, y-axis).
    All models lie below the $y=x$ line: their accuracy under this distribution shift is lower than on the standard test set.
    Nevertheless, improvements in accuracy on the standard test set almost perfectly predict a consistent improvement under distribution shift, as shown by the linear fit (red line).
    A hypothetical robustness intervention, shown in green, should provide \emph{effective robustness}, i.e., the intervention should improve the accuracy under distribution shift beyond what is predicted by the linear fit.\\\hspace{\textwidth}
    (Right) 
    We plot most of the \nummodels{} models in our testbed, highlighting those with the highest effective robustness using square markers.
    These models are still far from closing the accuracy gap induced by the distribution shift (ideally a robust model would fall on the $y=x$ line).
    Figure \ref{fig:dataset-shifts} shows a more detailed version of this plot with error bars for all points.
	} \label{fig:hypothetical-plots}
\end{figure*}

\textbf{Current synthetic robustness measures do not imply natural robustness.}
Prior work often evaluates on synthetic distribution shifts to measure robustness~\cite{hendrycks2019benchmarking,carlini2019evaluating,galloway2018adversarial}.
We find that current robustness measures for synthetic distribution shift are at most weakly predictive for robustness on the natural distribution shifts presently available. %
While there are good reasons to study synthetic forms of robustness -- for instance, adversarial examples are interesting from a security perspective -- synthetic distribution shifts alone do not provide a comprehensive measure of robustness at this time.
Moreover, as the right plot in Figure \ref{fig:hypothetical-plots} exemplifies, current robustness interventions are often (but not always) ineffective on the natural distribution shifts in our testbed.

\textbf{Training on more diverse data improves robustness.}
Across all of our experiments, the only intervention that improves robustness to multiple natural distribution shifts is training with a more diverse dataset.
This overarching trend has not previously been identified and stands out only through our large testbed.
Quantifying when and why training with more data helps is an interesting open question: while more data is generally helpful, we find some models that are trained on 100 times more data than the standard ImageNet training set but do not provide any robustness.

The goal of our paper is specifically \emph{not} to introduce a new classification method or image dataset.
Instead, our paper is a meta-study of current robustness research to identify overarching trends that span multiple evaluation settings.
This is particularly important if the ultimate goal of a research direction is to produce models that function reliably in a wide variety of contexts.
Our findings highlight robustness on real data as a clear challenge for future work.
Due to the diminishing returns of larger training datasets, addressing this robustness challenge will likely require new algorithmic ideas and more evaluations on natural distribution shifts. %

%% file: sec2-robustness-defn.tex
We first discuss how to measure robustness as a quantity distinct from accuracy.
In our experiments, we always have two evaluation settings: the ``standard'' test set, and the test set with distribution shift.
For a model $f$, we denote the two accuracies with $\acca(f)$ and $\accb(f)$, respectively.

When comparing the robustness of two models $f_a$ and $f_b$, one approach would be to rank the models by their accuracy under distribution shift.
However, this approach does not disentangle the robustness of a model from its accuracy on the standard test set.
As an example, consider a pair of models with accuracy $\acca(f_a) = 0.8$, $\accb(f_a) = 0.75$ (i.e., a 5\% drop in accuracy from the distribution shift), and  $\acca(f_b) = 0.9$, $\accb(f_b) = 0.76$ (a 14\% drop).
Model $f_b$ has higher accuracy on the second test set, but overall sees a drop of 14\% from the standard to the shifted test set.
In contrast, the first model sees only a 5\% drop.
Hence we would like to refer to the first model as more robust, even though it achieves lower accuracy on the shifted test set.

\textbf{Effective robustness.} The core issue in the preceding example is that standard accuracy (\acca) acts as a confounder.
Instead of directly comparing accuracies under distribution shift, we would like to understand if a model $f_b$ offers higher accuracy on the shifted test set \emph{beyond what is expected from having higher accuracy on the original test set}.
We call this notion of robustness beyond a baseline \emph{effective robustness}.
Graphically, effective robustness corresponds to a model being above the linear trend (red line) given by our testbed of standard models in Figure \ref{fig:hypothetical-plots} (left).

To precisely define effective robustness, we introduce $\beta(x)$, the baseline accuracy on the shifted test set for a given accuracy $x$ on the standard test set.
On the distribution shifts in our testbed, we instantiate $\beta$ by computing the parameters of a log-linear fit for the models without a robustness intervention (the red line in Figure~\ref{fig:hypothetical-plots}).
Empirically, this approach yields a good fit to the data.
For other distribution shifts, the baseline accuracy may follow different trends and may also depend on properties beyond the standard accuracy, e.g., model architecture.
\Cref{app:recipe-beta} contains detailed information on how to compute $\beta$.

Given the accuracy baseline $\beta$, we define the effective robustness of a model as
\[ \rho(f) = \accb(f) - \beta(\acca(f)) \; . \]
A model without special robustness properties falls on the linear fit and hence has $\rho(f)$ = 0.
The main goal of a robustness intervention is to increase $\rho$.
Models with large $\rho$ offer robustness beyond what we can currently achieve with standard models.

\textbf{Relative robustness.}
Effective robustness alone does not imply that a robustness intervention is useful.
In particular, a robustness intervention could increase $\rho$ for a model it is applied to, but at the same time \emph{decrease} both $\acca$ and $\accb$.
Such a robustness intervention would offer no benefits.
So to complement effective robustness, we also introduce \emph{relative robustness} to directly quantify the effect of an intervention on the accuracy under distribution shift.
For a model $f'$ with robustness intervention, derived from a model $f$ without the intervention, the relative robustness is $\tau(f') = \accb(f') - \accb(f)$.
We graphically illustrate this notion of robustness in Appendix \ref{app:rel-eff-robustness-plot}.

Overall, a useful robustness intervention should obtain \emph{both} positive effective and relative robustness. 
As we will see, only few classification models currently achieve this goal, and no models achieve both large effective and relative robustness.

%% file: sec3-testbed.tex
We now describe our experimental setup. 
A model $f$ is first trained on a fixed training set.
We then evaluate this model on two test sets: the ``standard'' test set (denoted $S_{1}$) and the test set with a distribution shift (denoted $S_{2}$).

A crucial question in this setup is what accuracy the model $f$ can possibly achieve on the test set with distribution shift.
In order to ensure that the accuracy on the two test sets are comparable, we focus on natural distribution shifts where  humans have thoroughly reviewed the test sets to include only correctly labeled images \cite{shankar2019systematic,recht2019imagenet,hendrycks2019natural,objectnet,imagenet}.\footnote{For ObjectNet \cite{objectnet}, \citet{borji2020objectnet} has pointed out potential label quality issues, but also found that a substantial accuracy drop remains when taking these issues into account.}
This implies that an ideal robust classifier does not have a substantial accuracy gap between the two test sets.
Indeed, recent work experimentally confirms that humans achieve similar classification accuracy on the original ImageNet test set and the ImageNetV2 replication study (one of the distribution shifts in our testbed) \cite{humanaccuracy}.

\subsection{Types of distribution shifts}
At a high level, we distinguish between two main types of distribution shift.
We use the term \emph{natural} distribution shift for datasets that rely only on unmodified images.
In contrast, we refer to distribution shifts as \emph{synthetic} if they involve modifications of existing images specifically to test robustness.
To be concrete, we next provide an overview of the distribution shifts in our robustness evaluation, with further details in Appendix \ref{app:eval-settings} and visual overviews in Appendices \ref{app:testbed-overview} and \ref{app:example-images}.

\subsubsection{Natural distribution shifts}
We evaluate on seven natural distribution shifts that we classify into three categories.

\textbf{Consistency shifts.} 
To evaluate a notion of robustness similar to $\ell_p$-adversarial examples but without synthetic perturbations, we measure robustness to small changes across video frames as introduced by \citet{gu2019using} and \citet{shankar2019systematic}.
The authors assembled sets of contiguous video frames that appear perceptually similar to humans,
but produce inconsistent predictions for classifiers. 
We define $S_1$ to be the set of ``anchor'' frames in each video, and evaluate the accuracy under distribution shift by choosing the worst frame from each frame set for a classifier.
This is the ``\texttt{pm-k}'' metric introduced by \citet{shankar2019systematic}.

\textbf{Dataset shifts.}
Next, we consider datasets $S_{2}$ that are collected in a different manner
from $S_{1}$ but still evaluate a classification task with a compatible set of classes.
These distribution shifts test to what extent current robustness interventions help with natural variations between datasets that are hard to model explicitly.
We consider four datasets of this variety:
(i) ImageNetV2, a reproduction of the ImageNet test set collected by \citet{recht2019imagenet};
(ii) ObjectNet, a test set of objects in a variety of scenes with 113 classes that overlap with ImageNet \cite{objectnet};
and, (iii) ImageNetVid-Robust-anchor and YTBB-Robust-anchor  \cite{shankar2019systematic}, which are the datasets constructed from only the anchor frames in the consistency datasets described above.
These two datasets contain 30 and 24 super-classes of the ImageNet class hierarchy, respectively.
For each of these distribution shifts, we define $S_1$ to be a subset of the ImageNet test set with the same label set as $S_{2}$ so that the accuracies are comparable.

\textbf{Adversarially filtered shifts.}
Finally, we consider an adversarially collected dataset, \imageneta{} \cite{hendrycks2019natural}.
\citet{hendrycks2019natural} assembled the dataset by downloading a large number of labeled images from Flickr, DuckDuckGo, iNaturalist, and other sites,
and then selected the subset that was misclassified by a ResNet-50 model.
We include \imageneta{} in our testbed to investigate whether the adversarial filtering process leads to qualitatively different results.
Since \imageneta{} contains only 200 classes, the standard test set $S_{1}$ here is again a subset of the ImageNet test set that has the same 200 classes as \imageneta{}.

\iftoggle{isneurips}{\input{figs-2-3}}{}

\subsubsection{Synthetic distribution shifts}
The research community has developed a wide range of synthetic robustness notions for image classification over the past five years.
In our study, we consider the following classes of synthetic distribution shifts, which cover the most common types of image perturbations.

\textbf{Image corruptions.} 
We include all corruptions from \citep{hendrycks2019benchmarking}, as well as some corruptions from \citep{Geirhos2018}.
These include common examples of image noise (Gaussian, shot noise), various blurs (Gaussian, motion), simulated weather conditions (fog, snow), and ``digital'' corruptions such as various JPEG compression levels.
We refer the reader to Appendix \ref{app:corruptions} for a full list of the \numcorruptions{} corruptions.

\textbf{Style transfer.} We use a stylized version of the ImageNet test set \citep{Huang_2017, geirhos2019imagenettrained}.

\textbf{Adversarial examples.} 
We include untargeted adversarial perturbations bounded in $\ell_\infty$- or $\ell_2$-norm by running projected gradient descent as described in \citep{madry2017towards}.
We use $\eps = \{\frac{0.5}{255}, \frac{2}{255}\}$ for $\ell_\infty$ and $\eps = \{0.1, 0.5\}$ for $\ell_2$ (further details in Appendix \ref{app:pgd}).

\subsection{Classification models}
Our model testbed includes \nummodels{} ImageNet models covering a variety of different architectures and training methods.
The models can be divided into the following three categories (see Appendix \ref{app:models} for a full list of all models and their categories).

\textbf{Standard models.} We refer to models trained on the ILSVRC 2012 training set without a specific robustness focus as \emph{standard} models.
This category includes \numstandardmodels{} models with architectures ranging from AlexNet to EfficietNet, e.g., \citep{alexnet,vgg,resnet,inceptionv1,tan2019efficientnet}.

\textbf{Robust models.} This category includes \numrobustnessmodels{} models with an explicit robustness intervention such as adversarially robust models \citep{xie2019feature,shafahi2019adversarial,cohen2019certified,salman2019provably,robustness}, models with special data augmentation \citep{geirhos2019imagenettrained,yun2019cutmix,xie2019adversarial,hendrycks2019augmix,engstrom2019exploring,devries2017improved,zhang2017mixup}, and models with architecture modifications \citep{zhang2019making}.

\textbf{Models trained on more data.} Finally, our testbed contains \nummoredatamodels{} models that utilize substantially more training data than the standard ImageNet training set.
This subset includes models trained on (i) Facebook's collection of 1 billion Instagram images \citep{wslimageseccv2018,yalniz2019billionscale}, (ii) the YFCC 100 million dataset \citep{yalniz2019billionscale}, (iii) Google's JFT 300 million dataset \citep{sun2017revisiting,xie2019selftraining}, (iv) a subset of OpenImages \citep{tencent-ml-images-2019}, or (v) a subset of the full ImageNet dataset of 21,841 classes \citep{imagenet11kmodel,chen2017dual,alex2019big}.

%% file: figs-2-3.tex
\begin{figure*}[t!]
    \centering
  \begin{subfigure}[t]{0.48\textwidth}
    \includegraphics[width=\textwidth]{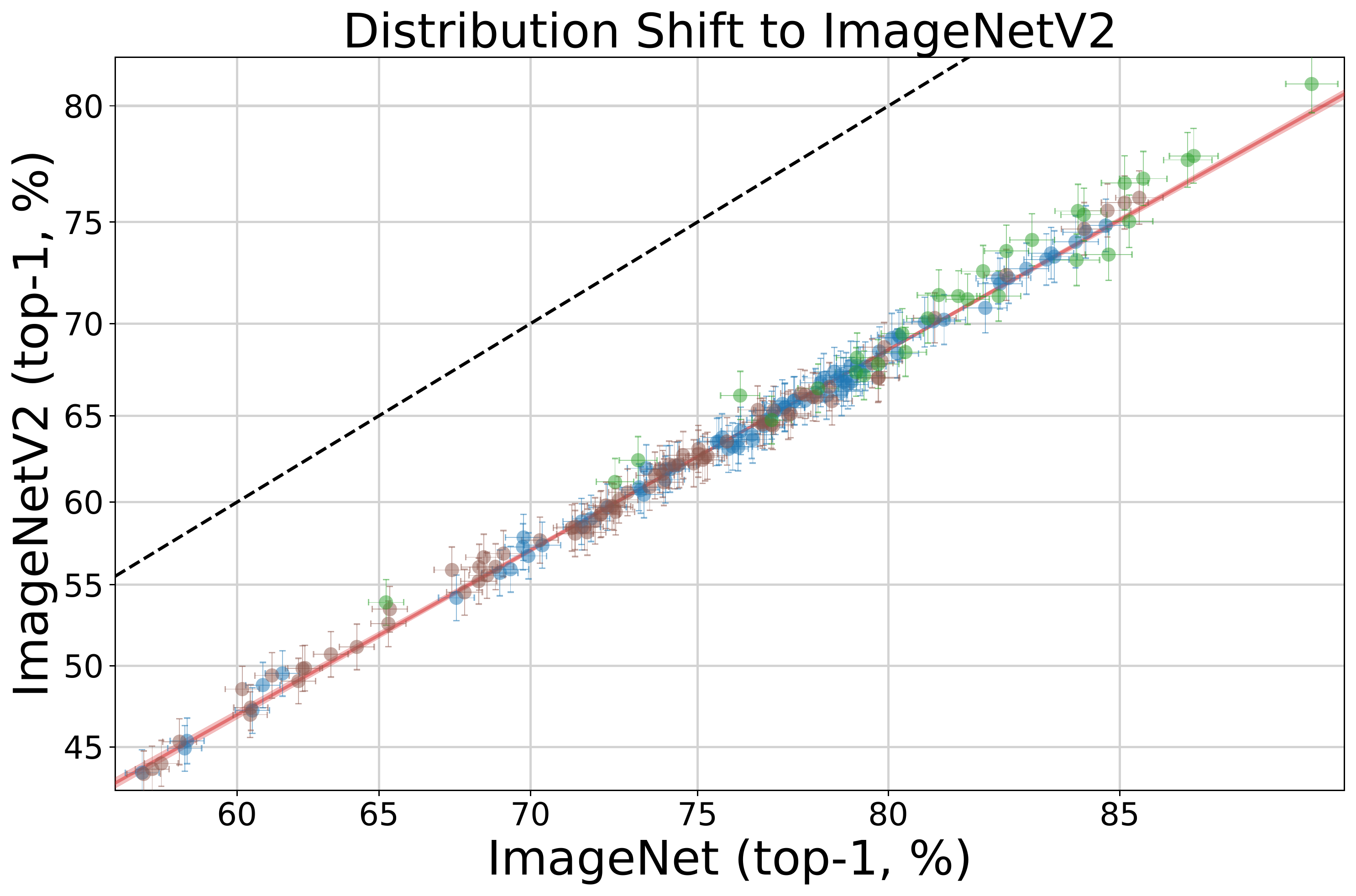}
  \end{subfigure}
  \hfill
  \begin{subfigure}[t]{0.48\textwidth}
    \includegraphics[width=\textwidth]{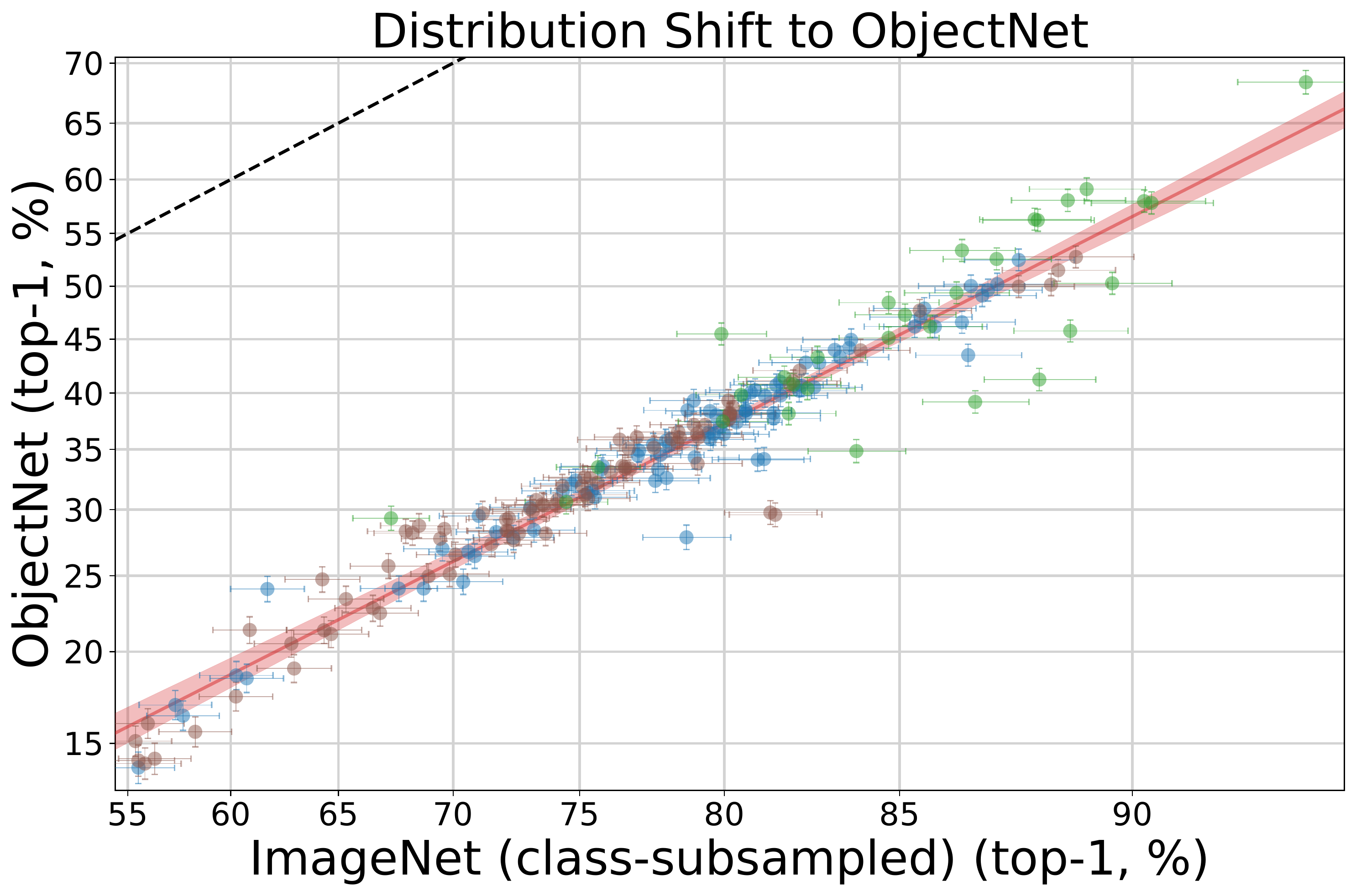}
  \end{subfigure}
  \par\medskip
  \begin{subfigure}[t]{0.48\textwidth}
    \includegraphics[width=\textwidth]{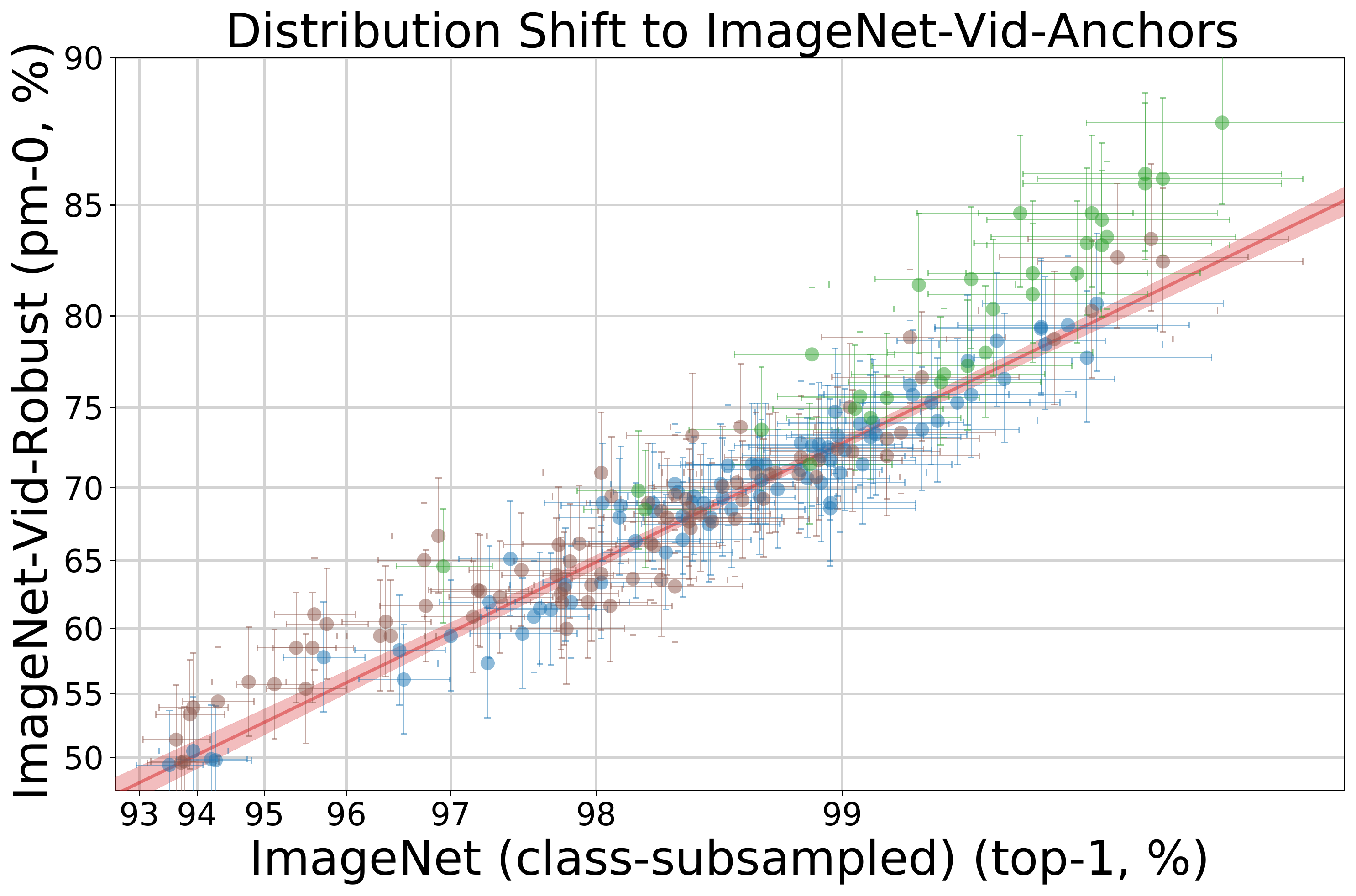}
  \end{subfigure}
  \hfill
  \begin{subfigure}[t]{0.48\textwidth}
    \includegraphics[width=\textwidth]{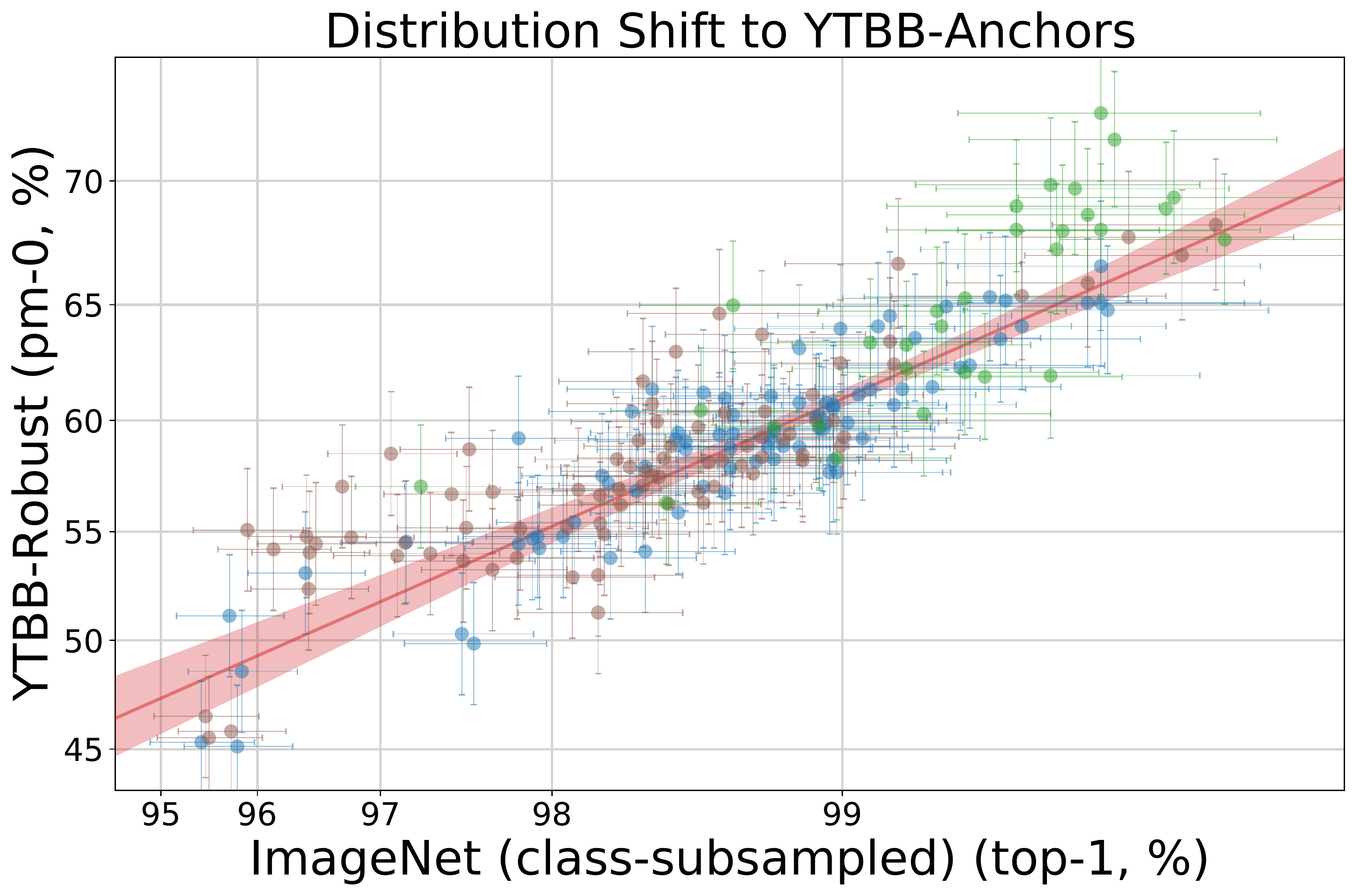}
  \end{subfigure}
  \begin{subfigure}[t]{\textwidth}
    \centering \includegraphics[width=0.7\textwidth]{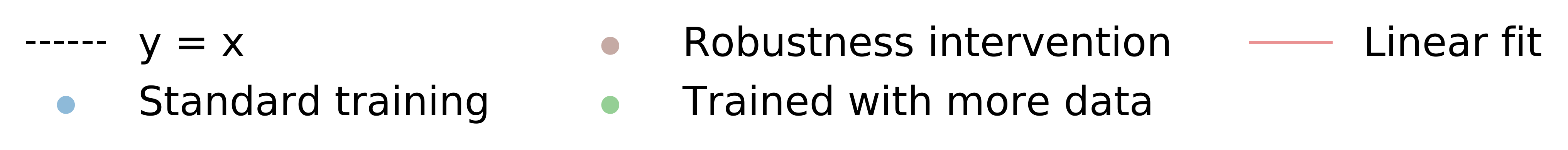}
  \end{subfigure}
  \iftoggle{isneurips}{\vspace{-.65cm}}{}
  \caption{
    Model accuracies on the four natural dataset shifts: \imagenetnew{} (top left), \objectnet{} (top right), \imagenetvidrobust{}-anchor (bottom left), and \ytbbrobust{}-anchor (bottom right).
    These plots demonstrate that the standard test accuracy (x-axis) is a reliable predictor for the test accuracy under distribution shift (y-axis), especially for models trained without a robustness intervention.
    The notable outliers to this trend are some models trained on substantially more data.
    For \objectnet{}, \imagenetvidrobust{}-anchor, and \ytbbrobust{}-anchor, we show the accuracy on a subset of the ImageNet classes on the x-axis to match the label space of the target task (y-axis). 
    Each data point corresponds to one model in our testbed and is shown with 99.5\% Clopper-Pearson confidence intervals.
    The axes were adjusted using logit scaling and the linear fit was computed in the scaled space on only the standard models.
    The red shaded region is a 95\% confidence region for the linear fit from 1,000 bootstrap samples.
  }
  \label{fig:dataset-shifts}
  \iftoggle{isneurips}{\vspace{-.39cm}}{}
\end{figure*}

\begin{figure*}[ht!]
    \centering
  \begin{subfigure}[t]{0.48\textwidth}
    \includegraphics[width=\linewidth]{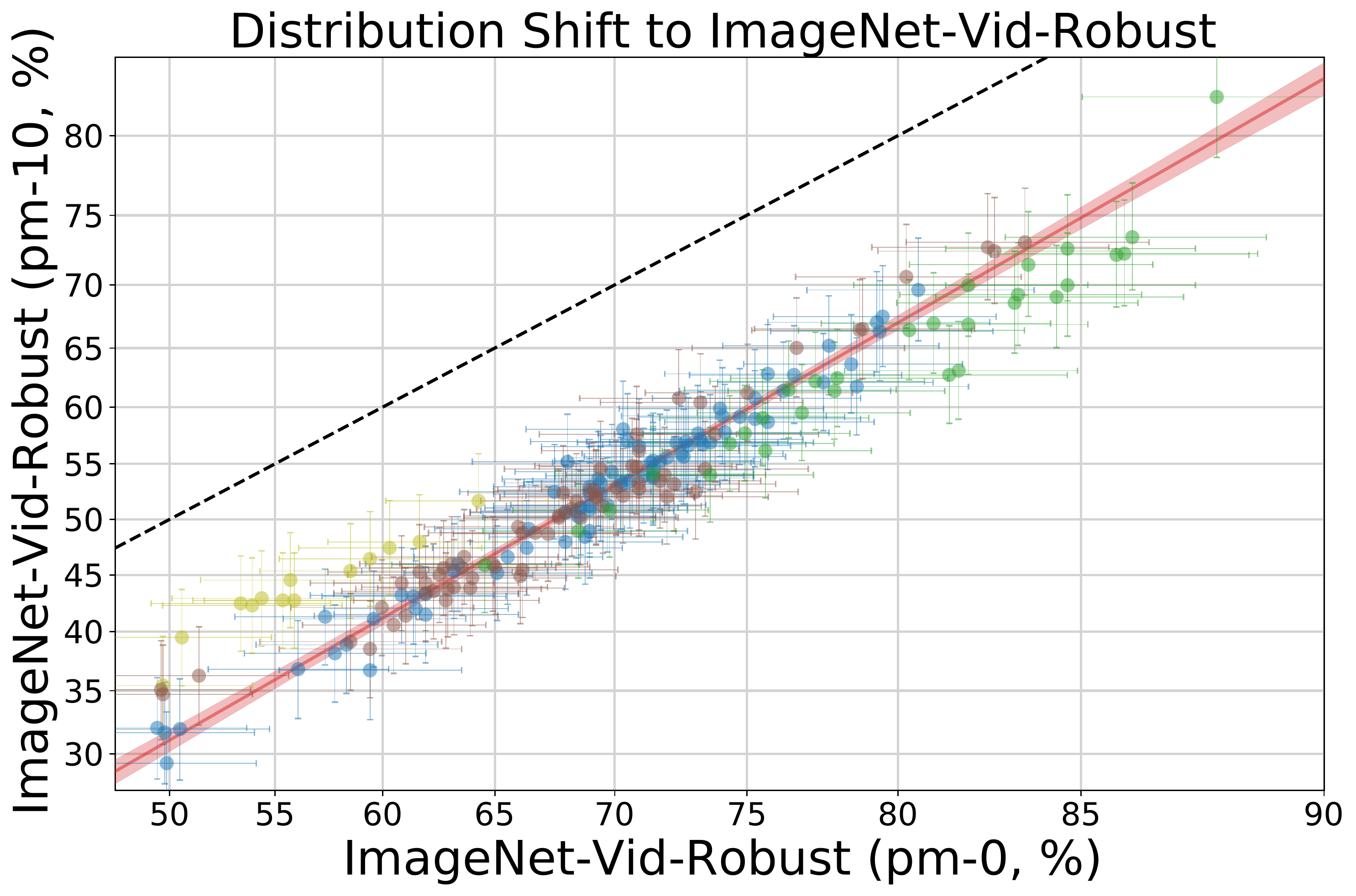}
  \end{subfigure}
  \hfill
  \begin{subfigure}[t]{0.48\textwidth}
    \includegraphics[width=\linewidth]{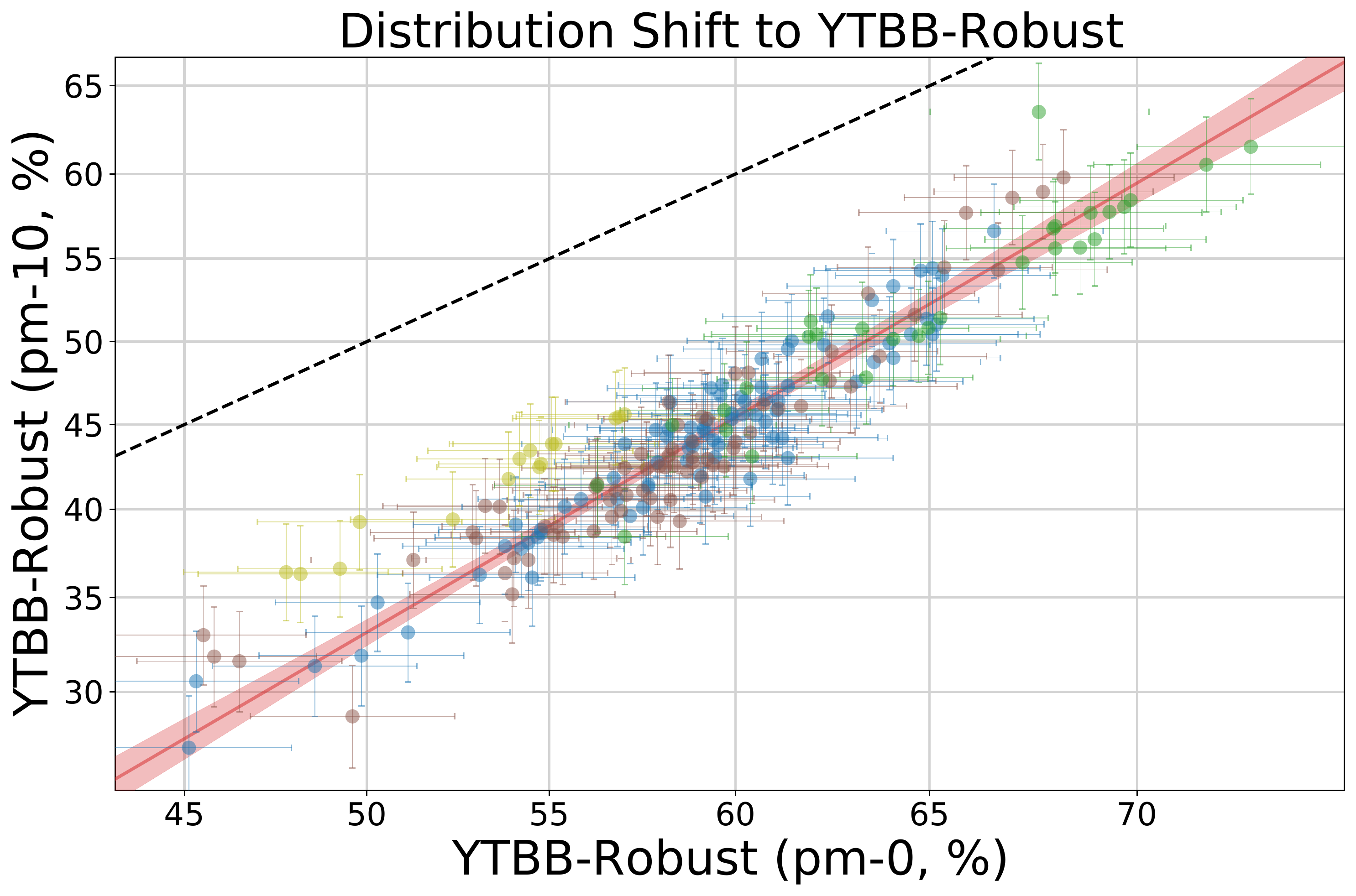}
  \end{subfigure}
  \begin{subfigure}[t]{\textwidth}
    \centering \includegraphics[width=0.9\textwidth]{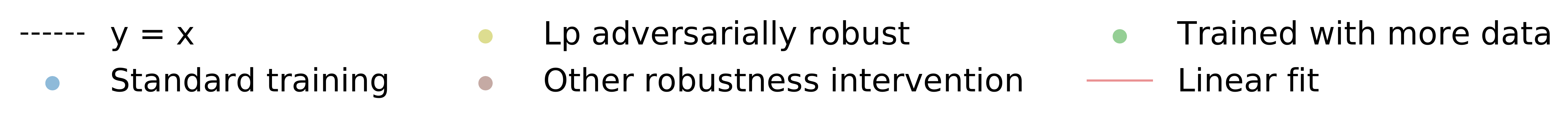}
  \end{subfigure}
  \iftoggle{isneurips}{\vspace{-.5cm}}{}
  \caption{
    Model accuracies on the two consistency shifts: \imagenetvidrobust{} (left), and \ytbbrobust{} (right).
    Both plots are shown with evaluation on pm-0 (anchor frames) on the x-axis and pm-10 (worst case prediction in a 20-frame neighborhood) on the y-axis.
    This plot shows that most current robustness interventions do not provide robustness to consistency distribution shifts.
    The notable outliers to this trend are $\ell_p$-adversarially robust models and EfficientNet-L2 (NoisyStudent). We color the adversarially robust models separately in this figure to illustrate this phenomenon.
    Confidence intervals, axis scaling, and the linear fit are computed similarly to Figure \ref{fig:dataset-shifts}.
    }
  \label{fig:consistency-shifts}
  \iftoggle{isneurips}{\vspace{-.1cm}}{}
\end{figure*}

%% file: sec4-results.tex
\iftoggle{isneurips}{}{\input{figs-2-3}}

We now present our main experiments.
First, we measure how much effective and relative robustness models achieve on the natural distribution shifts in our testbed.
Then we investigate to what extent robustness on synthetic distribution shift is predictive of robustness on natural distribution shift.

\subsection{Results on natural distribution shifts}
\label{sec:natshiftresults}

Following the categorization in Section \ref{sec:setup}, we measure the robustness of classification models on three types of natural distribution shift.
Appendix \ref{app:ablations} contains variations of the figures referenced in this section.
For further detail, we have made interactive plots available at \href{http://robustness.imagenetv2.org/}{\color{blue}http://robustness.imagenetv2.org/}.

\textbf{Dataset shifts.} Figure \ref{fig:dataset-shifts} shows the effective robustness of models on the four dataset shifts in our testbed.
In each case, we find that the standard test accuracy (x-axis) is a good predictor for the test accuracy under distribution shift (y-axis).
The linear fit is best for \imagenetnew{}, \objectnet{}, and \imagenetvidrobust{} with respective $r^2$ scores of 1.00, 0.95, and 0.95, but is more noisy for \ytbbrobust{} ($r^2=0.83$). 
The noisy fit on \ytbbrobust{} is likely due to the fact that the categories in \ytbbrobust{} are not well aligned with those of ImageNet, where the models were trained \cite{shankar2019systematic}.
Another potential reason is that the video test sets are significantly smaller (2,530 images in YTBB and 1,109 images in ImageNet-Vid-Robust).

In the high accuracy regime (above the 76\% achieved by a ResNet-50), the main outliers in terms of positive effective robustness are models trained on substantially more data than the standard ImageNet training set. 
This includes a ResNet152 model trained on 11,000 ImageNet classes ($\rho = 2.1$\%) \citep{imagenet11kmodel}, several ResNeXt models trained on 1 billion images from Instagram ($\rho = 1.5$\%) \citep{wslimageseccv2018}, and the EfficientNet-L2 (NoisyStudent) model trained on a Google-internal \jft{} dataset of 300 million images ($\rho = 1.1$\%) \citep{xie2019selftraining}.
However, not all models trained on more data display positive effective robustness.
For instance, a ResNet101 trained on the same \jft{} dataset has an effective robustness of $\rho = -0.23$\% \citep{sun2017revisiting}.
We conduct additional experiments to investigate the effect of training data in \Cref{sec:more-data}.
Appendix \ref{app:model-accuracies} contains a full list of models with their effective robustness numbers. 
On \ytbbrobust{}, a few data augmentation strategies and $\ell_p$-robust models display positive effective robustness; we investigate this further in Appendix \ref{app:rel-eff-robustness-resnet50}.

\textbf{Consistency shifts.} We plot the effective robustness of models on consistency shifts in ~\Cref{fig:consistency-shifts}.
Interestingly, we observe that $\ell_p$-adversarially robust models display substantial effective robustness to \imagenetvidrobust{} (average $\rho = 6.7$\%) and \ytbbrobust{} (average $\rho = 4.9$\%).
This suggests that these models are not only more robust to synthetic perturbations, but also offer robustness for the perceptually small variations between consecutive video frames.

However, these gains in effective robustness do not necessarily lead to relative robustness.
On average, relative robustness on both datasets is negative (average $\tau = -8.5$\% on \imagenetvidrobust{} and average $\tau = -0.7$\% on \ytbbrobust{} for ResNet50 models).
See Appendix \ref{app:rel-eff-robustness-resnet50} (\Cref{fig:rel_eff_robutness_consistency_shift}) for a visual comparison. 
Among the models trained on more data, only one achieves both high accuracy and substantial effective robustness: EfficientNet-L2 (NoisyStudent) \cite{xie2019selftraining} has $\rho = 2.4$\% and $\rho = 7.4$\% on \imagenetvidrobust{} and \ytbbrobust{}, respectively.

\iftoggle{isneurips}{
  \begin{figure}[t!]
    \begin{minipage}[c]{0.48\textwidth}
      \includegraphics[width=\linewidth]{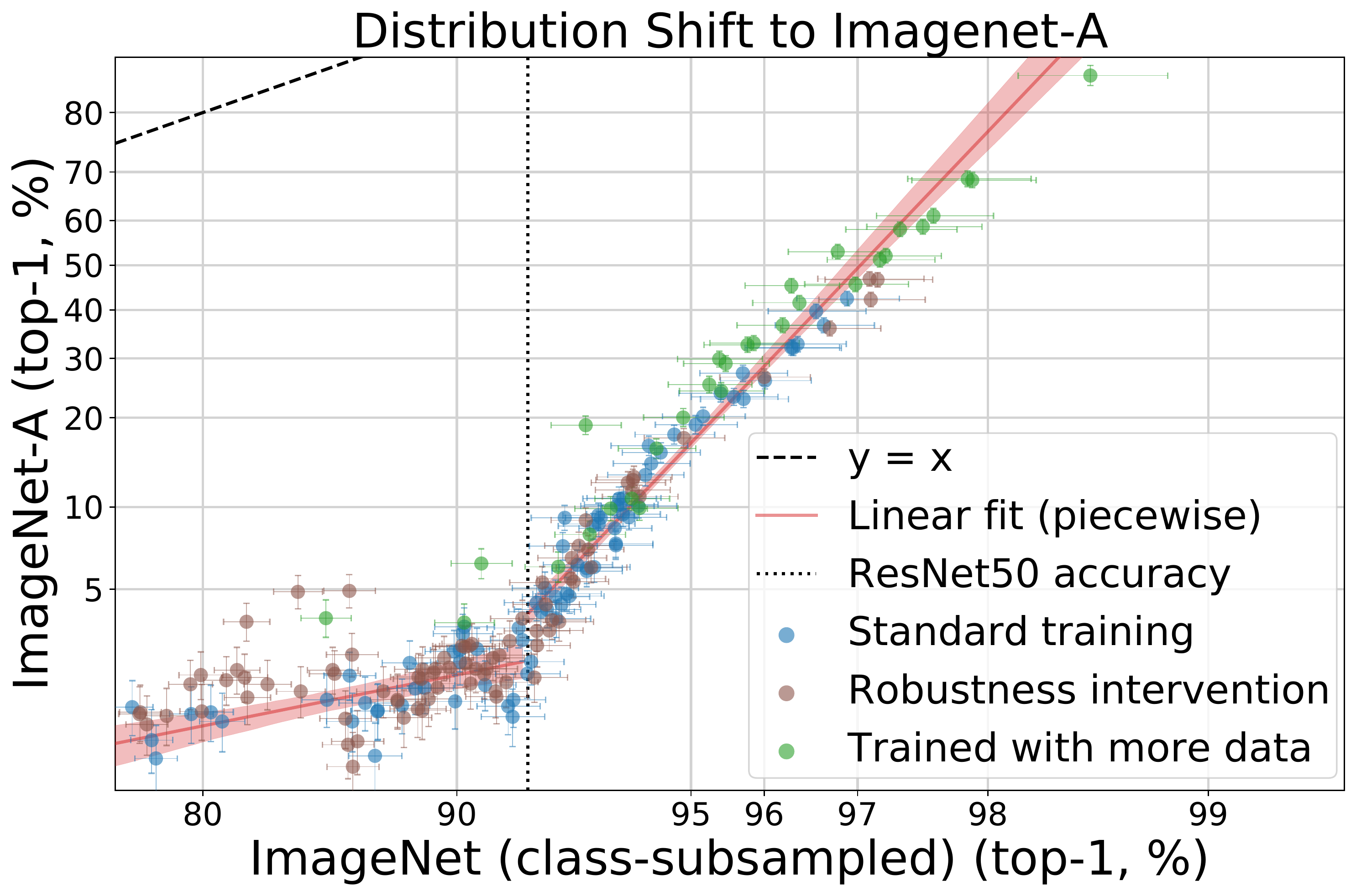}
    \end{minipage}
    \hfill
    \begin{minipage}[c]{0.48\textwidth}
      \caption{
        Model accuracies on \imageneta{}, a dataset adversarially filtered to contain only images incorrectly classified by a ResNet50 trained on ImageNet.
        This filtering results in a `knee' curve: models with lower ImageNet accuracy than ResNet-50 have near-chance performance on \imageneta{}, while models with higher ImageNet accuracy improve drastically on \imageneta{}.
        The linear fit is computed piecewise around the ResNet50 model accuracy.
      } \label{fig:adversarial-shifts}
    \end{minipage}
    \iftoggle{isneurips}{\vspace{-.3cm}}{}
  \end{figure}
}{
  \begin{figure*}[t!]
  \centering
    \includegraphics[width=0.6\linewidth]{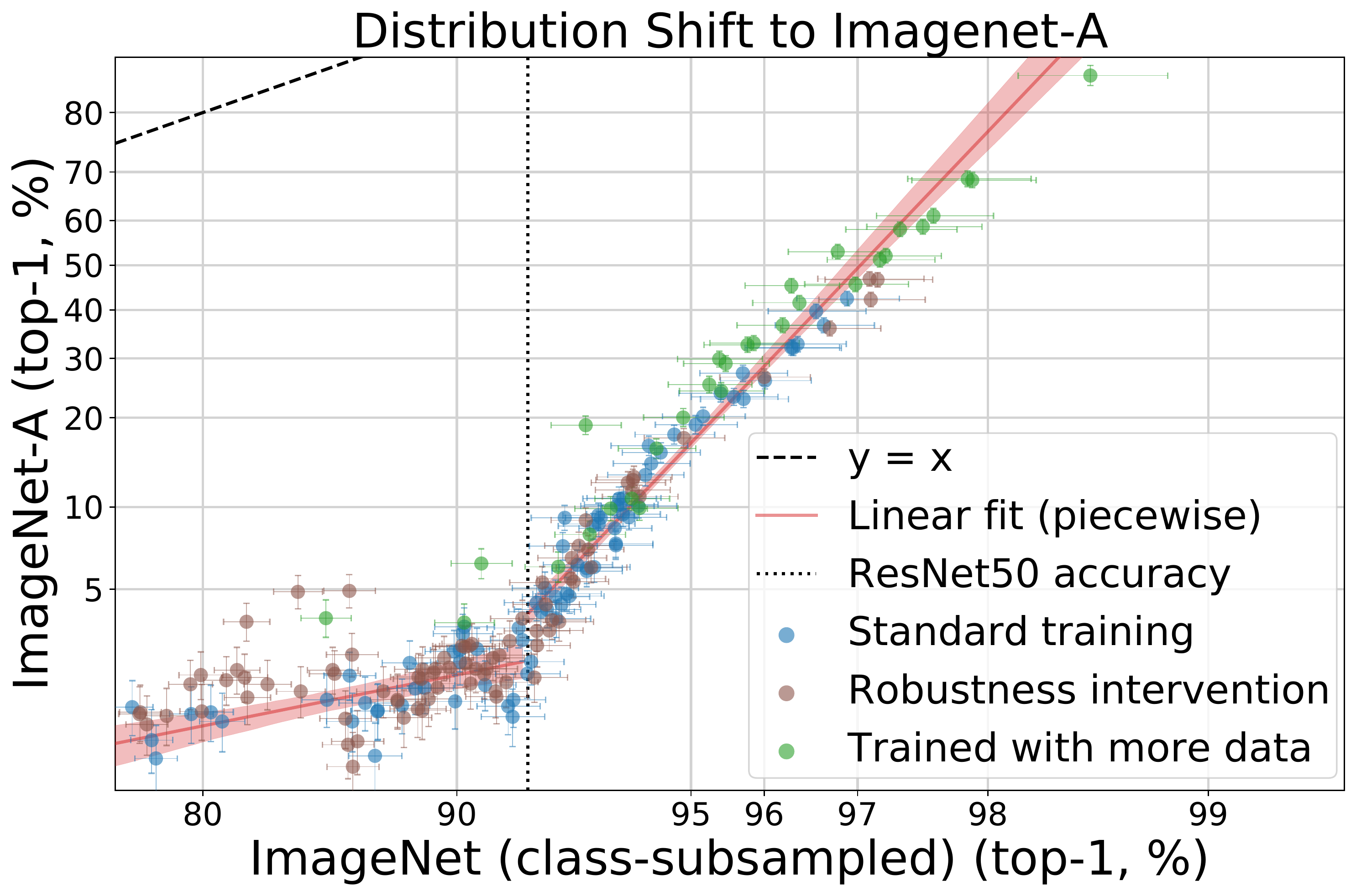}
  \caption{
      Model accuracies on \imageneta{}, a dataset adversarially filtered to contain only images incorrectly classified by a ResNet50 trained on ImageNet.
      This filtering results in a `knee' curve: models with lower ImageNet accuracy than ResNet-50 have near-chance performance on \imageneta{}, while models with higher ImageNet accuracy improve drastically on \imageneta{}.
      ImageNet classes were subsampled to match the class distribution of \imageneta{}.
      Confidence intervals and axis scaling are computed similarly to Figure \ref{fig:dataset-shifts}. 
      The linear fit is computed piecewise around the ResNet50 model accuracy.
  } \label{fig:adversarial-shifts}
  \end{figure*}
}

\textbf{Adversarially filtered shifts.} \imageneta{} \cite{hendrycks2019natural} was created by classifying a set of images with a ResNet50 and only keeping the misclassified images. 
Interestingly, this approach creates a ``knee'' in the resulting scatter plot (see Figure \ref{fig:adversarial-shifts}): models below a ResNet50's standard accuracy have close to chance performance on \imageneta{},\footnote{Chance performance is 0.5\% as \imageneta{} contains 200 classes.} and models above a ResNet50's standard accuracy quickly close the accuracy gap.
In the high accuracy regime, every percentage point improvement on ImageNet brings at least an 8\% improvement on ImageNet-A.
This is in contrast to datasets that are not constructed adversarially, where the initial accuracy drops are smaller, but later models make slow progress on closing the gap. 
These results demonstrate that adversarial filtering does not necessarily lead to harder distribution shifts.

\subsection{Results on synthetic distribution shifts}
Given the difficulty of collecting real world data to measure a model's robustness to natural distribution shifts, an important question is whether there are synthetic proxies.
We now study to what extent robustness to the above synthetic distribution shifts predicts robustness on these natural distribution shifts.

\iftoggle{isneurips}{
In Figure \ref{fig:synthetic-shifts}, we analyze the predictiveness of one commonly studied synthetic robustness metric: average accuracy on image corruptions \cite{hendrycks2019benchmarking}. %
We compare this metric with effective robustness on \imagenetnew.
While effective robustness is only one aspect (c.f.\ Section \ref{sec:robustness-defn}), it is a necessary prerequisite for a model to have helpful robustness properties.

The plots show that robustness under this synthetic distribution shift does not imply that the
corresponding model has effective robustness on \imagenetnew{} (the Pearson correlation coefficient is $r = 0.24$).
In Appendix \ref{app:pgd-robustness}, we repeat the above experiment for accuracy drop under PGD adversarial attacks \cite{madry2017towards} and also find a weak correlation ($r = -0.05$).
Appendix \ref{app:eff-robust-syn-nat-robustness} further extends the experiment by comparing both synthetic distribution shift measures with the remaining natural distribution shifts in our testbed and reaches similiar conclusions.

}{

\begin{figure*}[t!]
    \centering
  \begin{subfigure}[t]{0.48\textwidth}
    \includegraphics[width=\linewidth]{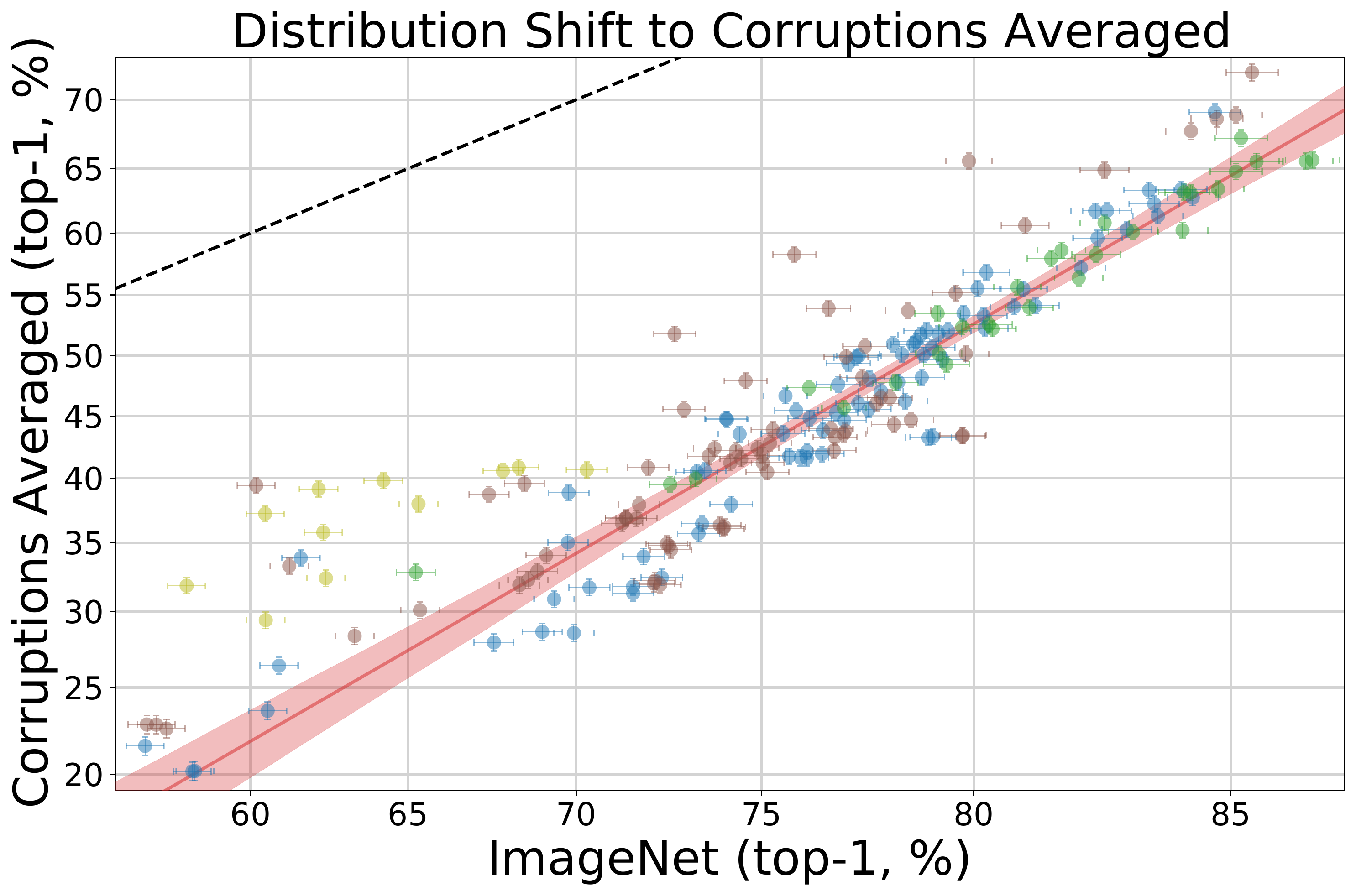}
  \end{subfigure}
    \hfill
  \begin{subfigure}[t]{0.48\textwidth}
    \includegraphics[width=\linewidth]{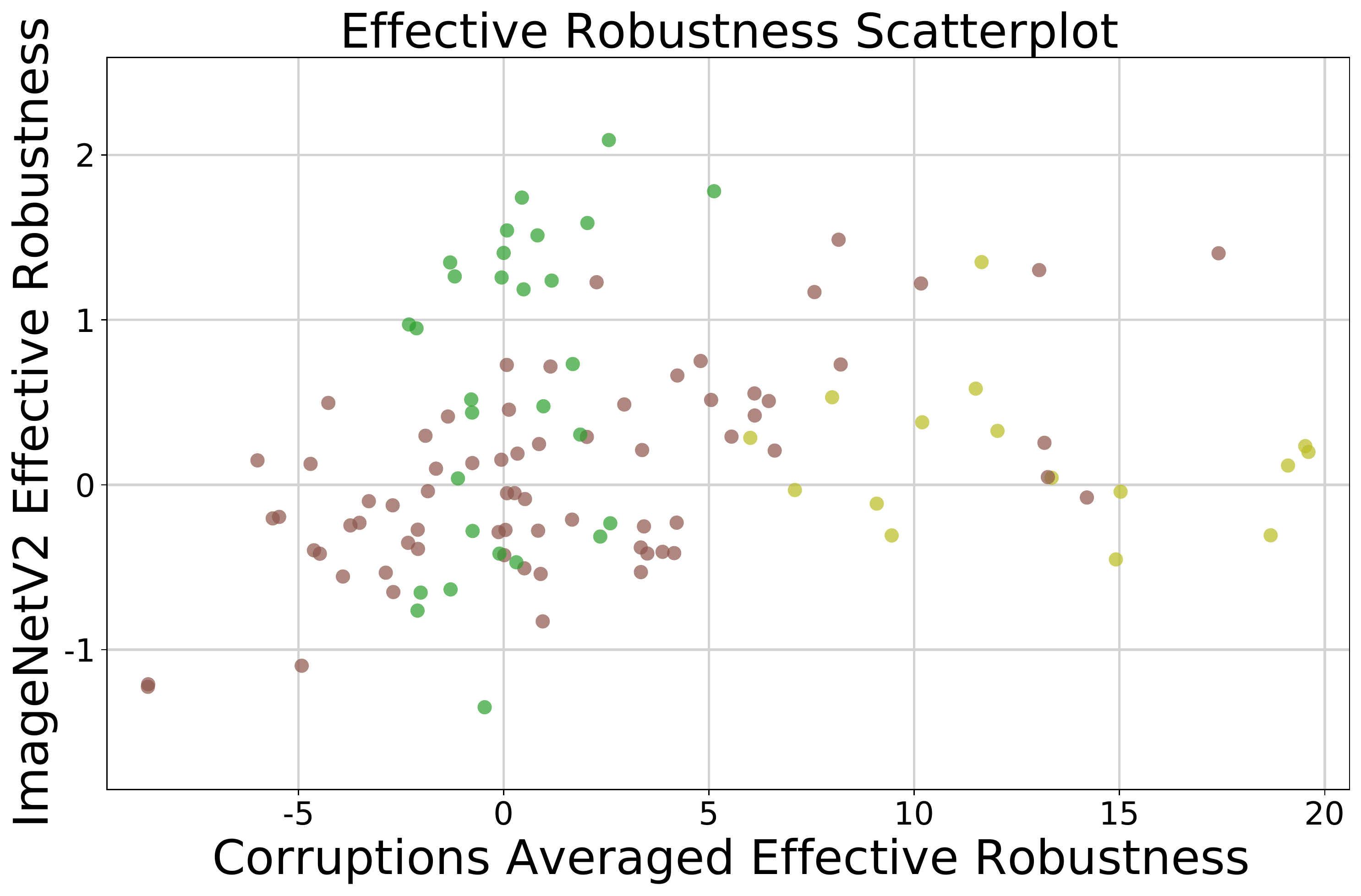}
  \end{subfigure}
  \par\medskip
  \begin{subfigure}[t]{0.48\textwidth}
    \includegraphics[width=\linewidth]{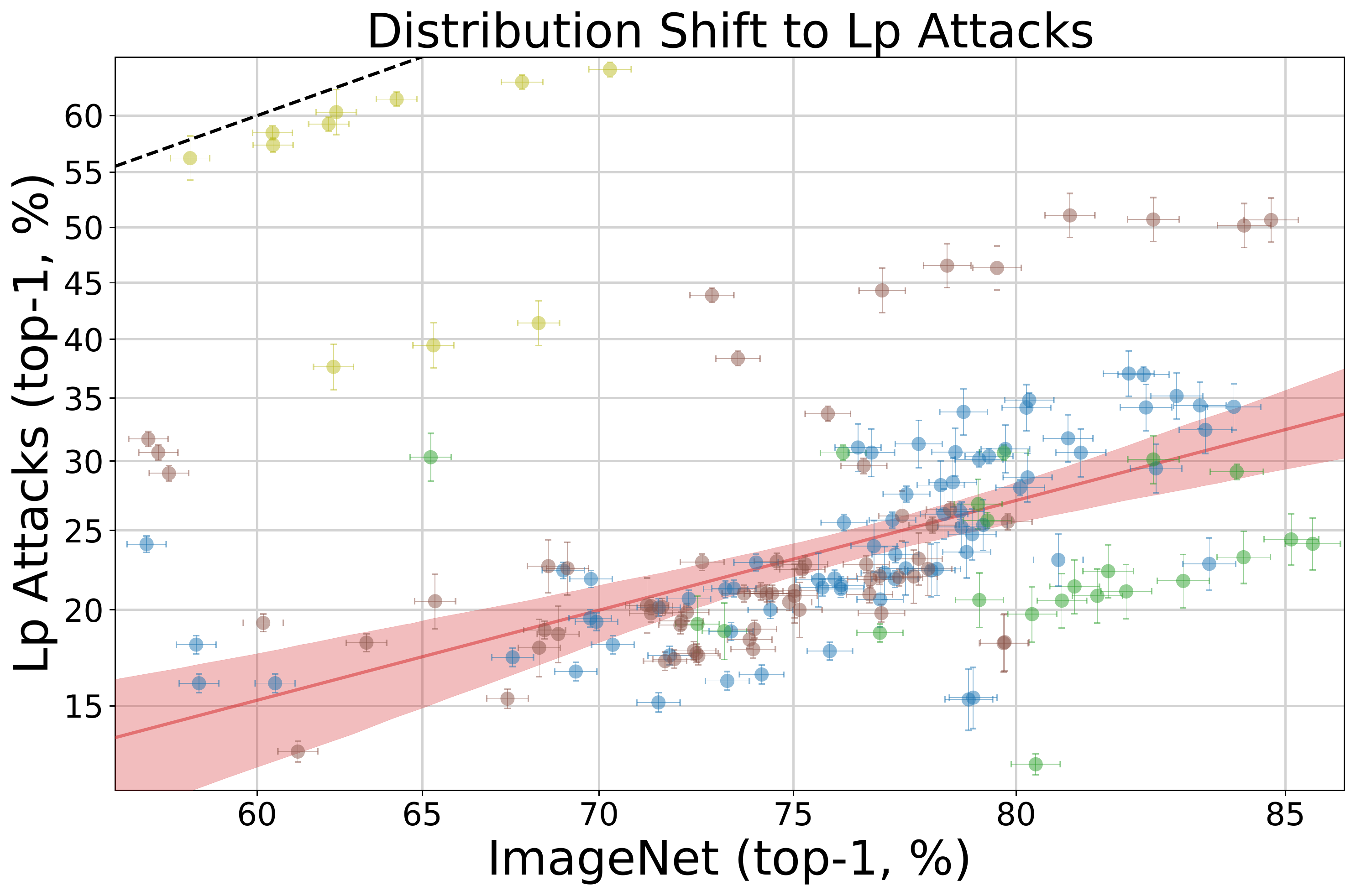}
  \end{subfigure}
    \hfill
  \begin{subfigure}[t]{0.48\textwidth}
    \includegraphics[width=\linewidth]{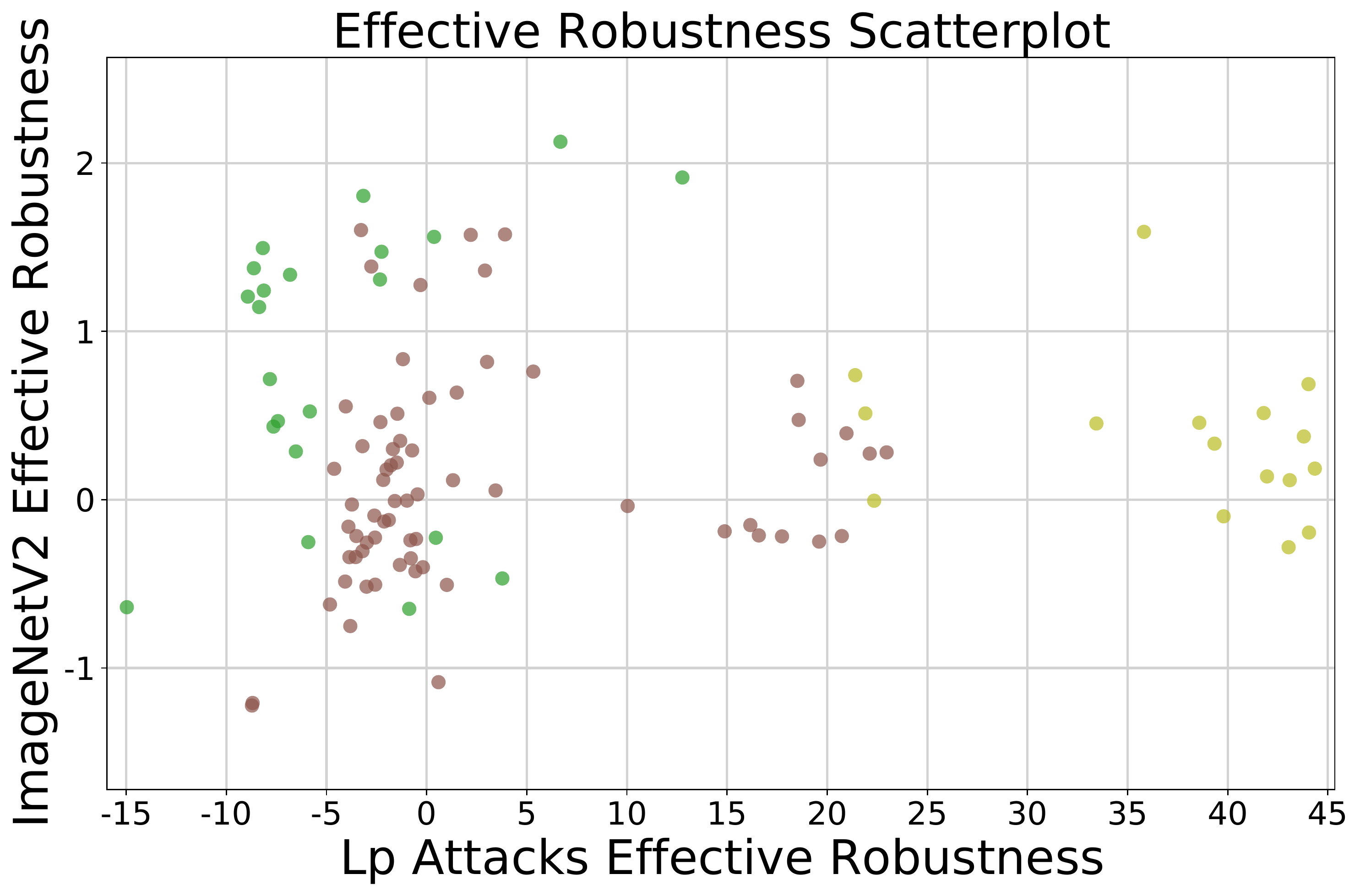}
  \end{subfigure}
  \begin{subfigure}[t]{\textwidth}
    \centering \includegraphics[width=0.9\textwidth]{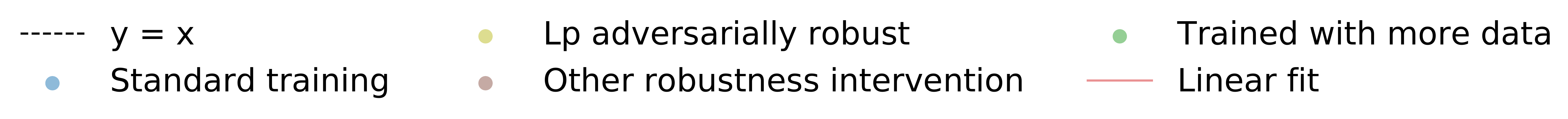}
  \end{subfigure}

  \caption{
    Model accuracies under image corruptions (top row) and $\ell_p$-attacks (bottom row).
    Similar to Figure \ref{fig:dataset-shifts}, the left plots show the effective robustness for each synthetic distribution shift.
    Multiple non-standard models achieve substantial effective robustness, corroborating recent research progress on creating models robust to synthetic shift. %
    The right plots show the correlation between the effective robustness for each synthetic shift and the \imagenetnew{} distribution shift (top left in Figure \ref{fig:dataset-shifts}) for the non-standard models.
    Both image corruptions and $\ell_p$-attacks are very weakly predictive of effective robustness on \imagenetnew: there are several models that achieve high effective robustness under the synthetic measures but little to no effective
    robustness on \imagenetnew.
    }
  \label{fig:synthetic-shifts}
\end{figure*}

In Figure \ref{fig:synthetic-shifts}, we analyze the predictiveness of two commonly studied synthetic robustness metrics: average accuracy on image corruptions \cite{hendrycks2019benchmarking}, and  average accuracy drop under a range of PGD adversarial attacks \cite{madry2017towards}.
We compare these metrics with effective robustness on \imagenetnew.
While effective robustness is only one aspect (c.f.\ Section \ref{sec:robustness-defn}), it is a necessary prerequisite for a model to have helpful robustness properties.

The plots show that robustness under either of these synthetic distribution shifts does not imply that the
corresponding model has effective robustness on \imagenetnew{} (the Pearson correlation coefficients are $r = 0.24$ and $r = -0.05$ for image corruptions and $\ell _p$-adversarial attacks, respectively). 
Appendix \ref{app:eff-robust-syn-nat-robustness} further extends the experiment by comparing both synthetic distribution shift measures with the remaining natural distribution shifts in our testbed and reaches similiar conclusions.
}

Our analysis of the aggregate measures proposed in prior work does not preclude that specific synthetic distribution shifts do predict behavior on natural distribution shifts.
Instead, our results show that averaging a large number of synthetic corruptions does not yield a comprehensive robustness measure that also predicts robustness on natural distribution shift.

To extend on this analysis, in Appendix \ref{app:correlation} we find that no individual synthetic measure in our testbed is a consistent predictor of natural distribution shift, but some synthetic shifts are substantially more predictive than others.
For instance, $\ell_p$-robustness has the highest correlation with consistency shifts,
and some image corruptions such as brightness or Gaussian blur have higher correlation with dataset shifts.
However, our testbed indicates that these synthetic measures are not necessarily causal, i.e., models trained with brightness or Gaussian blur do not have substantial effective robustness on dataset shifts.
Further analyzing relationships between individual synthetic and natural distribution shifts is an interesting avenue for future work.

\iftoggle{isneurips}{
\begin{figure*}[t!]
    \centering
  \begin{subfigure}[t]{0.48\textwidth}
    \includegraphics[width=\linewidth]{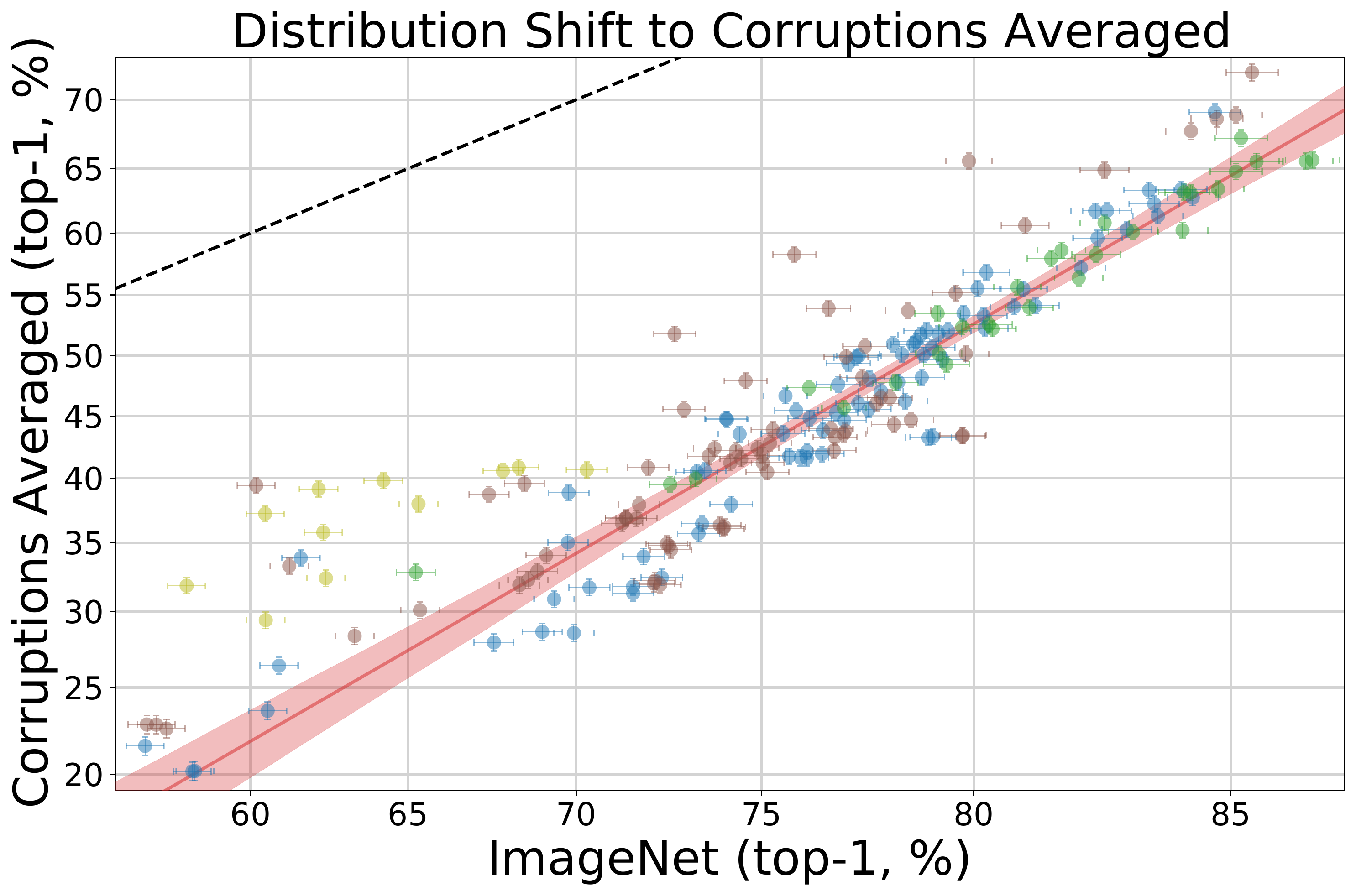}
  \end{subfigure}
    \hfill
  \begin{subfigure}[t]{0.48\textwidth}
    \includegraphics[width=\linewidth]{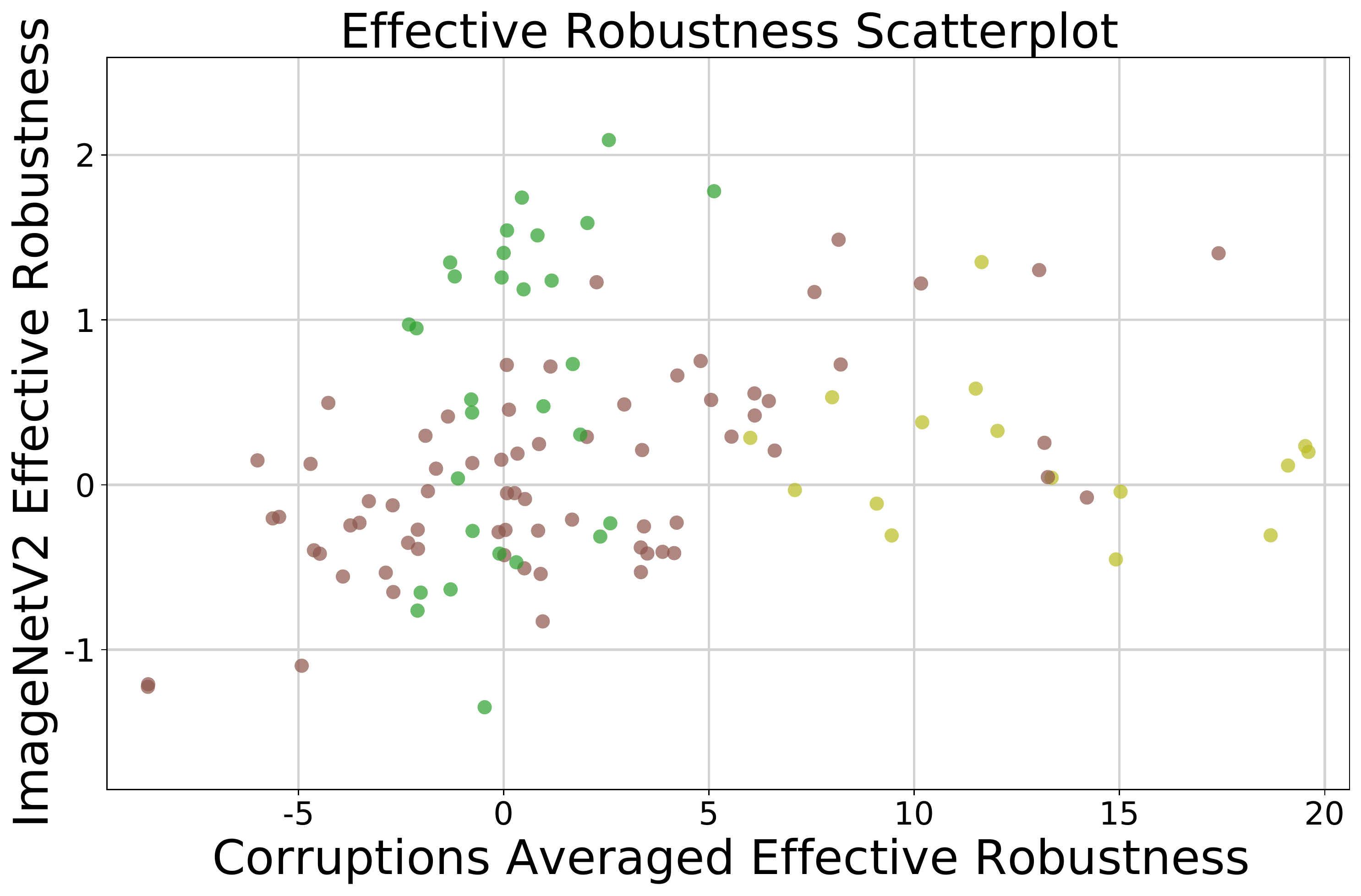}
  \end{subfigure}
  \begin{subfigure}[t]{\textwidth}
    \centering \includegraphics[width=0.7\textwidth]{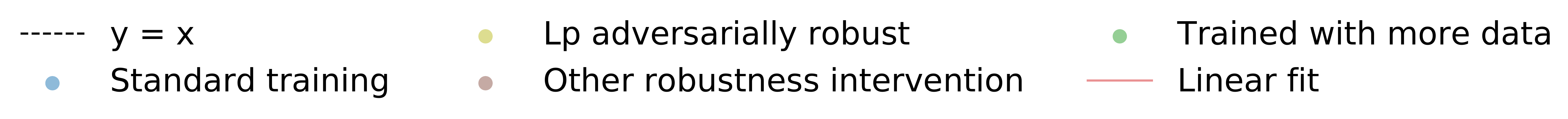}
  \end{subfigure}
  \vspace{-.6cm}
  \caption{
    Model accuracies under image corruptions.
    Similar to Figure \ref{fig:dataset-shifts}, the left plot shows the effective robustness for this synthetic distribution shift.
    Multiple non-standard models achieve substantial effective robustness, corroborating recent research progress on creating models robust to synthetic shifts. %
    The right plot shows the correlation between the effective robustness for image corruptions  and the \imagenetnew{} distribution shift (top left in Figure \ref{fig:dataset-shifts}) for the non-standard models.
    Image corruptions are only weakly predictive of effective robustness on \imagenetnew: there are several models that achieve high effective robustness under image corruptions but little to no effective
    robustness on \imagenetnew.
    }
  \label{fig:synthetic-shifts}
  \vspace{-.4cm}
\end{figure*}
}

\subsection{Takeaways and discussion}

To recap our results, we now discuss two of the central questions in our paper:
Do current robustness interventions help on real data? And is synthetic robustness correlated with natural robustness?

Across our study, current robustness interventions offer little to no improvement on the natural distribution shifts presently available.

For dataset shifts, we find that models trained with substantially more data yield a small improvement.
However, the amount of extra data needed is orders of magnitude larger than the standard ImageNet training set, and the models show only small gains (in the best case improving the accuracy drop from 8.6\% to 7.5\% on \imagenetnew{} for EfficientNet-L2 NoisyStudent).
These results suggest that current robustness interventions methods do not provide benefits on the dataset shifts in our study.

For consistency shifts, adversarially trained models generally have effective robustness, but usually little or no relative robustness.
On \imagenetvidrobust, the baseline models without adversarial training still achieve higher accuracy under distribution shift.
A notable outlier is EfficietNet-L2 (NoisyStudent) \cite{xie2019selftraining}, which utilizes self-training and exhibits high effective robustness in the high accuracy regime.
Self-training has recently been shown to help adversarial robustness as well \cite{carmon2019unlabeled,uesato2019labels,najafi2019robustness}.
Investigating the effect of self-training on robustness is an interesting direction for future work.

Moreover, we find that current aggregate metrics for synthetic robustness are at most weakly correlated with natural robustness.
Effective robustness under non-adversarial image corruptions or $\ell_p$-attacks does not imply effective robustness to natural distribution shifts.
While much progress has been made on creating models robust to synthetic distribution shift, new methods may be needed to handle natural shifts.

%% file: secX-more-data.tex
\section{How does the amount of training data impact robustness?}
\label{sec:more-data}

As discussed in Section \ref{sec:natshiftresults}, multiple models trained on more data achieve positive effective robustness on dataset shifts.
However, this effect is not uniform.
Among others, the ResNet101 model trained on \jft{} has negligible effective robustness ($\rho = -0.23$\%) despite being trained on $300\times$ more data than standard ImageNet models.
A possible explanation is that differences in label diversity or quality play a role in promoting robustness.
We investigate the role of data in more detail with two experiments.

\textbf{Varying the number of images per class.}
We start by subsampling the ILSVRC-2012 training set by factors of \{2, 4, 8, 16, 32\} and show the impact on accuracy and robustness on \imagenetnew{} in~\Cref{fig:subsample-iid}.
While larger training subsets yield higher accuracies, they do not improve effective robustness, at least for \imagenetnew.

\textbf{Varying the number of classes.}
Next, we subsample ImageNet in a more biased way by varying the set of classes.
First, we create three subsets of the ILSVRC training set with 500, 250, and 125 classes and train models on these subsets.
We then evaluate all models on the 125 class subset and show the results in \Cref{fig:subsample-iid}.
Varying the number of classes again affects accuracies, but does not impact effective robustness.

\iftoggle{isneurips}{
	\begin{figure*}[h]
	  \centering
	    \includegraphics[width=0.45\linewidth]{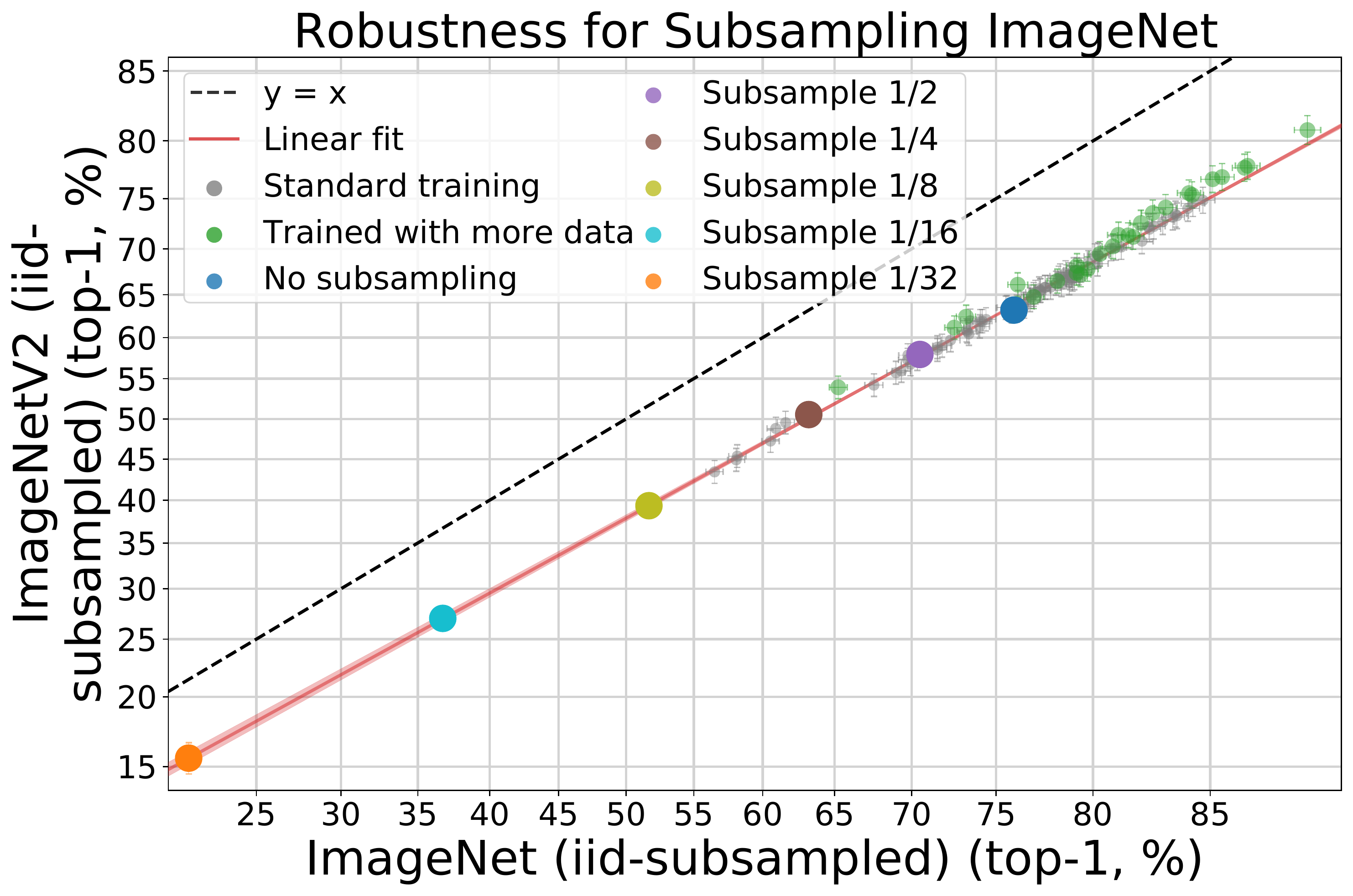}
	  \hspace{0.5cm}
	    \includegraphics[width=0.45\linewidth]{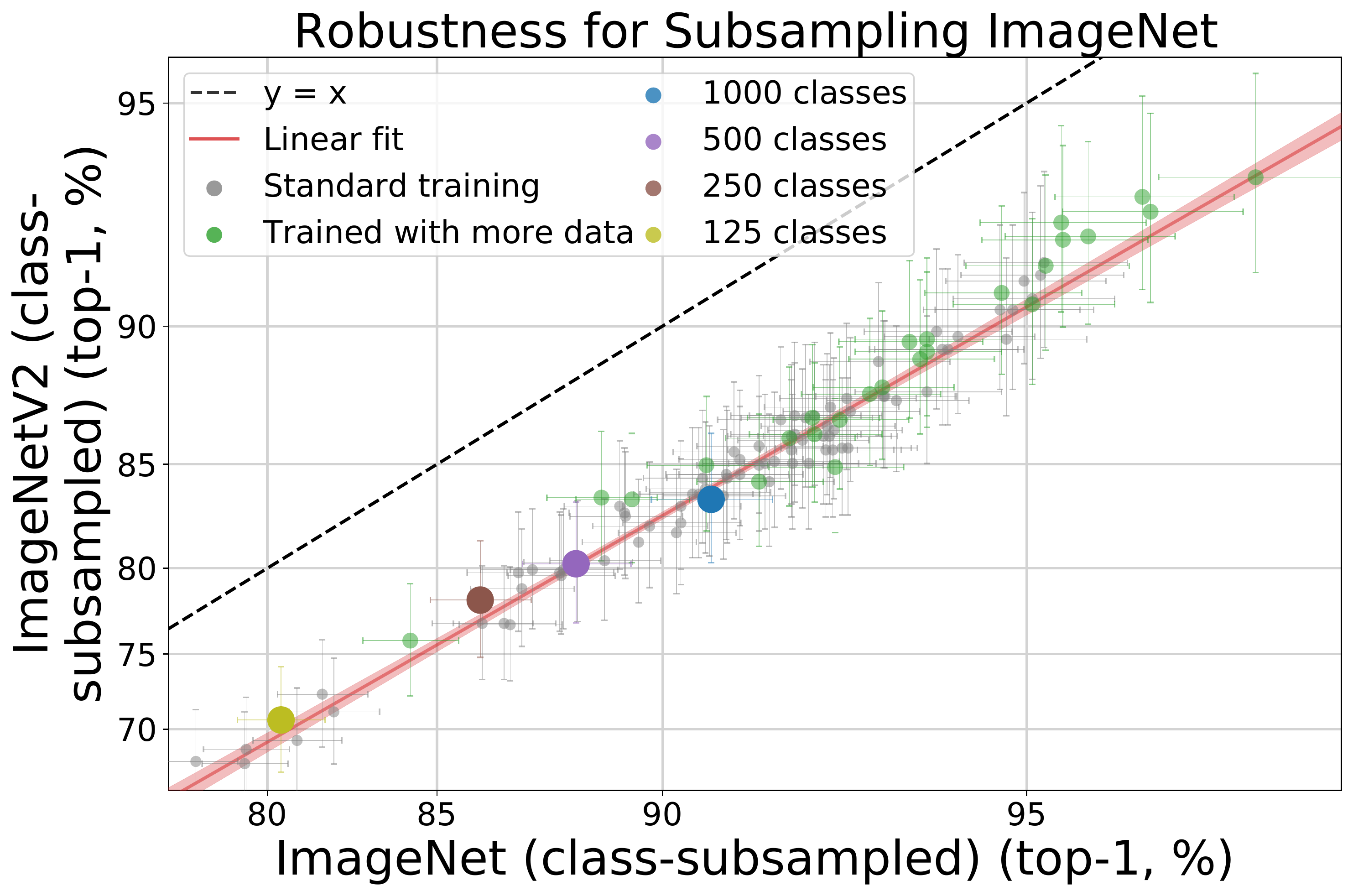}

	  \caption{To investigate the impact of training data on robustness, we vary the training data along two axes: the number of images per class (left), and the number of classes (right).
	  Although models trained on more data provide improvements in effective robustness, we find that subsampling the training set has no impact on effective robustness.
	  Confidence intervals, axis scaling, and the linear fit are computed similarly to Figure \ref{fig:dataset-shifts}.
	  }
	  \label{fig:subsample-iid}
	\end{figure*}
}{
	\begin{figure*}[t!]
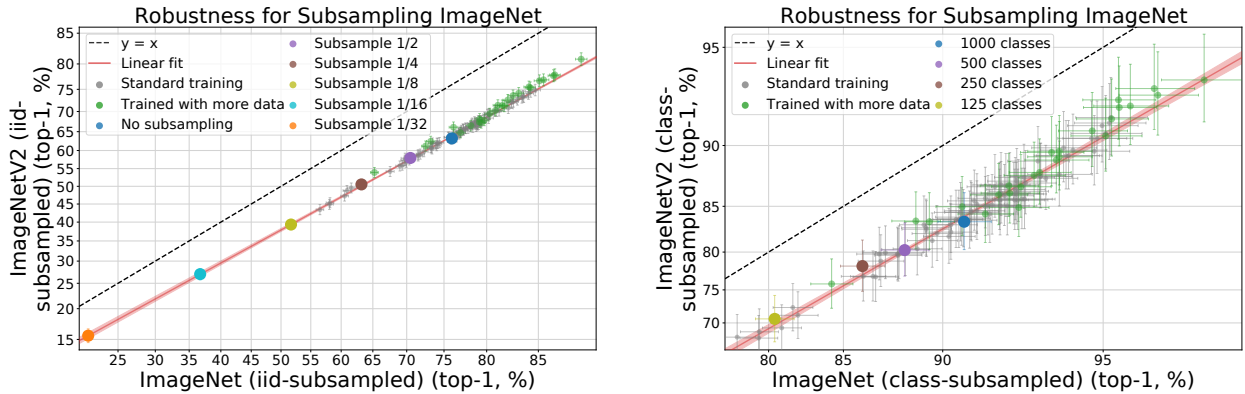

	  \centering
	    \includegraphics[width=0.48\linewidth]{appendix/subsample_iid}
	  \hfill
	    \includegraphics[width=0.48\linewidth]{appendix/subsample_classes}

	  \caption{To investigate the impact of training data on robustness, we vary the training data along two axes: the number of images per class (left), and the number of classes (right).
	  Although models trained on more data provide improvements in effective robustness, we find that subsampling the training set has no impact on effective robustness.
	  Confidence intervals, axis scaling, and the linear fit are computed similarly to Figure \ref{fig:dataset-shifts}.
	  }
	  \label{fig:subsample-iid}
	\end{figure*}
}

Our experiments suggest that neither growing the number of images nor classes in an i.i.d.\ fashion are effective robustness interventions.
Nevertheless, \Cref{fig:dataset-shifts} shows that larger datasets can provide meaningful robustness improvements.
This disparity may be due to limitations of emulating dataset growth by subsampling ILSVRC.
For one, our experiments consider only i.i.d.\ subsets of the training images or classes.
Another possibility is that increases in dataset size may only improve robustness after the dataset is large enough so that the accuracy on the original distribution is nearly saturated.
Our experiments only observe dataset sizes smaller than ILSVRC, which may fall below this inflection point.
Studying the effect of data on robustness is an important direction for future work.

%% file: sec5-related-work.tex
Our work is best seen as a unification of two independent lines of
research---synthetic and natural distribution shift---not
previously studied together.
Synthetic distribution shifts have been studied extensively in the literature
\citep{Geirhos2018,engstrom2019exploring,hendrycks2019benchmarking,kang2019testing,TB19b,maini2019adversarial}.
We incorporate as many prior synthetic measures of robustness as possible. 
Our dataset largely confirms the high-level results from these papers (see Appendix \ref{app:corr-robustness} for additional discussion).
For example, \citet{ford2019} provide evidence for the relationship between adversarial robustness
and robustness to Gaussian noise.
The study of natural distribution shifts has been an equally extensive
research direction \citep{recht2019imagenet,shankar2019systematic,torralba2011unbiased,objectnet}.
When examining each natural distribution shift individually, we
confirm the findings of earlier work that there is a consistent
drop with a linear trend going from ImageNet to each of the other test sets \cite{shankar2019systematic,recht2019imagenet,objectnet}.

We study the relationship between these two previously independent lines of work.
By creating a testbed $100\times$ larger
than prior work \cite{kang2019transfer,recht2019imagenet,robustness,Geirhos2018}, we are able to make several new
observations.
For instance, we show that robustness to synthetic distribution shift often behaves differently from robustness to natural distribution shift.
We argue that it is important to control for accuracy when measuring
the efficacy of a robustness intervention.
Viewed in this light, most interventions do not provide effective robustness.
The main exception is training with
more data, which improves robustness across natural distribution shifts.
In some situations, $\ell_p$-adversarial robustness helps with natural distribution shift that asks for consistency across similar looking images.

\Cref{app:related-work} contains additional discussion of specific related work in more detail.
For instance, \Cref{app:video-robustness} revisits consistency shifts and explains why, in contrast to previous work \cite{gu2019using}, we find consistency robustness is only weakly correlated with color corruption robustness.

\paragraph{Concurrent and subsequent work.} An early version of this paper with results on \imagenetnew{} and \imagenetvidrobust{} appeared on OpenReview in late 2019 \cite{taori2019robustness}.
Since then, two closely related papers have been published concurrently with the updated version of this paper.

\citet{djolonga2020robustness} evaluate 40 models on the same natural distribution shifts as our paper.
Our testbed is larger and contains 200 models with more robustness interventions.
Overall both papers reach similar conclusions.
Their focus is more on the connections to transfer learning while we focus more on comparisons between synthetic and natural distribution shifts.
\citet{djolonga2020robustness} also explore the performance of various models with a synthetic image dataset.

\citet{hendrycks2020faces} also study the connections between synthetic robustness and robustness to natural distribution shifts.
This paper introduces a new dataset, \imagenetr{}, that contains various renditions (sculptures, paintings, etc.) of 200 ImageNet classes as a new example of natural distribution shift.
The paper then introduces DeepAugment, a new data augmentation technique based on synthetic image transformations, and find that this robustness intervention is effective on \imagenetr{}.
In \Cref{app:imagenet-r}, we analyze the \imagenetr{} test set and DeepAugment models, as well as the closely related \imagenetsketch{} test set \cite{wang2019learning}, in more detail.

At a high-level, \imagenetr{} and \imagenetsketch{} follow the trends of the other dataset shifts in our testbed, with models trained on extra data providing the most robustness (up to $\rho = 29.1$\% on \imagenetr{}, though the effect is not uniform, similar to the other dataset shifts).
After the models trained on more data, we find that DeepAugment (in combination with AugMix \cite{hendrycks2019augmix}) achieves substantial effective robustness ($\rho = 11.2$\%).
Interestingly, adversarial robustness also leads to effective robustness on \imagenetr{}.
An AdvProp model \cite{xie2019adversarial} achieves the highest absolute accuracy on \imagenetr{} for a model trained without extra data ($57.8$\%) and has effective robustness $\rho = 7.5$\%.
A model with feature denoising and trained with PGD-style robust optimization \cite{xie2019feature,madry2017towards} achieves the highest effective robustness on ImageNet-R ($\rho = 22.7$\%) and also positive relative robustness ($\tau = 5.7$\%).

\paragraph{Domain adaptation / transfer learning.} Our work is focused on generalizing to out-of-distribution data \emph{without} fine-tuning on the target distribution.
A complementary approach uses data from the target domain in order to improve generalization on that particular domain \cite{transfer_survey}.
Depending on the scenario, robustness (without fine-tuning) or domain adaptation may be more appropriate.
For instance, it may be challenging to record data from the distribution shift, which would prevent fine-tuning before deployment.
In some scenarios, we also expect our model to generalize without extra data (e.g., because humans can do so \cite{humanaccuracy}).
Concurrent work by \citet{djolonga2020robustness} studies connections between robustness to distribution shifts and transfer learning.
Investigating our testbed from the perspective of transfer learning is an interesting direction for future research.

\paragraph{Domain generalization.} Out-of-distribution generalization as measured in our robustness testbed is closely related to domain generalization \cite{blanchard2011,muandet2013}.
In domain generalization, the training algorithm has access to samples drawn from multiple different distributions (domains).
At test time, the model is evaluated on samples from a new domain that was not present in training.
The idea is that having explicit knowledge of multiple domains at training time may help generalization to a new domain at test time.

Several papers have proposed algorithms for domain generalization; we refer to \citet{lostgeneralization} for a comprehensive survey.
Our testbed currently does not contain any algorithms explicitly following the domain generalization paradigm (though pre-training on a different distribution and then fine-tuning on ImageNet has similarities to domain generalization).
A recent meta-study of domain generalization found that standard empirical risk minimization performs as well or better than the eight domain generalization algorithms they compared to \cite{lostgeneralization}.
This result of \citet{lostgeneralization} has similarities to our finding that robustness interventions currently rarely improve over the trend given by standard (ERM) models trained without a robustness intervention.
Evaluating domain generalization approaches on the distribution shifts in our testbed may yield new insights into the performance characteristics of these algorithms.

\paragraph{Distributionally robust optimization.}
Distributionally robust optimization (DRO) is another recently proposed technique to increase robustness to distribution shift \cite{duchi2018learning,duchi2019variancebased}.
The DRO objective minimizes the worst case risk over all distributions close to the data distribution (or in the group DRO setting, the worst case risk over all defined groups).
DRO has been used to train adversarially robust models \cite{sinha2017certifying}, vision models with higher worst-group accuracies \cite{sagawa2019distributionally}, models less reliant on spurious correlations \cite{srivastava2020robustness}, and many others \cite{oren2019distributionally,duchi2020distributionally}.
For a more thorough discussion on DRO and related work, we refer the reader to \cite{duchi2018learning}.
We are currently unable to include DRO models as we are not aware of any pre-trained DRO models for ImageNet.
We will add DRO models to our testbed as they become available.

\paragraph{Adversarial filtering.} One of the distribution shifts in our testbed was obtained via adversarial filtering (\imageneta, \cite{hendrycks2019natural}).
Architectures introduced after the model used to filter ImageNet-A made quick progress in closing the accuracy gap (see Section \ref{sec:natshiftresults}).
A similar phenomenon occurred in natural language processing.
\citet{swag} introduced Swag, an adversarially filtered test for grounded commonsense inference, a combination of natural language inference and commonsense reasoning.
At the time of publication, the best model achieved 59\% accuracy, while a human expert achieved 85\%.
Two months later, \citet{bert} introduced the BERT model which achieves 86\% accuracy on Swag.
This provides further evidence that adversarial filtering can create test sets that are only hard for a specific (existing) class of models.

In the context of training sets, adversarial filtering is similar to hard negative mining, which is often used to generate training data for detection models \cite{dpm,hog,sungpoggio,rowley}.
\citet{bras2020adversarial} propose AFLite, an adversarial filtering algorithm for both refining training sets and creating harder test sets.
They evaluate AFLite on natural language inference tasks and ImageNet classification.
An interesting question is whether combining their algorithm with a ResNet-50 and evaluating later models leads to similar phenomena as on \imageneta{} \cite{hendrycks2019natural} and Swag \cite{swag}.

\paragraph{Fairness in machine learning.} \citet{modelcards} proposed model cards to document the performance of machine learning models in a variety of conditions.
Their focus is on human-centered models and distribution shifts arising from demographic groups (race, gender, etc.).
Our focus here is on ImageNet due to the large number of available models and distribution shifts, but the underlying problem is similar: machine learning models are often brittle under distribution shift.
We remark that ImageNet is known to have geo-diversity deficiencies \cite{shankar2017classification}, among other issues \cite{excavatingai,dulhanty2019auditing}.
In the context of OpenImages \cite{Kuznetsova_2020}, researchers have proposed the Inclusive Image dataset \cite{inclusiveimages}.
Adding OpenImages and Inclusive Images to our testbed and comparing these distribution shifts to our existing examples is an interesting direction for future work.

\paragraph{Further domains.} Our work is focused on the domain of image classification.
There is a long line of work considering robustness (either natural or synthetic) on other domains \cite{dunn2017searchqa,sperber2017toward,yang2018hotpotqa,kwiatkowski2019natural,miller2020effect}. In the context of natural language processing, \citet{belinkov2017synthetic} explore language model robustness to synthetic versus natural one-word substitutions and reach similar high-level results, finding there is limited robustness transfer between the two distributions.

%% file: sec6-conclusion.tex
The goal of robust machine learning is to develop methods that function reliably in a wide variety of settings.
So far, this research direction has focused mainly on synthetic perturbations of existing test sets, highlighting important failure cases and initiating progress towards more robust models.
Ultimately, the hope is that the resulting techniques also provide benefits on real data.
Our paper takes a step in this direction and complements the current synthetic robustness tests with comprehensive experiments on distribution shifts arising from real data.

We find that current image classification models still suffer from substantial accuracy drops on natural distribution shifts.
Moreover, current robustness interventions -- while effective against synthetic perturbations -- yield little to no consistent improvements on real data. The only approach providing broad benefits is training on larger datasets, but the gains are small and inconsistent.

Overall, our results show a clear challenge for future research.
Even training on 1,000 times more data is far from closing the accuracy gaps, so robustness on real data will likely require new algorithmic ideas and better understanding of how training data affects robustness.
Our results indicate two immediate steps for work in this area: robustness metrics should control for baseline accuracy, and robust models should additionally be evaluated on natural distribution shifts.
We hope that our comprehensive testbed with nuanced robustness metrics and multiple types of distribution shift will provide a clear indicator of progress on the path towards reliable machine learning on real data.

%% file: broader_impact.tex
Robustness is one of the key problems that prevents deploying machine learning in the real world and harnessing the associated benefits. 
A canonical example is image classification for medical diagnosis. As was found when researchers attempted to deploy a neural network to detect diabetes from retina images, ``an accuracy assessment from a lab goes only so far. It says nothing of how the AI will perform in the chaos of a real-world environment'' \cite{beede_2020}. 
Similarly, researcher also found that current methods for chest X-ray classification are brittle even in the absence of recognized confounders \cite{zech_2018}. 
If models were robust, then this transfer to the real world would be straightforward. 
Unfortunately, achieving robustness on real data is still a substantial challenge for machine learning.

Our work studies how robust current image classification methods are to distribution shifts arising in real data. 
We hope that our paper will have a positive effect on the study of distribution shifts and allow researchers to more accurately evaluate to what extent a proposed technique increases the robustness to particular forms of distribution shift. 
This will allow researchers to better understand how a deployed system will work in practice, without actually having to deploy it first and users potentially suffering negative consequences.

However, there are several potential ways in which our study could cause unintended harm. 
It is possible that our paper might be used as an argument to stop performing research on some synthetic forms of robustness, e.g., adversarial examples or common corruptions. 
This is not our intention. 
These forms of corruption are interesting independent of any correlation to existing natural distribution shift (e.g., adversarial examples are a genuine security problem).

We only capture a small number of natural distribution shifts among all the possible distribution shifts. 
We selected these shifts because they have been used extensively in the literature and are concrete examples of the types of distribution shift we would like models to be robust to. 
It is likely that there are shifts that we do not capture, and so even if the shifts we define were to be completely solved, other shifts would remain a concern.

One significant form of distribution shift we do not evaluate is dataset bias in representing different demographic groups.
For example, the Inclusive Images dataset \cite{shankar2017classification} attempts to correct for the geographical bias introduced in the Open Images dataset \cite{Kuznetsova_2020} by including a more balanced representation of images from Africa, Asia, and South America. 
Neglecting such implicit biases in the data distribution can harm underrepresented demographic groups. 
Ultimately, evaluating on fixed datasets may not be enough, and validating the fairness and safety of deployable machine learning requires careful analysis in the application domain.

Finally, more reliable machine learning can also enable negative uses cases, e.g., widespread surveillance or autonomous weapon systems.
As with many technologies, these risks require careful regulation and awareness of unintended consequences arising from technological advances.

%% file: acknowledgements.tex
We would like to thank 
Logan Engstrom,
Justin Gilmer,
Moritz Hardt,
Daniel Kang,
Jerry Li,
Percy Liang,
Nelson Liu,
John Miller,
Preetum Nakkiran, 
Rebecca Roelofs,
Aman Sinha,
Jacob Steinhardt, and
Dimitris Tsipras
for helpful conversations while working on this paper.

This research was generously supported in part by ONR awards N00014-17-1-2191, N00014-17-1-2401, and N00014-18-1-2833, the DARPA Assured Autonomy (FA8750-18-C-0101) and Lagrange (W911NF-16-1-0552) programs, a Siemens Futuremakers Fellowship, an Amazon AWS AI Research Award.

%% file: appendix.tex
\section{Testbed overview}
\label{app:testbed-overview}

\begin{figure}[h!]
  \centering
    \includegraphics[width=\linewidth]{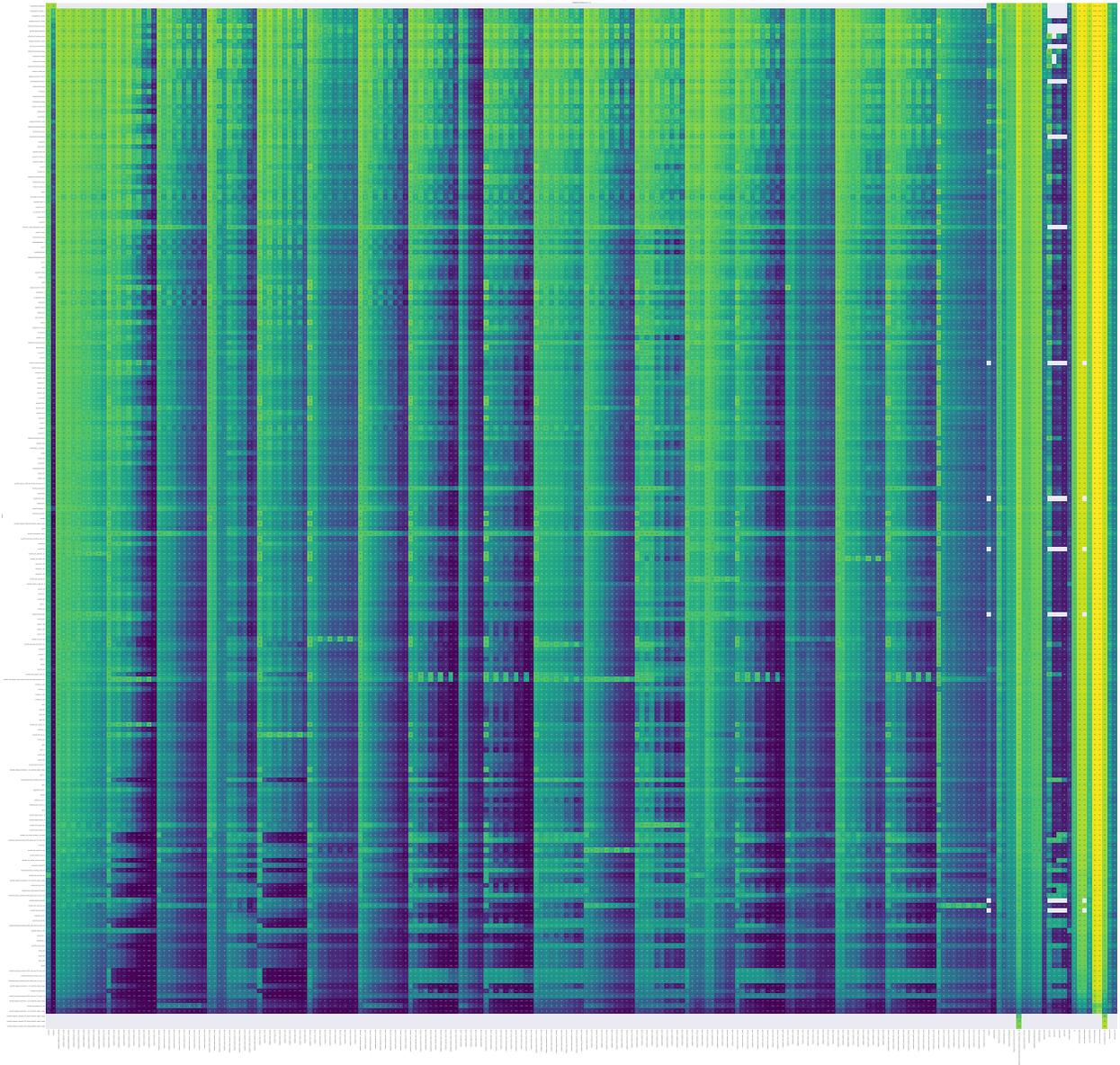}
  \caption{
    An overview of our testbed. 
    Each row is a model, and each column is an evaluation setting. 
    For the corruptions, we display each of the five severities defined in \citep{hendrycks2019benchmarking}. 
    We also plot in-memory and on-disk versions of each corruption as jpeg compression was found to be a confounding factor in \citep{ford2019}.
    A few cells are empty due to resource constraints.
    Testbed code and data is provided at \href{https://modestyachts.github.io/imagenet-testbed/}{\color{blue}https://modestyachts.github.io/imagenet-testbed/}.
  } \label{fig:grid}
\end{figure}

\clearpage

\iftoggle{isneurips} {
  \input{secX-more-data}
  \clearpage
}{}

\section{Relative and effective robustness}
\label{app:rel-eff-robustness}  

\subsection{Relative and effective robustness graphical sketch} 
\label{app:rel-eff-robustness-plot} 

\begin{figure}[ht!] 
  \begin{minipage}[c]{0.45\textwidth} 
    \includegraphics[width=\linewidth]{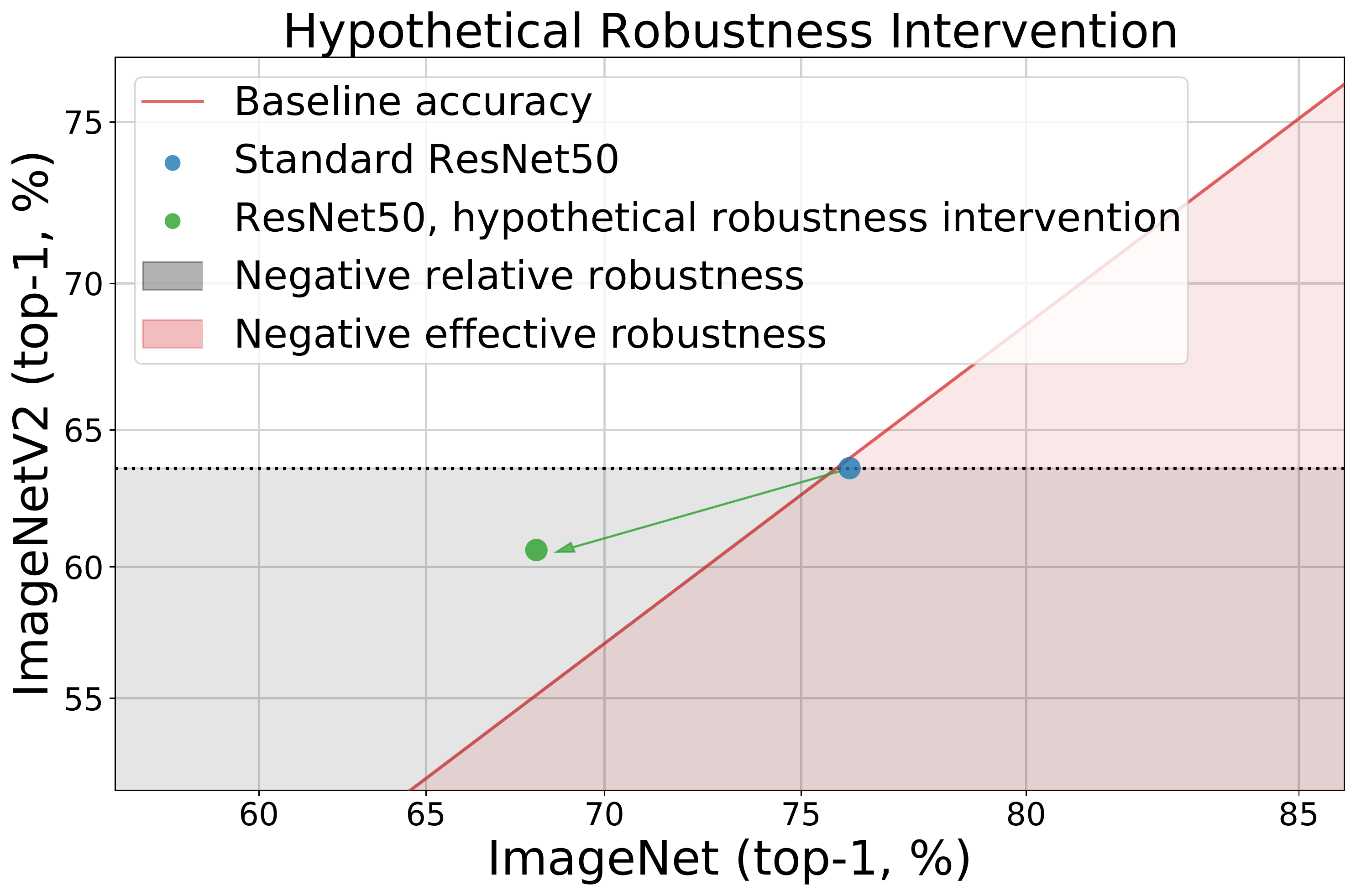} 
  \end{minipage}  
  \hspace{0.5cm}  
  \begin{minipage}[c]{0.45\textwidth} 
    \caption{ 
      While a hypothetical intervention (green), applied to a baseline model (blue), leads to effective robustness (it is above the red line), it reduces the model's accuracy under distribution shift.  
      Hence it fails to provide \textit{relative} robustness. 
      An ideal intervention would place the model in the white quadrant - positive effective and relative robustness. 
    }  \label{fig:rel-robustness} 
  \end{minipage}  
\end{figure}

A central question we address in our paper is whether current methodologies provide meaningful robustness to natural distribution shifts. We discuss how both relative robustness and effective robustness are needed to disentangle the confounding effect of original model accuracy. In Figure \ref{fig:rel-robustness}, we graphically illustrate this notion of relative robustness.   

\subsection{Relative and effective robustness for ResNet50 models}  
\label{app:rel-eff-robustness-resnet50}

We provide additional plots depicting a subset of the models in our testbed. 
In order to make an equal comparison, we only plot ResNeet50 variants, models which slightly modify the training data or architecture of a base ResNet50.
The plots in this section thus describe what the relative and effective robustness properties of various robustness interventions look like on a standard ResNet50. 
The models can be directly compared with each other since the base model before intervention is the same. 

\begin{figure*}[ht!]
  \centering
  \begin{subfigure}[t]{0.45\textwidth}
    \includegraphics[width=\textwidth]{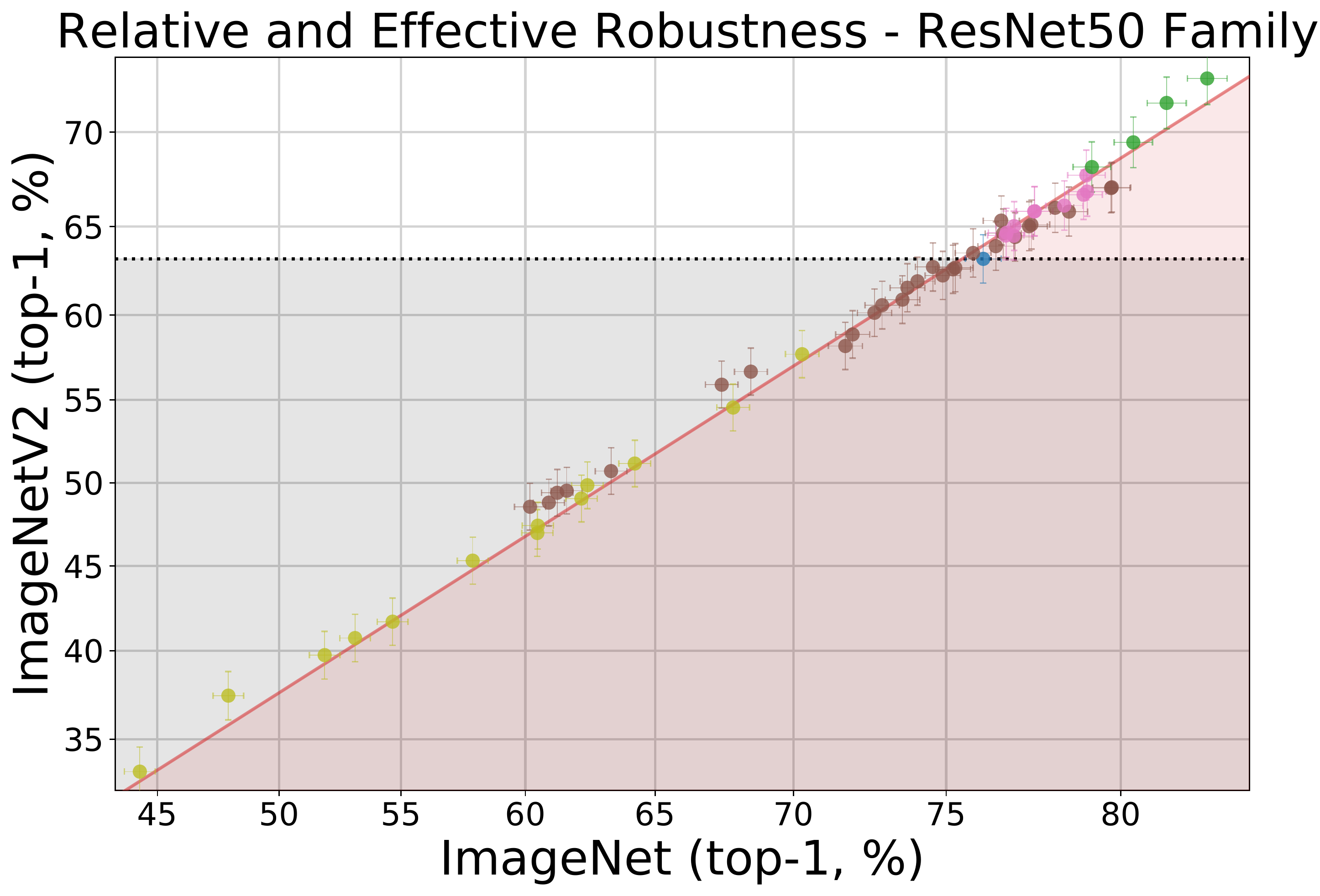}
  \end{subfigure}
  \hspace{0.5cm}
  \begin{subfigure}[t]{0.45\textwidth}
    \includegraphics[width=\textwidth]{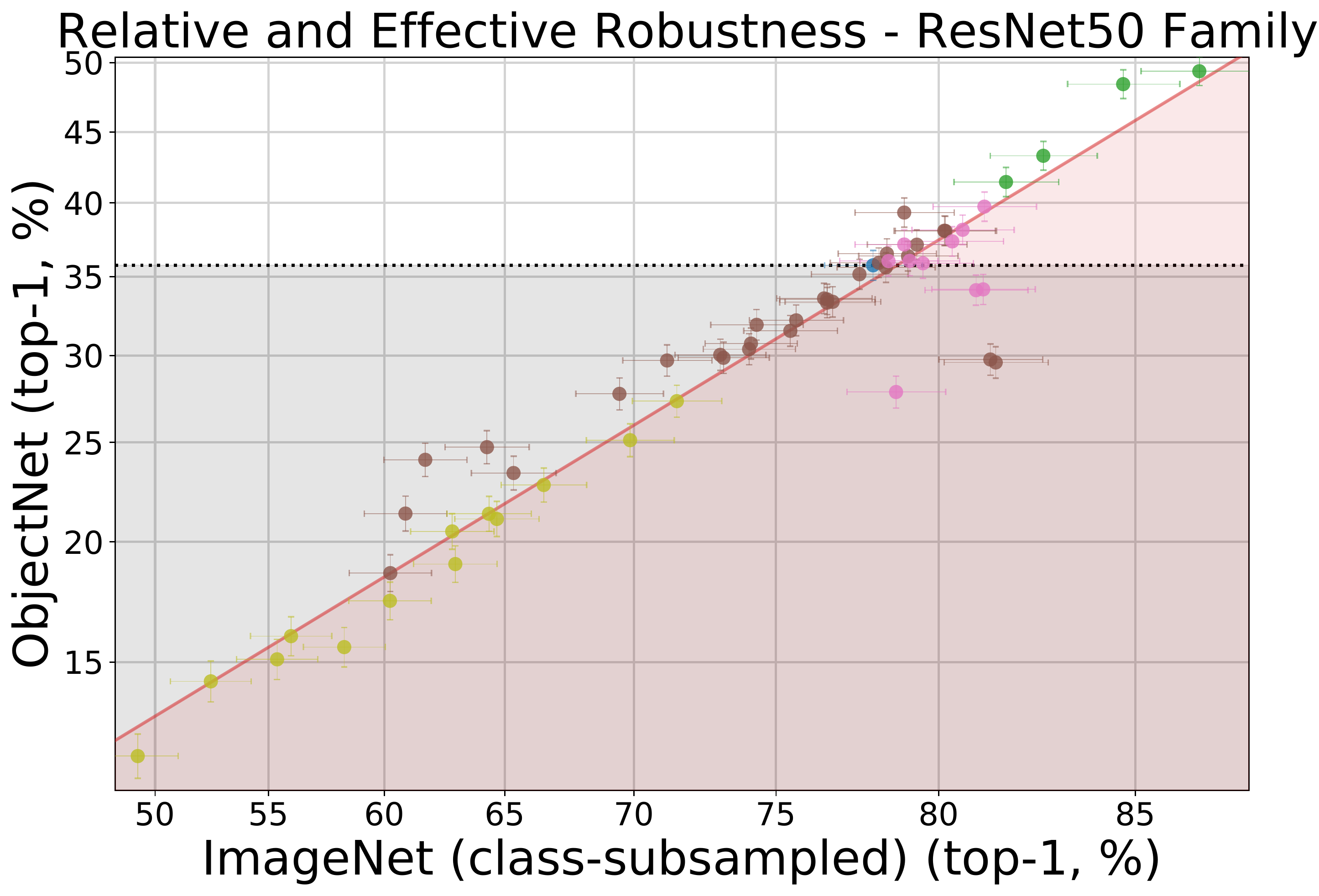}
  \end{subfigure}

  \par\medskip
  \begin{subfigure}[t]{0.45\textwidth}
    \includegraphics[width=\textwidth]{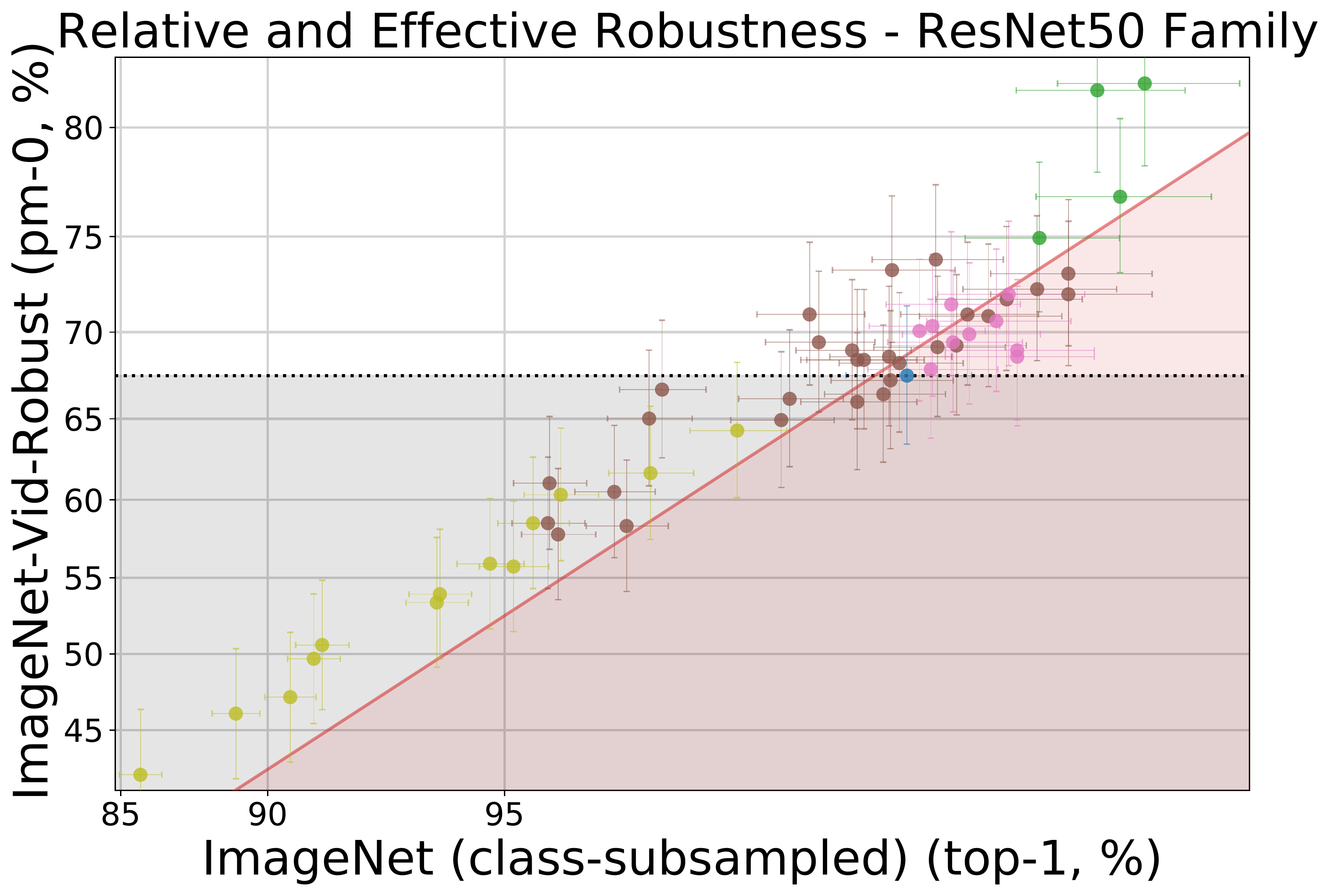}
  \end{subfigure}
  \hspace{0.5cm}
  \begin{subfigure}[t]{0.45\textwidth}
    \includegraphics[width=\textwidth]{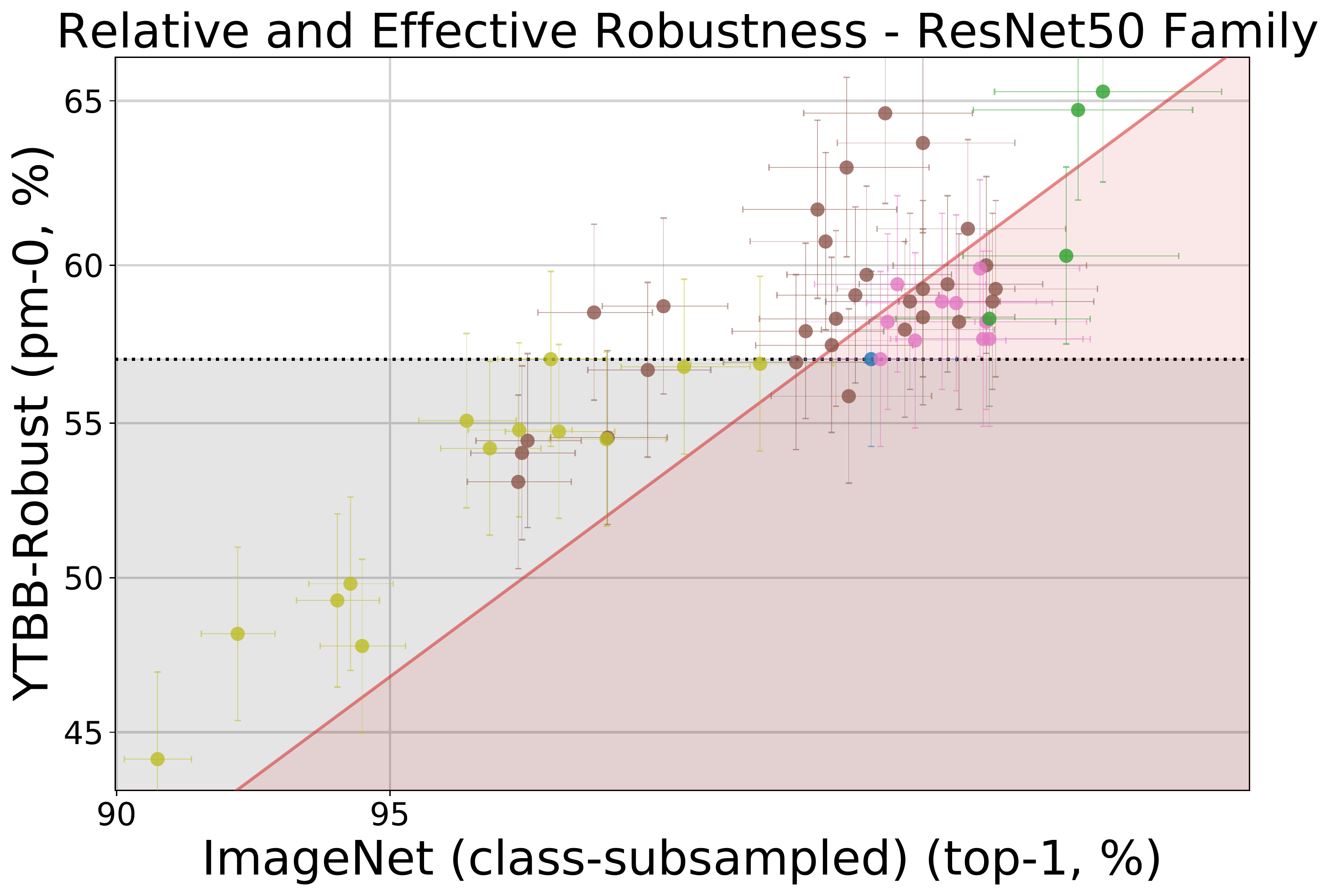}
  \end{subfigure}

  \begin{subfigure}[t]{\textwidth}
    \centering \includegraphics[width=1.0\textwidth]{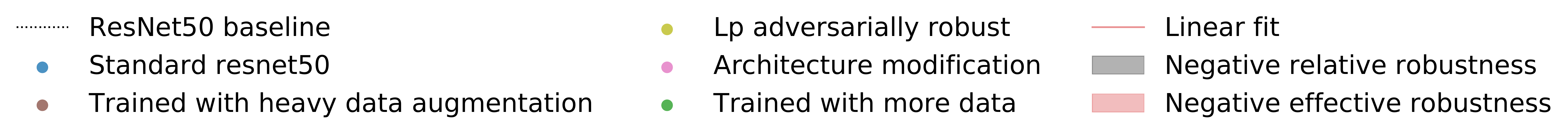}
  \end{subfigure}

  \caption{
    Relative and effective robustness for models that are variants of a ResNet50. 
    Model accuracies are displayed on the four natural dataset shifts: \imagenetnew{} (top left), \objectnet{} (top right), \imagenetvidrobust{}-anchor (bottom left), and \ytbbrobust{}-anchor (bottom right).
    These plots demonstrate that the only models that have consistently positive relative and positive effective robustness are models that are trained on more data.
    However, the effect is small, and not all models trained on more data are more robust.
    Confidence intervals, axis scaling, and the linear fit are computed similarly to Figure \ref{fig:dataset-shifts}.
  }
  \label{fig:rel_eff_robutness_dataset_shift}
\end{figure*}

For natural dataset shifts, the plots in Figure \ref{fig:rel_eff_robutness_dataset_shift} demonstrate that the only models that have consistently positive relative and positive effective robustness are models that are trained on more data. 
However, the effect is small, and not all models trained on more data are more robust. 
On \ytbbrobust{} specifically, a few data augmentation strategies from ImageNet-C provide significant both effective and relative robustness:
training on greyscale ($\rho = 6.9$\%, $\tau = 1.8$\%); 
training on pixelate ($\rho = 5.4$\%, $\tau = 2.0$\%); 
training on jpeg compression ($\rho = 5.4$\%, $\tau = 6.3$\%); 
training on gaussian noise, contrast, motion blur, and jpeg compression ($\rho = 4.8$\%, $\tau = 5.0$\%); 
and training on gaussian noise ($\rho = 3.6$\%, $\tau = 4.0$\%). 
However, this performance is not consistent across the natural distribution shifts.
Exploring why these data augmentation strategies are helpful on \ytbbrobust{} is an interesting direction for future work.
Additionally, while some $\ell _p$-adversarially robust models display significant effective robustness on \ytbbrobust{} -
$\ell _2$ robust ResNet50 ($\rho = 6.4$\%),
$\ell _{\inf}$ robust ResNet50 ($\rho = 6.4$\%),
and ResNet50 smoothed with 0.25 gaussian noise and adversarially 1-step PGD trained ($\rho = 5.0$)
- in most cases, they fail to provide positive relative robustness.

\begin{figure*}[ht!]
  \centering
  \begin{subfigure}[t]{0.45\textwidth}
    \includegraphics[width=\textwidth]{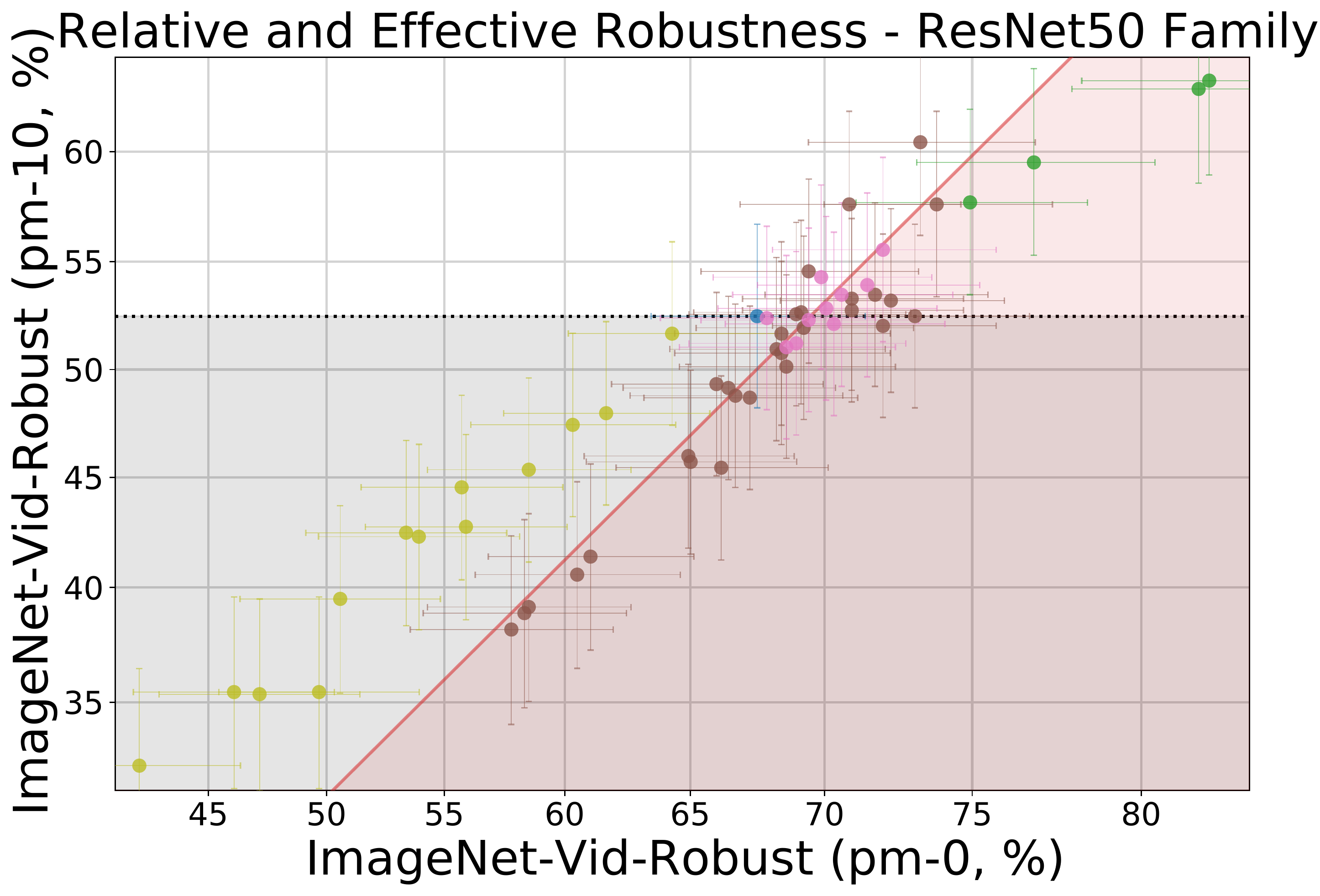}
  \end{subfigure}
  \hspace{0.5cm}
  \begin{subfigure}[t]{0.45\textwidth}
    \includegraphics[width=\textwidth]{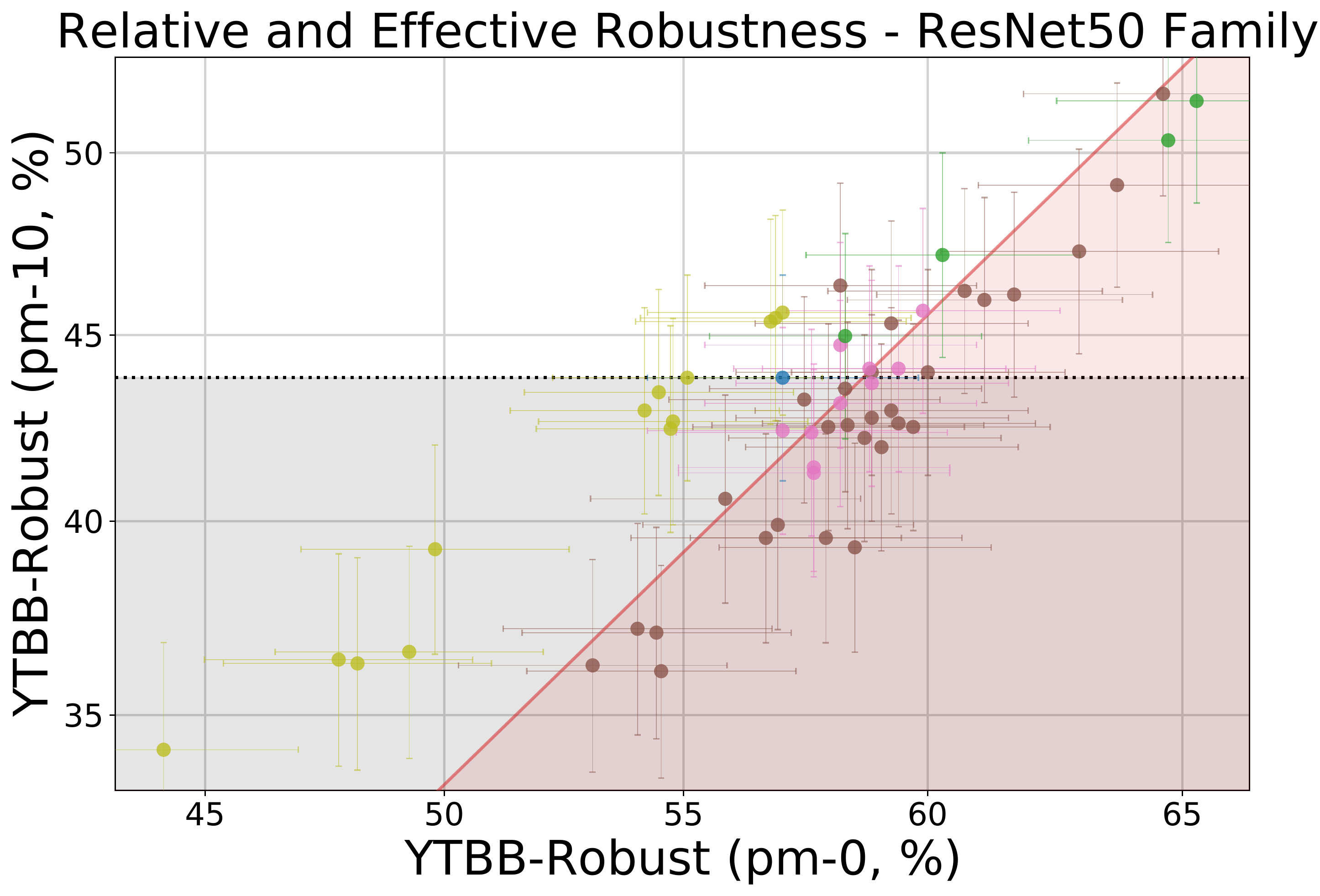}
  \end{subfigure}

  \begin{subfigure}[t]{\textwidth}
    \centering \includegraphics[width=1.0\textwidth]{appendix/resnet50_legend}
  \end{subfigure}

  \caption{
    Relative and effective robustness for models that are variants of a ResNet50. 
    Model accuracies are displayed the two consistency shifts: \imagenetvidrobust{} (left), and \ytbbrobust{} (right).
    These plots demonstrate that while adversarially robust models provide effective robustness, they do not necessarily provide relative robustness.
    Confidence intervals, axis scaling, and the linear fit are computed similarly to Figure \ref{fig:dataset-shifts}.
  }
  \label{fig:rel_eff_robutness_consistency_shift}
\end{figure*}

For natural consistency shifts, the plots in Figure \ref{fig:rel_eff_robutness_consistency_shift} demonstrate that while adversarially robust models provide effective robustness (average $\rho = 4.3$\% on \imagenetvidrobust{} and average $\rho = 3.9$\% on \ytbbrobust{}), they only sometimes provide relative robustness on \ytbbrobust{}.

\begin{figure*}[ht!]
  \centering
  \begin{subfigure}[t]{0.45\textwidth}
    \includegraphics[width=\textwidth]{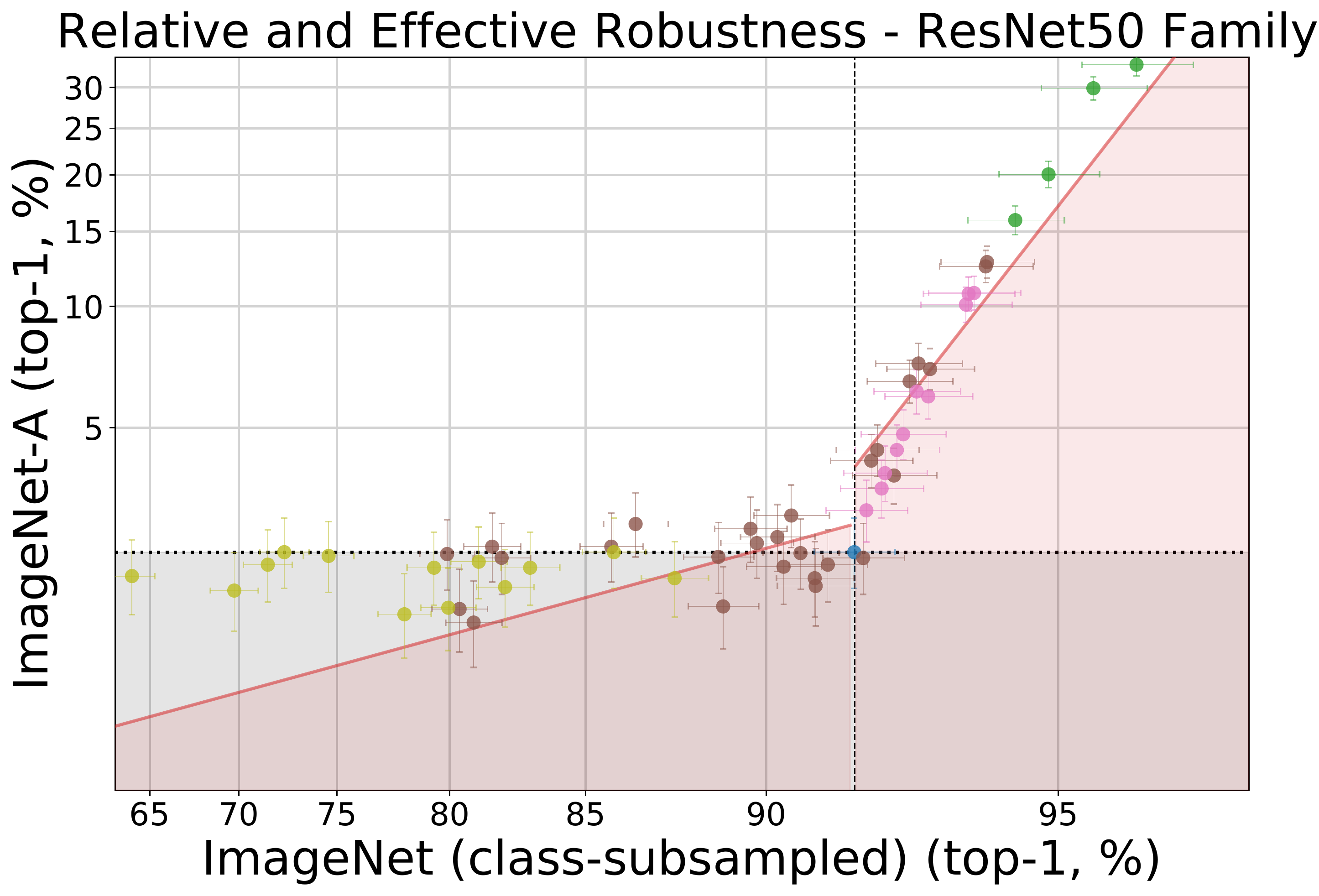}
  \end{subfigure}

  \begin{subfigure}[t]{\textwidth}
    \centering \includegraphics[width=1.0\textwidth]{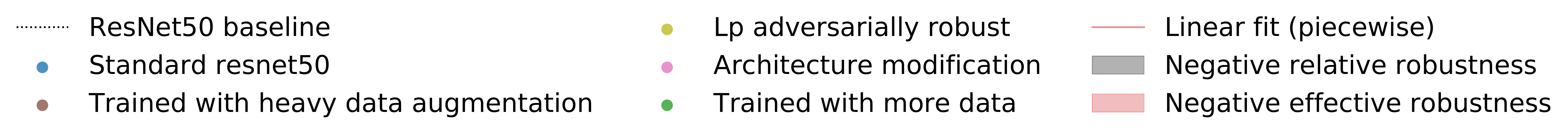}
  \end{subfigure}

  \caption{
    Relative and effective robustness for models that are variants of a ResNet50. 
    Model accuracies are displayed on \imageneta{}, a dataset adversarially filtered to contain only images incorrectly classified by a ResNet50 trained on ImageNet.
    Due to the "knee"-like response curve, an artifact of the adversarial filtering, effective robustness is defined piecewise around the ResNet50 model accuracy point. 
    The plot demonstrates that robustness interventions have little impact on \imageneta{} accuracy.
    However, the effect is small, and not all models trained on more data are more robust.
    Confidence intervals, axis scaling, and the linear fit are computed similarly to Figure \ref{fig:dataset-shifts}.
  }
  \label{fig:rel_eff_robutness_adversarially_filtered}
\end{figure*}

For the adversarially filtered shift, the plot in Figure \ref{fig:rel_eff_robutness_adversarially_filtered} demonstrates that robustness interventions have little impact on \imageneta{} accuracy. Most of the "knee"-like response curve can be explained as an artifact of the adversarial filtering, with the knee occuring at the ResNet50 model accuracy.

\clearpage

\iftoggle{isneurips}{

\section{Synthetic vs. natural robustness}
\label{app:syn-nat-robustness}

\subsection{Adversarial attacks vs. \imagenetnew{}}
\label{app:pgd-robustness}

In Figure \ref{fig:synthetic-shifts-pgd}, we analyze the predictiveness of accuracy under $\ell_p$ adversarial attacks and compare this metric with effective robustness on \imagenetnew. 
This plot is similar to Figure \ref{fig:synthetic-shifts}, but analyzes $\ell_p$ attacks instead of image corruptions. 
The plots show that robustness under $\ell_p$ attacks does not imply that the
corresponding model has effective robustness on \imagenetnew{} (the Pearson correlation coefficient is $r = -0.05$).

\begin{figure*}[ht!]
    \centering
  \begin{subfigure}[t]{0.45\textwidth}
    \includegraphics[width=\linewidth]{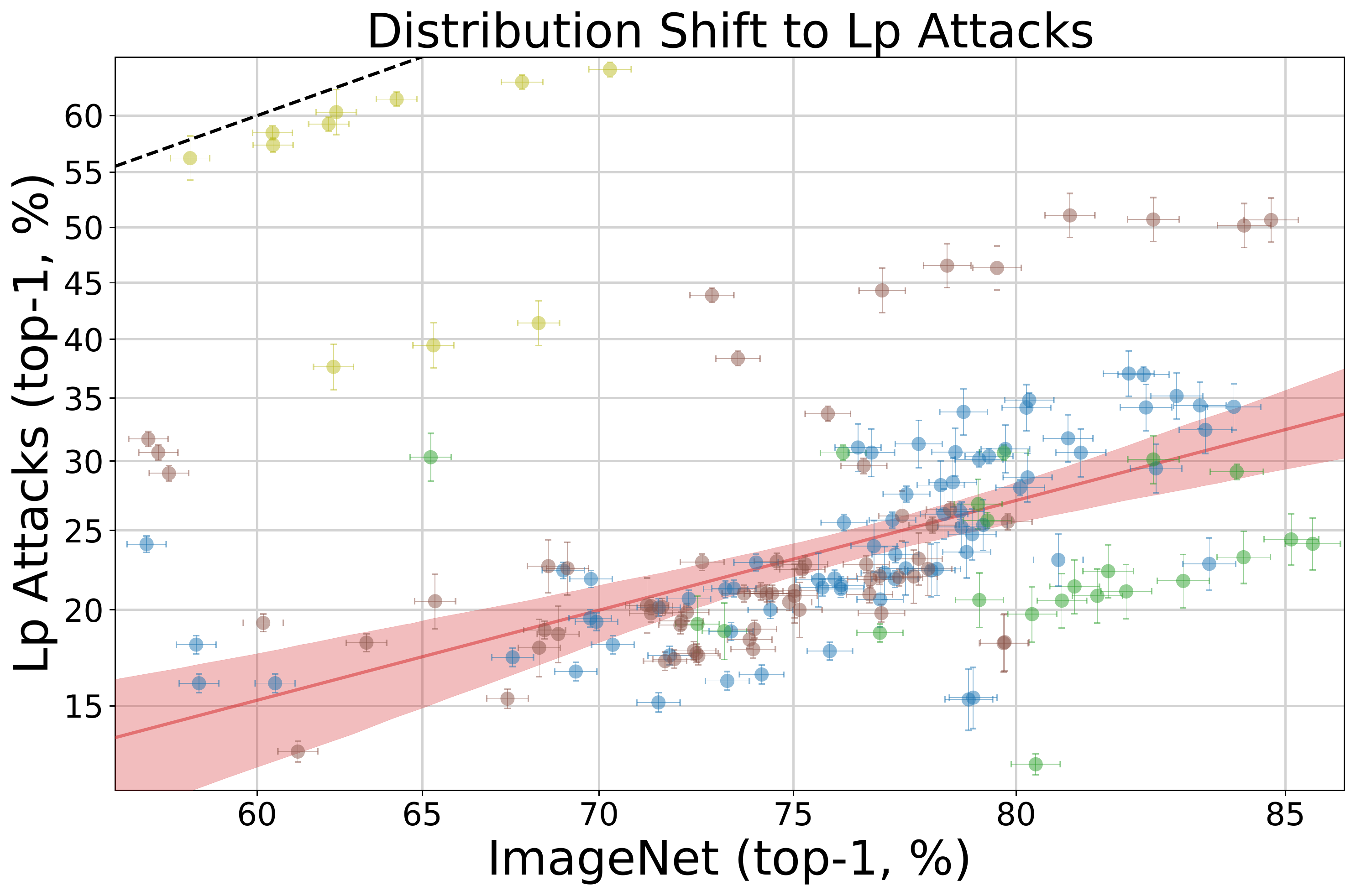}
  \end{subfigure}
    \hspace{0.5cm}
  \begin{subfigure}[t]{0.45\textwidth}
    \includegraphics[width=\linewidth]{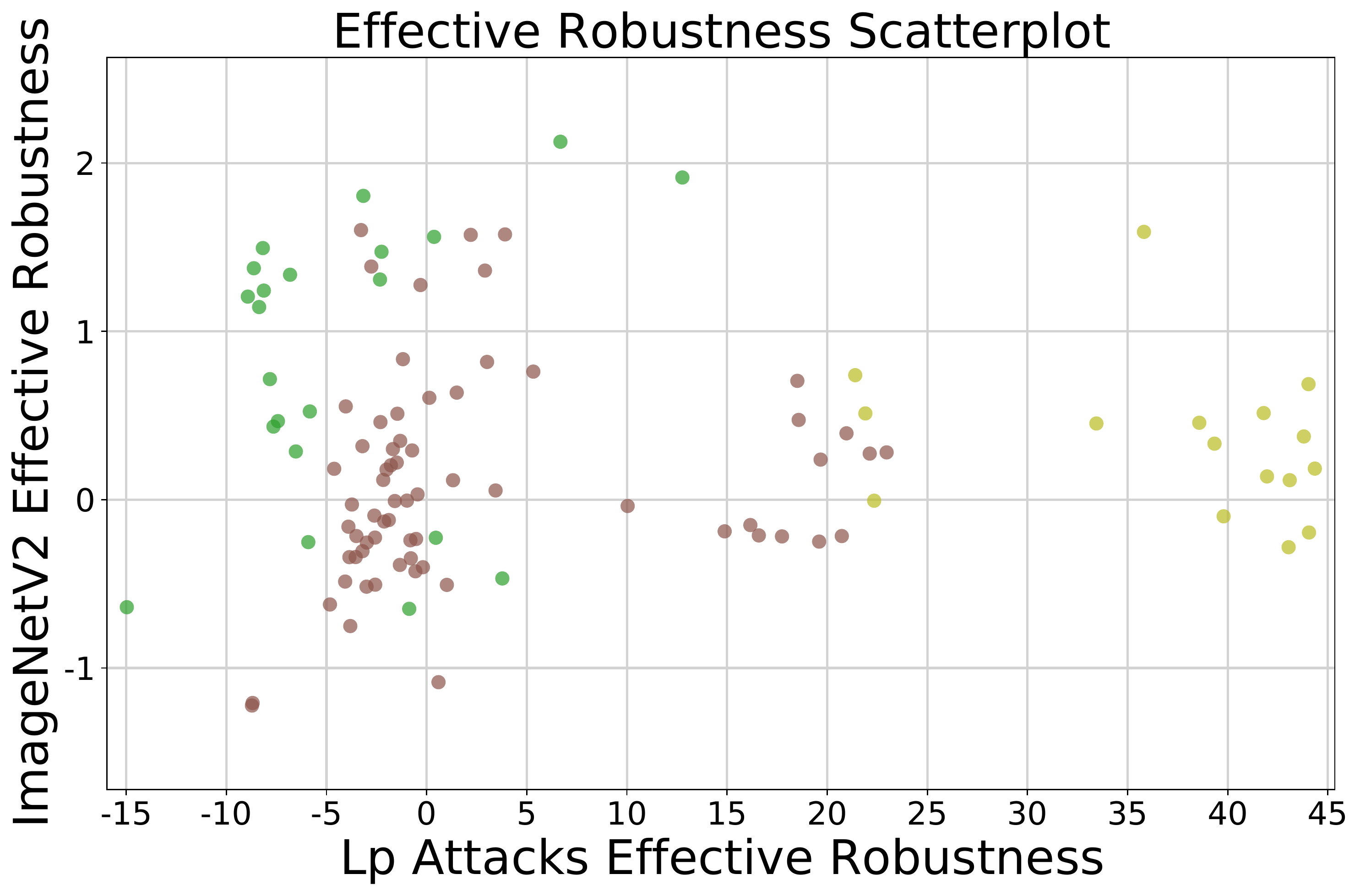}
  \end{subfigure}
  \begin{subfigure}[t]{\textwidth}
    \centering \includegraphics[width=0.7\textwidth]{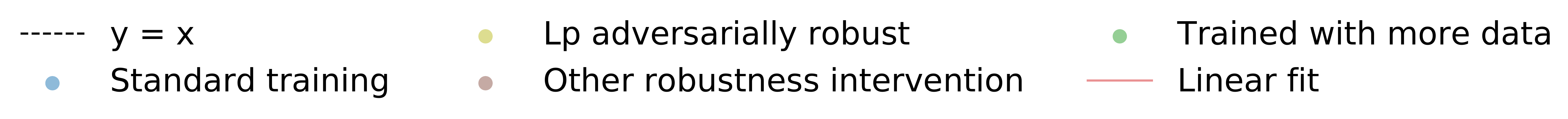}
  \end{subfigure}
  \caption{
    Model accuracies under $\ell_p$ adversarial PGD attacks.
    Similar to Figure \ref{fig:dataset-shifts}, the left plot shows the effective robustness for this synthetic distribution shift.
    Multiple non-standard models achieve substantial effective robustness, corroborating recent research progress on creating adversarially robust models.
    The right plot shows the correlation between the effective robustness for $\ell_p$ attacks and the \imagenetnew{} distribution shift (top left in Figure \ref{fig:dataset-shifts}) for the non-standard models.
    $\ell_p$ attacks are only weakly predictive of effective robustness on \imagenetnew: there are several models that achieve high effective robustness under $\ell_p$ but little to no effective
    robustness on \imagenetnew.
    }
  \label{fig:synthetic-shifts-pgd}
\end{figure*}

\subsection{Effective robustness scatterplots}
\label{app:eff-robust-syn-nat-robustness}

}{
  
\section{Effective robustness scatterplots}
\label{app:eff-robust-syn-nat-robustness}

}

In this section, we further explore to what extent robustness to synthetic distribution shifts predicts robustness on natural distribution shift. 
We extend the analysis in 
\iftoggle{isneurips}{Figures \ref{fig:synthetic-shifts} and \ref{fig:synthetic-shifts-pgd}}{Figure \ref{fig:synthetic-shifts}} 
by computing effective robustness on all natural distribution shifts and comparingn them against effective robustness on synthetic distribution shifts.

\begin{figure*}[ht!]
  \centering
  \begin{subfigure}[t]{\iftoggle{isneurips}{0.45}{0.40}\textwidth}
    \includegraphics[width=\textwidth]{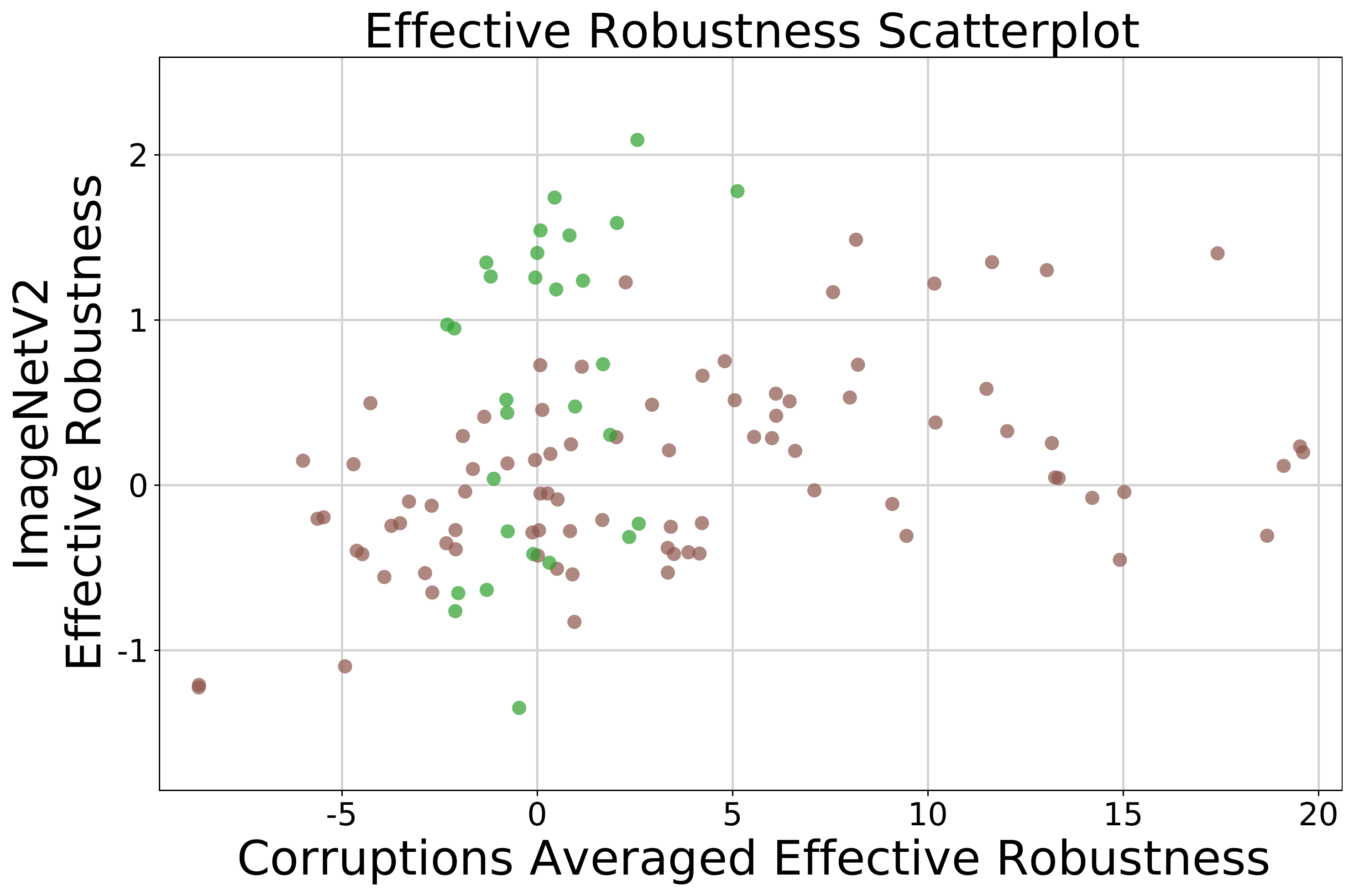}
  \end{subfigure}
  \hspace{0.5cm}
  \begin{subfigure}[t]{\iftoggle{isneurips}{0.45}{0.40}\textwidth}
    \includegraphics[width=\textwidth]{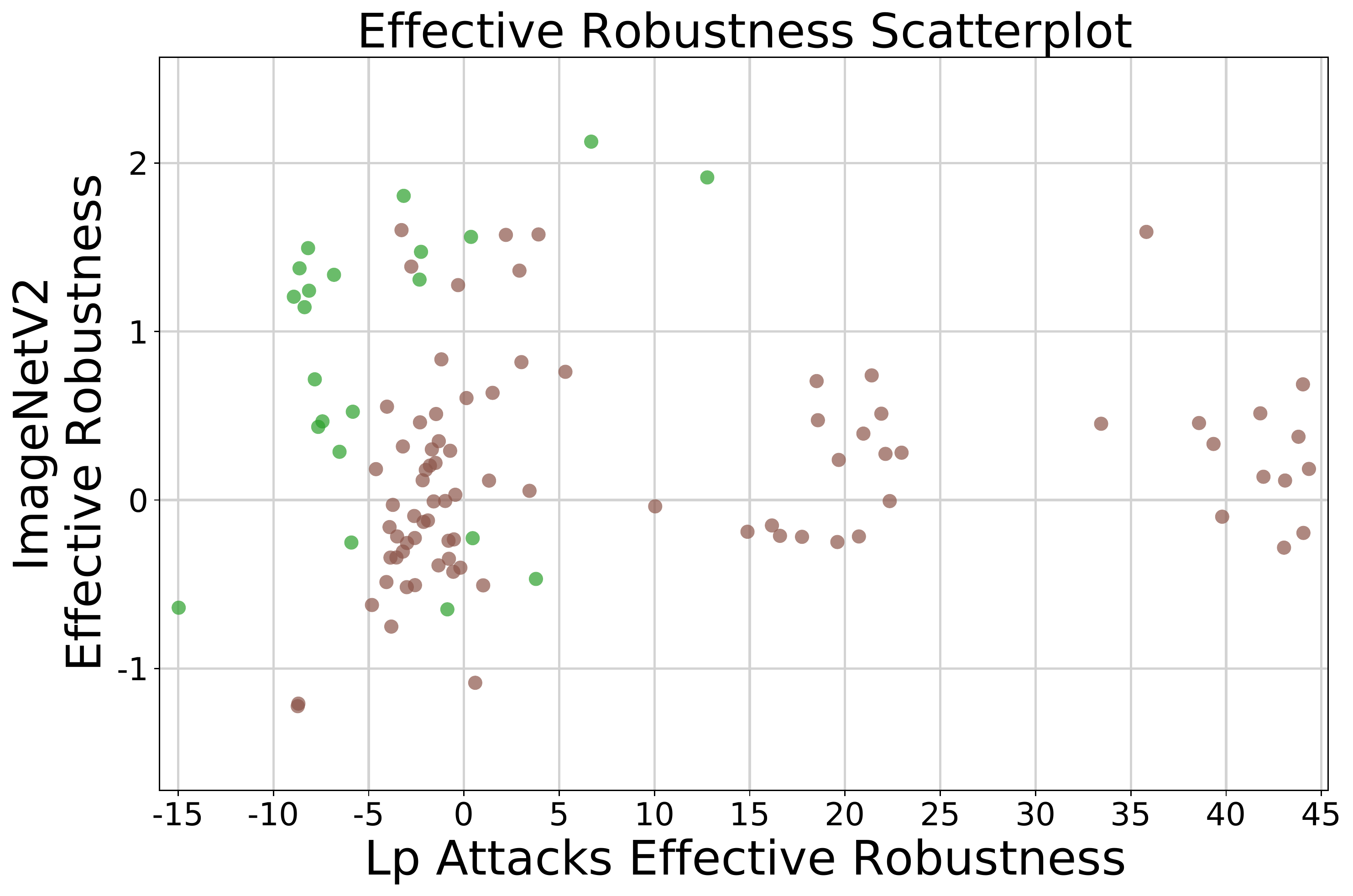}
  \end{subfigure}

  \par\medskip
  \begin{subfigure}[t]{\iftoggle{isneurips}{0.45}{0.40}\textwidth}
    \includegraphics[width=\textwidth]{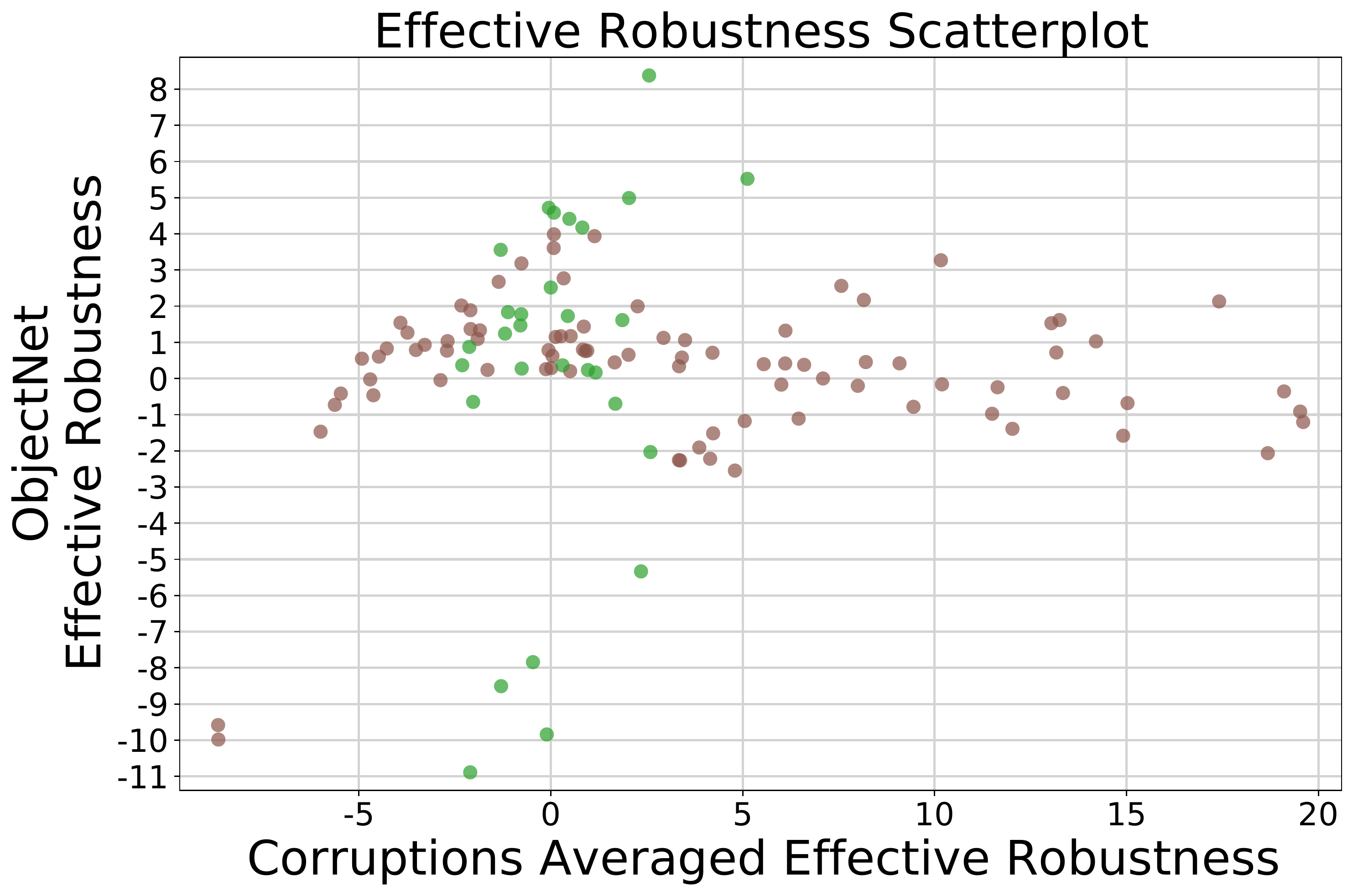}
  \end{subfigure}
  \hspace{0.5cm}
  \begin{subfigure}[t]{\iftoggle{isneurips}{0.45}{0.40}\textwidth}
    \includegraphics[width=\textwidth]{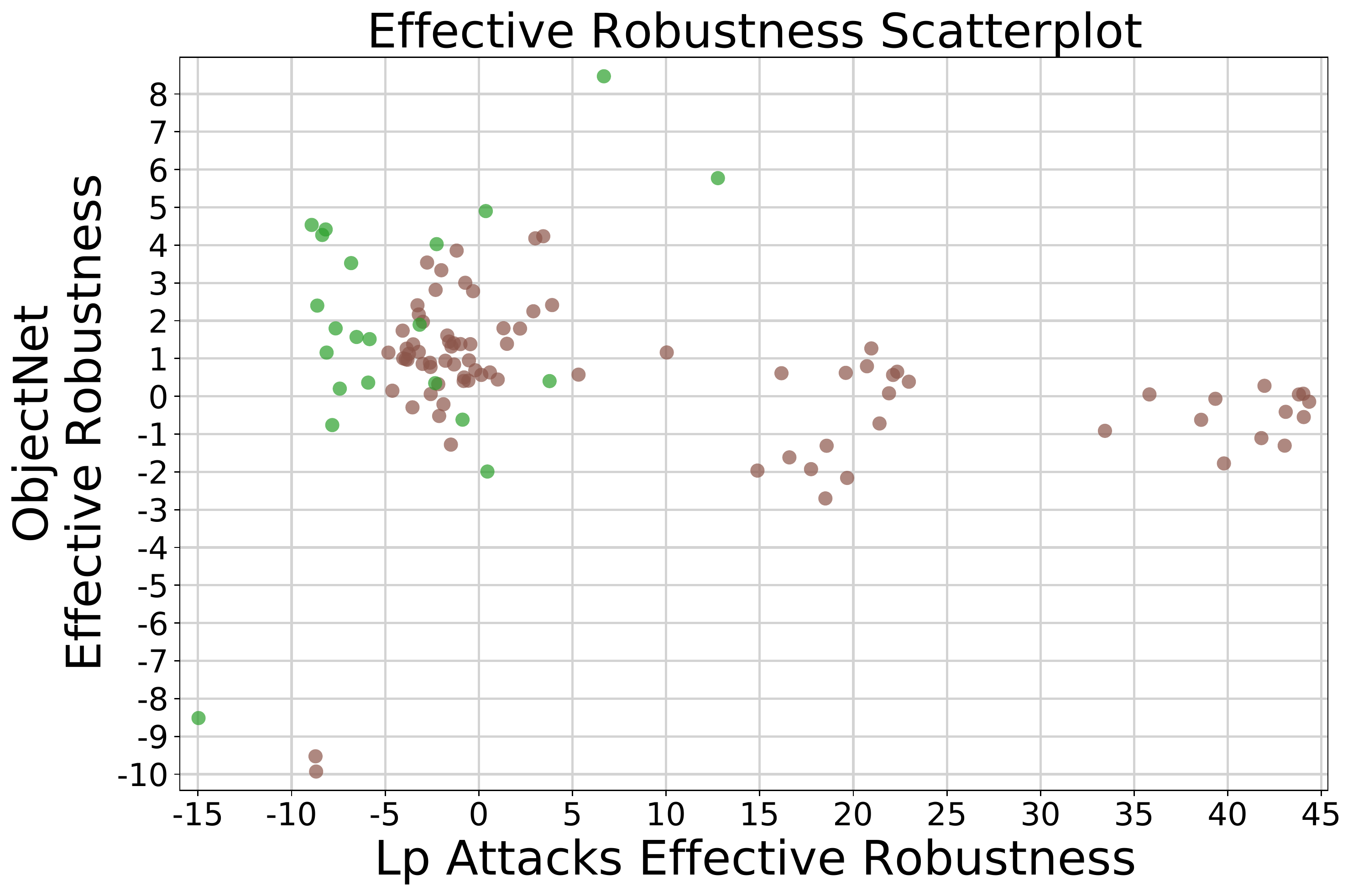}
  \end{subfigure}

  \par\medskip
  \begin{subfigure}[t]{\iftoggle{isneurips}{0.45}{0.40}\textwidth}
    \includegraphics[width=\textwidth]{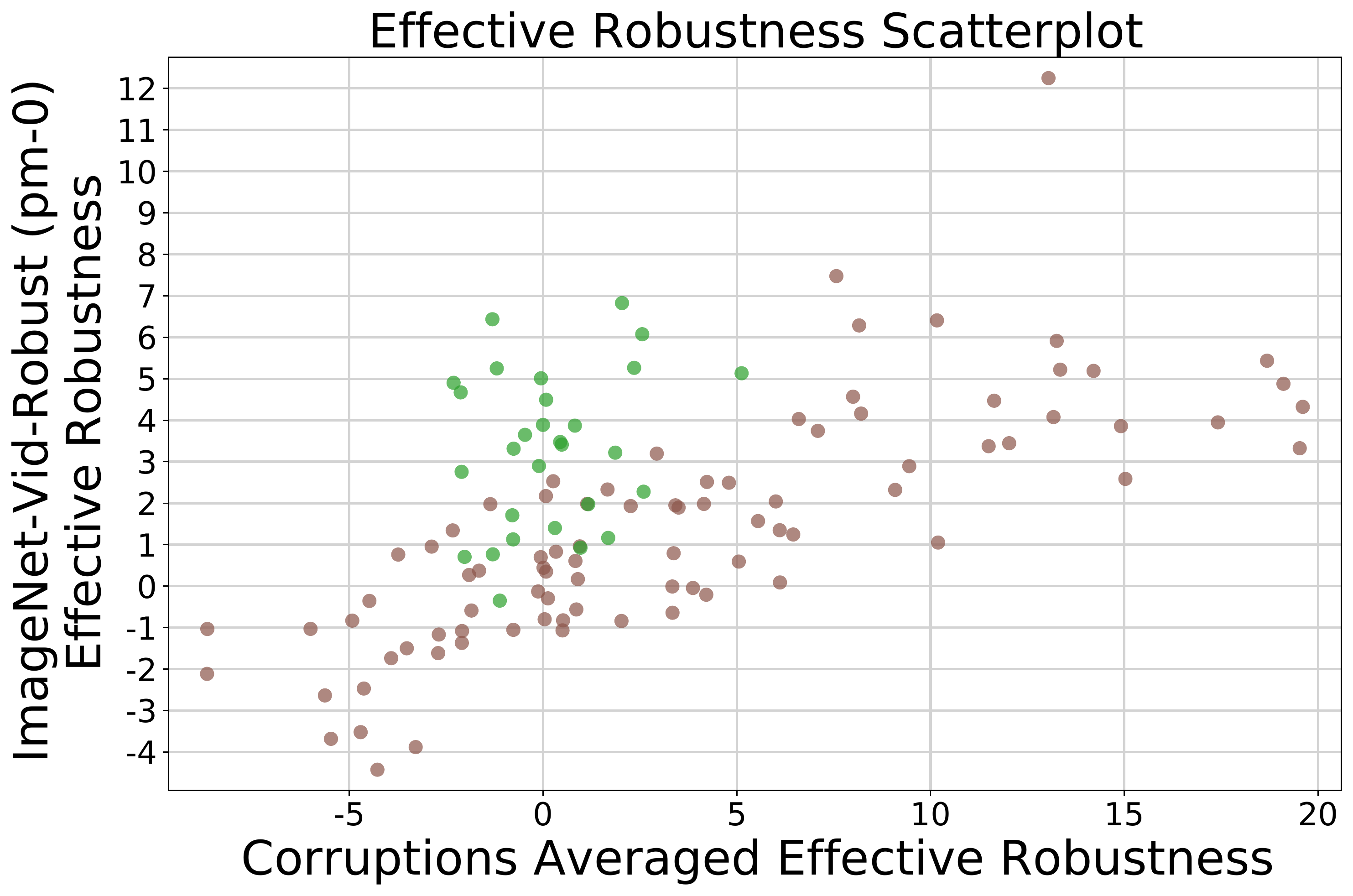}
  \end{subfigure}
  \hspace{0.5cm}
  \begin{subfigure}[t]{\iftoggle{isneurips}{0.45}{0.40}\textwidth}
    \includegraphics[width=\textwidth]{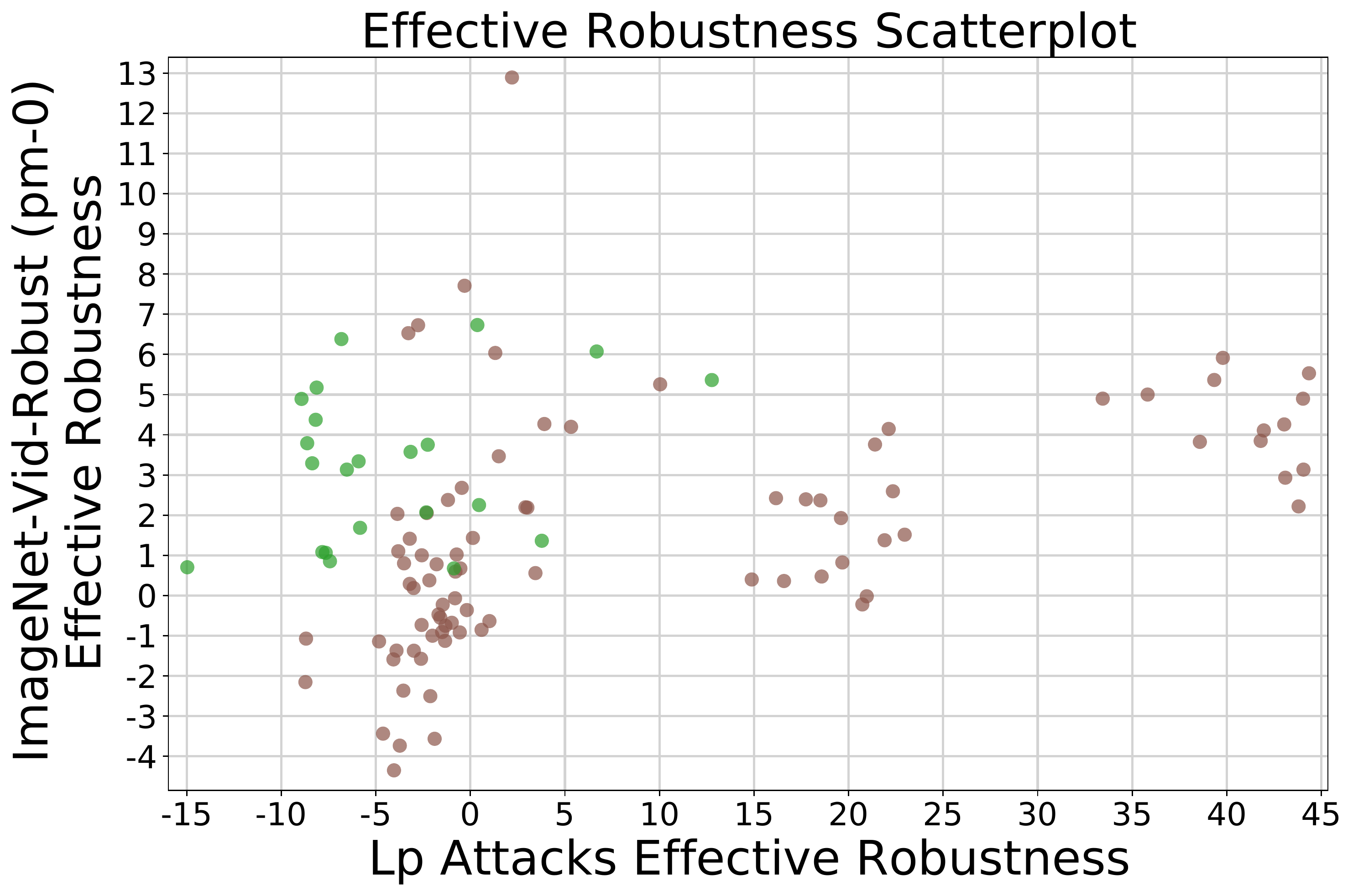}
  \end{subfigure}

  \par\medskip
  \begin{subfigure}[t]{\iftoggle{isneurips}{0.45}{0.40}\textwidth}
    \includegraphics[width=\textwidth]{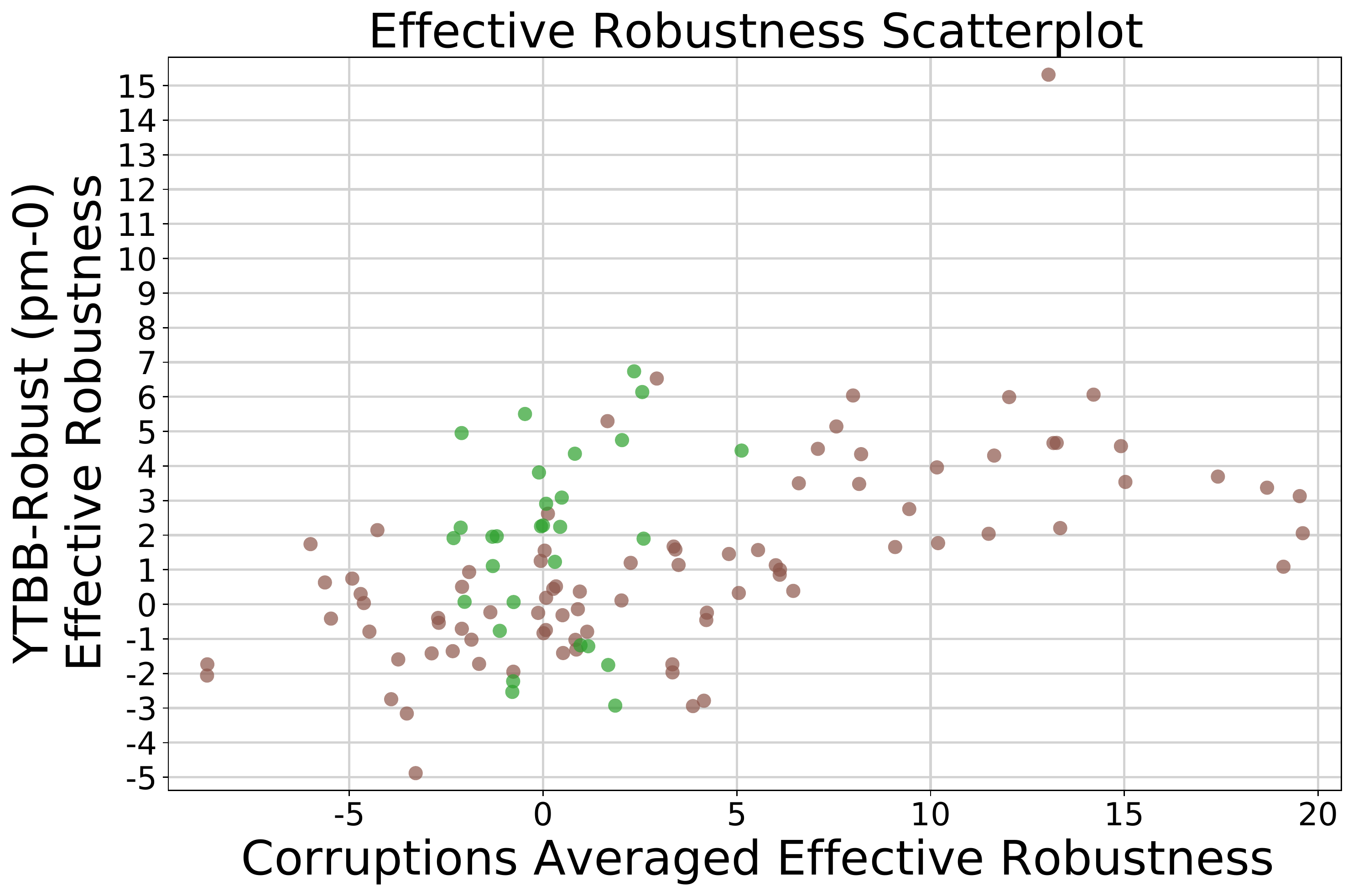}
  \end{subfigure}
  \hspace{0.5cm}
  \begin{subfigure}[t]{\iftoggle{isneurips}{0.45}{0.40}\textwidth}
    \includegraphics[width=\textwidth]{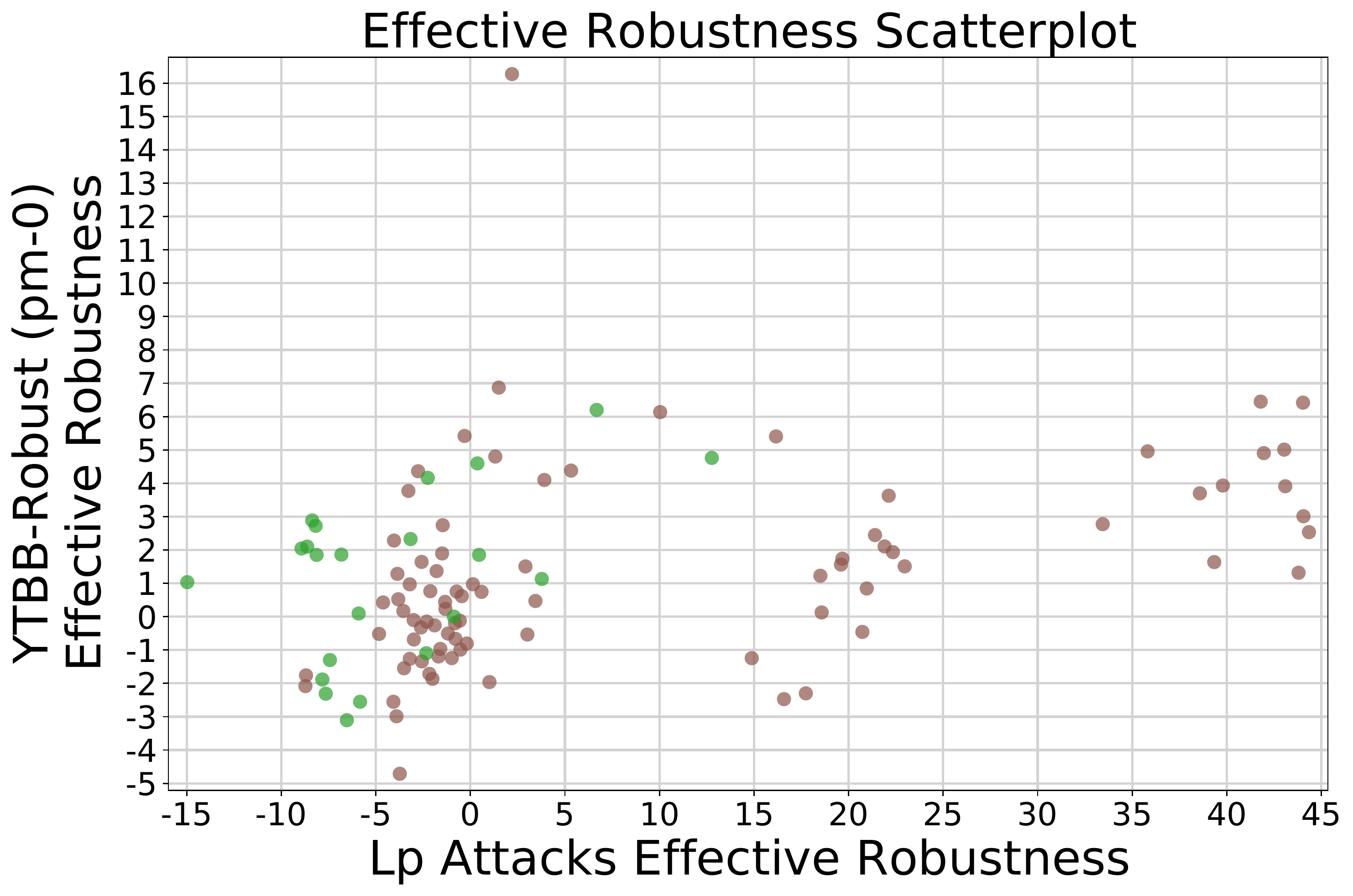}
  \end{subfigure}

  \begin{subfigure}[t]{\textwidth}
    \centering \includegraphics[width=0.55\textwidth]{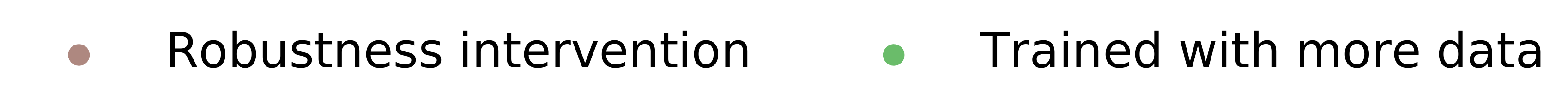}
  \end{subfigure}

  \caption{
    We compare the effective robustness of models with their accuracy drop due to corruptions (left column) and adversarial attacks (right column).
    The effective robustness is computed with respect to linear fits on the four natural dataset shifts: \imagenetnew{} (first row), \objectnet{} (second row), \imagenetvidrobust{}-anchor (third row), and \ytbbrobust{}-anchor (fourth row).
    The measures are largely uncorrelated, indicating that improved robustness to corruptions or adversarial attacks does not improve effective robustness under natural dataset shifts.
  }
  \label{fig:eff_robutness_dataset_shift}
\end{figure*}

For natural dataset shifts, the scatter plots in Figure \ref{fig:eff_robutness_dataset_shift} are weakly correlated (the Pearson correlation coefficients are $r = 0.24, -0.05, -0.01, -0.26, 0.61, 0.30, 0.52, 0.36$ in reading order), 
indicating that improved robustness to corruptions or adversarial attacks in general does not improve effective robustness under natural dataset shifts.
Of the group, the two strongest correlations are effective robustness between \imagenetvidrobust{} and image corruptions ($r = 0.61$) and between \ytbbrobust{} and image corruptions ($r = 0.52$). 
While not very strong, the correlations are significant, and exploring this phenomenon between image corruptions and video anchor frames is an interesting direction for future work.

\begin{figure*}[ht!]
  \centering
  \begin{subfigure}[t]{\iftoggle{isneurips}{0.45}{0.40}\textwidth}
    \includegraphics[width=\textwidth]{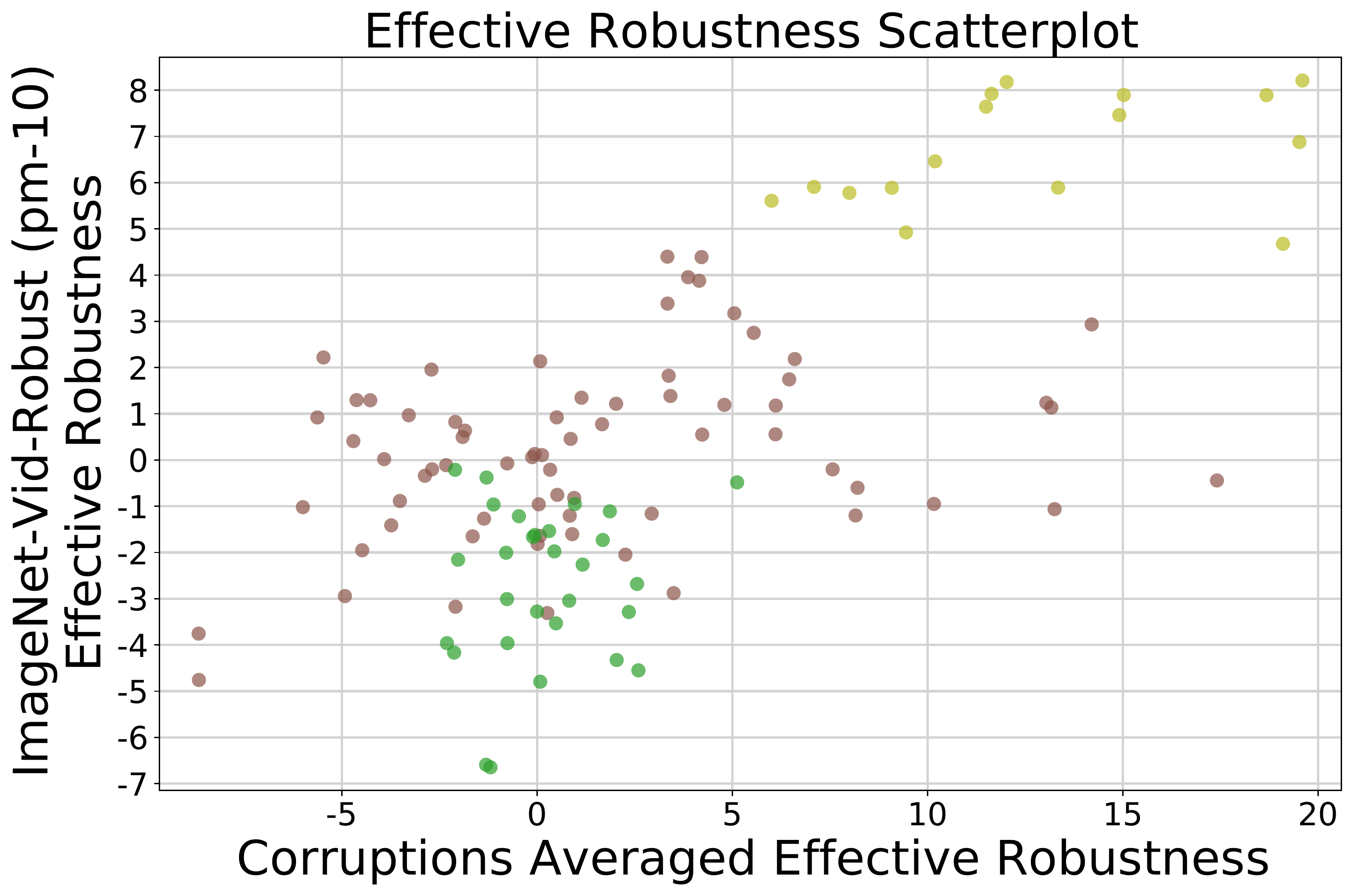}
  \end{subfigure}
  \hspace{0.5cm}
  \begin{subfigure}[t]{\iftoggle{isneurips}{0.45}{0.40}\textwidth}
    \includegraphics[width=\textwidth]{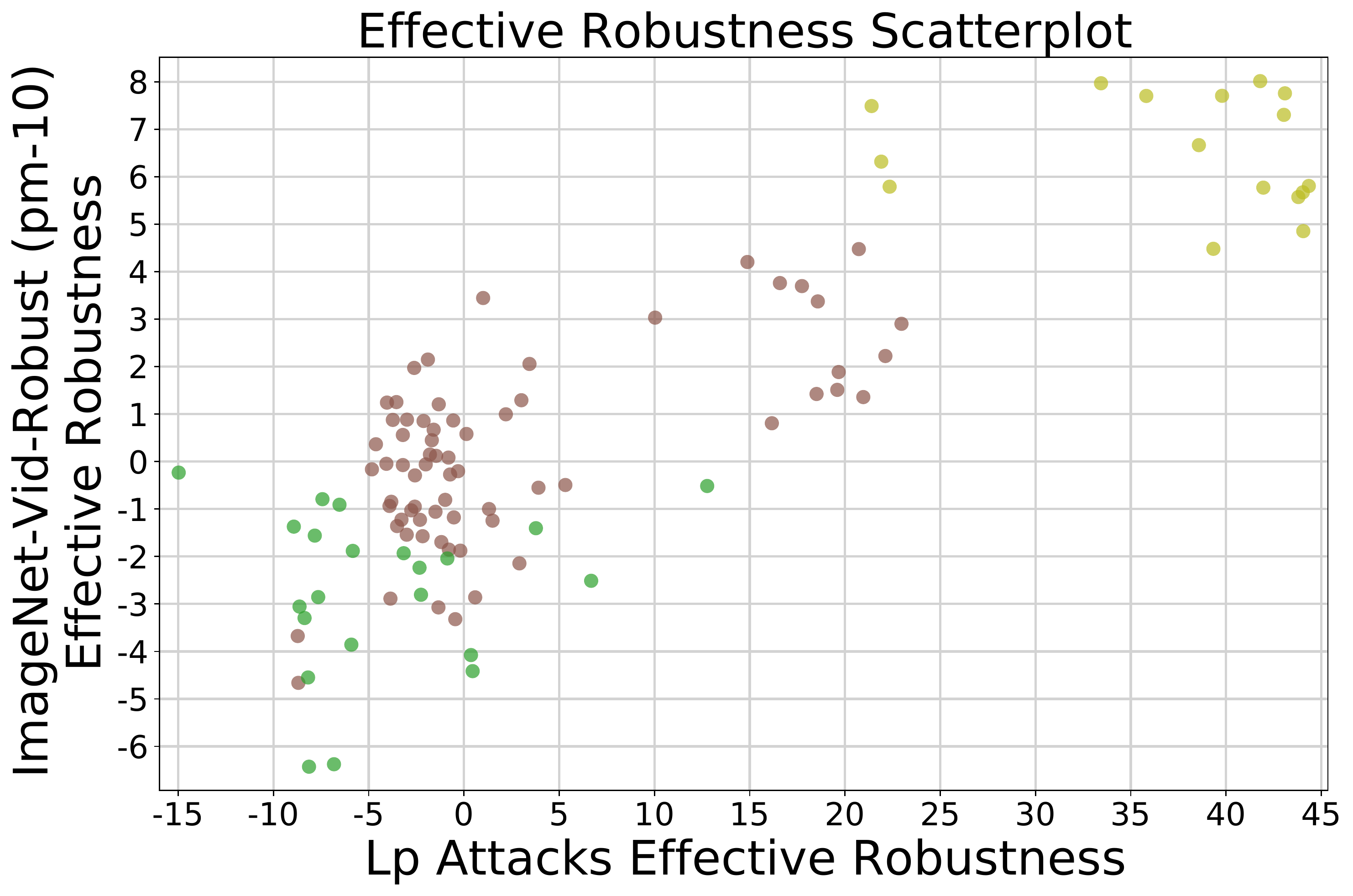}
  \end{subfigure}

  \par\medskip
  \begin{subfigure}[t]{\iftoggle{isneurips}{0.45}{0.40}\textwidth}
    \includegraphics[width=\textwidth]{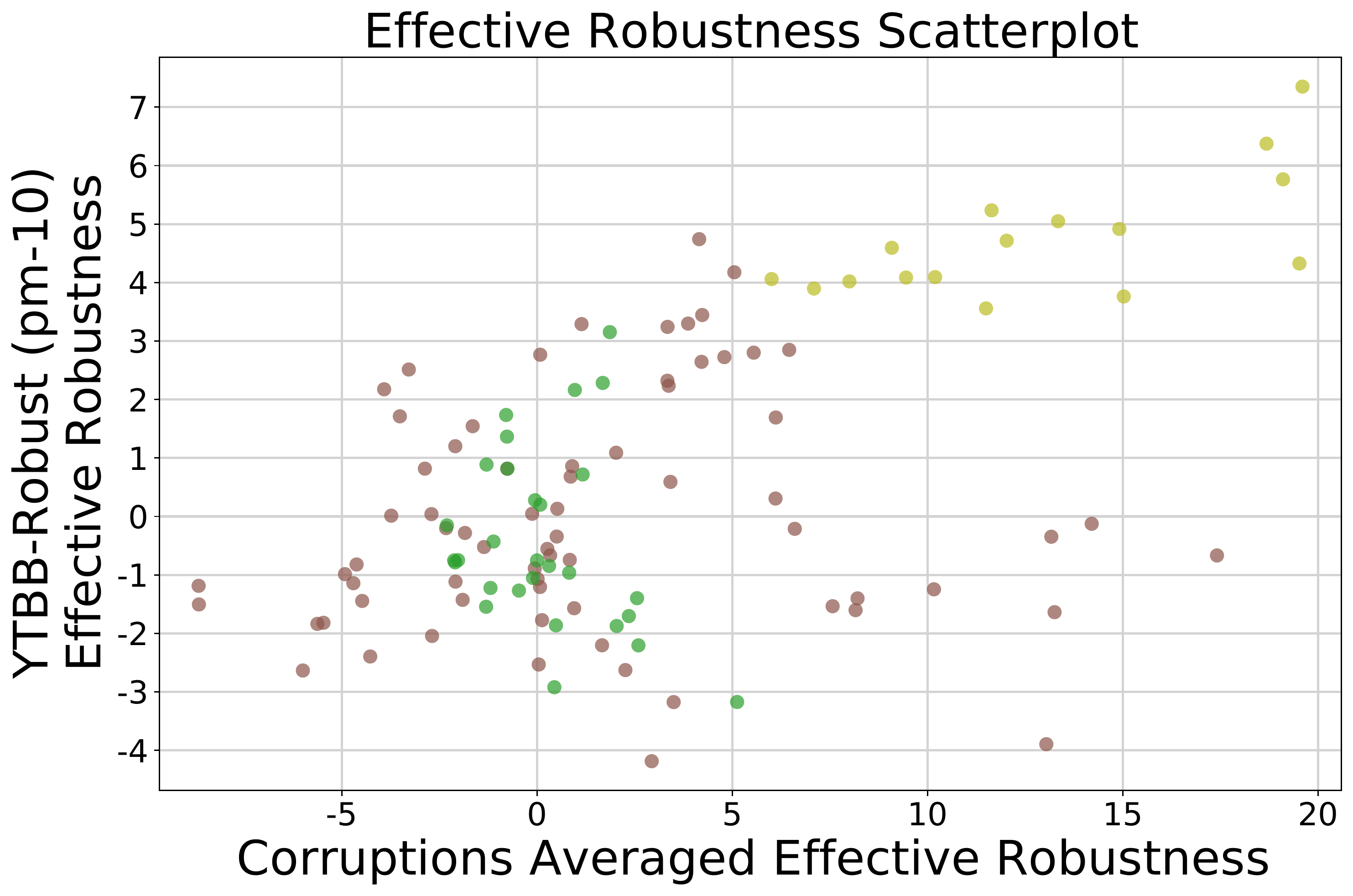}
  \end{subfigure}
  \hspace{0.5cm}
  \begin{subfigure}[t]{\iftoggle{isneurips}{0.45}{0.40}\textwidth}
    \includegraphics[width=\textwidth]{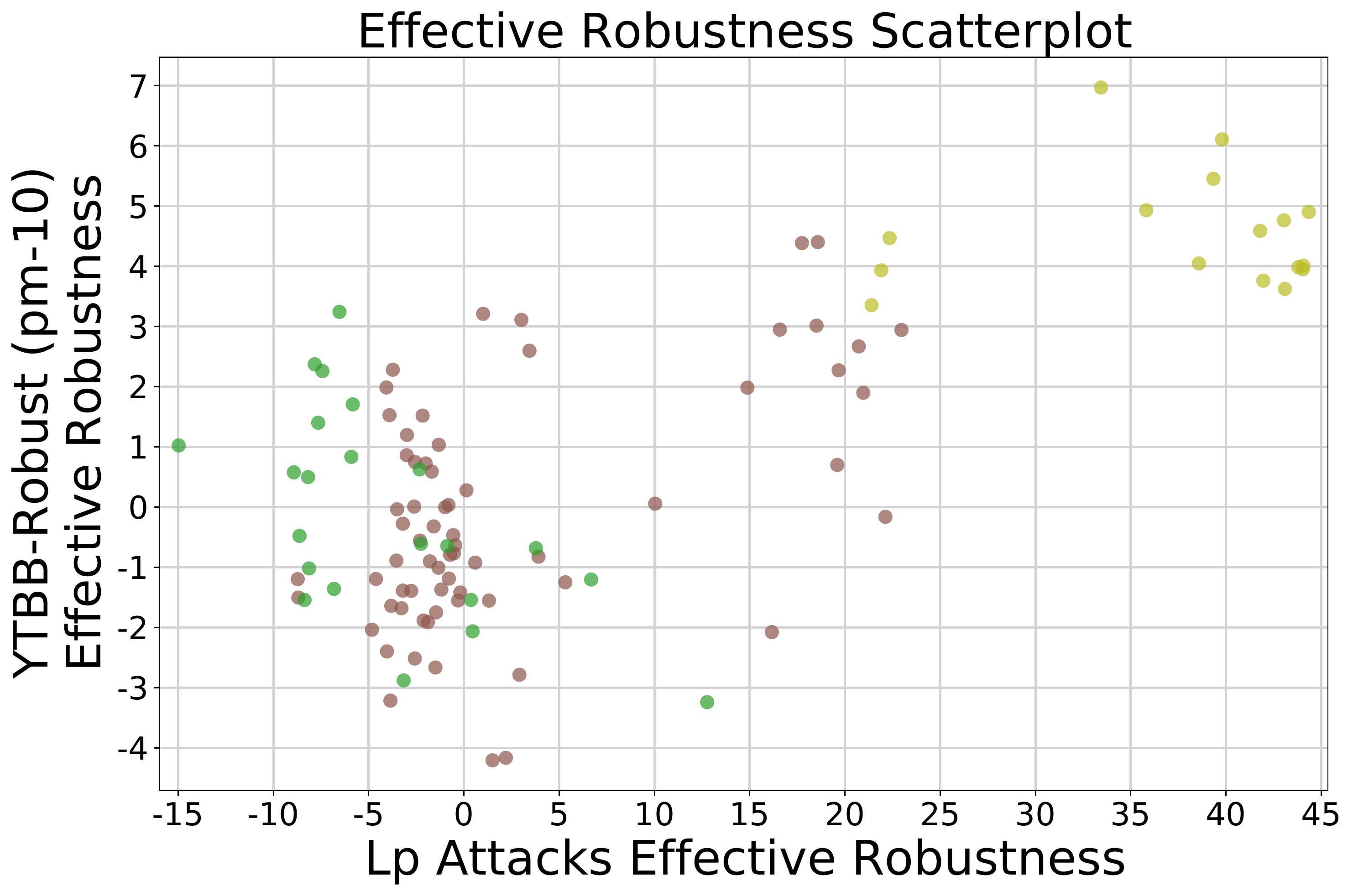}
  \end{subfigure}

  \begin{subfigure}[t]{\textwidth}
    \centering \includegraphics[width=0.9\textwidth]{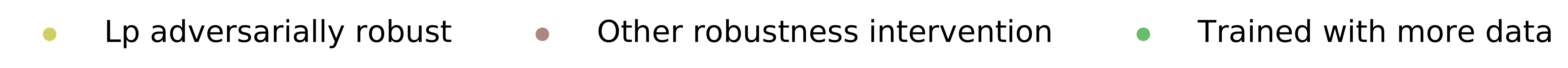}
  \end{subfigure}

  \caption{
    We compare the effective robustness of models with their accuracy drop due to corruptions (left column) and adversarial attacks (right column).
    The effective robustness is computed with respect to linear fits on the two consistency shifts: \imagenetvidrobust{} (first row), and \ytbbrobust{} (second row).
    The measures are largely uncorrelated, with the exception that accuracy on adversarial attacks is correlated with effective robustness on consistency shifts for lp adversarially models.
  }
  \label{fig:eff_robutness_consistency_shift}
\end{figure*}

For natural consistency shifts, the plots in Figure \ref{fig:eff_robutness_consistency_shift} are largely uncorrelated, with the exception that accuracy on adversarial attacks is correlated with effective robustness on consistency shifts for lp adversarially models. However, as explored in Appendix \ref{app:rel-eff-robustness-resnet50}, effective robustness on these shifts does not always imply relative robustness.

\begin{figure*}[ht!]
  \centering
  \begin{subfigure}[t]{\iftoggle{isneurips}{0.45}{0.40}\textwidth}
    \includegraphics[width=\textwidth]{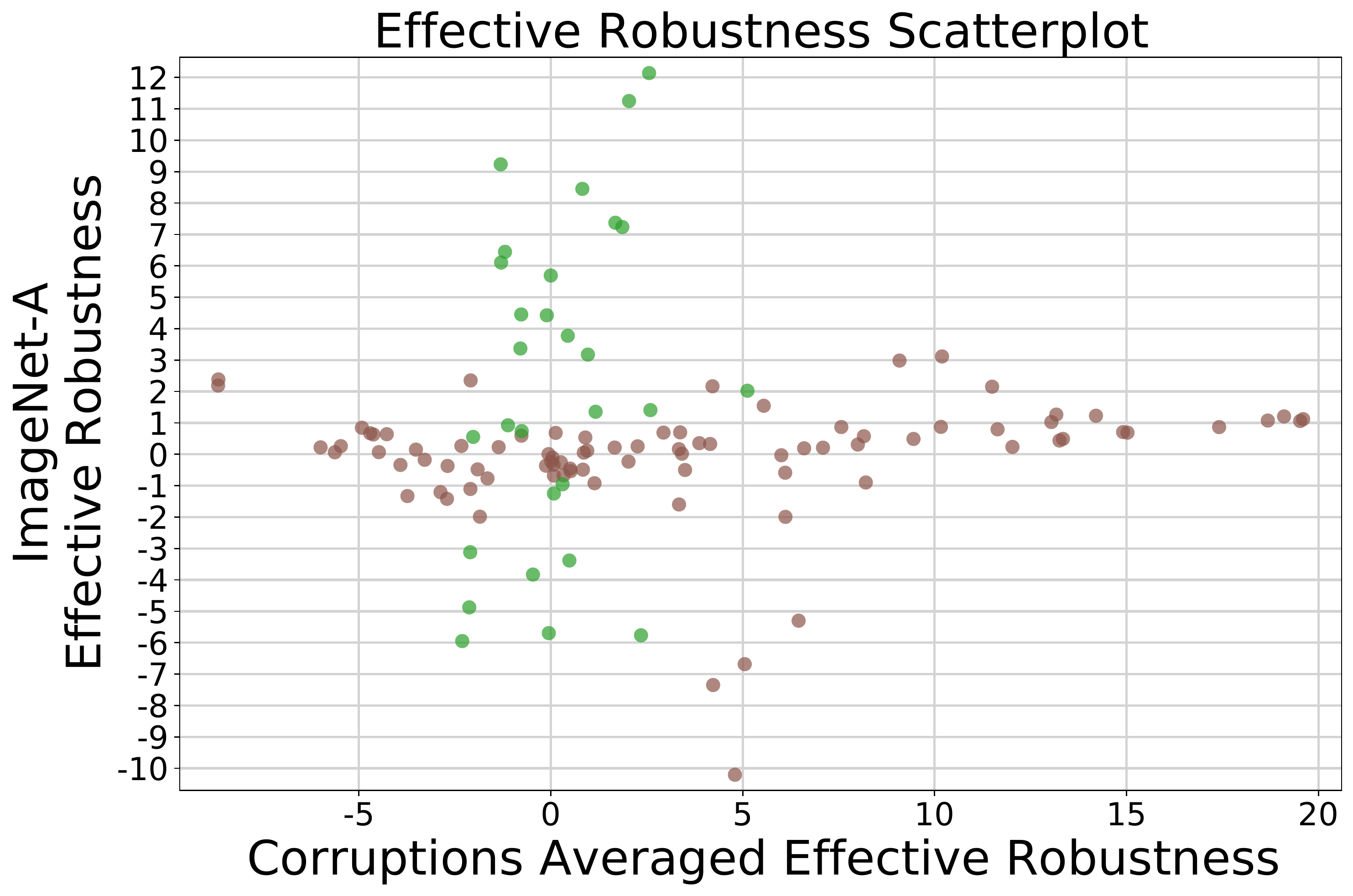}
  \end{subfigure}
  \hspace{0.5cm}
  \begin{subfigure}[t]{\iftoggle{isneurips}{0.45}{0.40}\textwidth}
    \includegraphics[width=\textwidth]{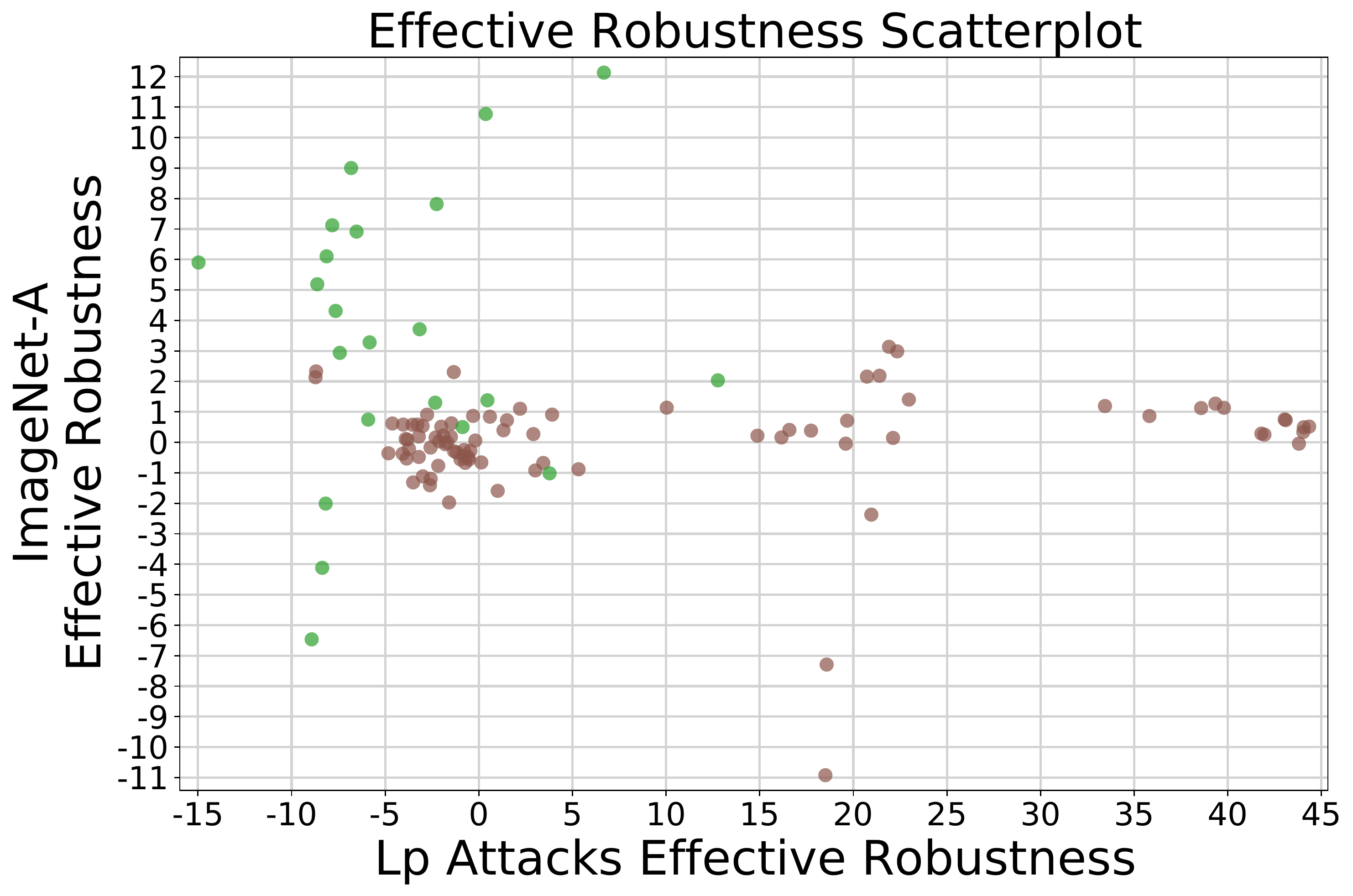}
  \end{subfigure}

  \begin{subfigure}[t]{\textwidth}
    \centering \includegraphics[width=0.55\textwidth]{appendix/eff_robust_legend}
  \end{subfigure}

  \caption{
    We compare the effective robustness of models with their accuracy drop due to corruptions (left column) and adversarial attacks (right column).
    The effective robustness is computed with respect to a linear fit on \imageneta{}, the adversarially filtered shift.
    After computing effective robustness piecewise around the ResNet50 accuracy, there is no observed correlation between the synthetic and natural robustness measures.
  }
  \label{fig:eff_robutness_adversarially_filtered}
\end{figure*}

For the adversarially filtered shift, as seen in Figure \ref{fig:eff_robutness_adversarially_filtered}, after computing effective robustness piecewise around the ResNet50 accuracy, there is no observed correlation between the synthetic and natural robustness measures on \imageneta{}.

\clearpage

\section{Corruption robustness}
\label{app:corr-robustness}

\begin{figure}[h]
  \centering \includegraphics[width=0.8\textwidth]{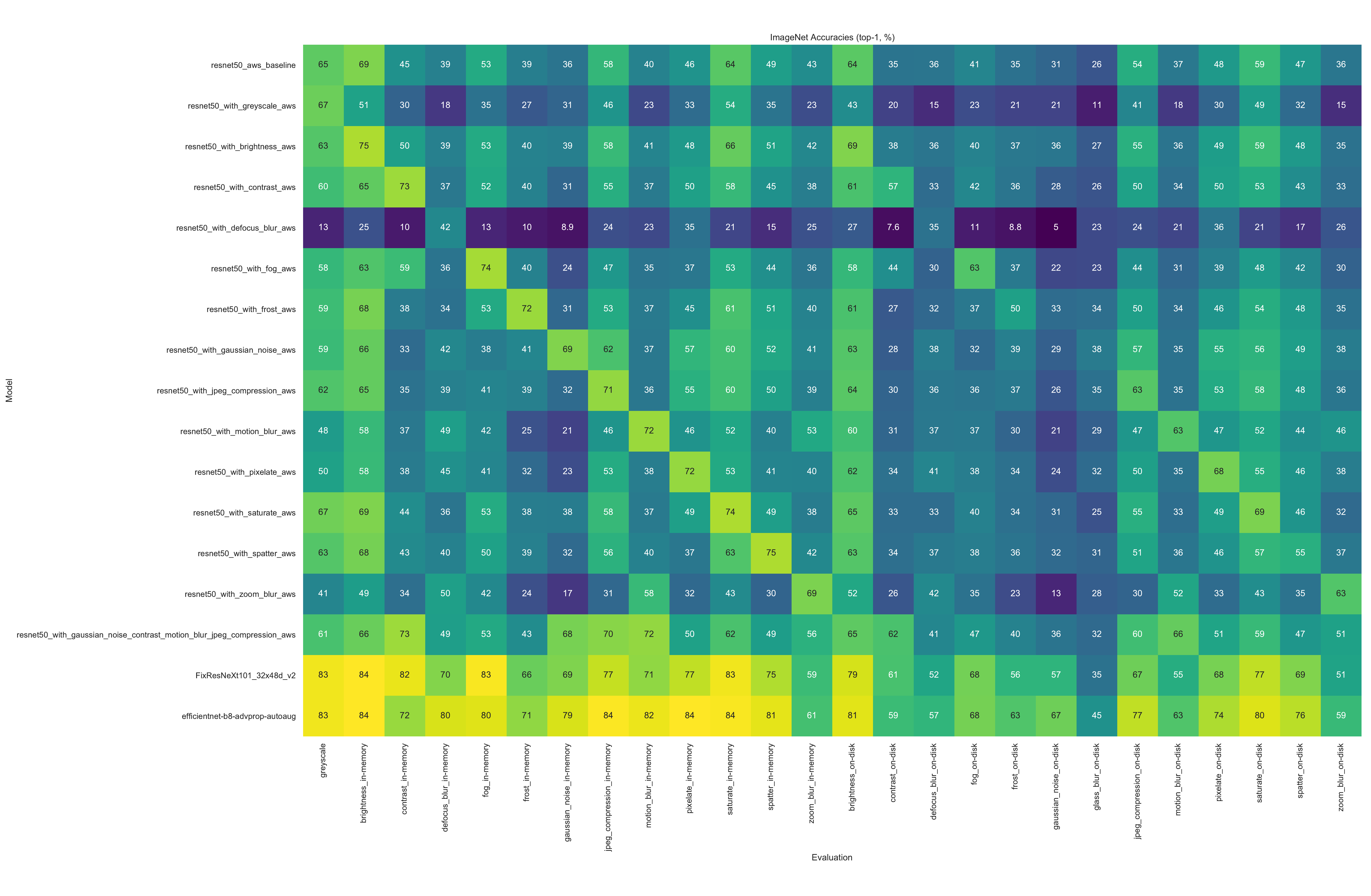}
  \caption{
  A detailed view of corruption robustness, with cells sampled from the main grid in Figure \ref{fig:grid}.
  Here we present ResNet50s trained on some of the corruptions from the ImageNet-C benchmark, as well as the best model trained on more data, FixResNeXt101\_32x48d\_v2, and the best model trained on just the standard training set, efficientnet-b8-advprop-autoaug.
  }
  \label{fig:zoom_grid}
\end{figure}

We have already seen that corruption robustness does not promote effective robustness, or robustness to real distribution shift. Here, we analyze whether robustness to some corruptions transfers to others, and what may contribute to corruption robustness. 
Figure \ref{fig:zoom_grid} shows the result of training various ResNet50s\footnote{Each ResNet50 was trained with a batch size of 256 for 120 epochs, starting with a learning rate of 0.1 and decaying by a factor of 10 every 30 epochs. For the ResNet50s trained on corruptions, we randomly sample a corruption and severity for each image. Refer to \ref{app:corruptions} for details on corruptions and severities. We use our custom fast gpu implementations of these corruptions for training.} on a few corruptions from ImageNet-C.

In line with prior work, this plot here tells us that training against one type of synthetic corruption or one set of synthetic corruption does not transfer well to other corruptions. There are cases where transfer does happen, but overall the models are only robust to the corruption they are trained on.

It is also interesting to note (from \Cref{fig:grid}) that PGD models actually see a drop in robustness to low frequency corruptions such as contrast, a phenomenon also observed in \citep{yin2019fourier}.

\clearpage

\section{Evaluation settings in the testbed}
\label{app:eval-settings}

\subsection{Natural distribution shifts}

For \imagenetnew{}, we evaluate on the following datasets: imagenetv2-matched-frequency, imagenetv2-matched-frequency-format-val, imagenetv2-threshold-0.7, imagenetv2-threshold-0.7-format-val, imagenetv2-top-images, imagenetv2-top-images-format-val. The format-val versions are variants of the original dataset encoded with jpeg settings similar to the original one. Unless otherwise stated, results in our paper referring to imagenetv2 are for imagenetv2-matched-frequency-format-val.

For \objectnet{}, we obtained a beta version of the dataset through personal correspondance. Each image in the dataset was then cropped by 2px on each side following the authors' instructions. Predictions were taken over only the classes that also appeared in the 1000 classes for the ImageNet validation set.

For \imagenetvidrobust{} and \ytbbrobust{}, we look at the anchor frames in the dataset and evaluate the benign accuracy for pm0. For pm10, we look at up to 20 nearest frames marked ``similar'' to the anchor frame in the dataset and count it as a misclassification if any one of the predictions is wrong.

For \imageneta{}, predictions were taken over only the classes that also appeared in the 1000 classes for the ImageNet validation set.

\subsection{Corruptions}
\label{app:corruptions}

We include \numcorruptions{} different corruption types: greyscale (in memory), gaussian noise (in memory and on disk), shot noise (in memory and on disk), impulse noise (in memory and on disk), speckle noise (in memory and on disk), gaussian blur (in memory and on disk), defocus blur (in memory and on disk), glass blur (on disk), motion blur (in memory and on disk), zoom blur (in memory and on disk), snow (in memory and on disk), frost (in memory and on disk), fog (in memory and on disk), spatter (in memory and on disk), brightness (in memory and on disk), contrast (in memory and on disk), saturate (in memory and on disk), pixelate (in memory and on disk), jpeg compression (in memory and on disk), elastic transform (in memory and on disk).

For each corruption, we average over the five severities.

We make sure to make the distinction between in memory corruptions, for which we provide custom fast gpu implementations, and on disk corruptions, for which we use the publicly available ImageNet-C dataset, since it was reported in \cite{ford2019} that jpeg compression can have a significant impact on model accuracies (indeed, as evidenced by Figure \ref{fig:zoom_grid}). 

\subsection{Adversarial attacks}
\label{app:pgd}

We run the following 4 pgd attacks one each model with these settings:

\texttt{pgd.linf.eps0.5} Norm: 0.5/255, Step size: 5.88e-5, Num steps: 100

\texttt{pgd.linf.eps2} Norm: 2/255, Step size: 2.35e-4, Num steps: 100

\texttt{pgd.l2.eps0.1} Norm: 0.1, Step size: 0.01, Num steps: 100

\texttt{pgd.l2.eps0.5} Norm: 0.5, Step size: 0.05, Num steps: 100

Most of the models were attacked with only 10\% of the dataset (in a class-balanced manner) due to computational constraints. These models are displayed with larger error bars in the plots.

\subsection{Stylized Imagenet}
We use the stylized imagenet dataset used by \citep{geirhos2019imagenettrained} as another evaluation dataset.

\subsection{125 class evaluation}
\label{sec:appendix-125-class}
For the 125 subsampled class evaluation, we evaluate on the following classes from ILSVRC:

\texttt{n01494475 n01630670 n01644373 n01644900 n01669191 n01677366 n01697457 n01742172 n01796340 n01829413 n01871265 n01924916 n01944390 n01978287 n01980166 n02007558 n02009229 n02017213 n02033041 n02037110 n02056570 n02071294 n02085936 n02086079 n02093428 n02093991 n02095314 n02095570 n02096294 n02096437 n02097474 n02100236 n02100583 n02102318 n02105056 n02107574 n02112706 n02113023 n02114855 n02128925 n02134418 n02138441 n02165105 n02219486 n02226429 n02264363 n02280649 n02441942 n02483708 n02486261 n02488291 n02492035 n02641379 n02730930 n02777292 n02790996 n02795169 n02808440 n02814533 n02814860 n02837789 n02859443 n02892201 n02895154 n02948072 n02951585 n02977058 n03000247 n03110669 n03201208 n03208938 n03216828 n03240683 n03250847 n03272562 n03297495 n03337140 n03376595 n03379051 n03447721 n03492542 n03527444 n03535780 n03642806 n03670208 n03673027 n03692522 n03710193 n03775071 n03832673 n03838899 n03840681 n03868242 n03873416 n03877845 n03884397 n03908714 n03920288 n03933933 n04004767 n04009552 n04037443 n04041544 n04067472 n04074963 n04099969 n04125021 n04141975 n04149813 n04204238 n04208210 n04229816 n04266014 n04310018 n04330267 n04335435 n04336792 n04355338 n04417672 n04479046 n04505470 n07715103 n07875152 n09256479 n12620546}

\clearpage

\section{Models in the testbed}
\label{app:models}

The following list contains all models we evaluated on ImageNet with references and links to the corresponding source code. 
Also noted is the model type used to color the plots in the paper.

\input{appendix/model_list}

\clearpage

\section{Model accuracies}
\label{app:model-accuracies}

\begin{footnotesize}
\rowcolors{1}{gray!15}{white}
\begin{longtable}[ht!]{W R R R R}
  \hiderowcolors
  \caption{
  Top-1 model accuracies on ImageNet validation set, effective robustness as calculated with respect to \imagenetnew{}, an average over all the corruptions, and an average over all the pgd attacks. 
  Note that since we take an average of many attacks, the PGD column can no longer be considered a worst-case attacker for the model (look to \ref{app:pgd} for specific attacks).
  }
  \label{tab:model_results}\\

  \hline
  \multicolumn{5}{c}{\textbf{Model accuracies}}\\
  \hline
  Model & ImageNet accuracy & ImageNetV2 eff. robust. & Avg. corr. accuracy & Avg. PGD accuracy\\
  \hline
  \endfirsthead

  \hline
  \multicolumn{5}{c}{\textbf{Model accuracies} (continued from previous page)}\\
  \hline
  Model & ImageNet accuracy & ImageNetV2 eff. robust. & Avg. corr. accuracy & Avg. PGD accuracy\\
  \hline
  \endhead

  \hline
  \multicolumn{5}{c}{Table continues onto next page}\\
  \hline
  \endfoot

  \hline
  \multicolumn{5}{c}{End of table}\\
  \hline\hline
  \endlastfoot
  \showrowcolors

  \input{appendix/model_table}
\end{longtable}
\end{footnotesize}

\clearpage

\section{Synthetic robustness correlation with natural robustness}
\label{app:correlation}

In this section, we investigate which individual synthetic robustness measures are most predictive of natural distribution shift.
For each of the synthetic shifts in our testbed, we compute the effective robustness for each model and measure the Pearson correlation coefficients against the effective robustness under each of the natural distribution shifts in our testbed.

\begin{figure*}[h]
    \centering
  \begin{subfigure}[t]{0.45\textwidth}
    \includegraphics[width=\linewidth]{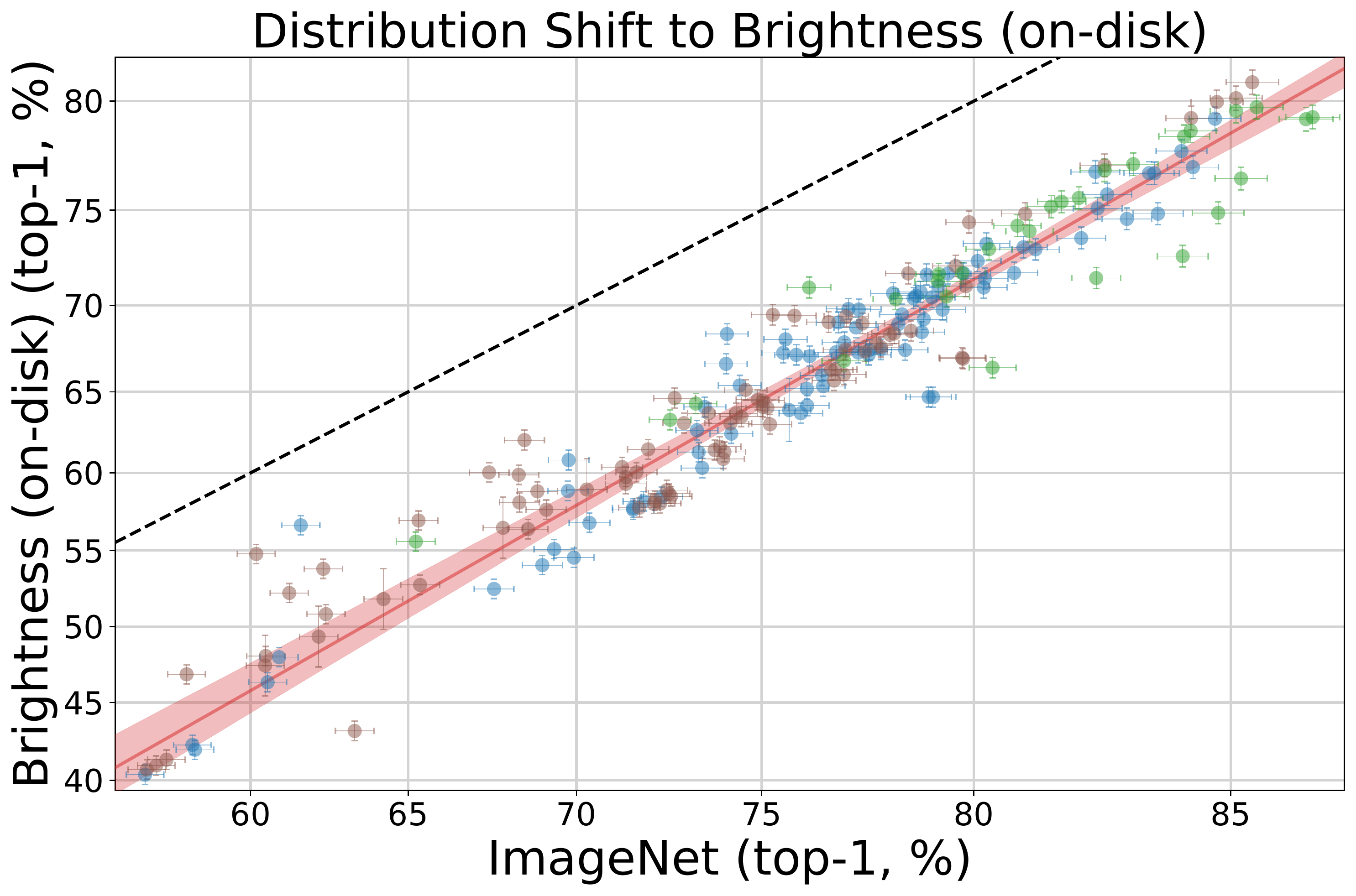}
  \end{subfigure}
    \hspace{0.5cm}
  \begin{subfigure}[t]{0.45\textwidth}
    \includegraphics[width=\linewidth]{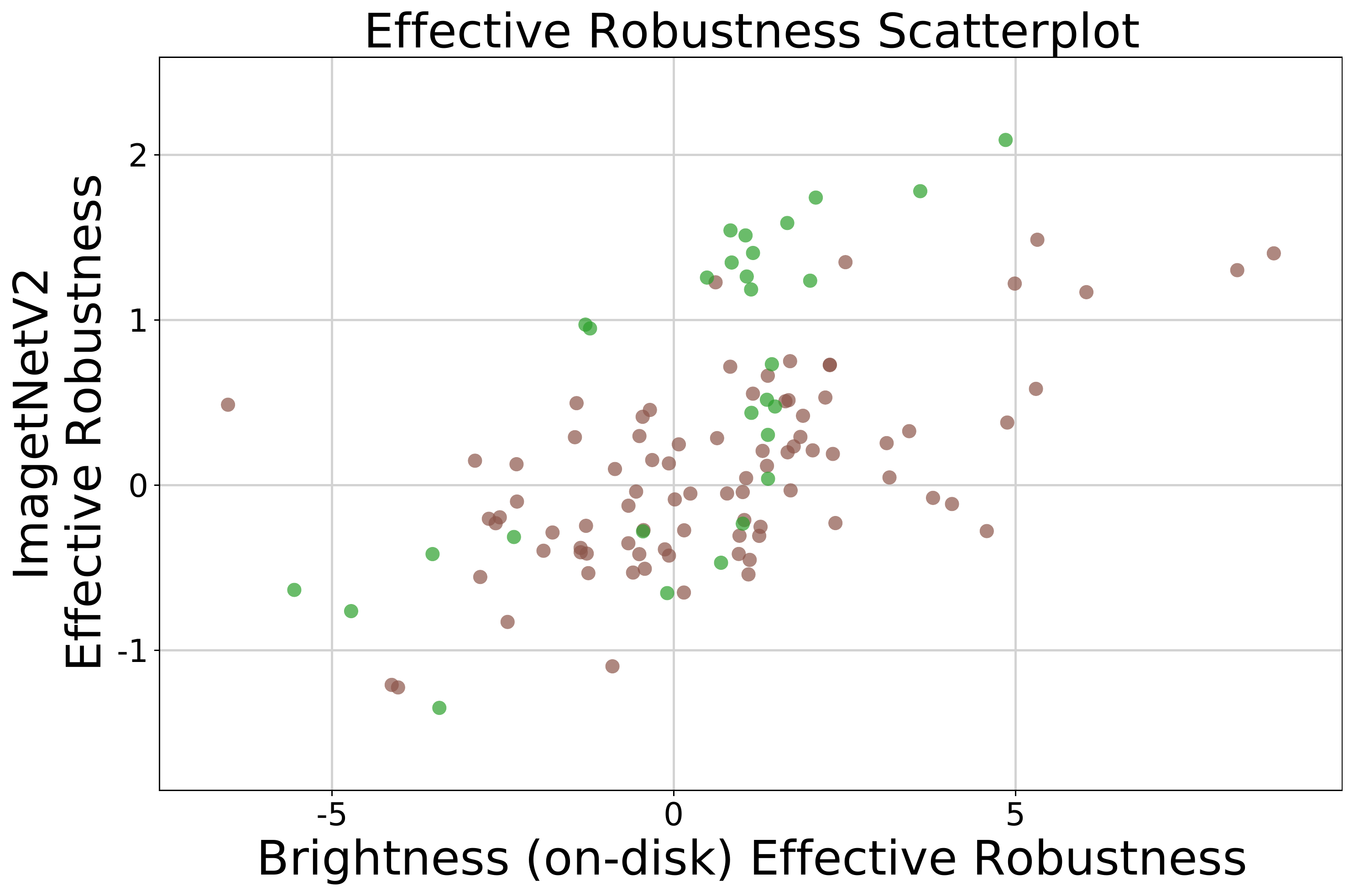}
  \end{subfigure}
  \begin{subfigure}[t]{0.45\textwidth}
    \includegraphics[width=\linewidth]{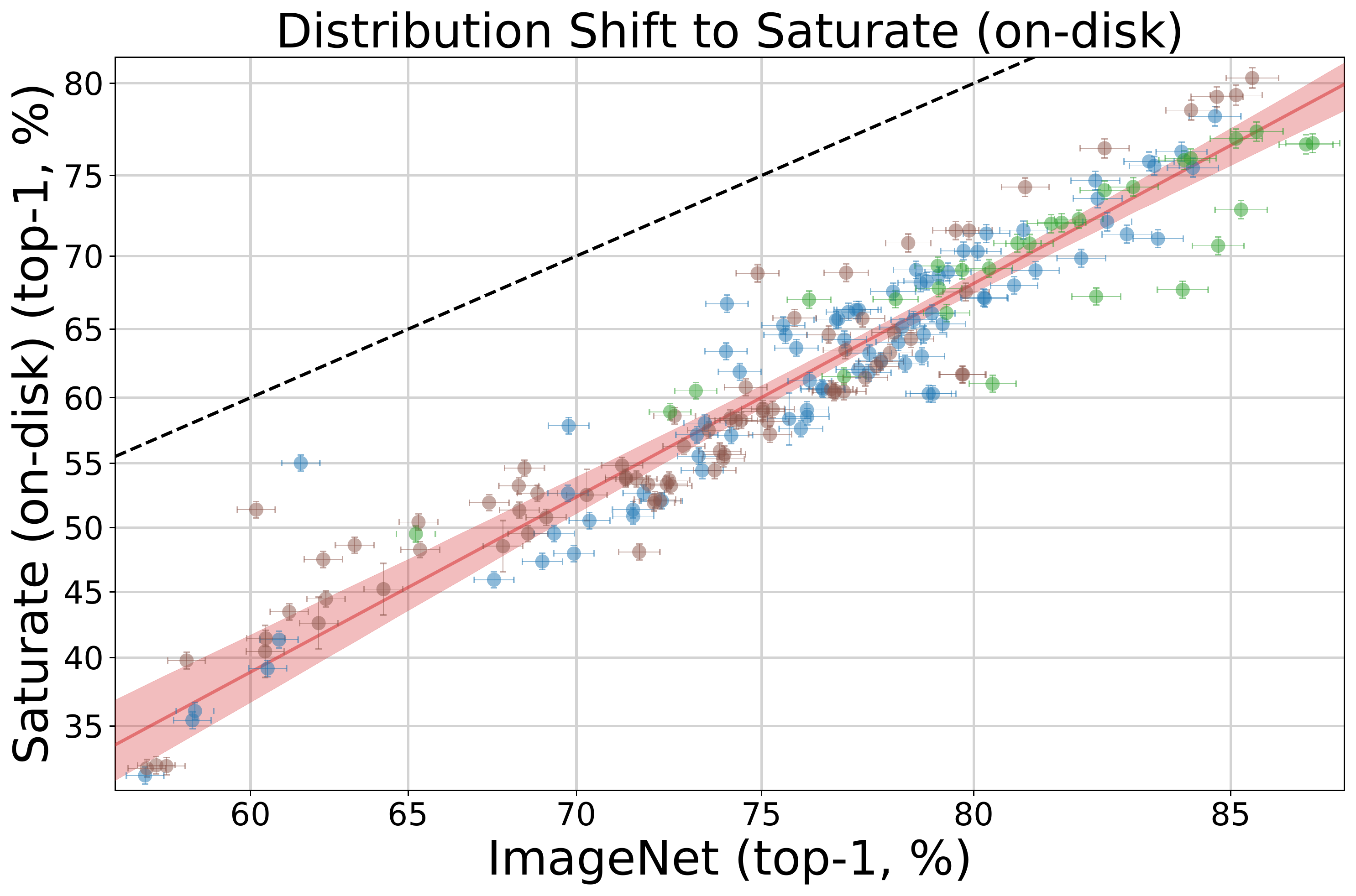}
  \end{subfigure}
    \hspace{0.5cm}
  \begin{subfigure}[t]{0.45\textwidth}
    \includegraphics[width=\linewidth]{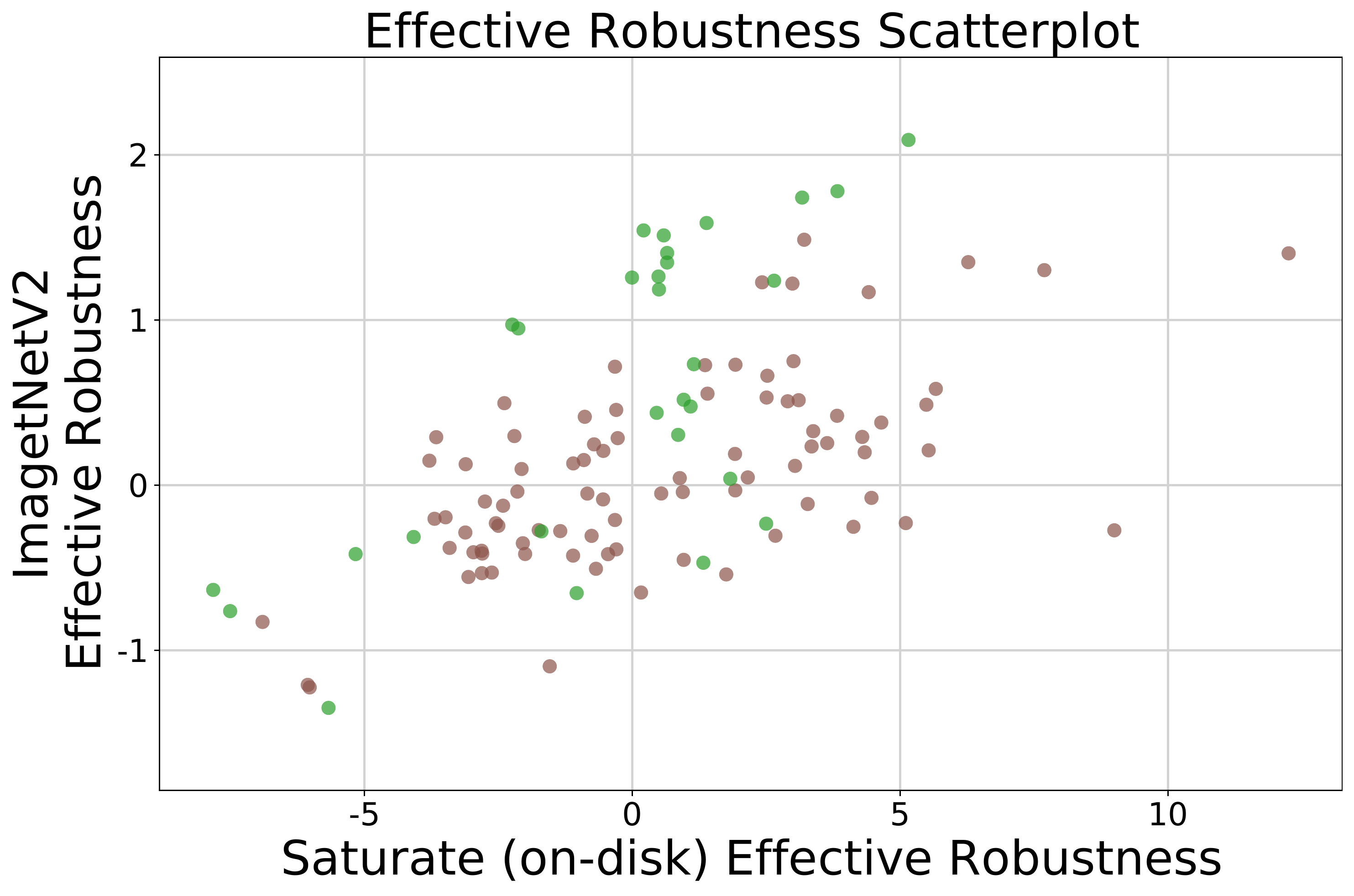}
  \end{subfigure}
  \begin{subfigure}[t]{\textwidth}
    \centering \includegraphics[width=0.7\textwidth]{figs/syn_shift_legend}
  \end{subfigure}
  \caption{
    Plots of the two synthetic distribution shifts with the highest correlation with \imagenetnew{}, compared similarly to \Cref{fig:synthetic-shifts}.
    }
  \label{fig:topk-correlation-imagenetv2}
\end{figure*}

\begin{figure*}[h]
    \centering
  \begin{subfigure}[t]{0.45\textwidth}
    \includegraphics[width=\linewidth]{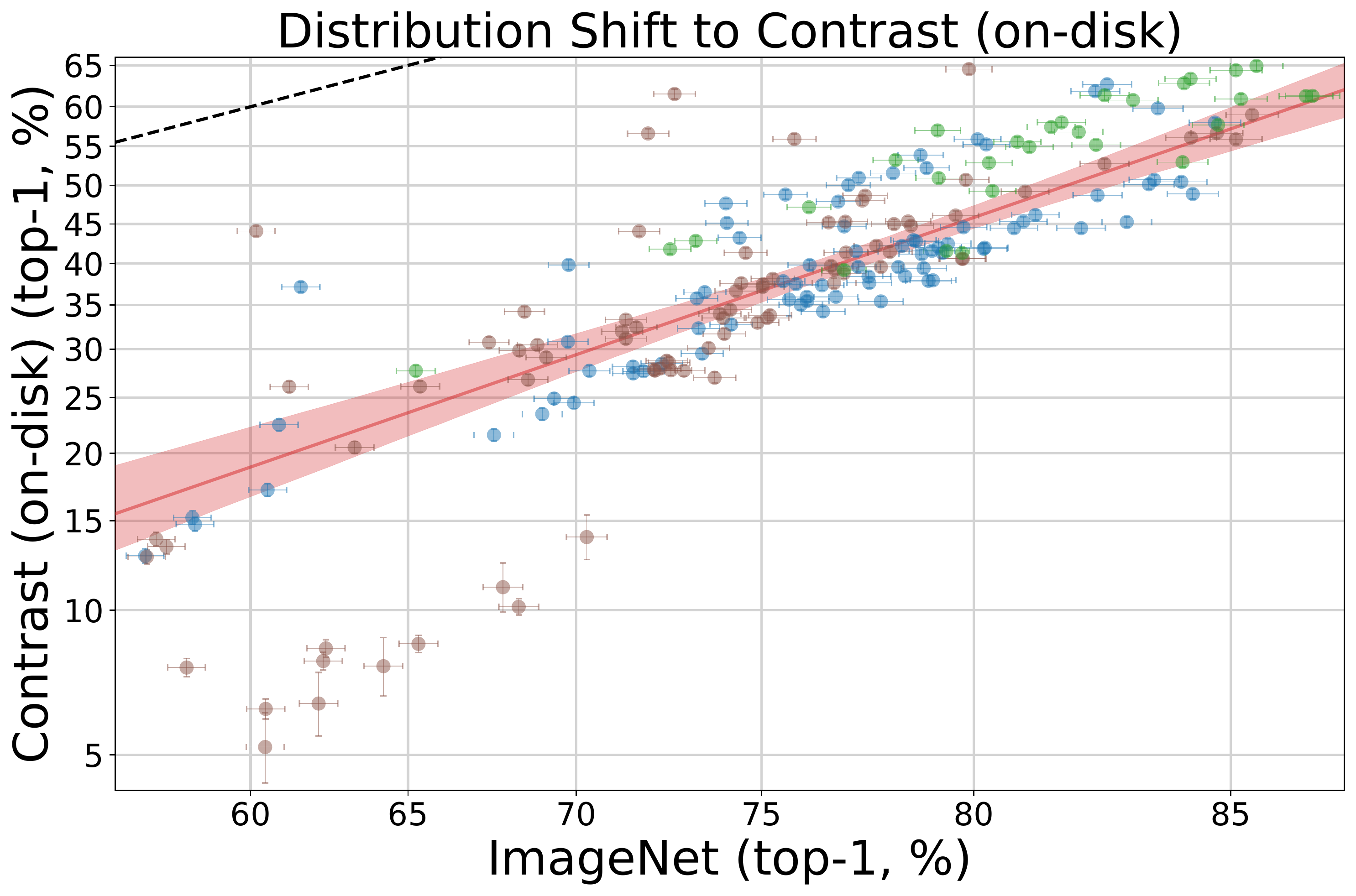}
  \end{subfigure}
    \hspace{0.5cm}
  \begin{subfigure}[t]{0.45\textwidth}
    \includegraphics[width=\linewidth]{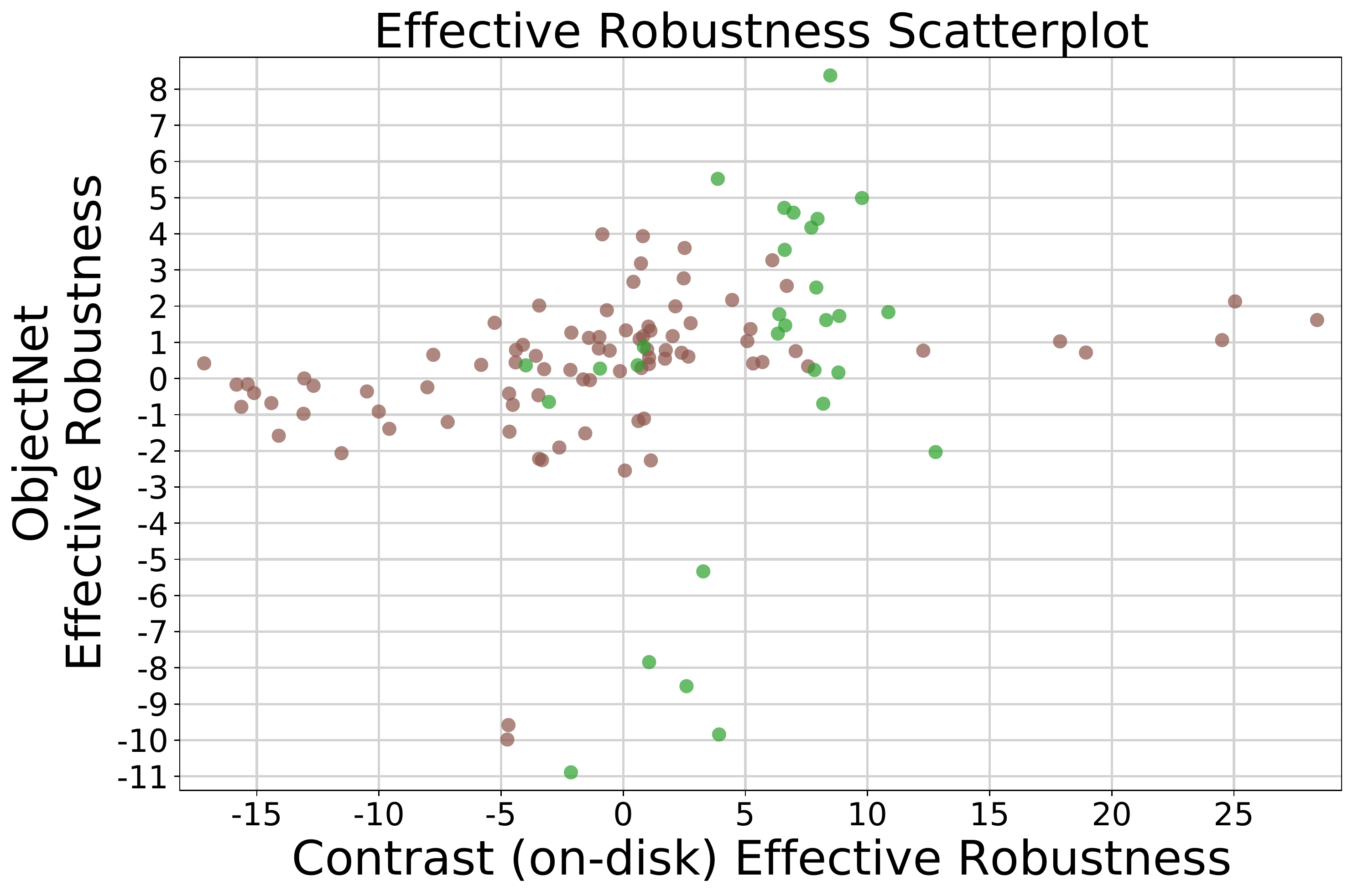}
  \end{subfigure}
  \begin{subfigure}[t]{0.45\textwidth}
    \includegraphics[width=\linewidth]{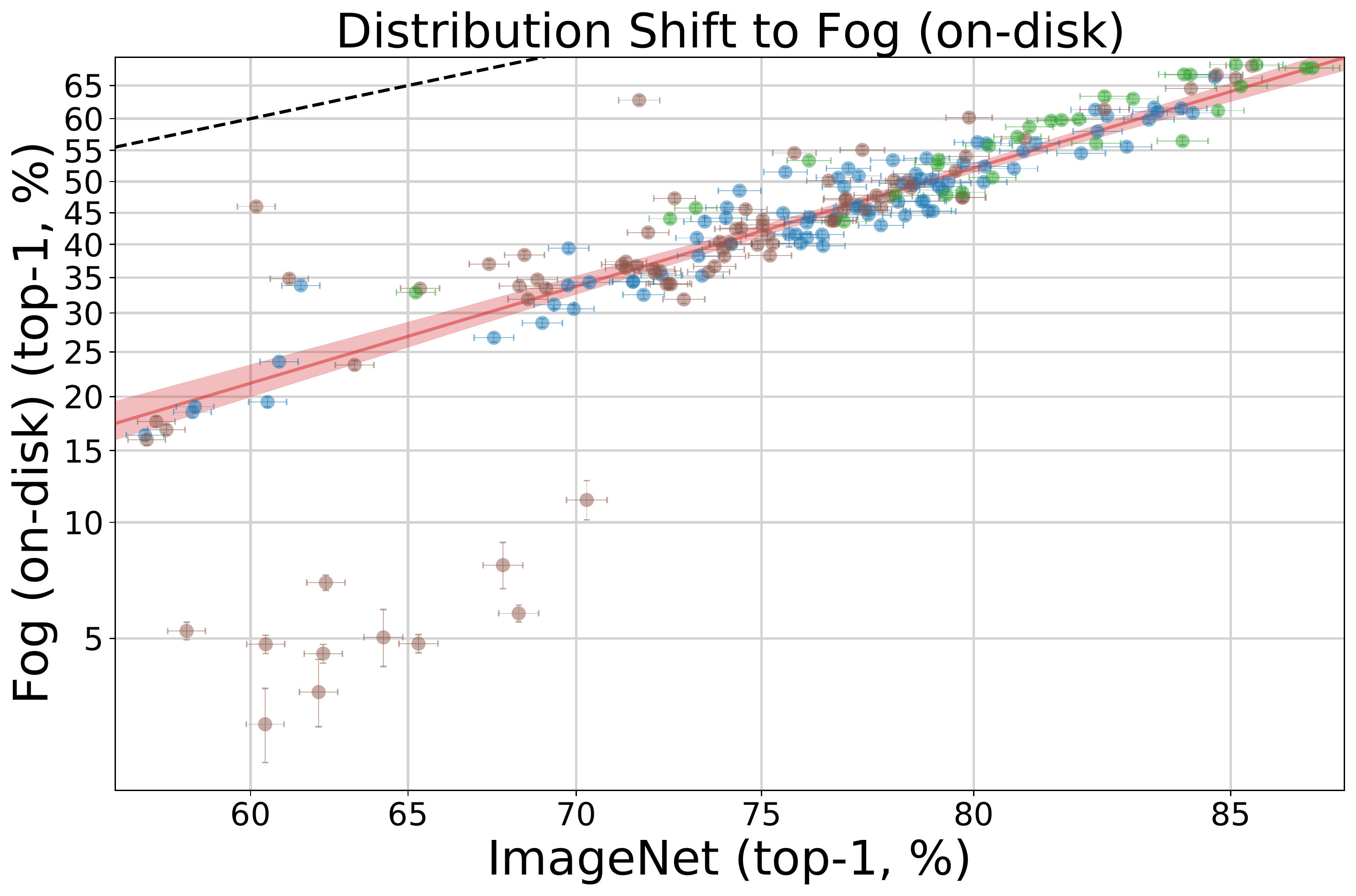}
  \end{subfigure}
    \hspace{0.5cm}
  \begin{subfigure}[t]{0.45\textwidth}
    \includegraphics[width=\linewidth]{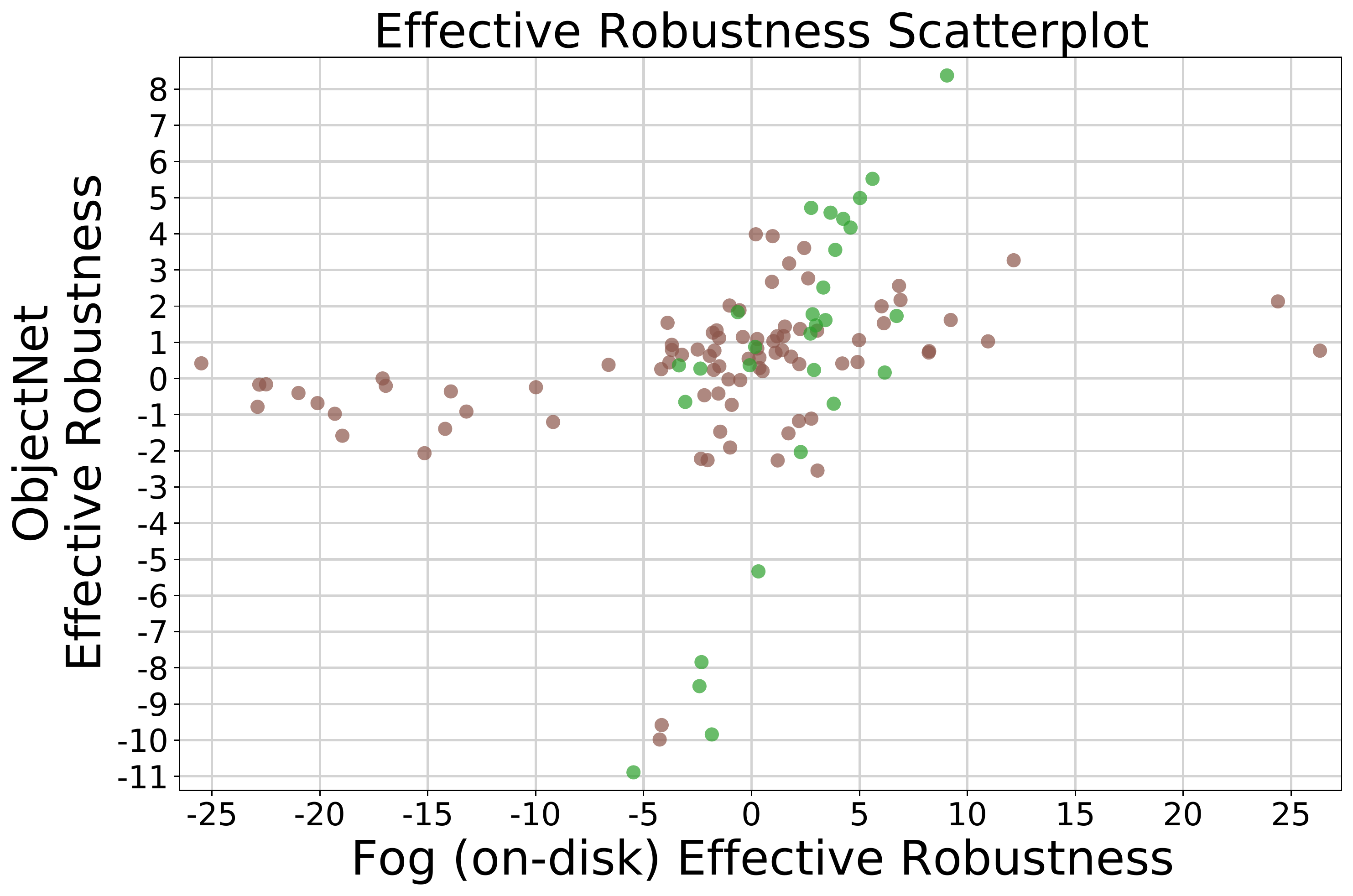}
  \end{subfigure}
  \begin{subfigure}[t]{\textwidth}
    \centering \includegraphics[width=0.7\textwidth]{figs/syn_shift_legend}
  \end{subfigure}
  \caption{
    Plots of the two synthetic distribution shifts with the highest correlation with \objectnet{}, compared similarly to \Cref{fig:synthetic-shifts}.
    }
  \label{fig:topk-correlation-objectnet}
\end{figure*}

\begin{figure*}[h]
    \centering
  \begin{subfigure}[t]{0.45\textwidth}
    \includegraphics[width=\linewidth]{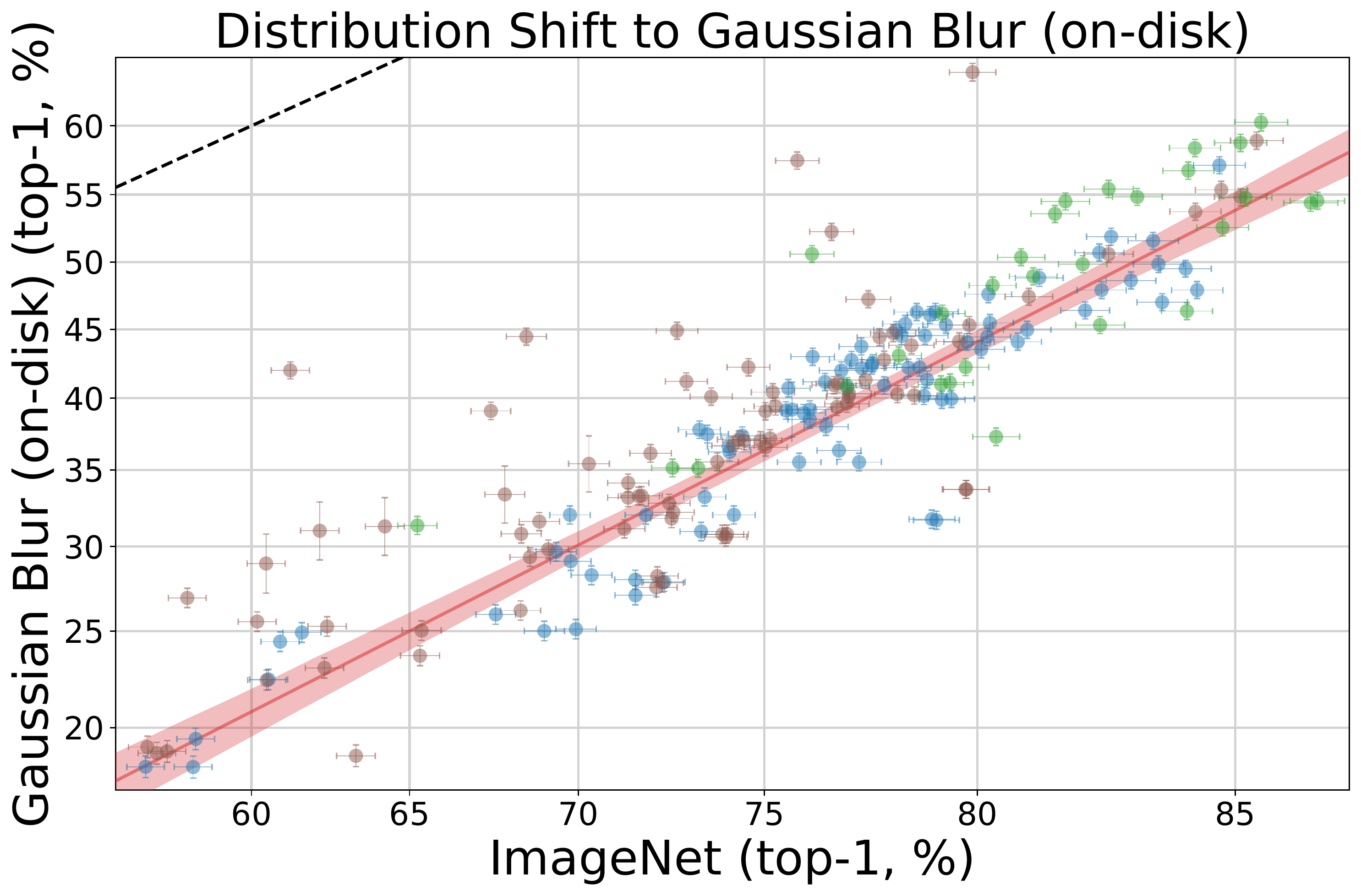}
  \end{subfigure}
    \hspace{0.5cm}
  \begin{subfigure}[t]{0.45\textwidth}
    \includegraphics[width=\linewidth]{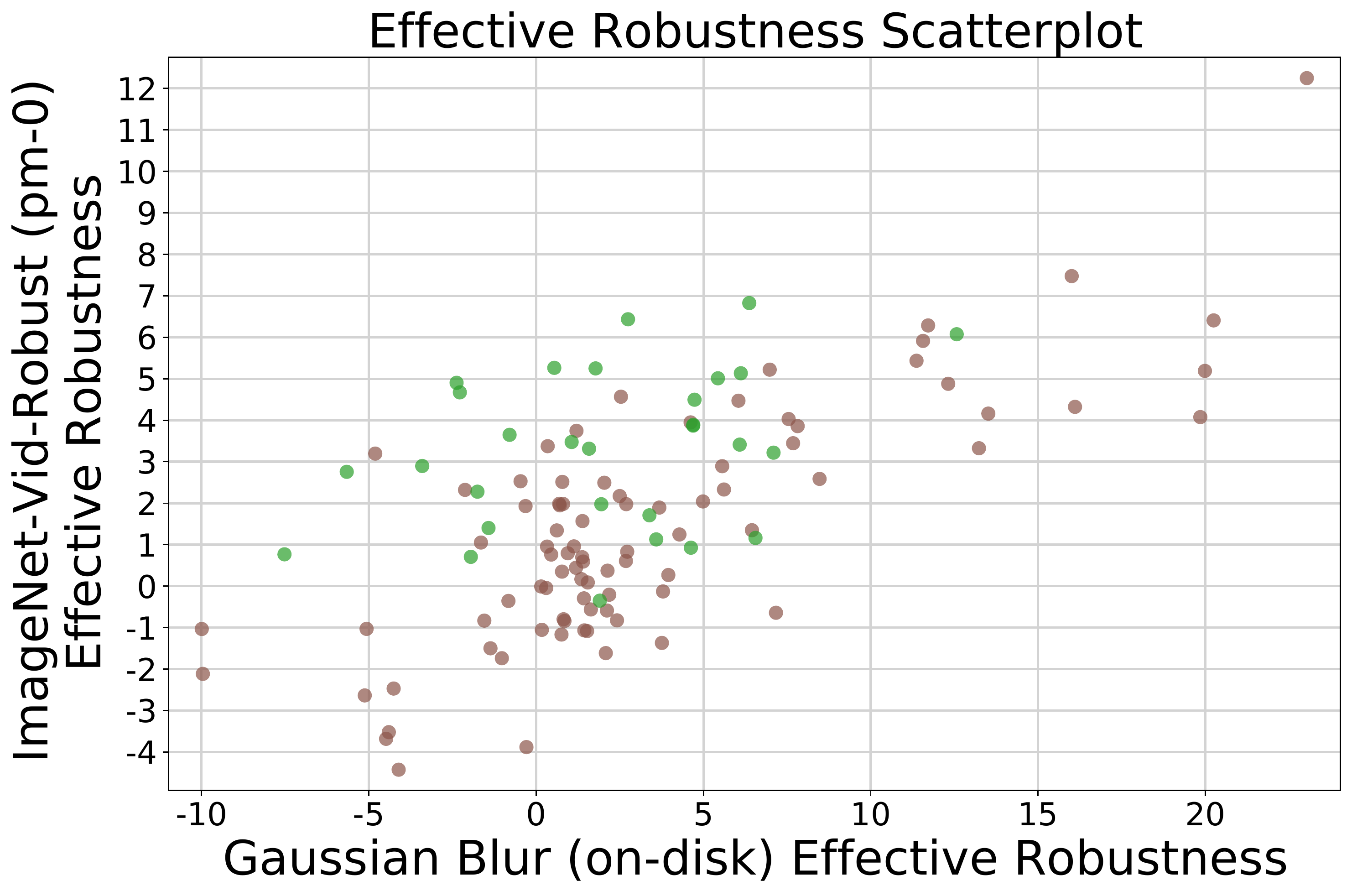}
  \end{subfigure}
  \begin{subfigure}[t]{0.45\textwidth}
    \includegraphics[width=\linewidth]{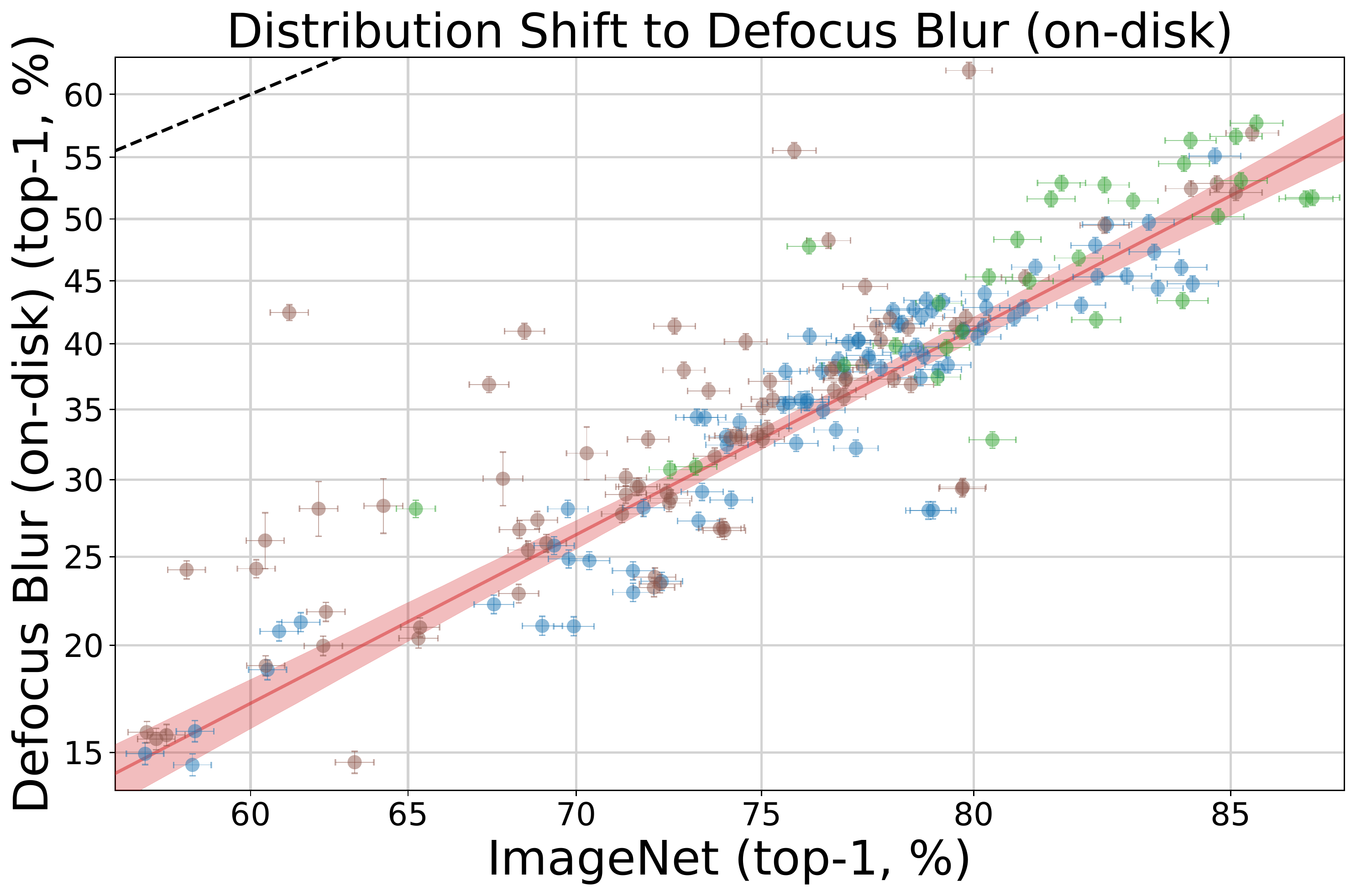}
  \end{subfigure}
    \hspace{0.5cm}
  \begin{subfigure}[t]{0.45\textwidth}
    \includegraphics[width=\linewidth]{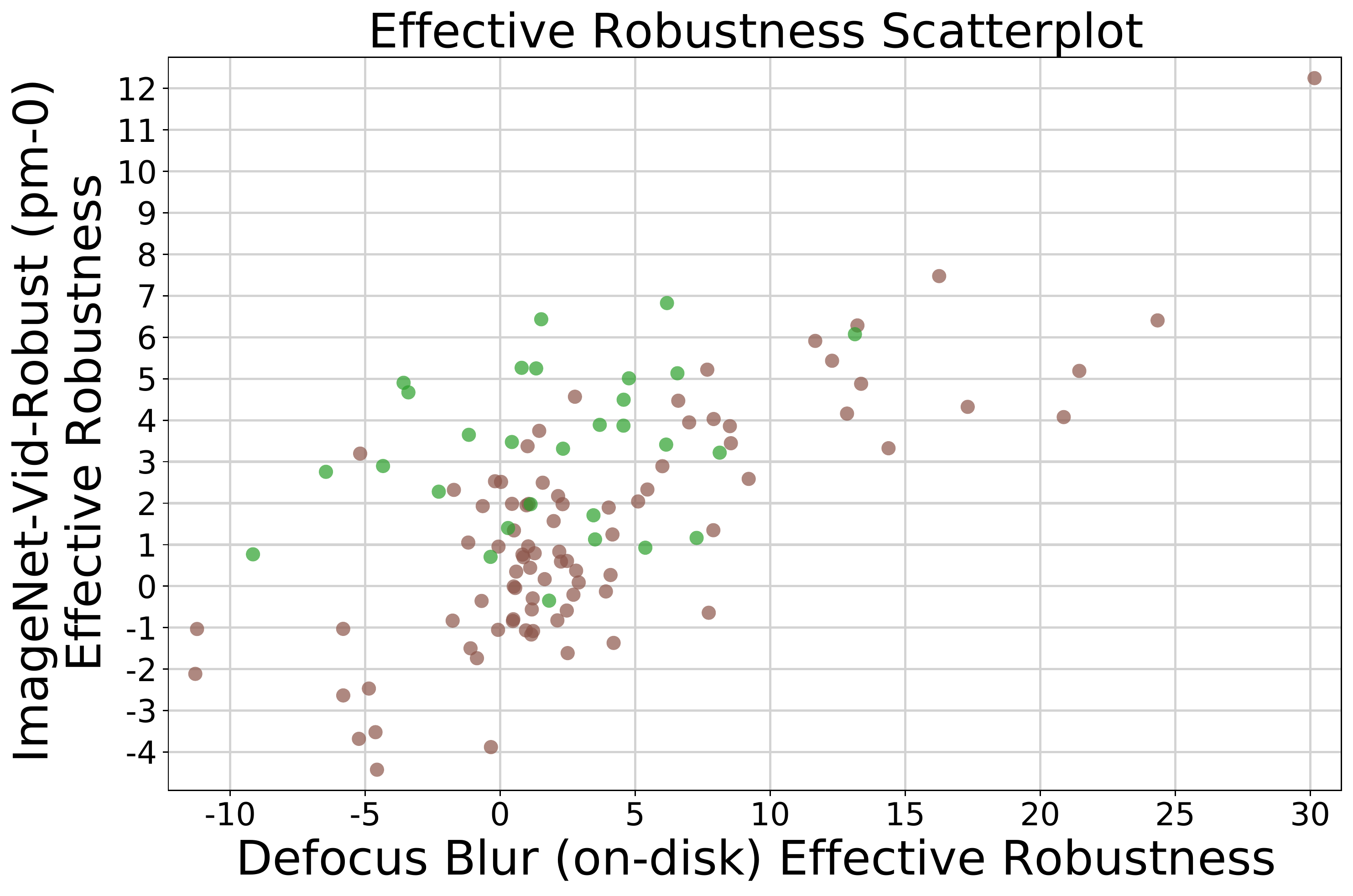}
  \end{subfigure}
  \begin{subfigure}[t]{\textwidth}
    \centering \includegraphics[width=0.7\textwidth]{figs/syn_shift_legend}
  \end{subfigure}
  \caption{
    Plots of the two synthetic distribution shifts with the highest correlation with \imagenetvidrobust{} pm-0, compared similarly to \Cref{fig:synthetic-shifts}.
    }
  \label{fig:topk-correlation-vidrobust-benign}
\end{figure*}

\begin{figure*}[h]
    \centering
  \begin{subfigure}[t]{0.45\textwidth}
    \includegraphics[width=\linewidth]{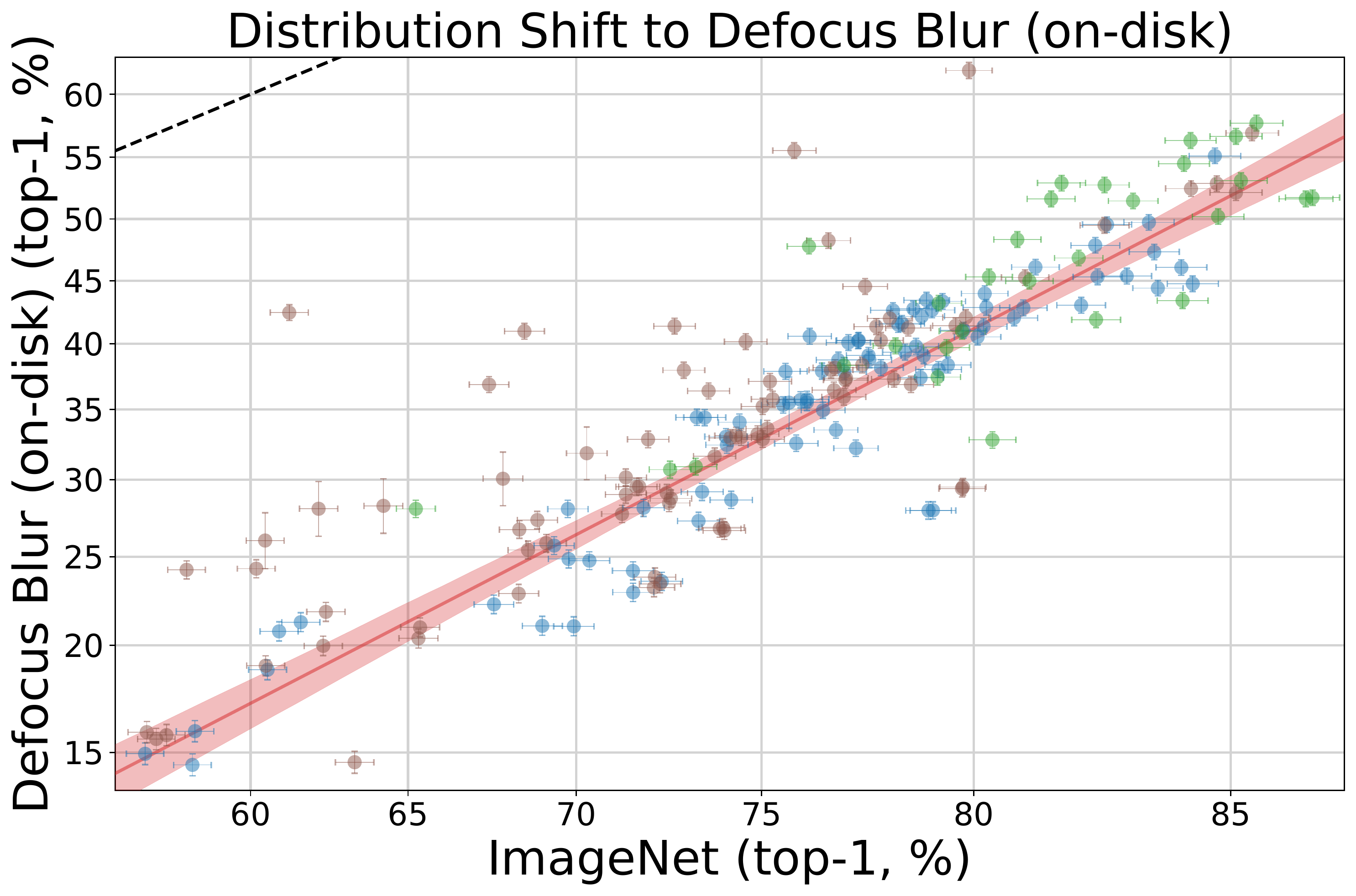}
  \end{subfigure}
    \hspace{0.5cm}
  \begin{subfigure}[t]{0.45\textwidth}
    \includegraphics[width=\linewidth]{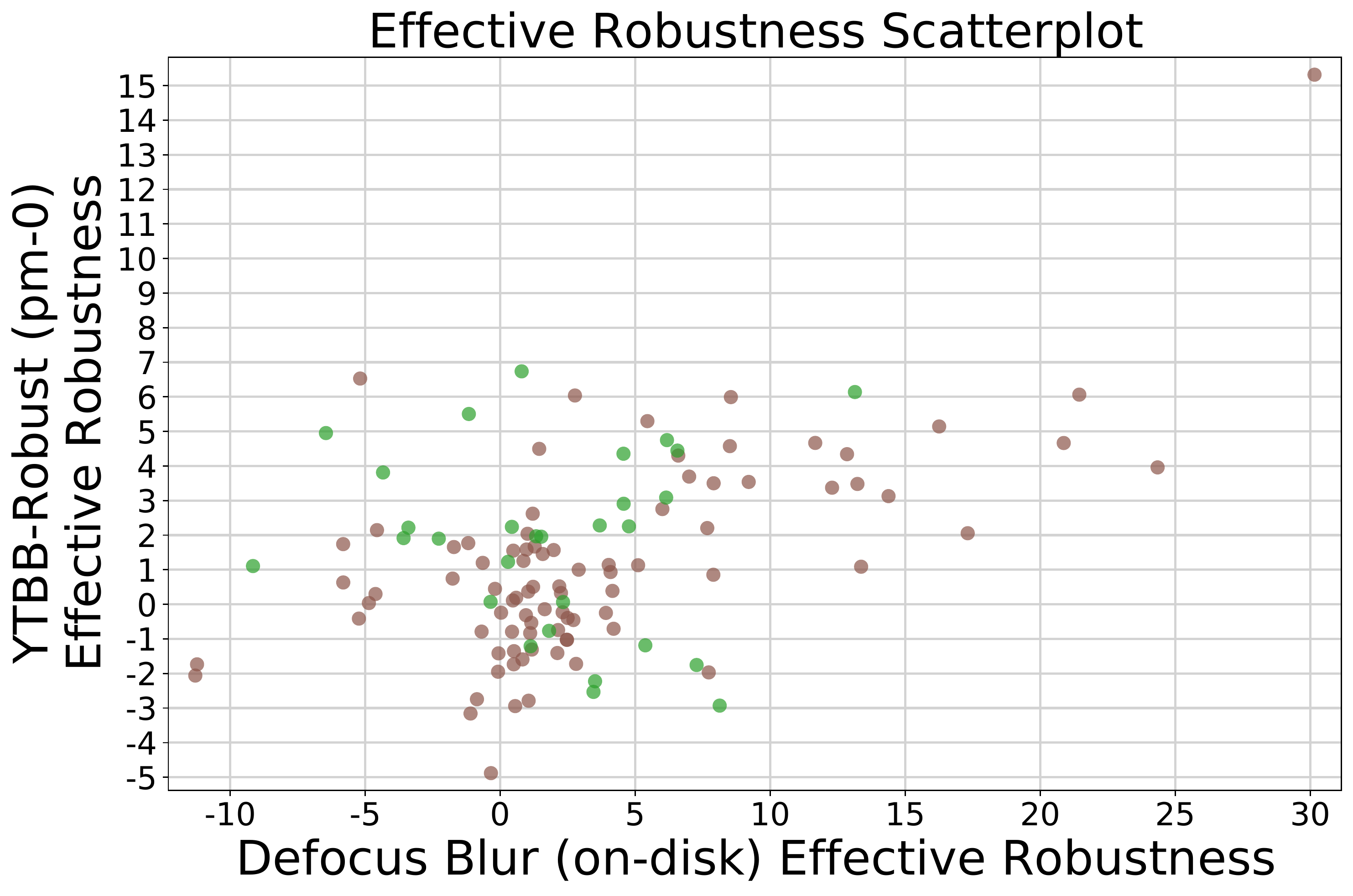}
  \end{subfigure}
  \begin{subfigure}[t]{0.45\textwidth}
    \includegraphics[width=\linewidth]{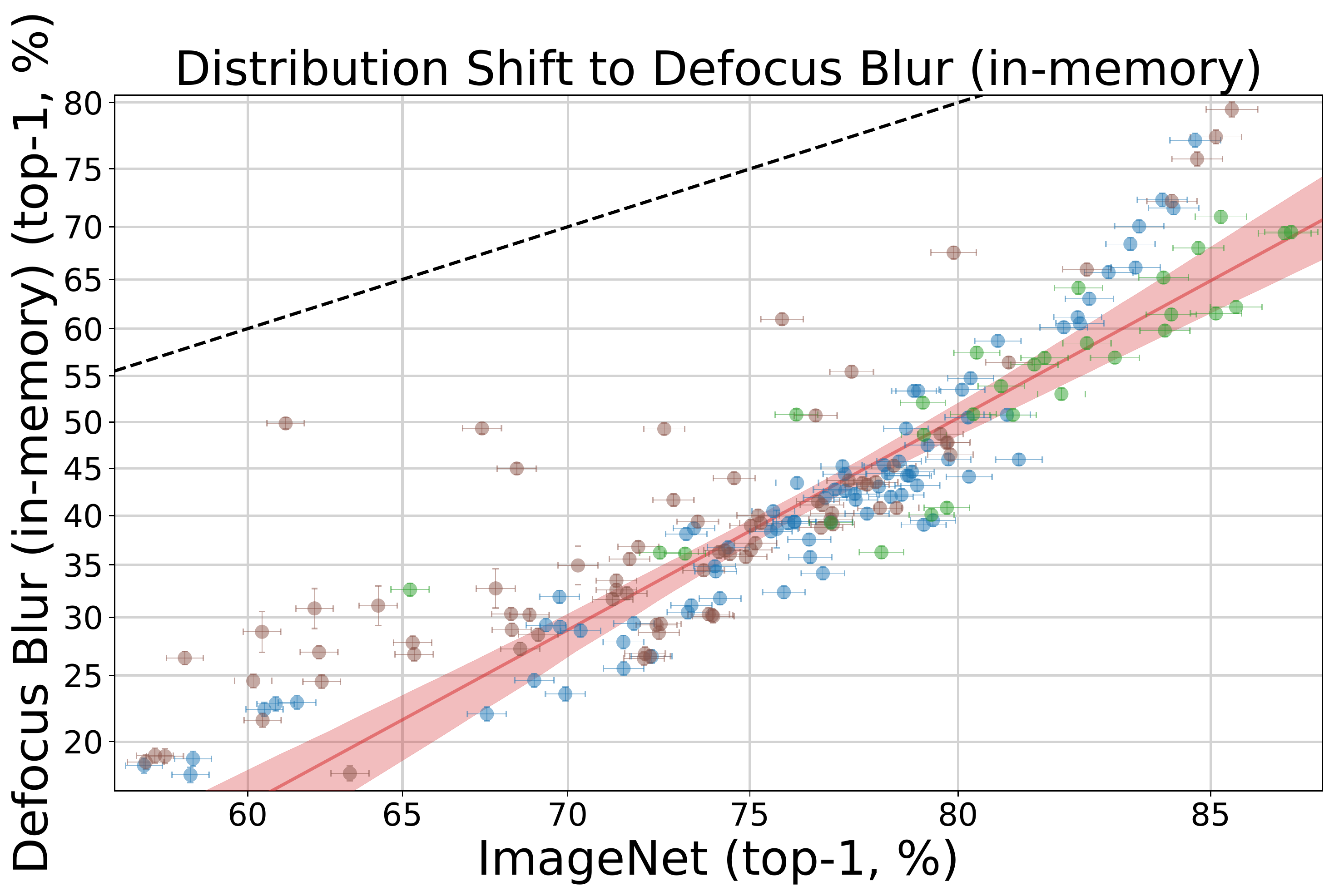}
  \end{subfigure}
    \hspace{0.5cm}
  \begin{subfigure}[t]{0.45\textwidth}
    \includegraphics[width=\linewidth]{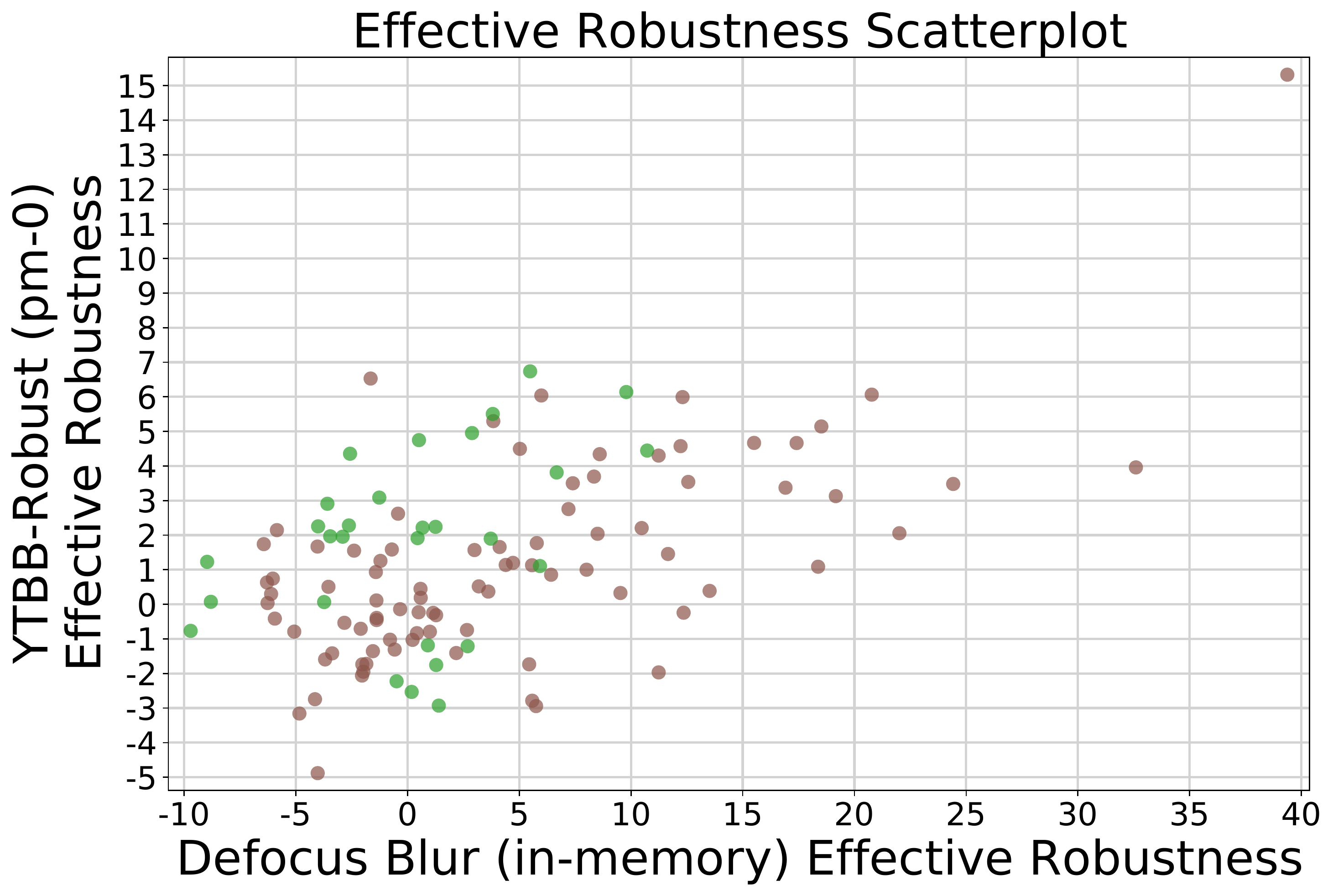}
  \end{subfigure}
  \begin{subfigure}[t]{\textwidth}
    \centering \includegraphics[width=0.7\textwidth]{figs/syn_shift_legend}
  \end{subfigure}
  \caption{
    Plots of the two synthetic distribution shifts with the highest correlation with \ytbbrobust{} pm-0, compared similarly to \Cref{fig:synthetic-shifts}.
    }
  \label{fig:topk-correlation-ytbb-benign}
\end{figure*}

\begin{figure*}[h]
    \centering
  \begin{subfigure}[t]{0.45\textwidth}
    \includegraphics[width=\linewidth]{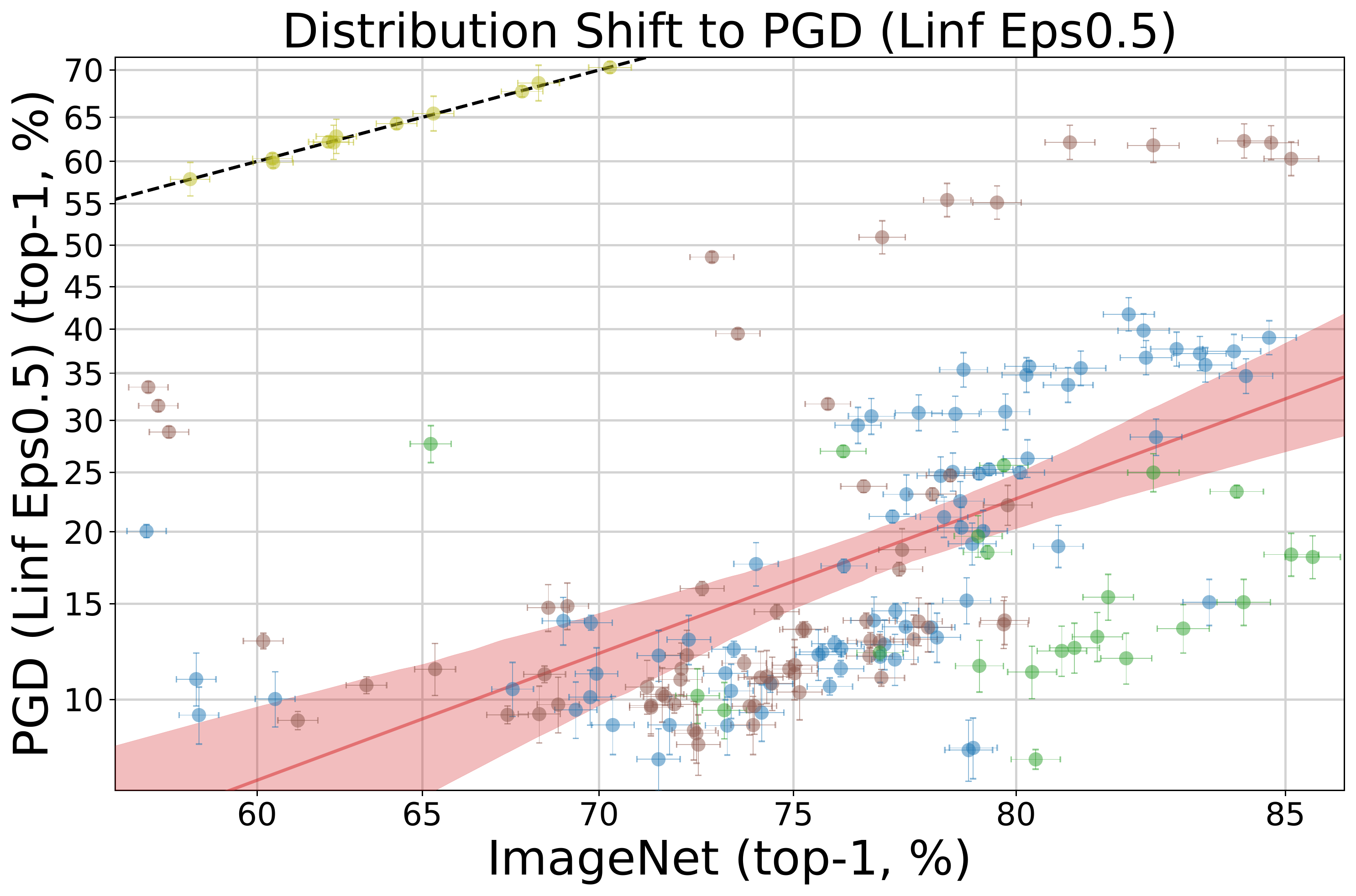}
  \end{subfigure}
    \hspace{0.5cm}
  \begin{subfigure}[t]{0.45\textwidth}
    \includegraphics[width=\linewidth]{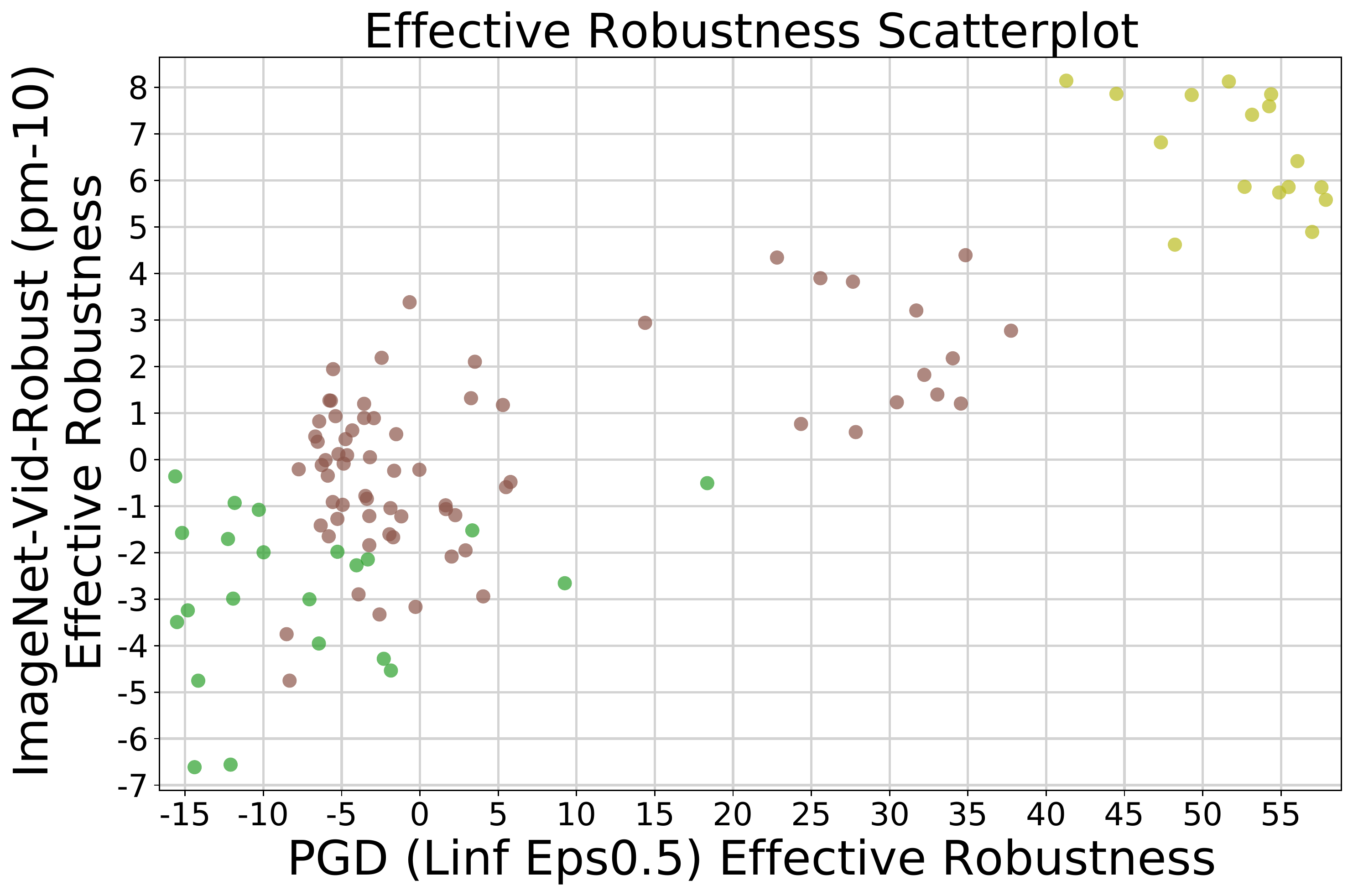}
  \end{subfigure}
  \begin{subfigure}[t]{0.45\textwidth}
    \includegraphics[width=\linewidth]{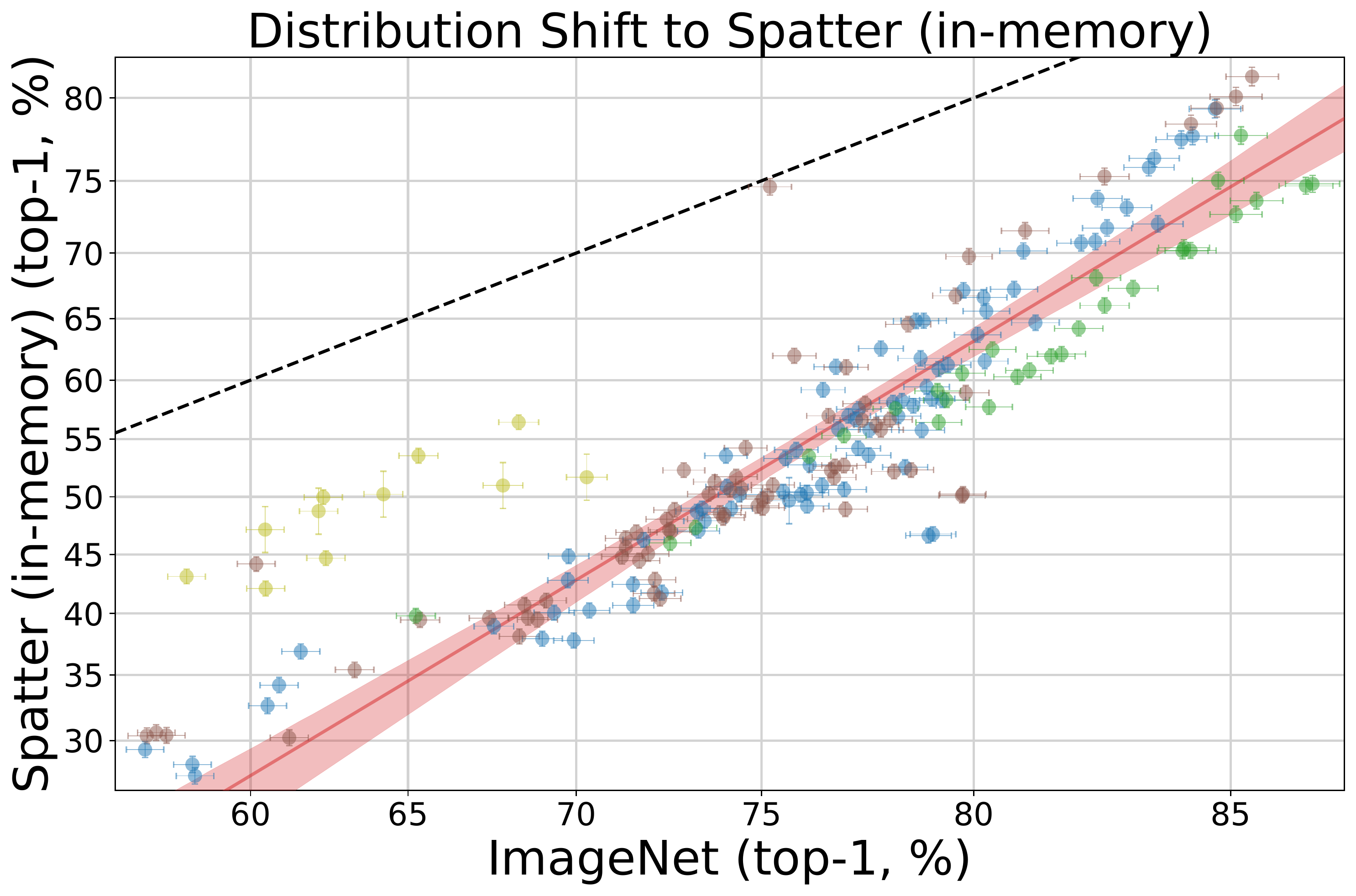}
  \end{subfigure}
    \hspace{0.5cm}
  \begin{subfigure}[t]{0.45\textwidth}
    \includegraphics[width=\linewidth]{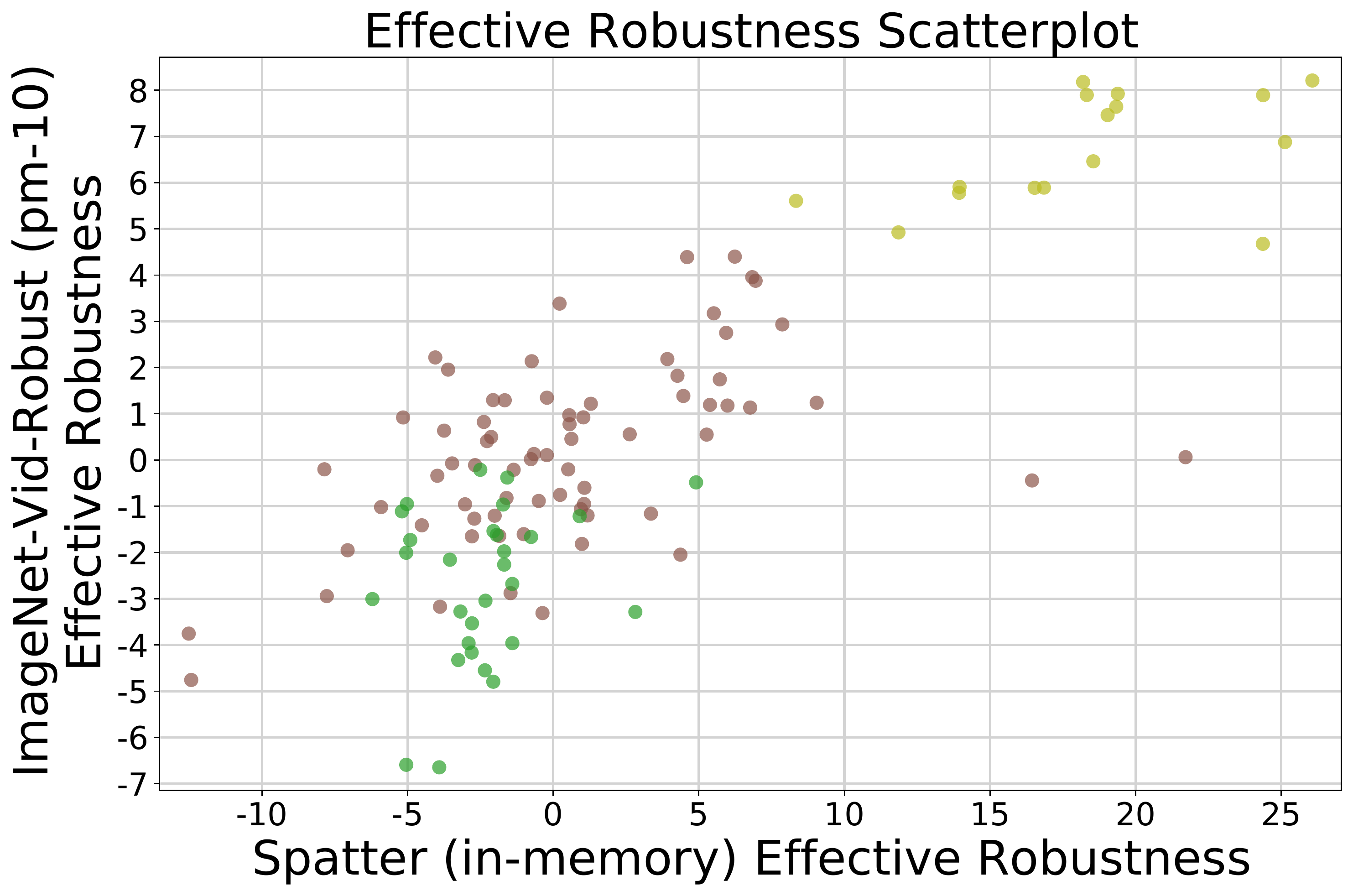}
  \end{subfigure}
  \begin{subfigure}[t]{\textwidth}
    \centering \includegraphics[width=0.8\textwidth]{figs/consistency_legend}
  \end{subfigure}
  \caption{
    Plots of the two synthetic distribution shifts with the highest correlation with \imagenetvidrobust{} pm-10, compared similarly to \Cref{fig:synthetic-shifts}.
    }
  \label{fig:topk-correlation-vidrobust-pmk}
\end{figure*}

\begin{figure*}[h]
    \centering
  \begin{subfigure}[t]{0.45\textwidth}
    \includegraphics[width=\linewidth]{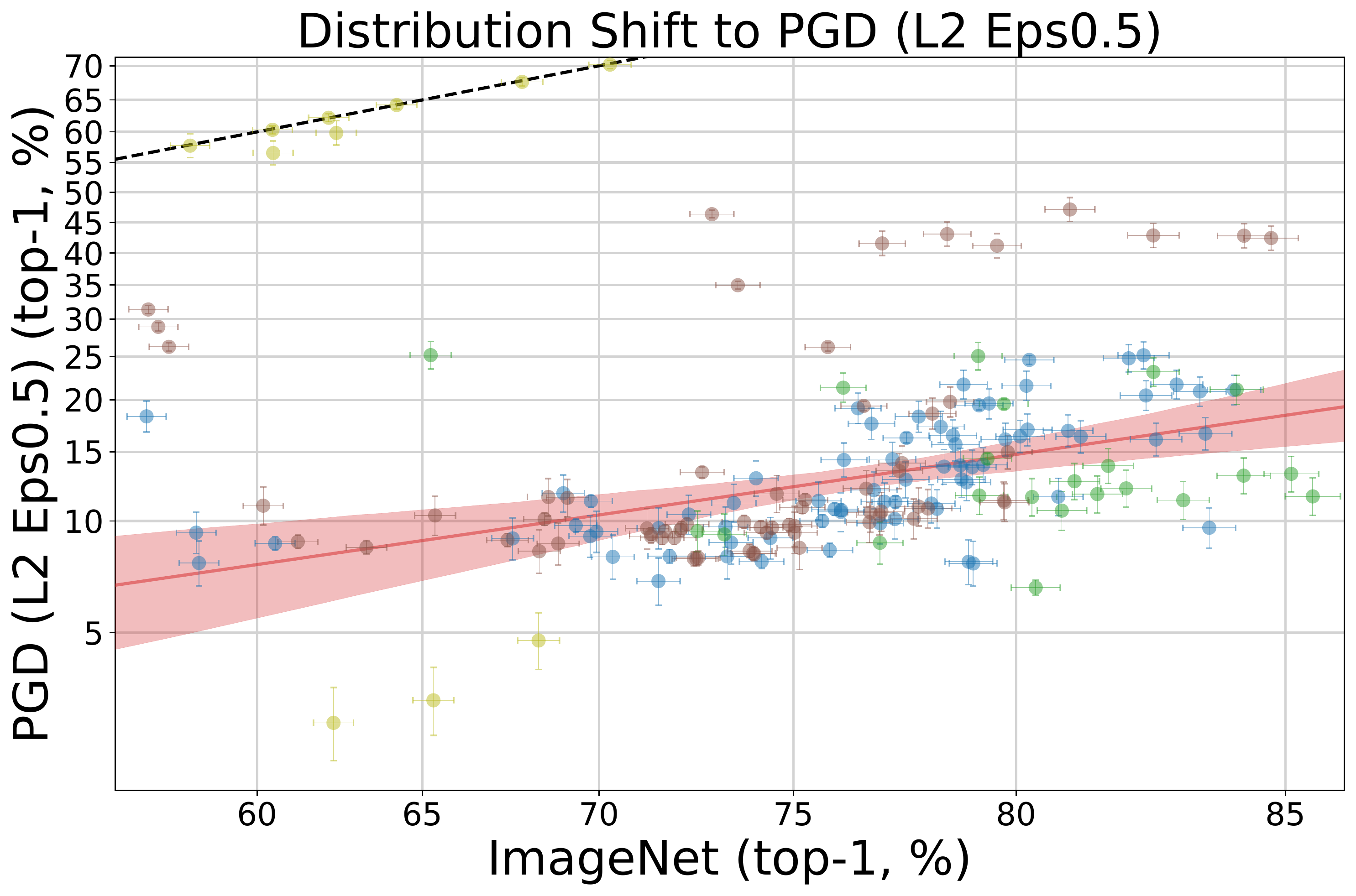}
  \end{subfigure}
    \hspace{0.5cm}
  \begin{subfigure}[t]{0.45\textwidth}
    \includegraphics[width=\linewidth]{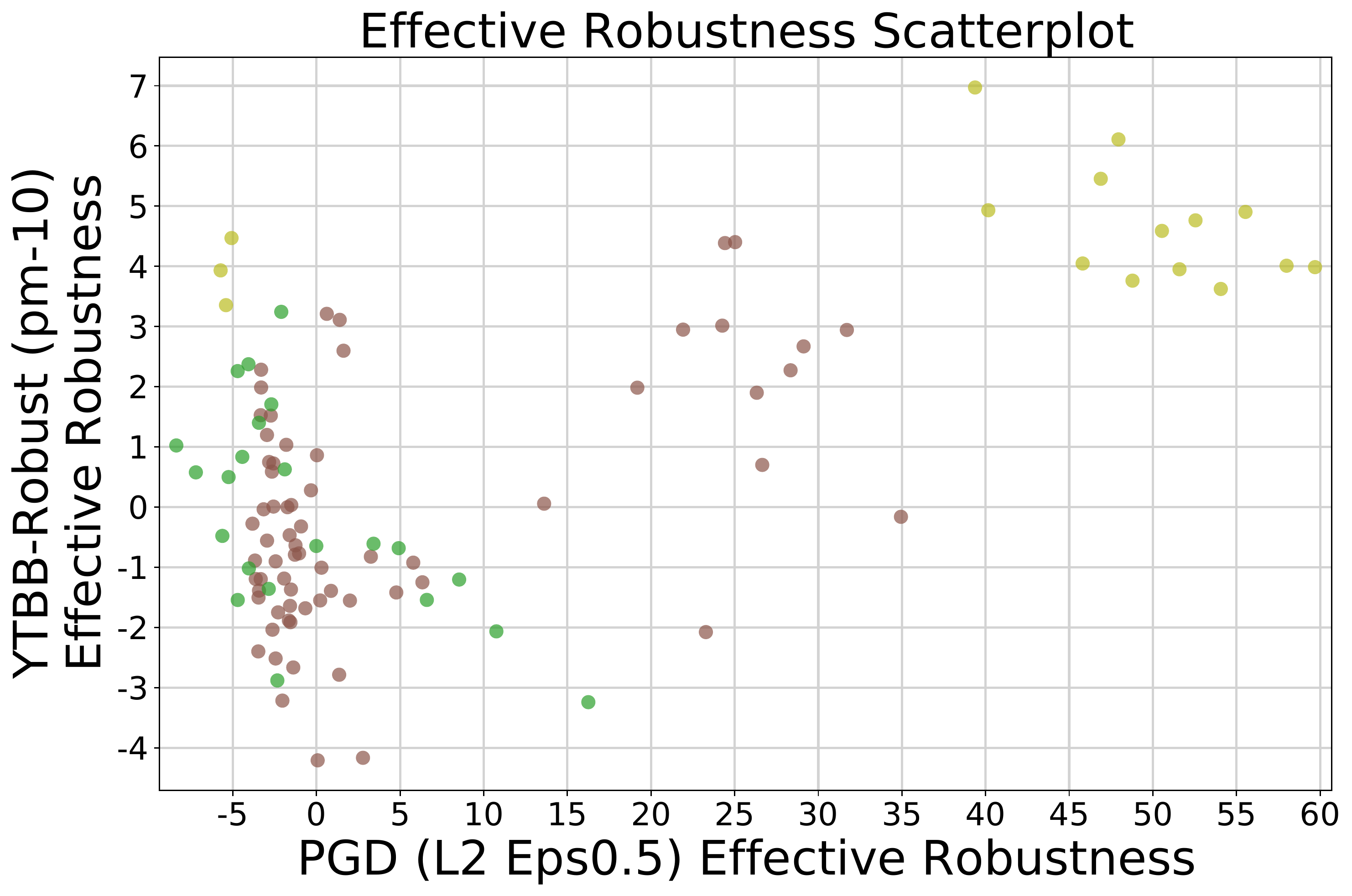}
  \end{subfigure}
  \begin{subfigure}[t]{0.45\textwidth}
    \includegraphics[width=\linewidth]{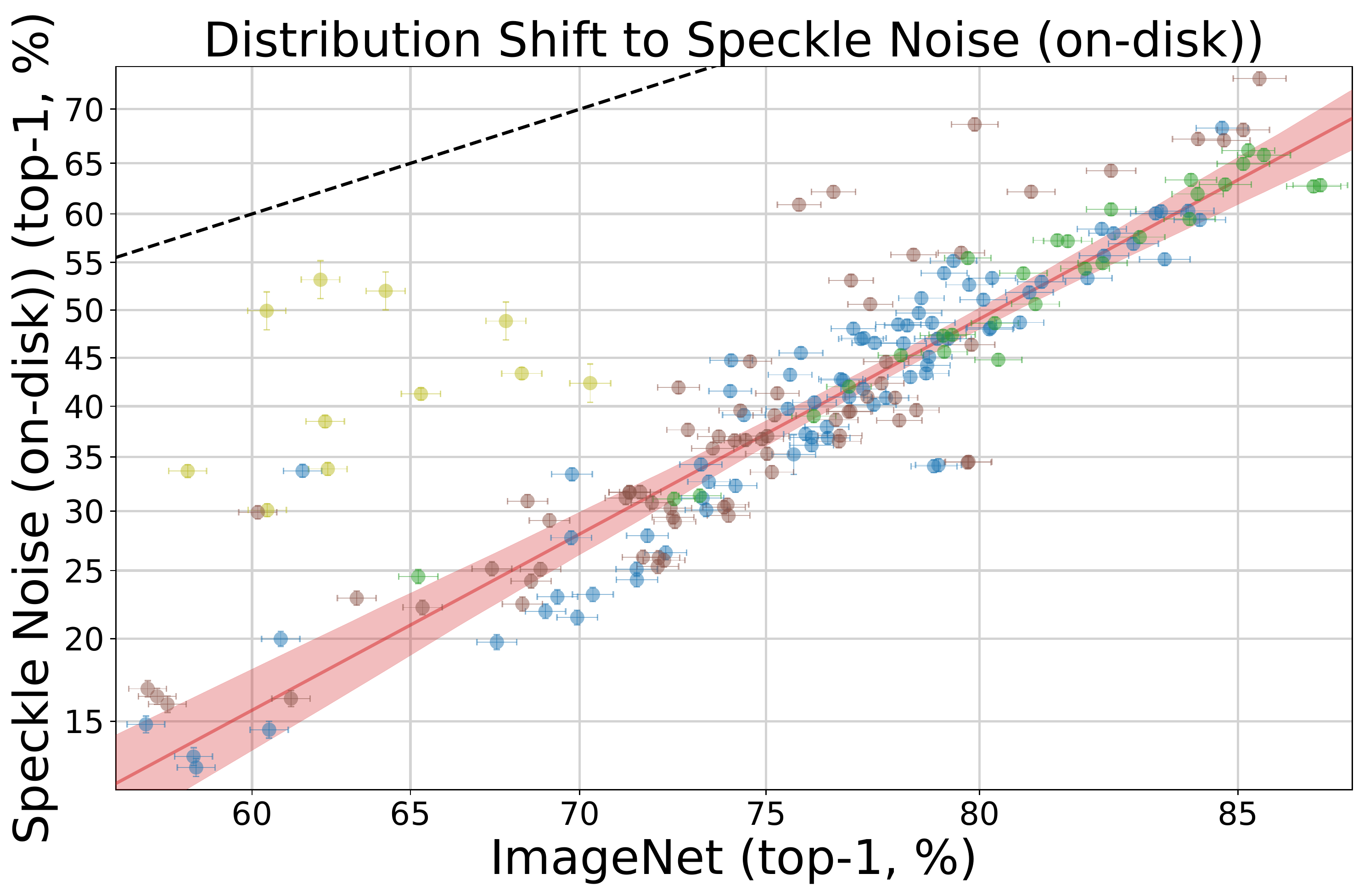}
  \end{subfigure}
    \hspace{0.5cm}
  \begin{subfigure}[t]{0.45\textwidth}
    \includegraphics[width=\linewidth]{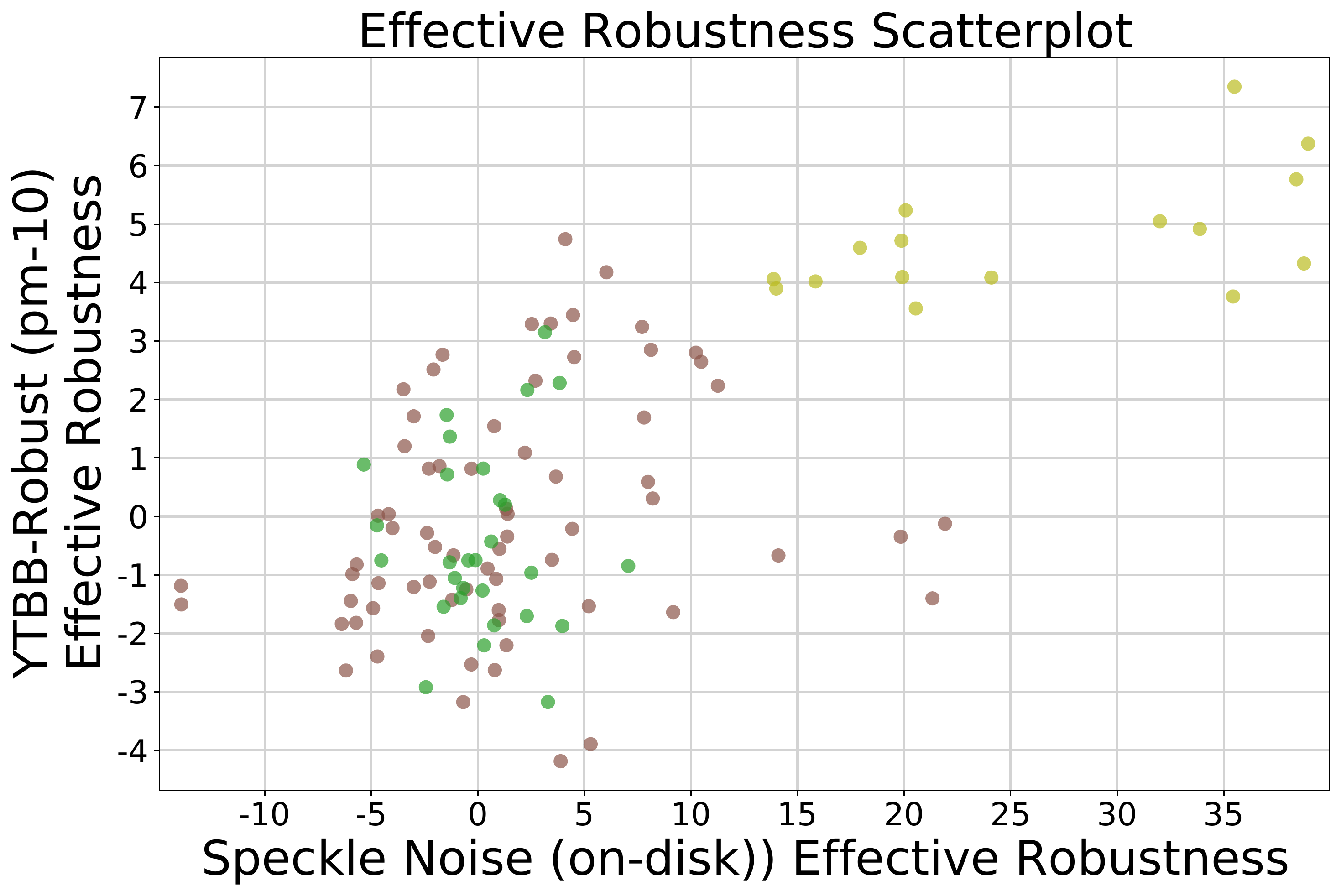}
  \end{subfigure}
  \begin{subfigure}[t]{\textwidth}
    \centering \includegraphics[width=0.8\textwidth]{figs/consistency_legend}
  \end{subfigure}
  \caption{
    Plots of the two synthetic distribution shifts with the highest correlation with \ytbbrobust{} pm-10, compared similarly to \Cref{fig:synthetic-shifts}.
    }
  \label{fig:topk-correlation-ytbb-pmk}
\end{figure*}

\begin{figure*}[h]
    \centering
  \begin{subfigure}[t]{0.45\textwidth}
    \includegraphics[width=\linewidth]{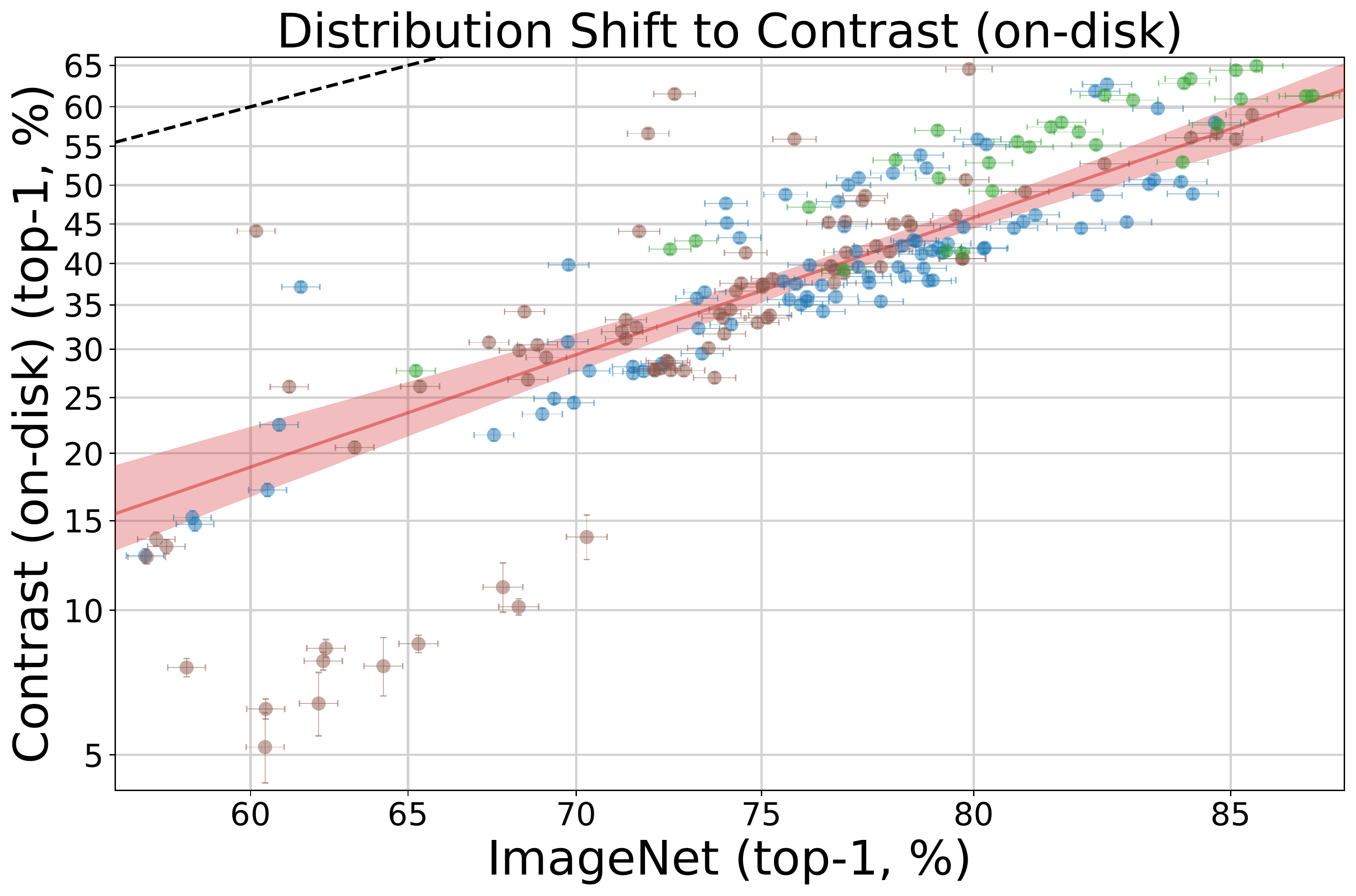}
  \end{subfigure}
    \hspace{0.5cm}
  \begin{subfigure}[t]{0.45\textwidth}
    \includegraphics[width=\linewidth]{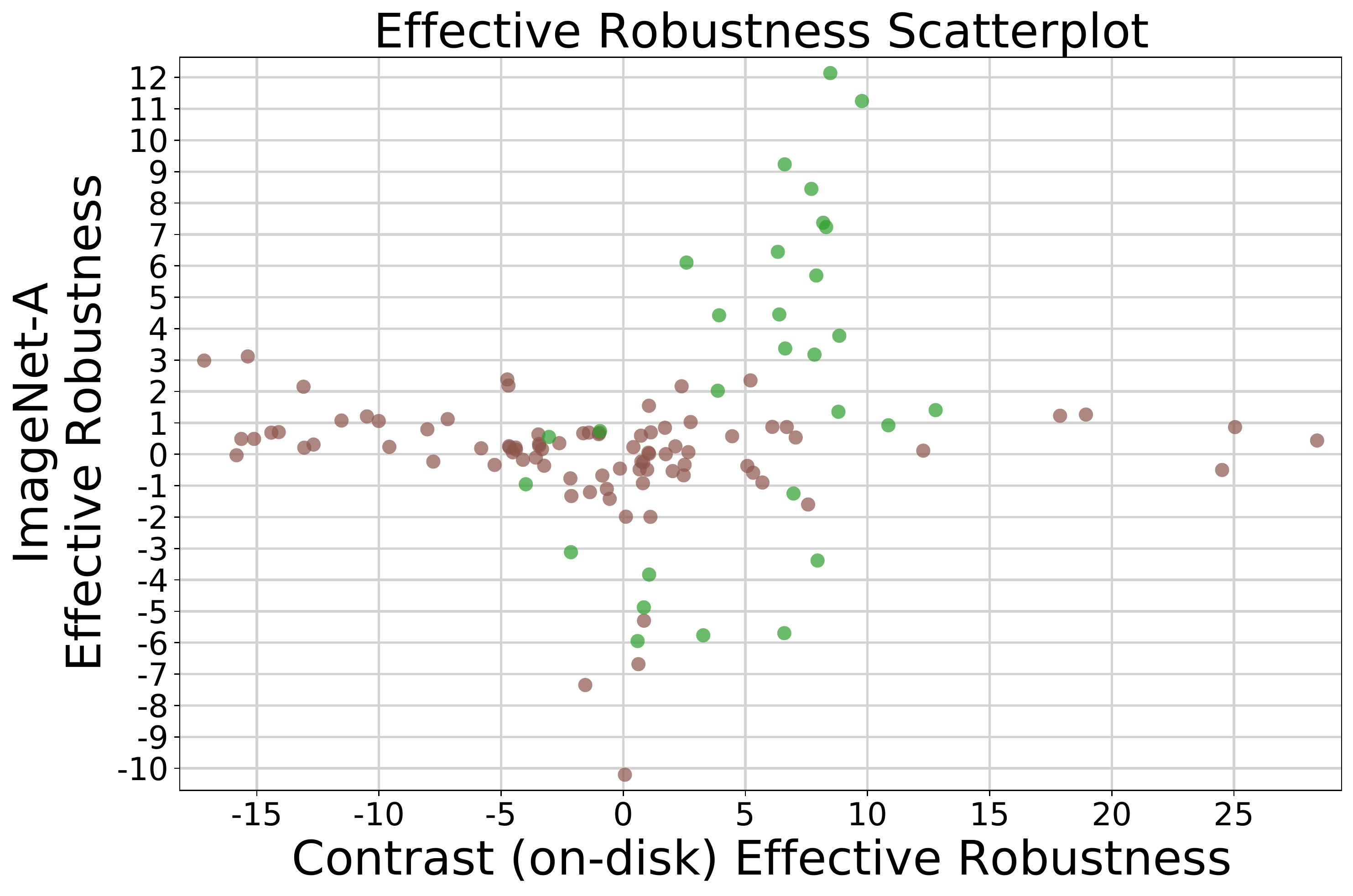}
  \end{subfigure}
  \begin{subfigure}[t]{0.45\textwidth}
    \includegraphics[width=\linewidth]{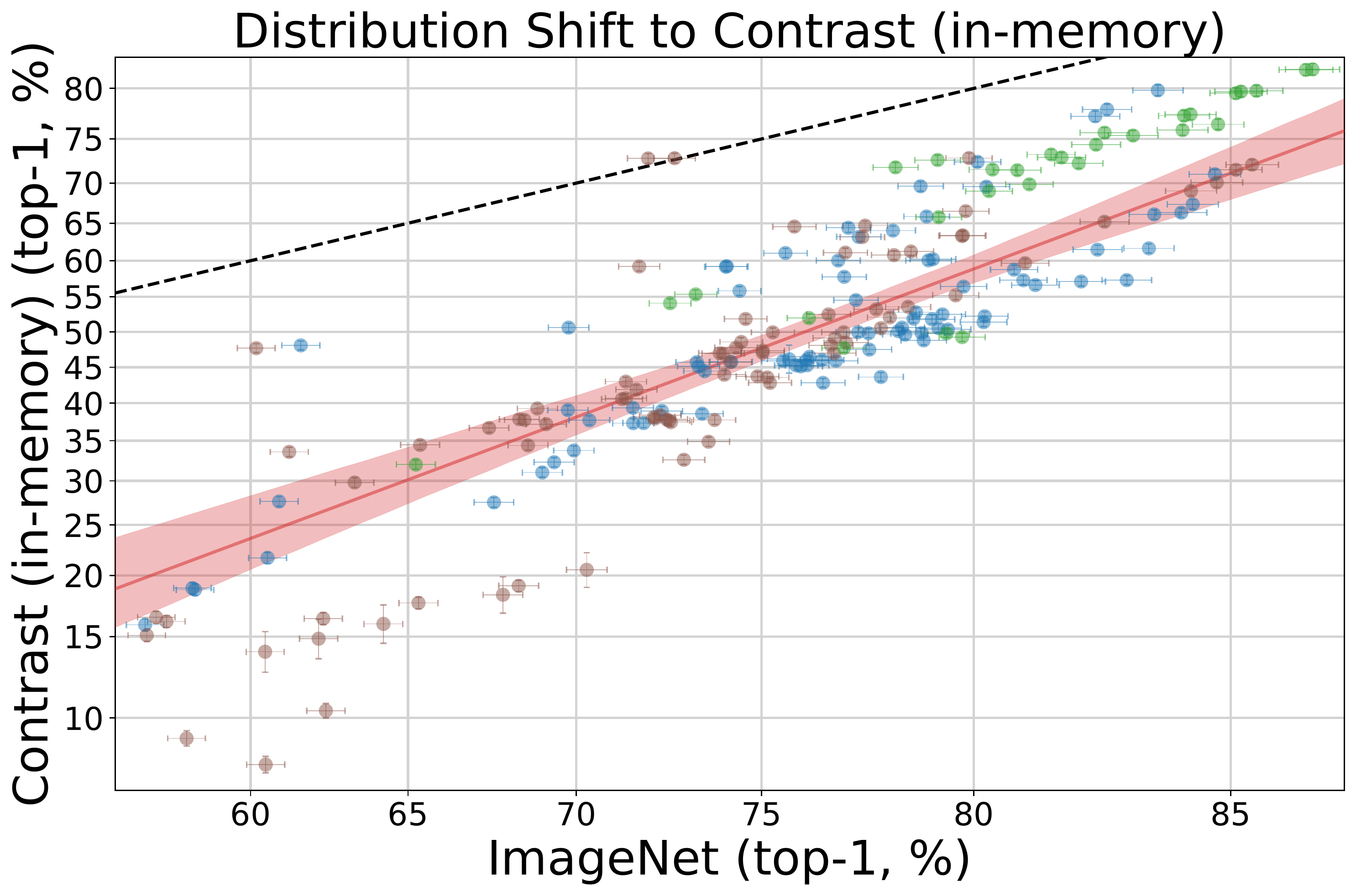}
  \end{subfigure}
    \hspace{0.5cm}
  \begin{subfigure}[t]{0.45\textwidth}
    \includegraphics[width=\linewidth]{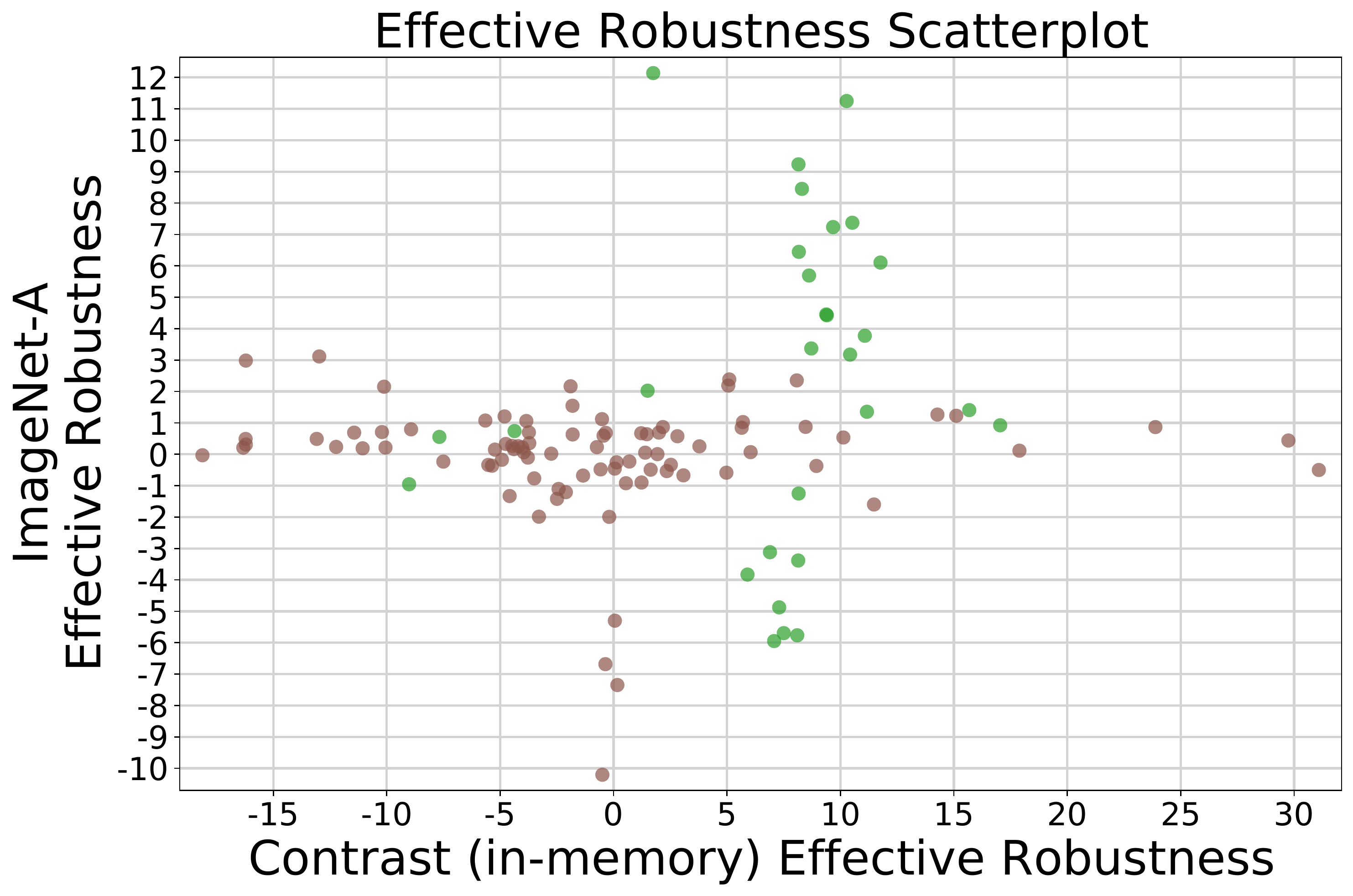}
  \end{subfigure}
  \begin{subfigure}[t]{\textwidth}
    \centering \includegraphics[width=0.7\textwidth]{figs/syn_shift_legend}
  \end{subfigure}
  \caption{
    Plots of the two synthetic distribution shifts with the highest correlation with \imageneta{}, compared similarly to \Cref{fig:synthetic-shifts}.
    }
  \label{fig:topk-correlation-imageneta}
\end{figure*}

Table \ref{tab:eff_robust_corr_table} provides a full list of the correlation numbers, and \Cref{fig:topk-correlation-imagenetv2,fig:topk-correlation-objectnet,fig:topk-correlation-vidrobust-benign,fig:topk-correlation-ytbb-benign,fig:topk-correlation-vidrobust-pmk,fig:topk-correlation-ytbb-pmk,fig:topk-correlation-imageneta} show scatter plots of the two highest correlated synthetic shifts for each natural distribution shift.
We find that some of the synthetic shifts are more predictive than others, but none have high correlation with all of the natural shifts.
For instance, $\ell_p$-robustness has the highest correlation with consistency shifts, but only low correlation with dataset shifts.
On the other hand, some image corruptions such as brightness, gaussian blur, defocus blur, and saturate have higher correlation with the dataset shifts.
It is worth nothing our testbed indicates that these synthetic measures are not causal, i.e., models trained on brightness, gaussian blur, defocus blur, or saturate do not have significant positive effective robustness on dataset shifts.
Further analyzing these fine-grained connections between synthetic and natural forms of distribution shift is an important direction for future work.

\begin{footnotesize}
\rowcolors{1}{gray!15}{white}
\begin{longtable}[ht!]{E T T T T T T T}
  \hiderowcolors
  \caption{
  Pearson correlation coefficients between all synthetic and natural distribution shifts in our testbed.
  For each distribution shift, effective robustness was calculated using a linear fit on the standard models.
  The correlation between synthetic and natural effective robustness was then only computed after filtering out the standard models.
  }
  \label{tab:eff_robust_corr_table}\\

  \hline
  \multicolumn{8}{c}{\textbf{Pearson correlation coefficients}}\\
  \hline
  Synthetic shift & \imagenetnew{} & \objectnet{} & \imagenetvid{} (pm-0) & \ytbb{} (pm-0) & \imagenetvid{} (pm-10) & \ytbb{} (pm-10) & \imageneta{} \\
  \hline
  \endfirsthead

  \hline
  \multicolumn{8}{c}{\textbf{Pearson correlation coefficients} (continued from previous page)}\\
  \hline
  Synthetic shift & \imagenetnew{} & \objectnet{} & \imagenetvid{} (pm-0) & \ytbb{} (pm-0) & \imagenetvid{} (pm-10) & \ytbb{} (pm-10) & \imageneta{} \\
  \hline
  \endhead

  \hline
  \multicolumn{8}{c}{Table continues onto next page}\\
  \hline
  \endfoot

  \hline
  \multicolumn{8}{c}{End of table}\\
  \hline\hline
  \endlastfoot
  \showrowcolors

  \input{appendix/eff_robust_corr_table}
\end{longtable}
\end{footnotesize}

\clearpage

\section{Information on our main figures}

\subsection{Constructing beta $\beta$}
\label{app:recipe-beta}

For each distribution shift, we construct the baseline accuracy function $\beta$ by analyzing the linear relationship between model performance on the original and shifted distributions. In particular, when constructing $\beta$ we only include "standard models," models that had not been designed with any robustness properties in mind or have not been trained on any data other than the standard 1,000-class ImageNet training set. Before constructing the predictor, model accuracies are then transformed according to the logit distribution; this transform assigns greater mass at the tails and experimentally provided the best linear fits. $\beta$ is then simply the linear predictor of the shifted distribution based on the independent variable (the original distribution), computed in this scaled space.

\subsection{Ablations on our main figures}
\label{app:ablations}

Here we provide various versions of the main figures in the main text. 
In each plot, we use logit scaling to demonstrate that gains in performance at higher accuracies represent greater progress.
The 95\% confidence intervals were empirically computed from the bootstrapped samples.
The bootstrapping was performed by computing 1,000 linear fits by sampling the models with replacement.

\begin{figure*}[ht!]
  \centering
  \begin{subfigure}[t]{0.4\textwidth}
    \includegraphics[width=\textwidth]{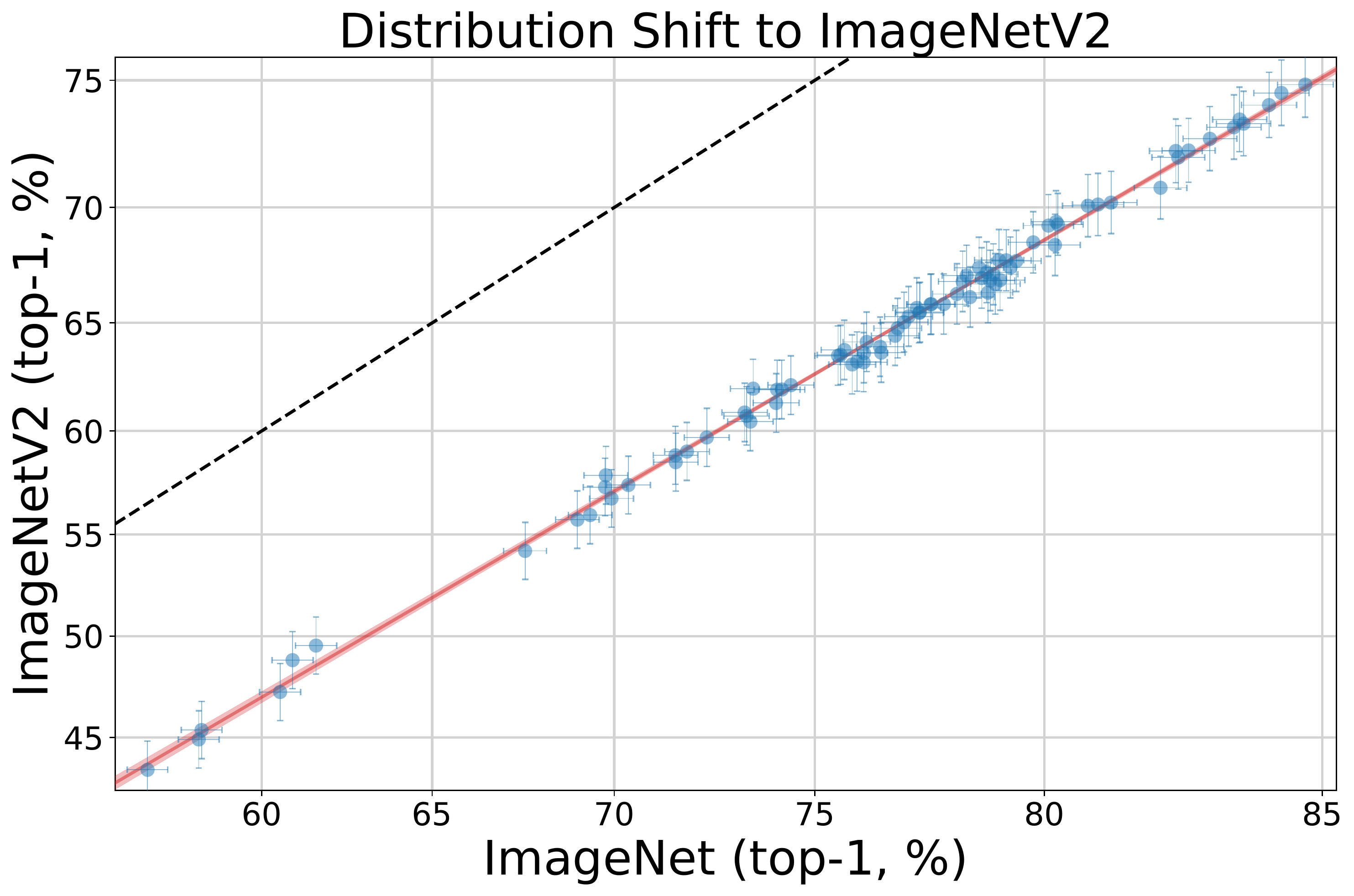}
  \end{subfigure}
  \hspace{0.5cm}
  \begin{subfigure}[t]{0.4\textwidth}
    \includegraphics[width=\textwidth]{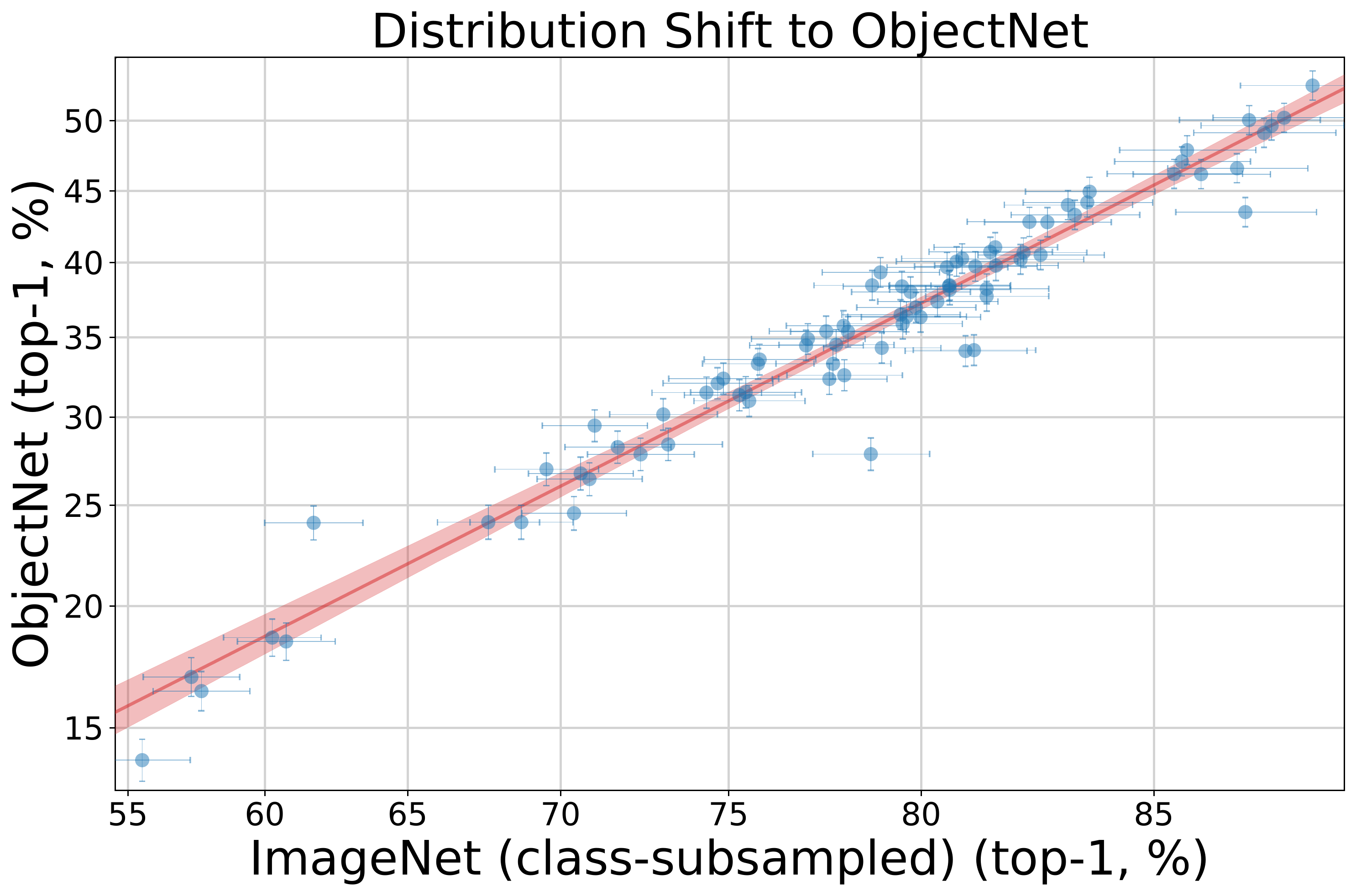}
  \end{subfigure}
  \par\medskip
  \begin{subfigure}[t]{0.4\textwidth}
    \includegraphics[width=\textwidth]{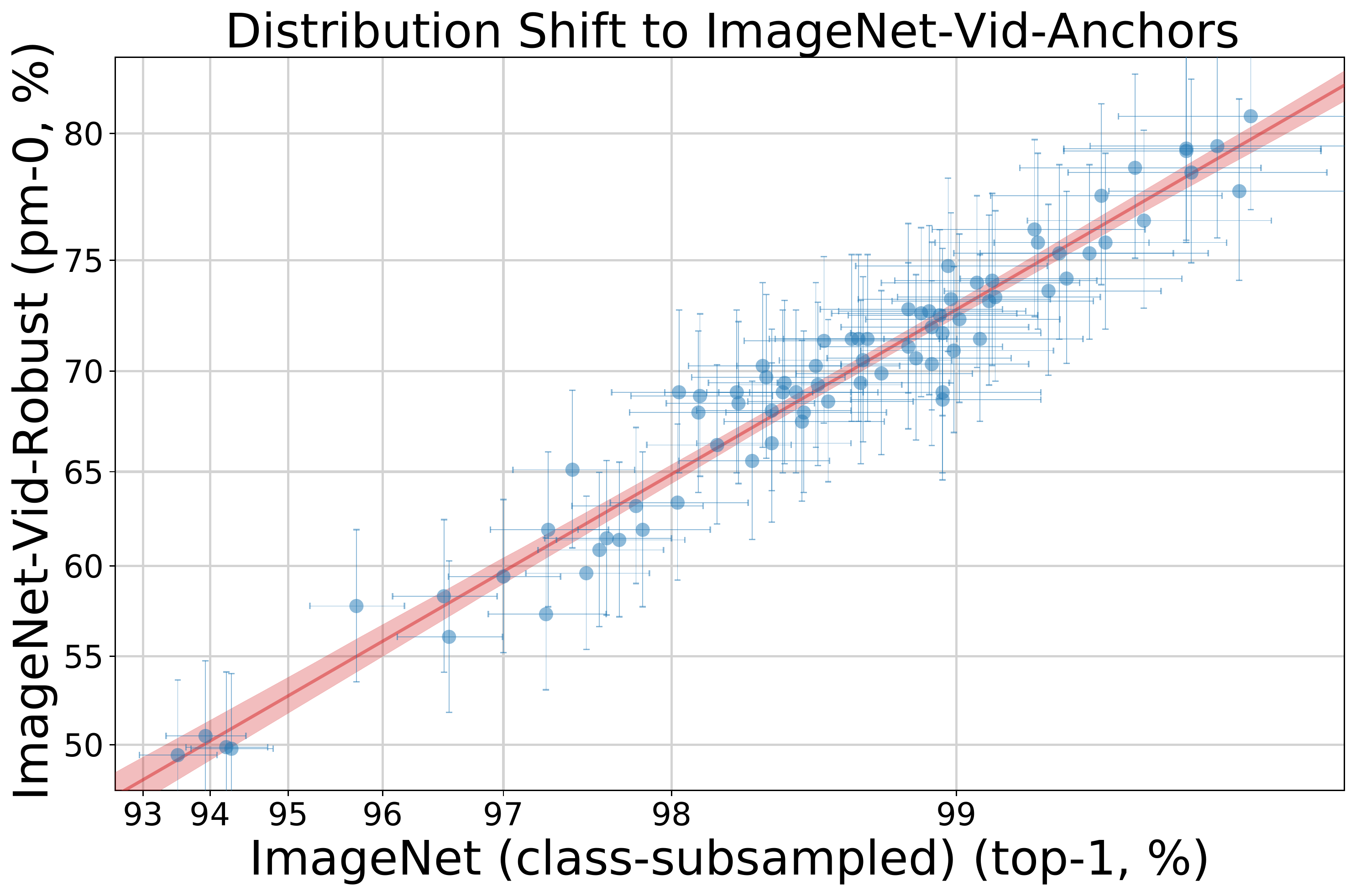}
  \end{subfigure}
  \hspace{0.5cm}
  \begin{subfigure}[t]{0.4\textwidth}
    \includegraphics[width=\textwidth]{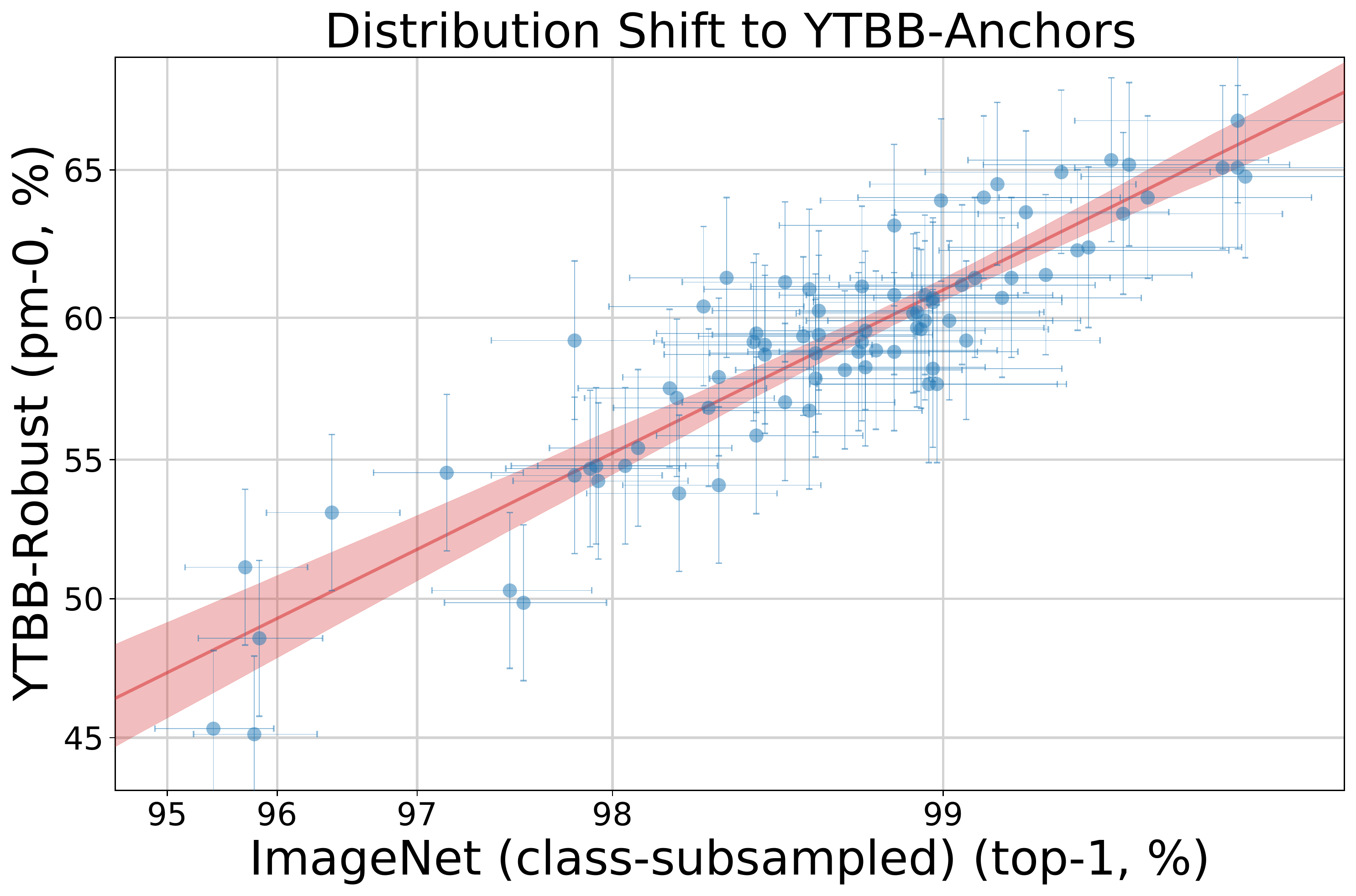}
  \end{subfigure}
  \par\medskip
  \begin{subfigure}[t]{0.4\textwidth}
    \includegraphics[width=\textwidth]{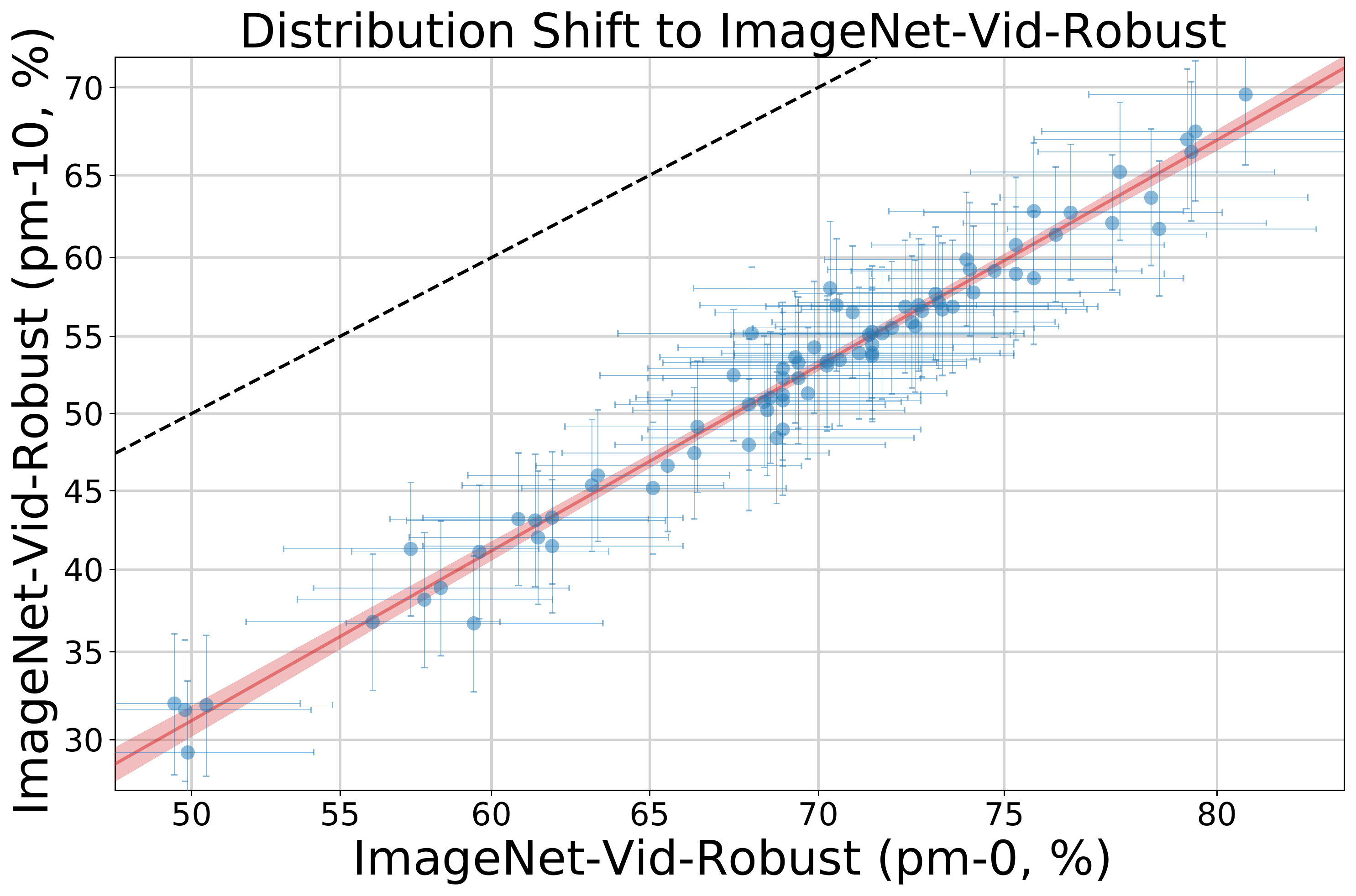}
  \end{subfigure}
  \hspace{0.5cm}
  \begin{subfigure}[t]{0.4\textwidth}
    \includegraphics[width=\textwidth]{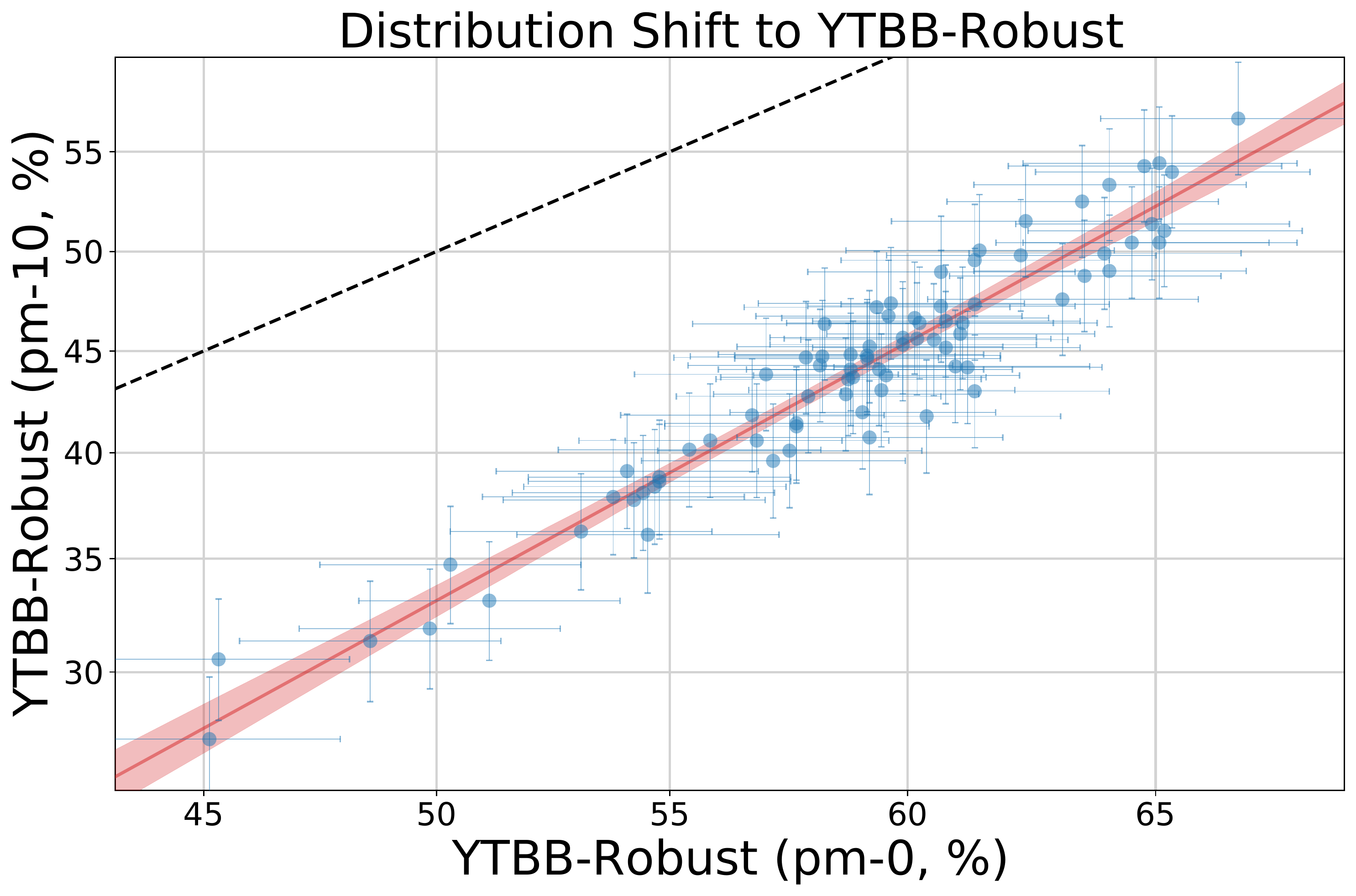}
  \end{subfigure}
  \par\medskip
  \begin{subfigure}[t]{0.4\textwidth}
    \includegraphics[width=\textwidth]{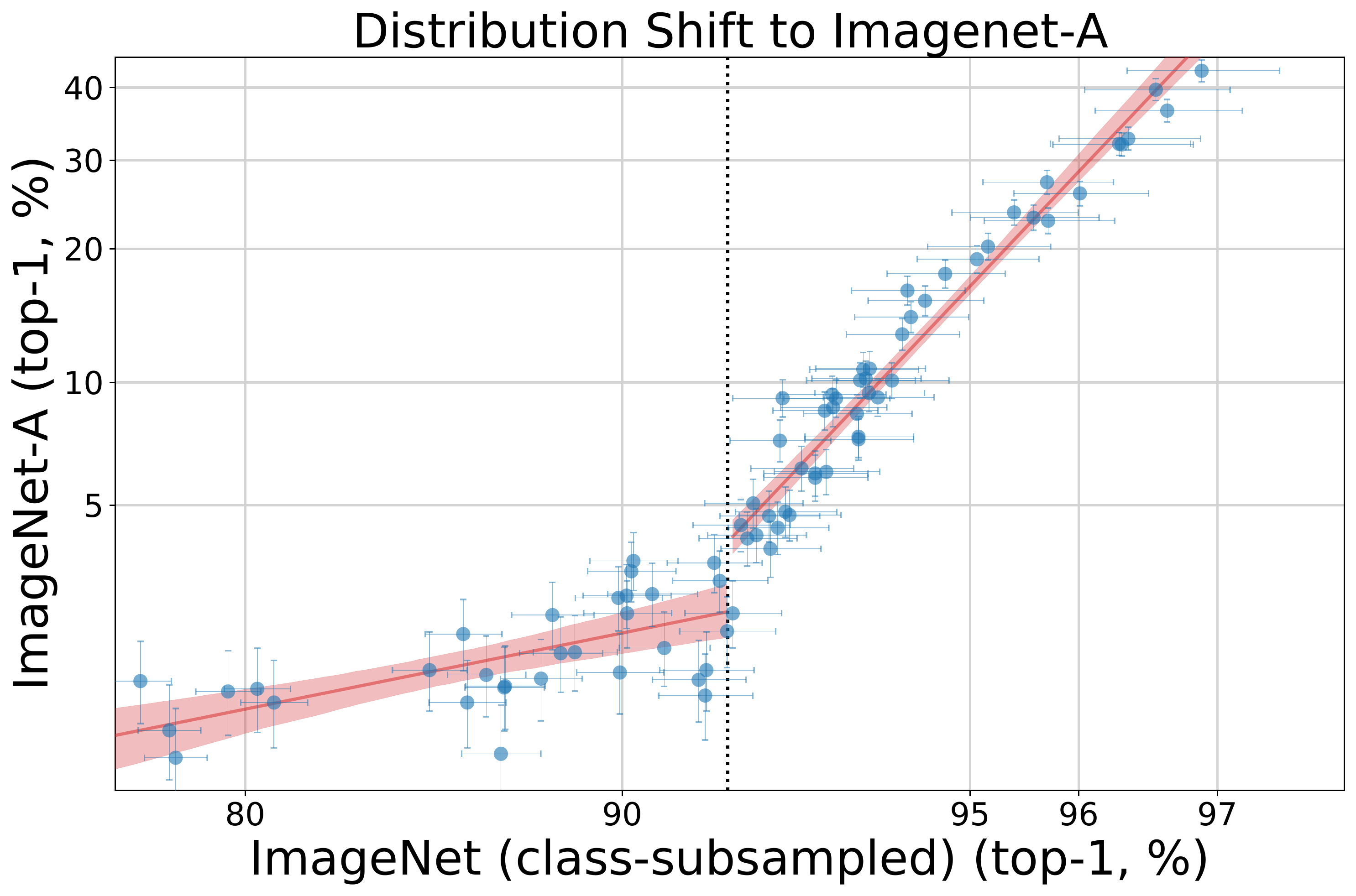}
  \end{subfigure}
  \begin{subfigure}[t]{\textwidth}
    \centering \includegraphics[width=0.5\textwidth]{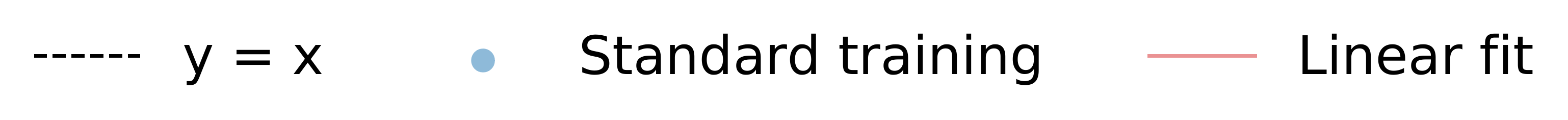}
  \end{subfigure}
  \caption{
    Only standard models are shown in these plots. 
    Otherwise, they are identical to the main plots in the main text.
    This is done to better illustrate the quality of the linear fit.
  }
  \label{fig:only-standard}
\end{figure*}

\begin{figure*}[ht!]
  \centering
  \begin{subfigure}[t]{0.4\textwidth}
    \includegraphics[width=\textwidth]{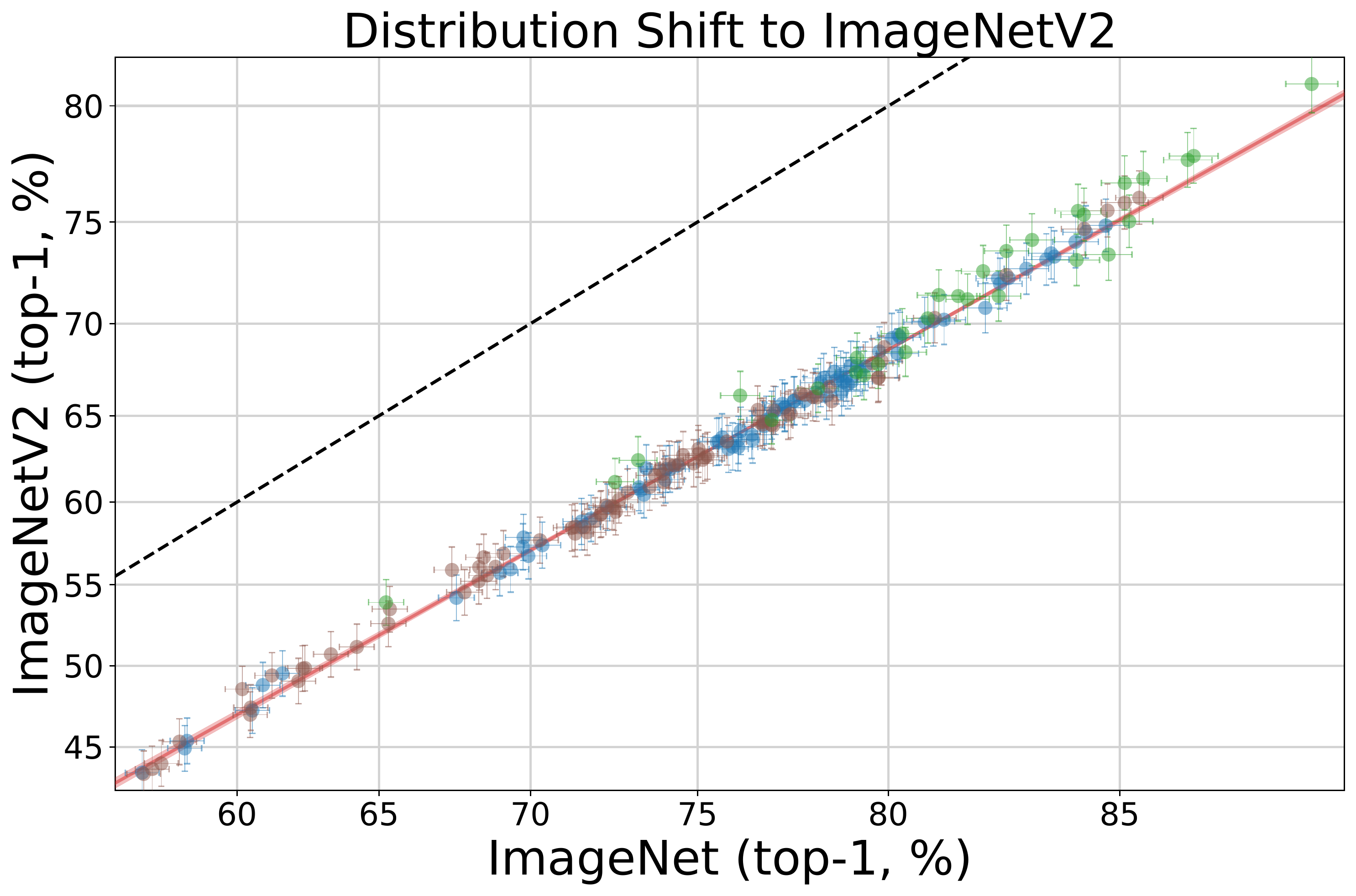}
  \end{subfigure}
  \hspace{0.5cm}
  \begin{subfigure}[t]{0.4\textwidth}
    \includegraphics[width=\textwidth]{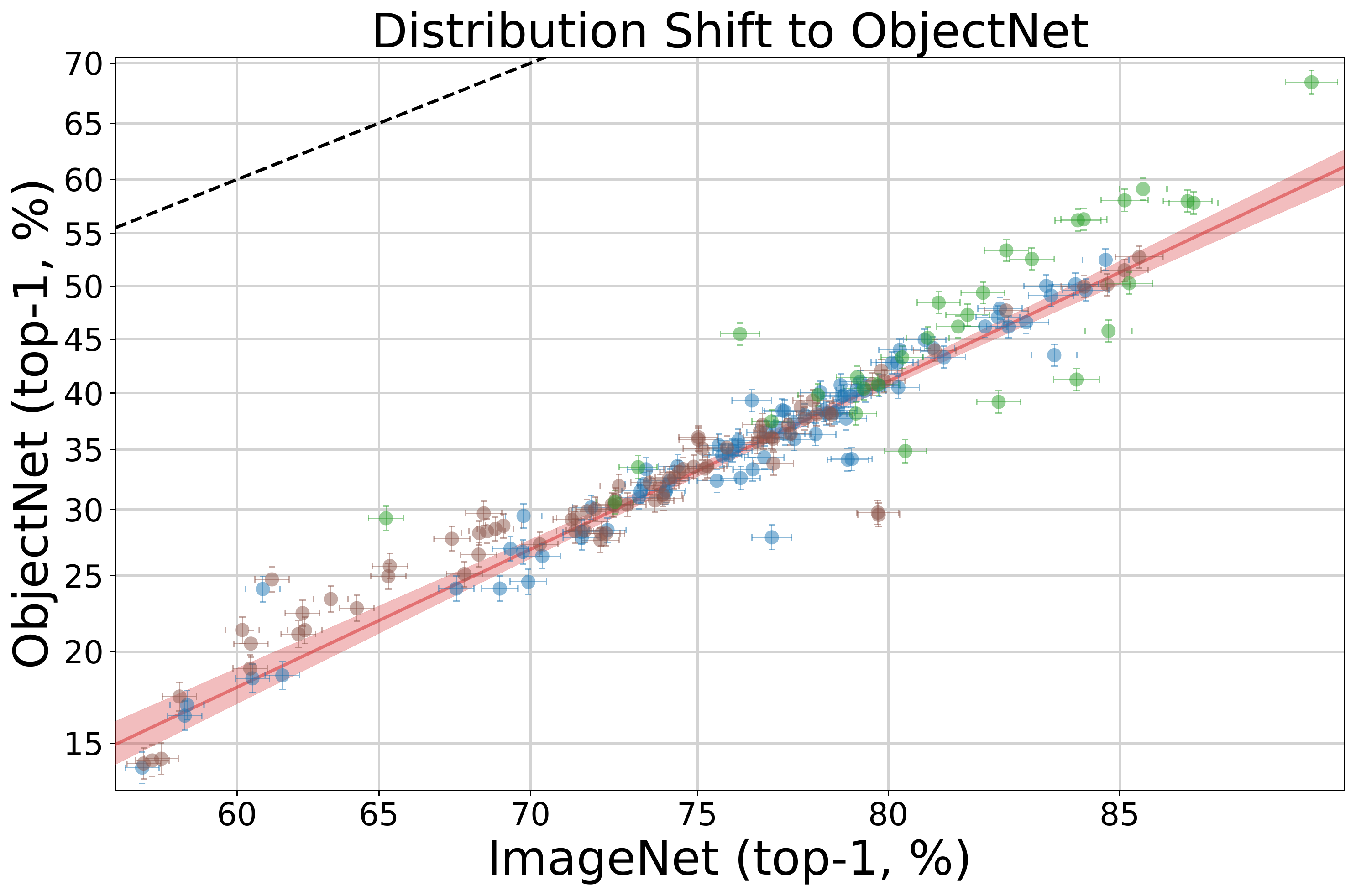}
  \end{subfigure}
  \par\medskip
  \begin{subfigure}[t]{0.4\textwidth}
    \includegraphics[width=\textwidth]{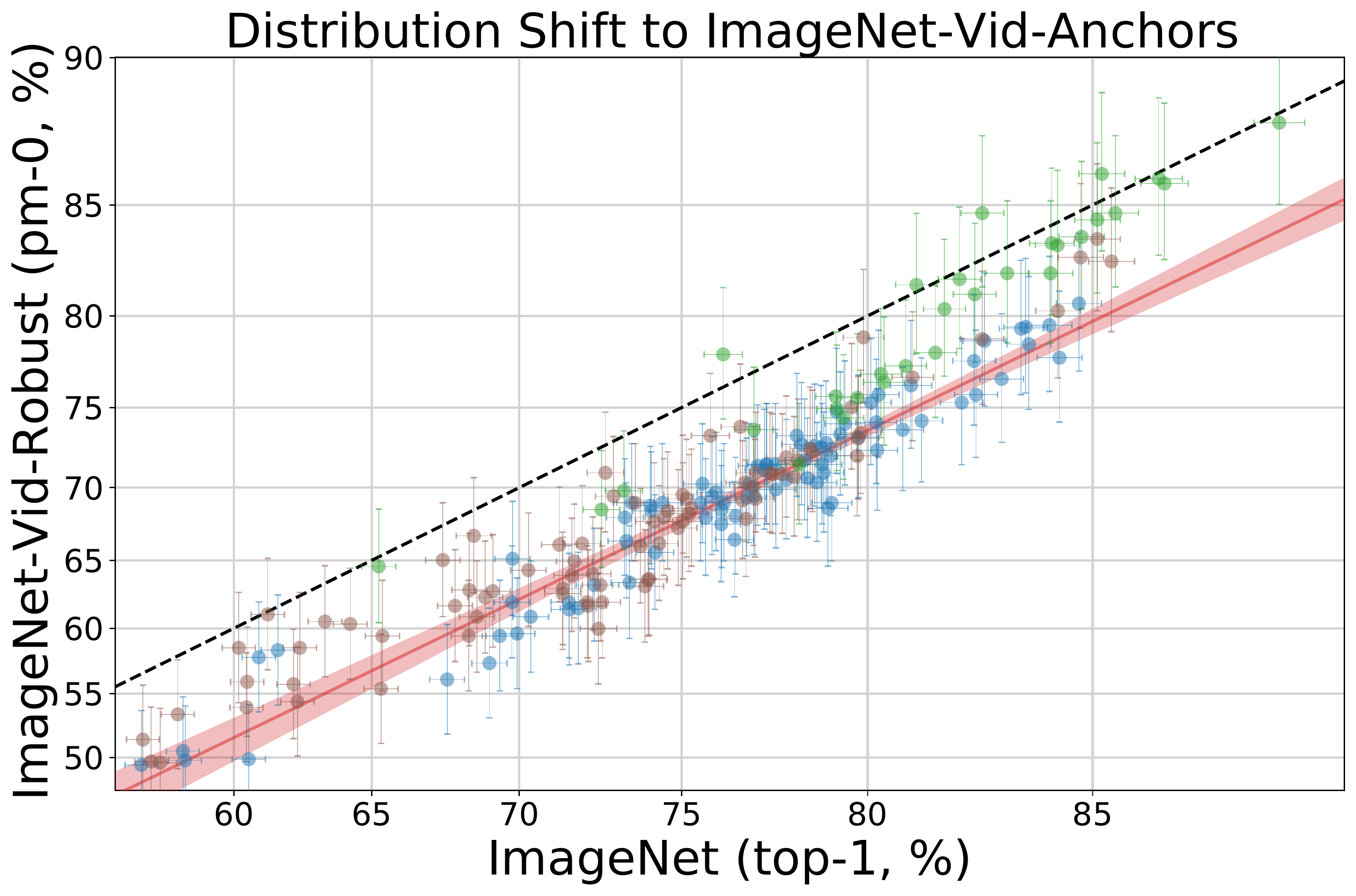}
  \end{subfigure}
  \hspace{0.5cm}
  \begin{subfigure}[t]{0.4\textwidth}
    \includegraphics[width=\textwidth]{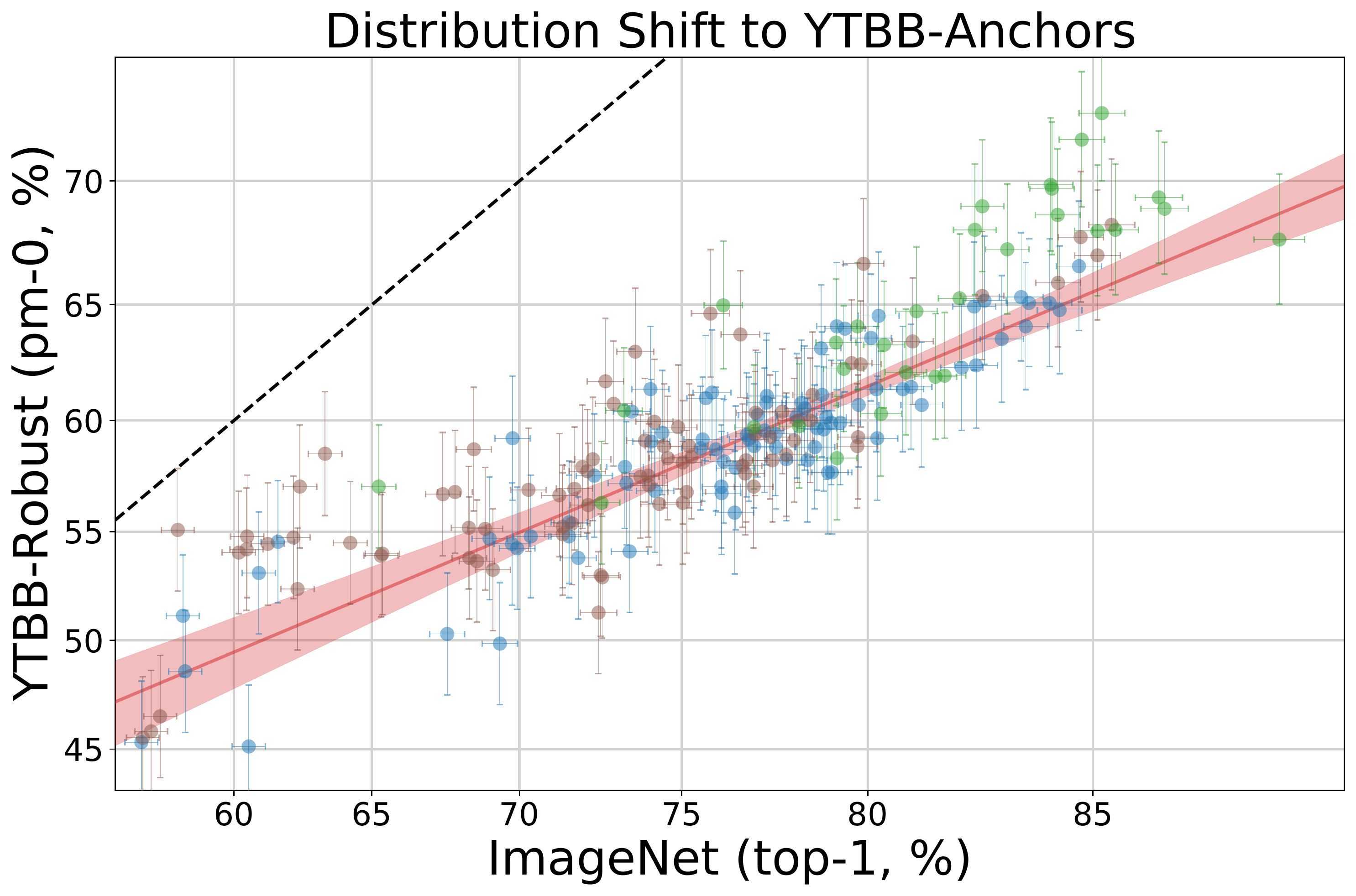}
  \end{subfigure}
  \par\medskip
  \begin{subfigure}[t]{0.4\textwidth}
    \includegraphics[width=\textwidth]{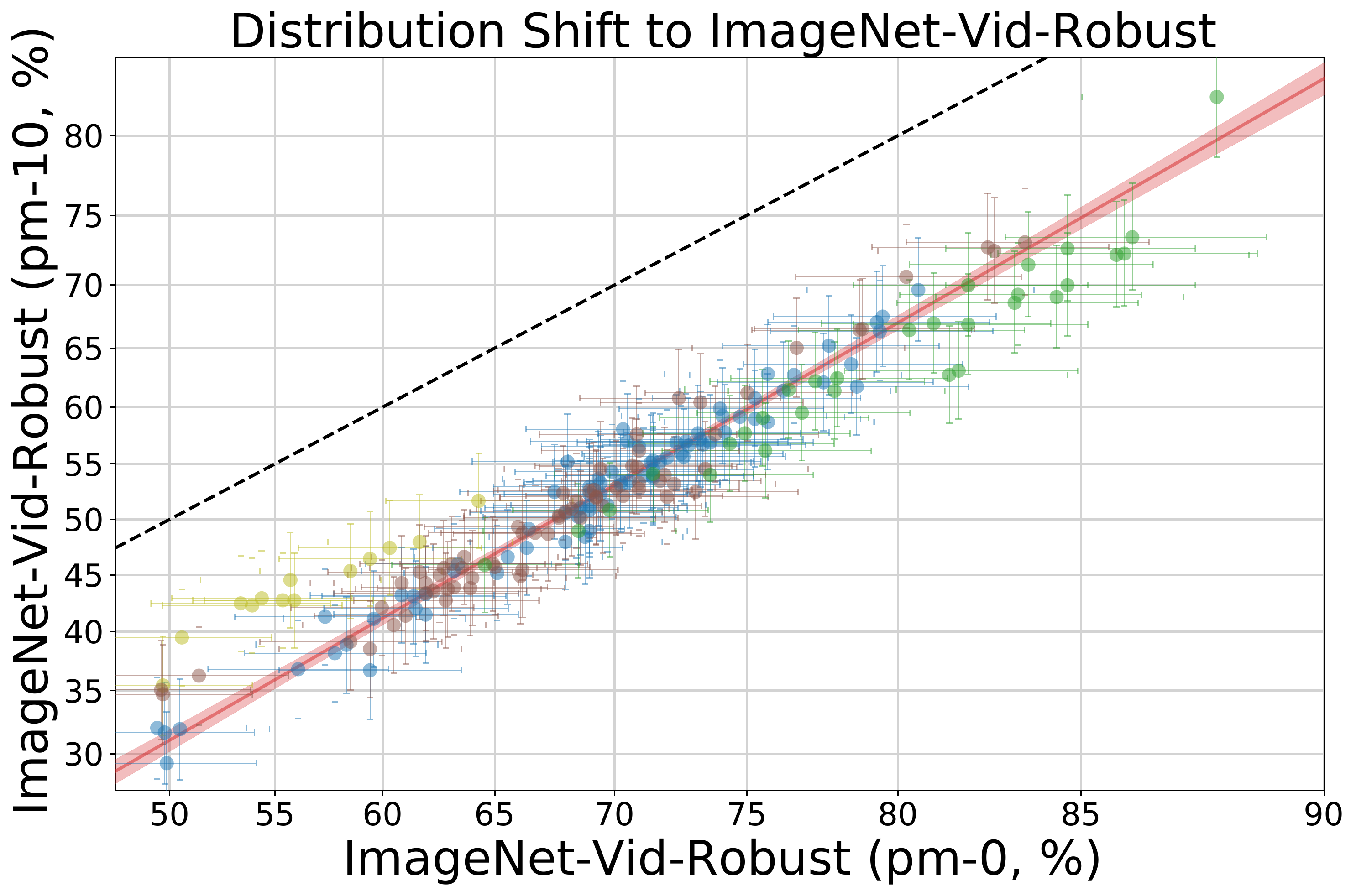}
  \end{subfigure}
  \hspace{0.5cm}
  \begin{subfigure}[t]{0.4\textwidth}
    \includegraphics[width=\textwidth]{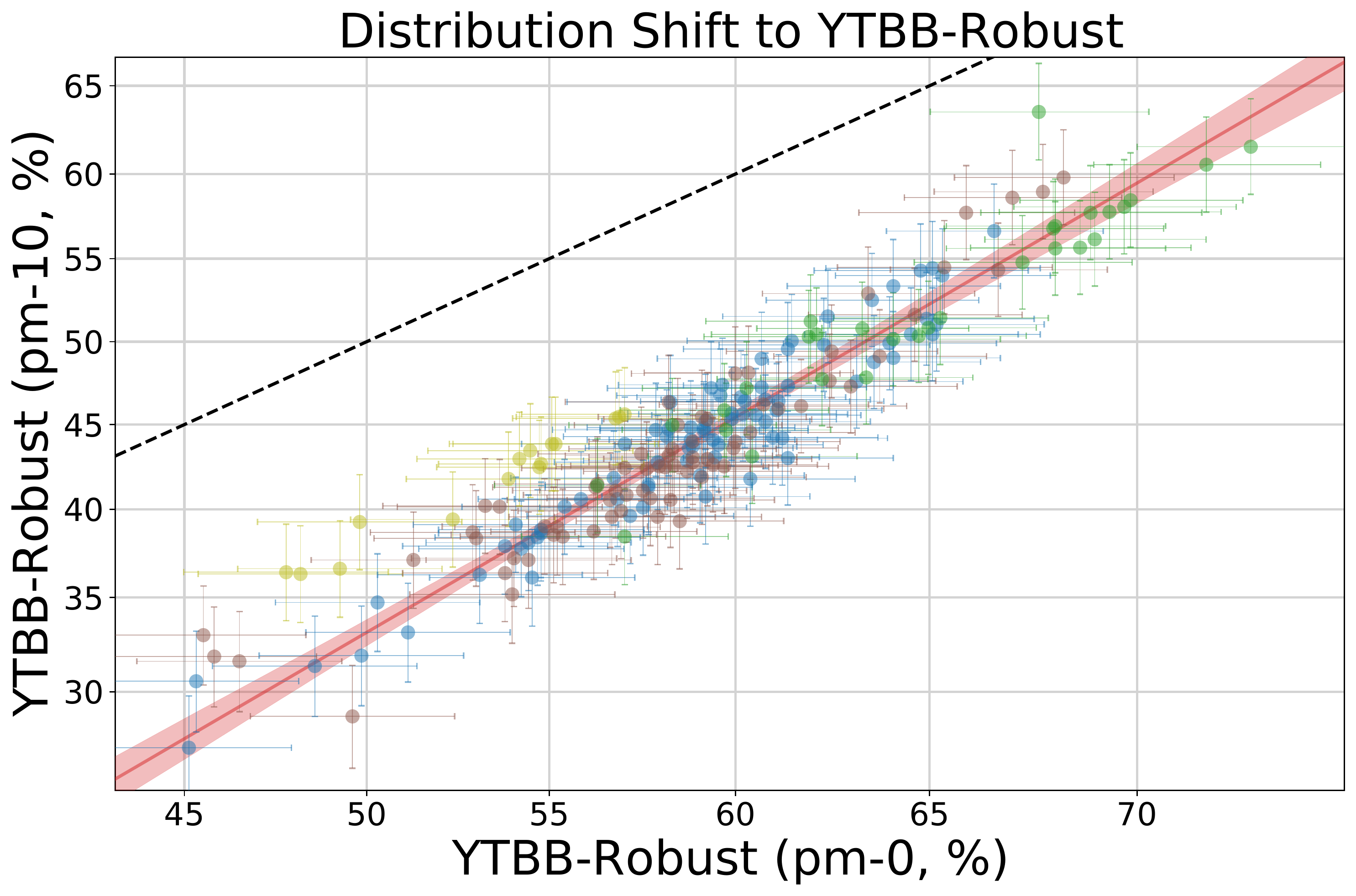}
  \end{subfigure}
  \par\medskip
  \begin{subfigure}[t]{0.4\textwidth}
    \includegraphics[width=\textwidth]{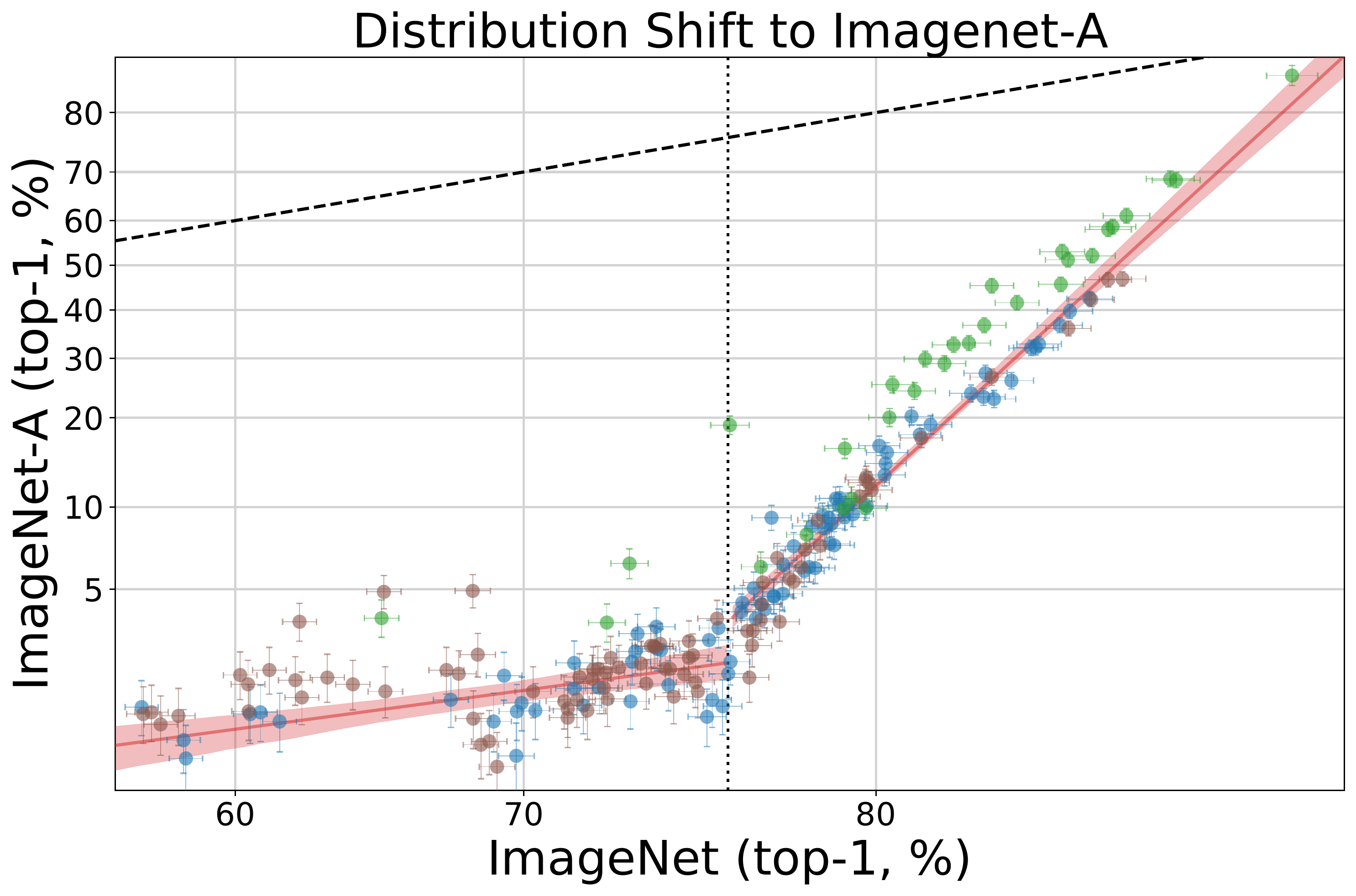}
  \end{subfigure}
  \begin{subfigure}[t]{\textwidth}
    \centering \includegraphics[width=0.7\textwidth]{figs/nat_shift_legend}
  \end{subfigure}
  \caption{
    The x-axes are not subsampled in these plots (they are performance on the full ImageNet validation set). 
    Otherwise, they are identical to the main plots in the main text.
    This is done to clarify that subsampling the axes does not skew the discussed results.
  }
  \label{fig:no-subsample}
\end{figure*}

\begin{figure*}[ht!]
  \centering
  \begin{subfigure}[t]{0.4\textwidth}
    \includegraphics[width=\textwidth]{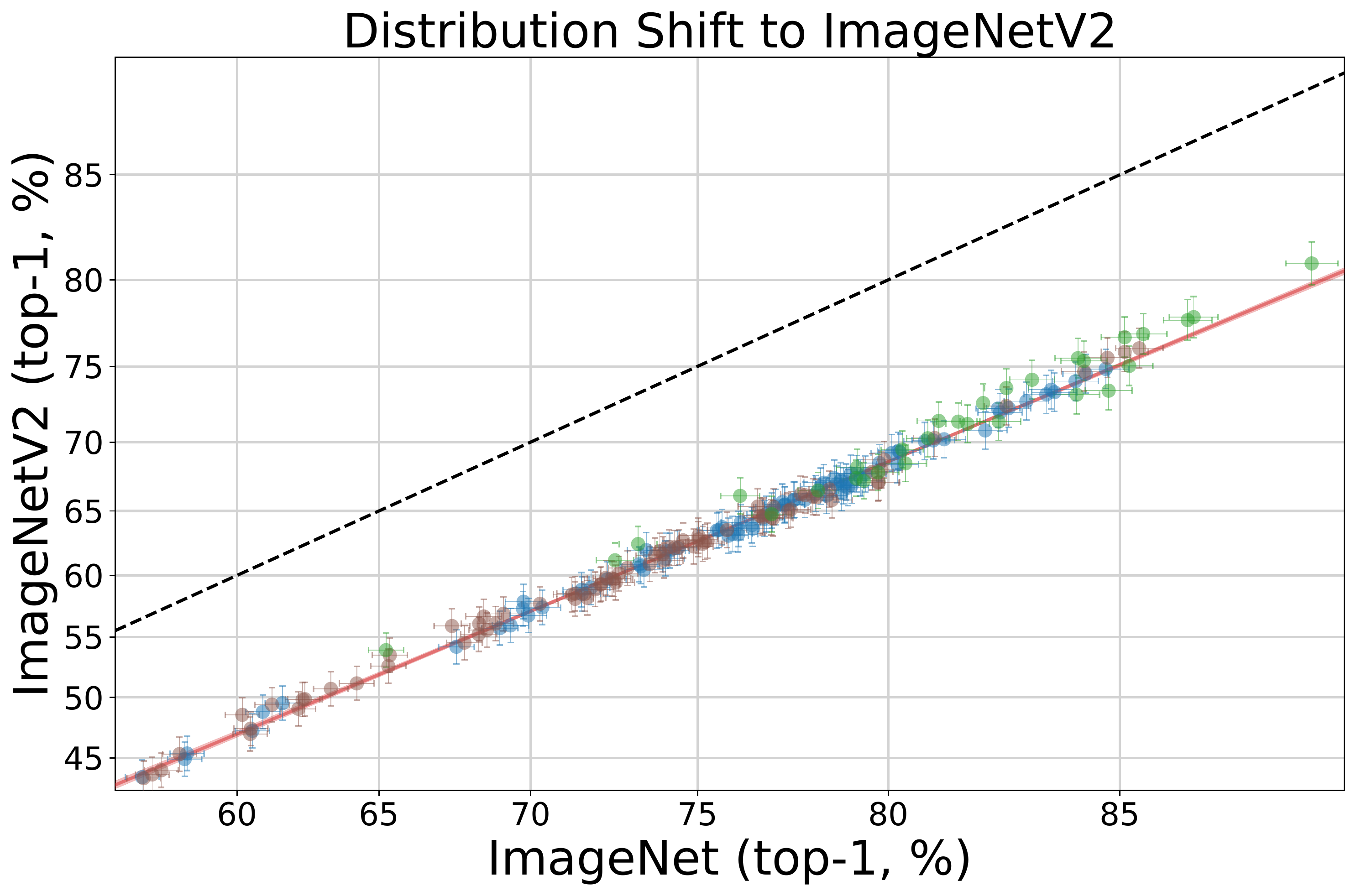}
  \end{subfigure}
  \hspace{0.5cm}
  \begin{subfigure}[t]{0.4\textwidth}
    \includegraphics[width=\textwidth]{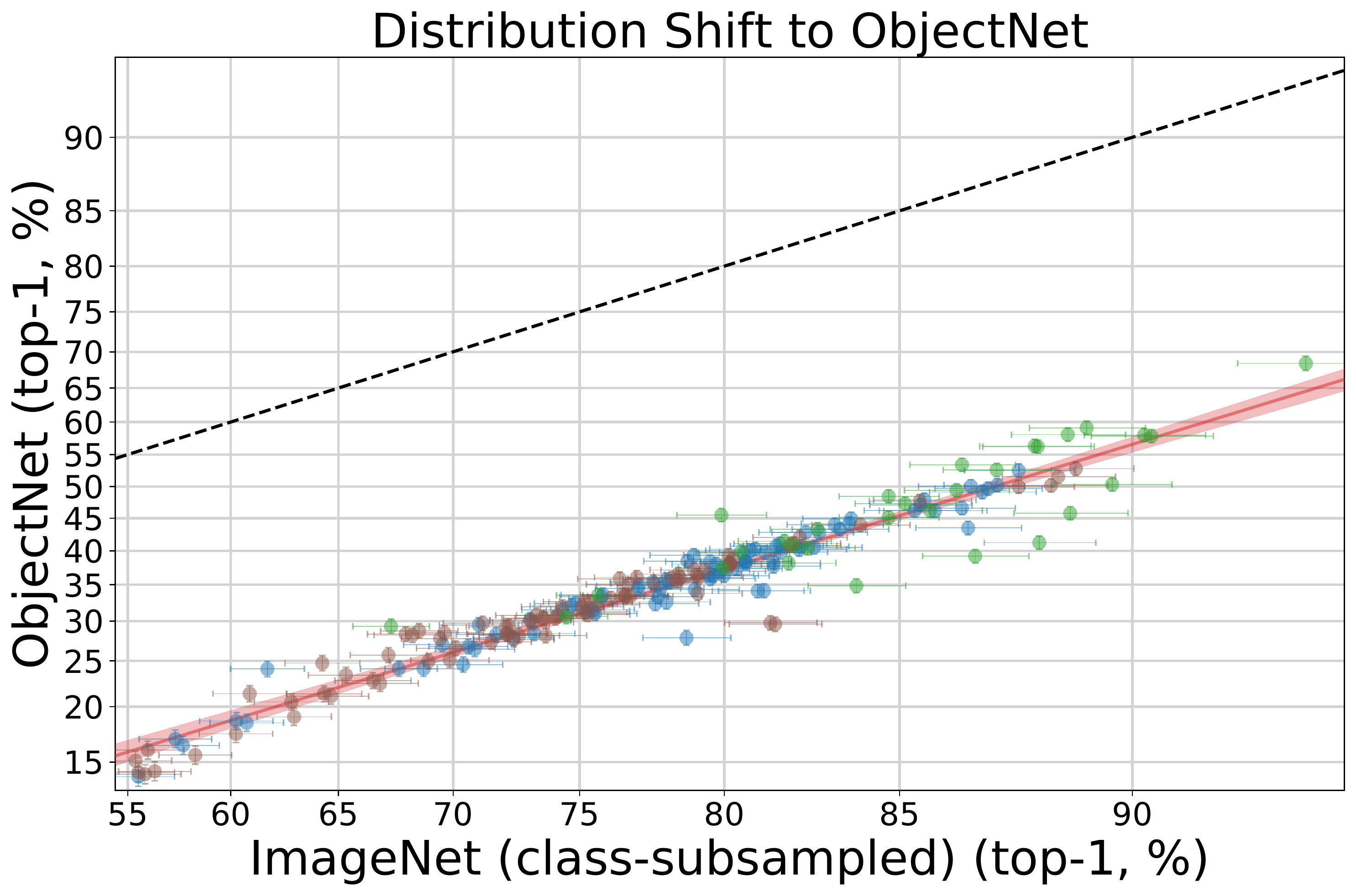}
  \end{subfigure}
  \par\medskip
  \begin{subfigure}[t]{0.4\textwidth}
    \includegraphics[width=\textwidth]{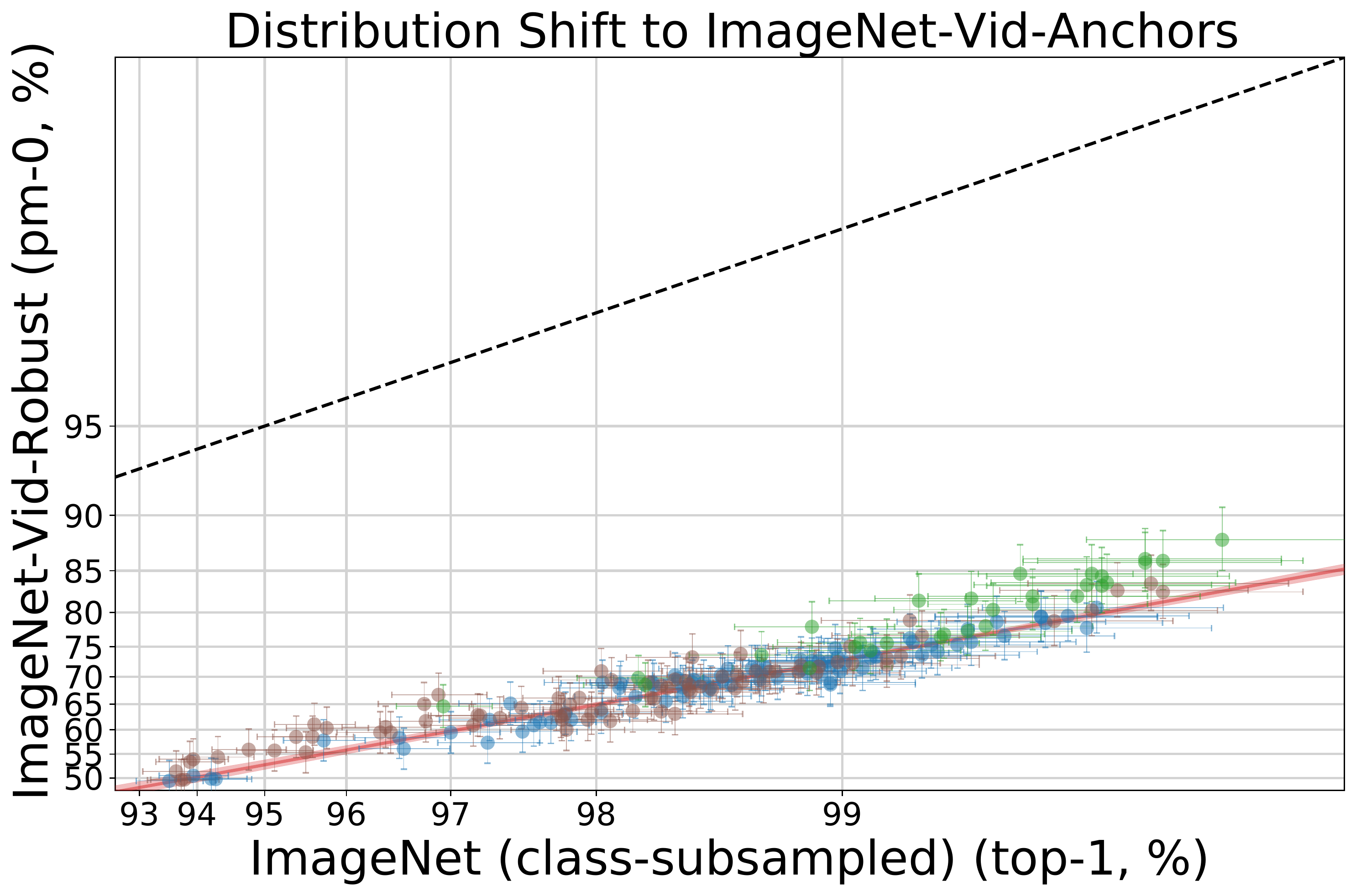}
  \end{subfigure}
  \hspace{0.5cm}
  \begin{subfigure}[t]{0.4\textwidth}
    \includegraphics[width=\textwidth]{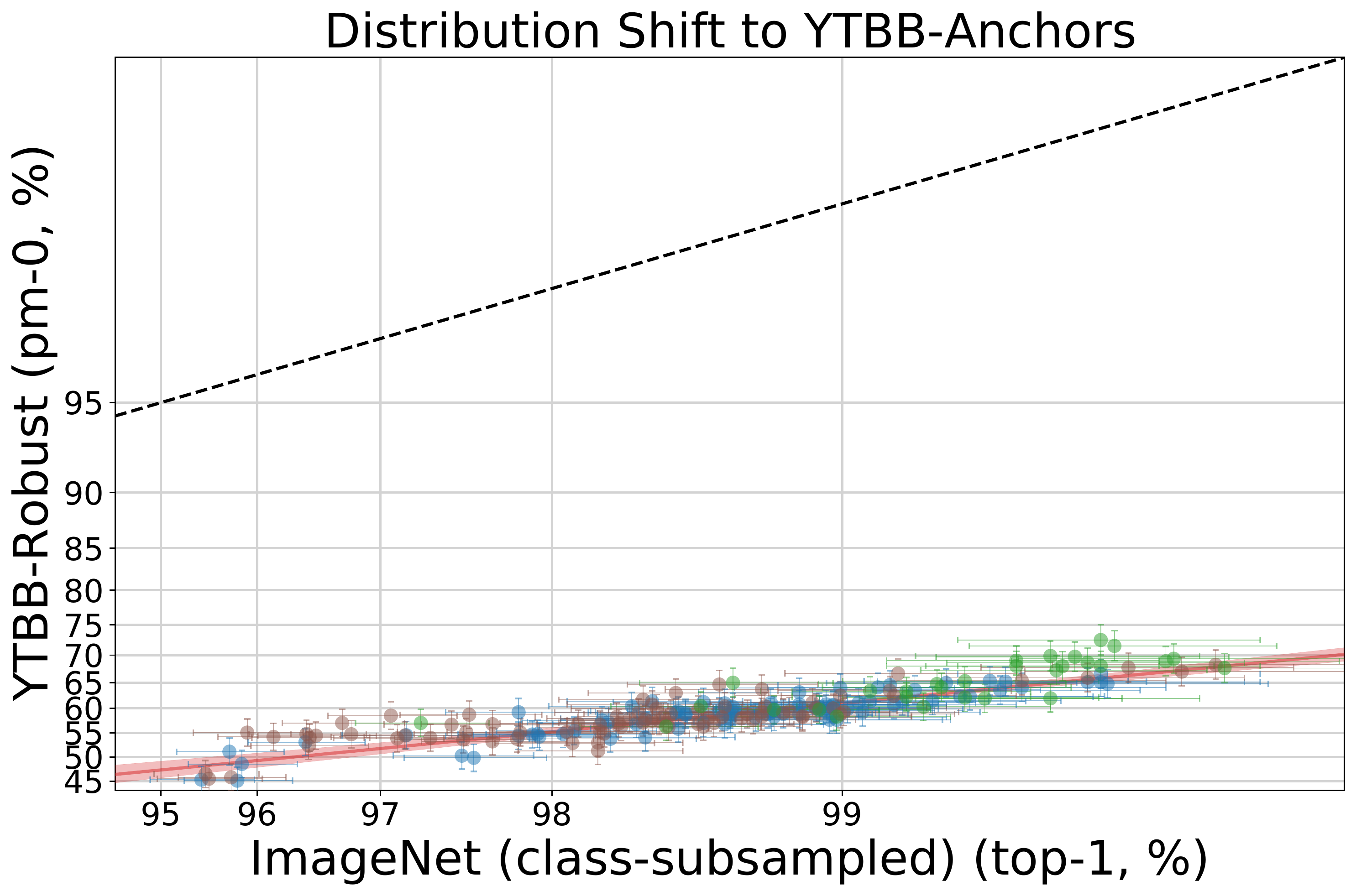}
  \end{subfigure}
  \par\medskip
  \begin{subfigure}[t]{0.4\textwidth}
    \includegraphics[width=\textwidth]{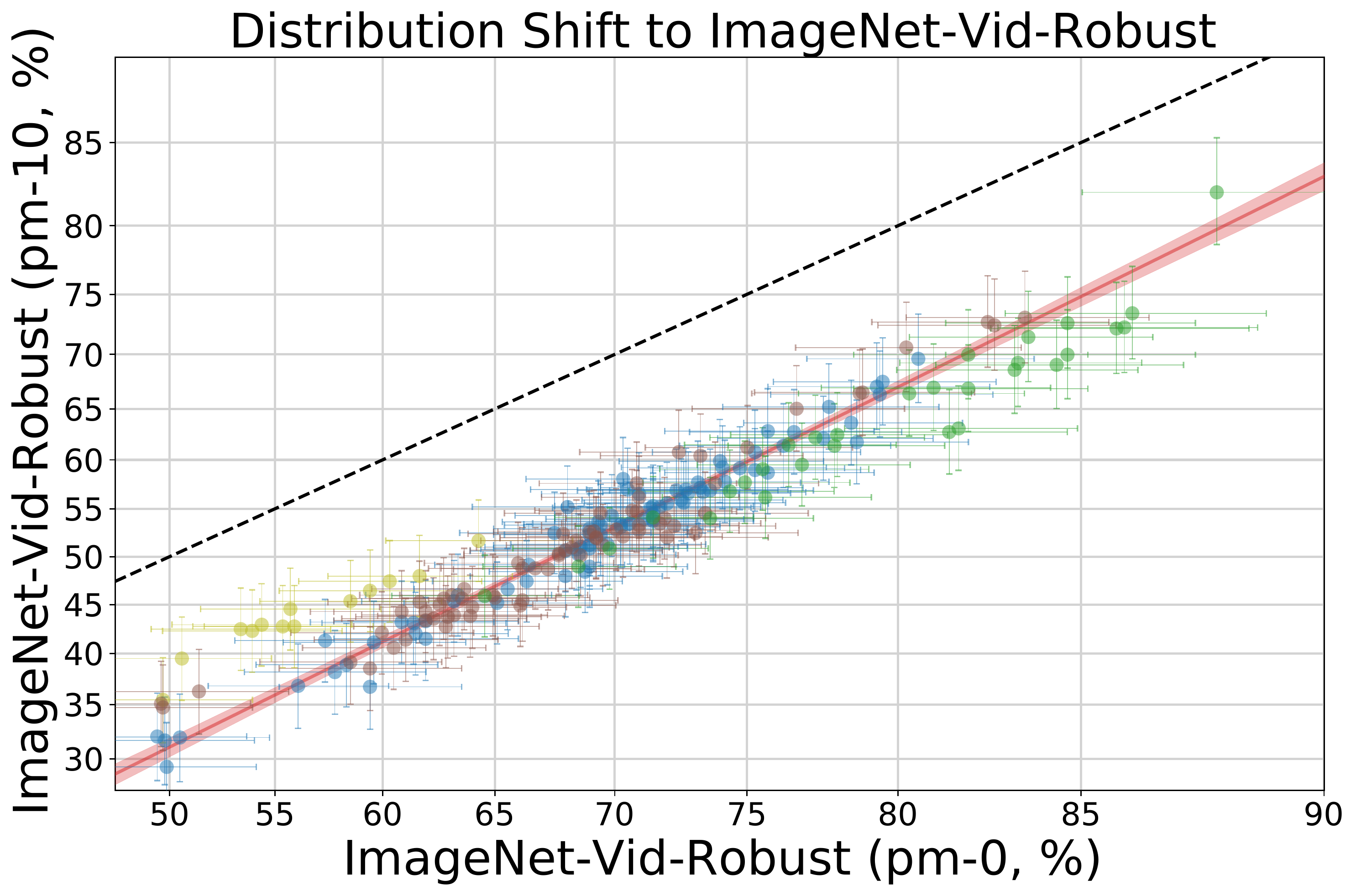}
  \end{subfigure}
  \hspace{0.5cm}
  \begin{subfigure}[t]{0.4\textwidth}
    \includegraphics[width=\textwidth]{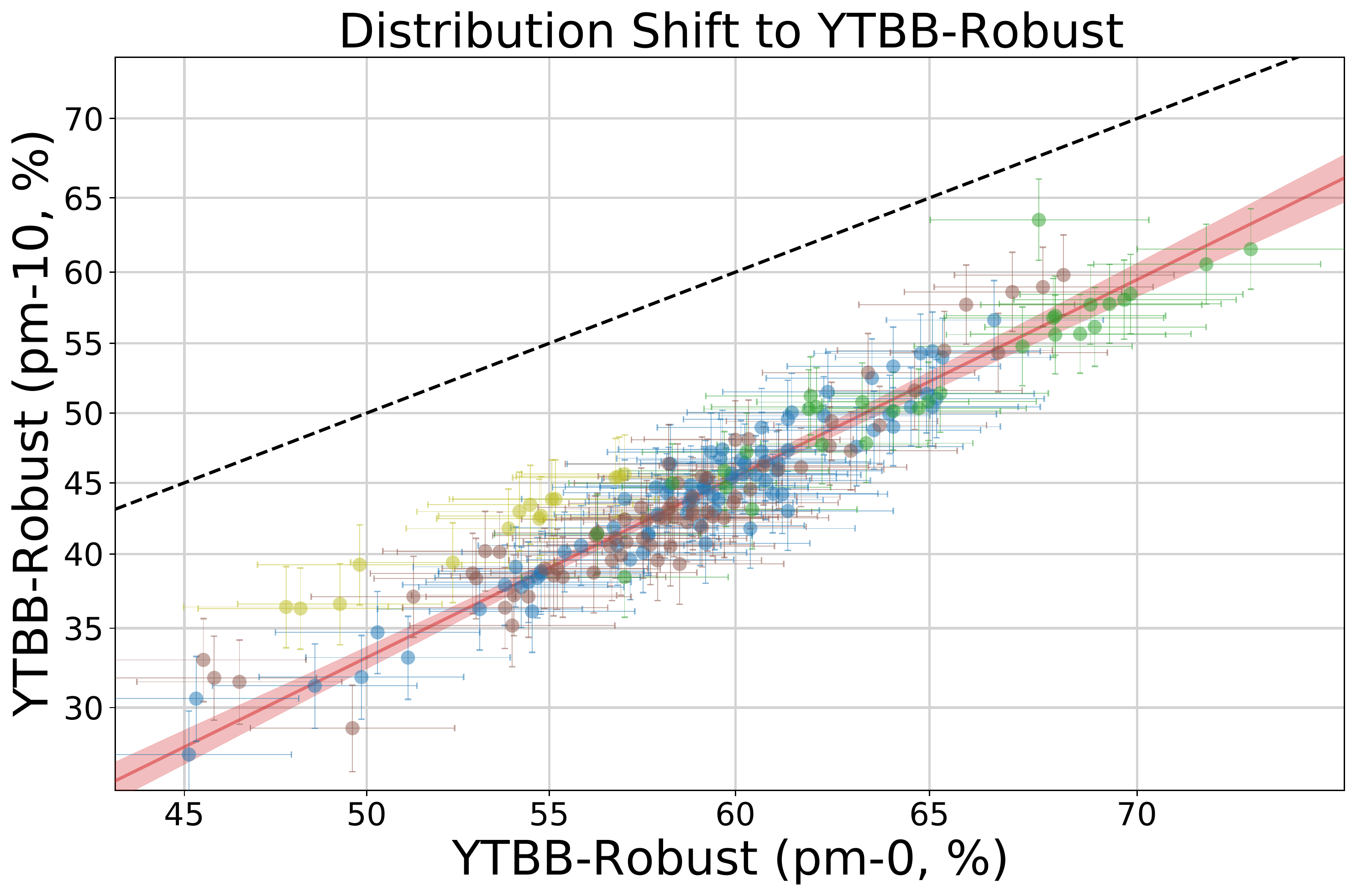}
  \end{subfigure}
  \par\medskip
  \begin{subfigure}[t]{0.4\textwidth}
    \includegraphics[width=\textwidth]{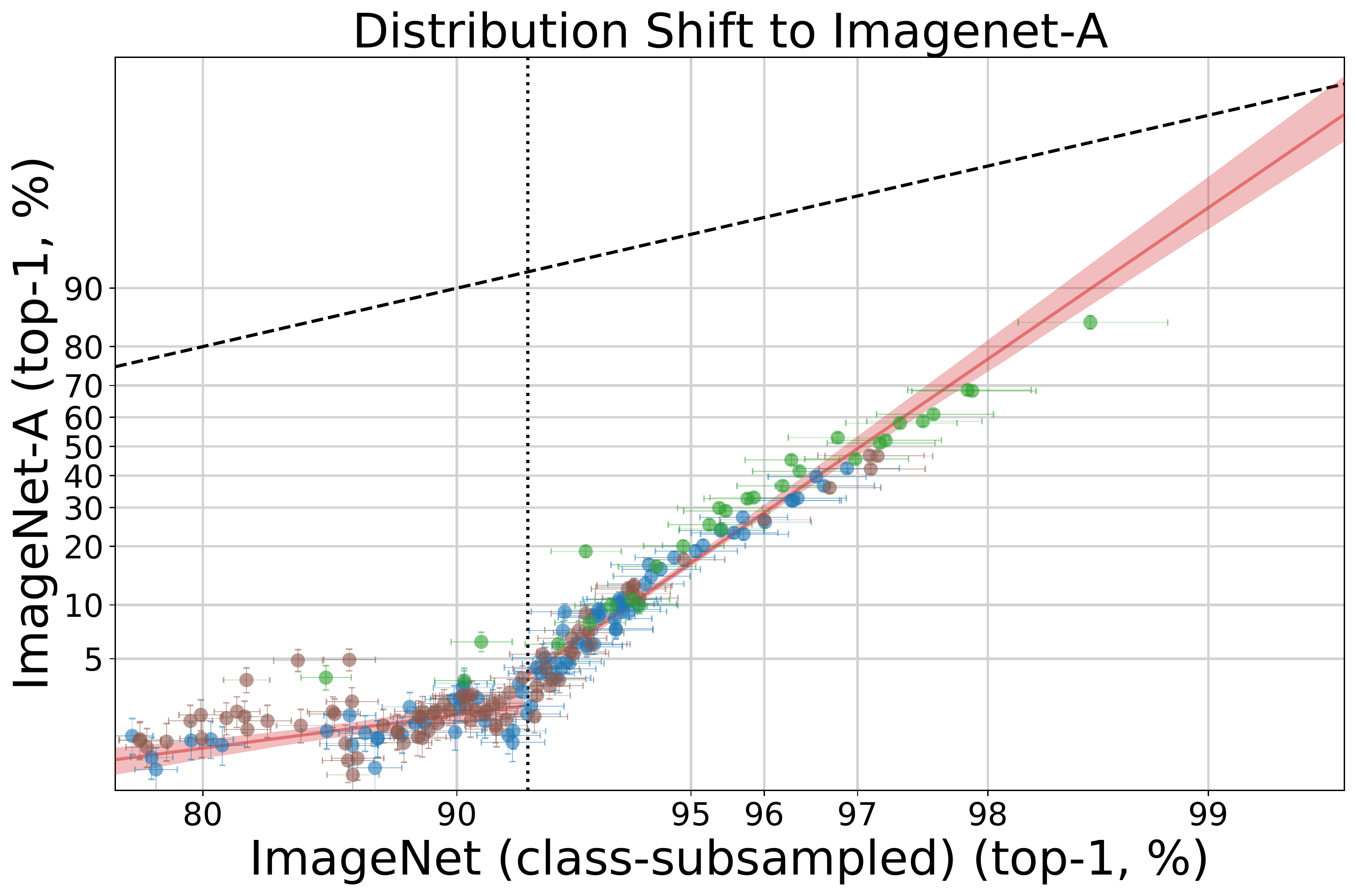}
  \end{subfigure}
  \begin{subfigure}[t]{\textwidth}
    \centering \includegraphics[width=0.7\textwidth]{figs/nat_shift_legend}
  \end{subfigure}
  \caption{
    The full y=x line is shown here in these plots.
    Otherwise, they are identical to the main plots in the main text.
    This is done to illustrate the performance gap due to distribution shift for each of the natural shifts.
  }
  \label{fig:show-yx}
\end{figure*}

\clearpage

\section{Example images of distribution shifts in our testbed}
\label{app:example-images}

\subsection{Natural distribution shift images}

\newcommand\width{0.2}
\newcommand\height{2.6cm}

\begin{figure}[ht!]
\begin{subfigure}[b]{\width\linewidth}
  \includegraphics[width=\linewidth, height=\height]{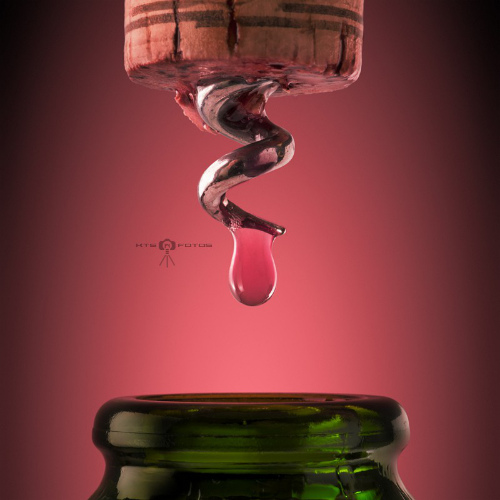}
\end{subfigure}
\hfill
\begin{subfigure}[b]{\width\linewidth}
  \includegraphics[width=\linewidth, height=\height]{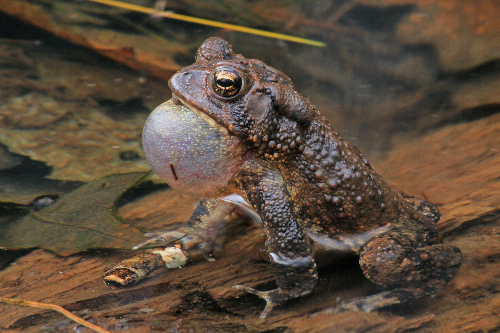}
\end{subfigure}
\hfill
\begin{subfigure}[b]{\width\linewidth}
  \includegraphics[width=\linewidth, height=\height]{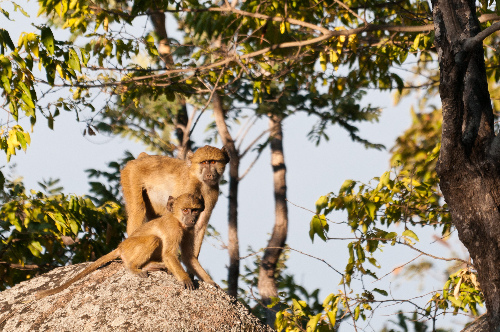}
\end{subfigure}
\hfill
\begin{subfigure}[b]{\width\linewidth}
  \includegraphics[width=\linewidth, height=\height]{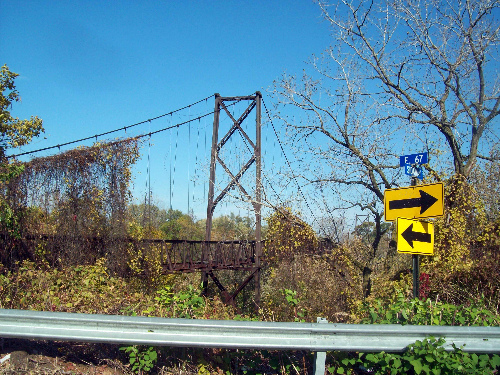}
\end{subfigure}
\begin{subfigure}[b]{\width\linewidth}
  \includegraphics[width=\linewidth, height=\height]{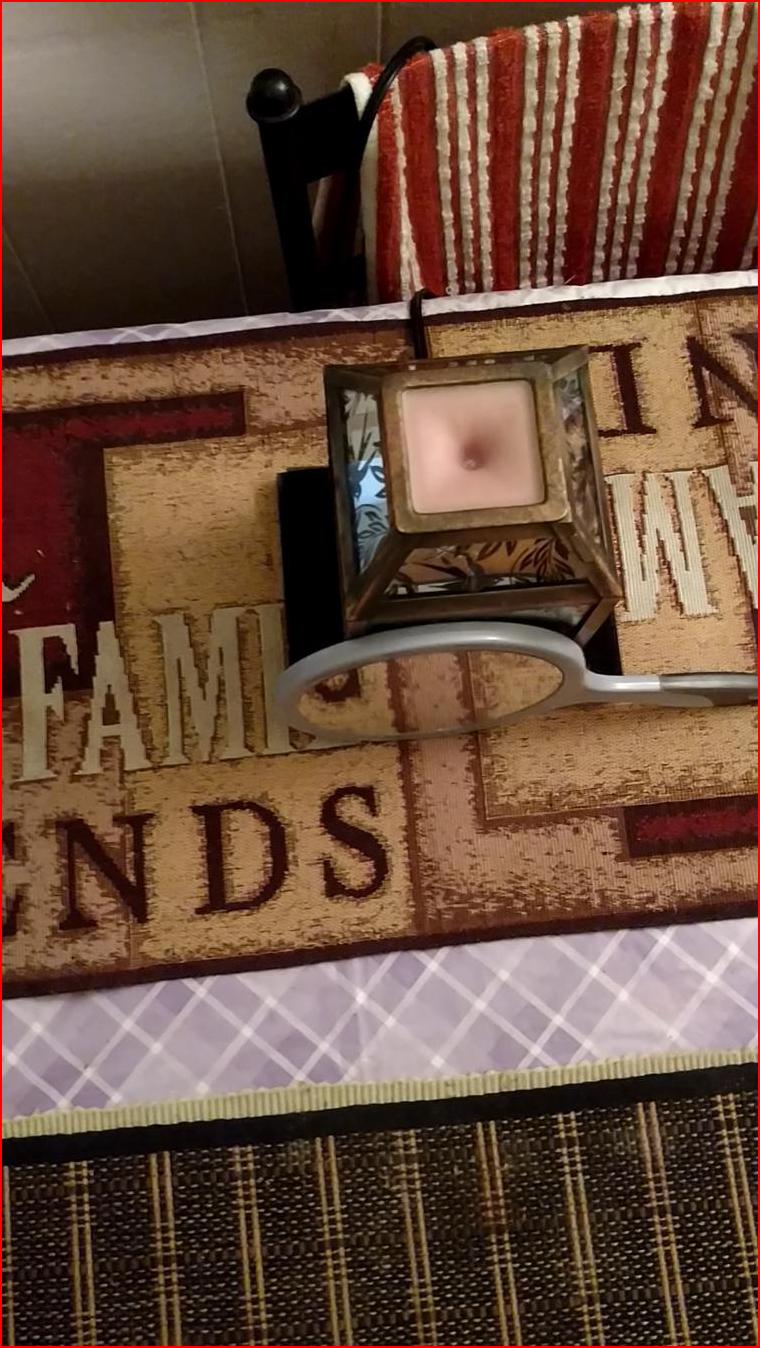}
\end{subfigure}
\hfill
\begin{subfigure}[b]{\width\linewidth}
  \includegraphics[width=\linewidth, height=\height]{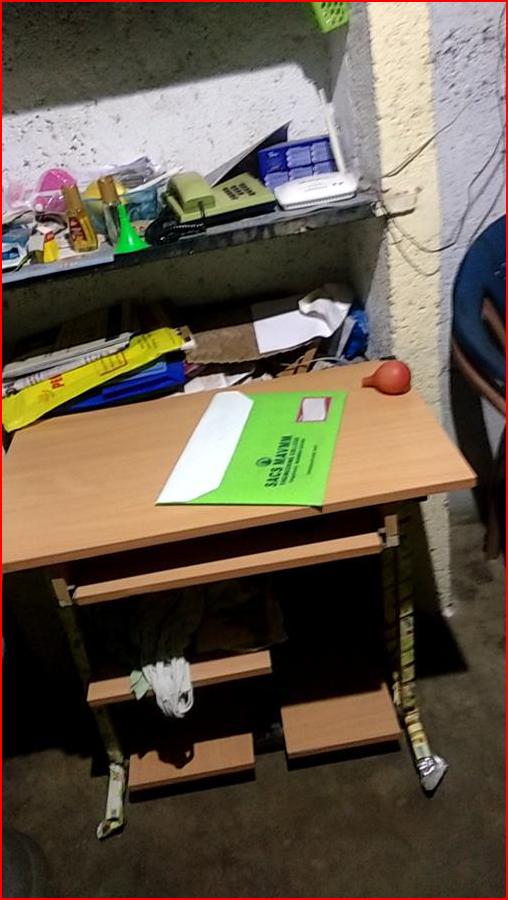}
\end{subfigure}
\hfill
\begin{subfigure}[b]{\width\linewidth}
  \includegraphics[width=\linewidth, height=\height]{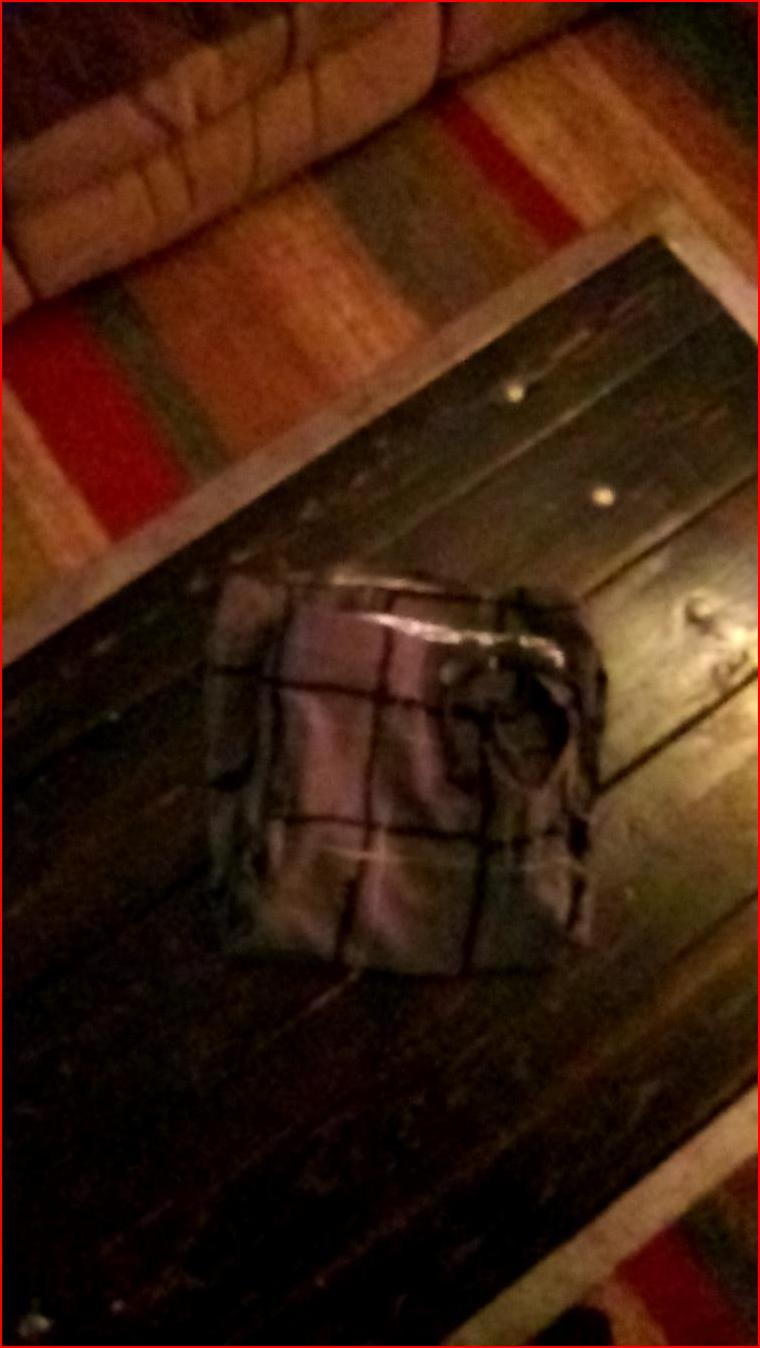}
\end{subfigure}
\hfill
\begin{subfigure}[b]{\width\linewidth}
  \includegraphics[width=\linewidth, height=\height]{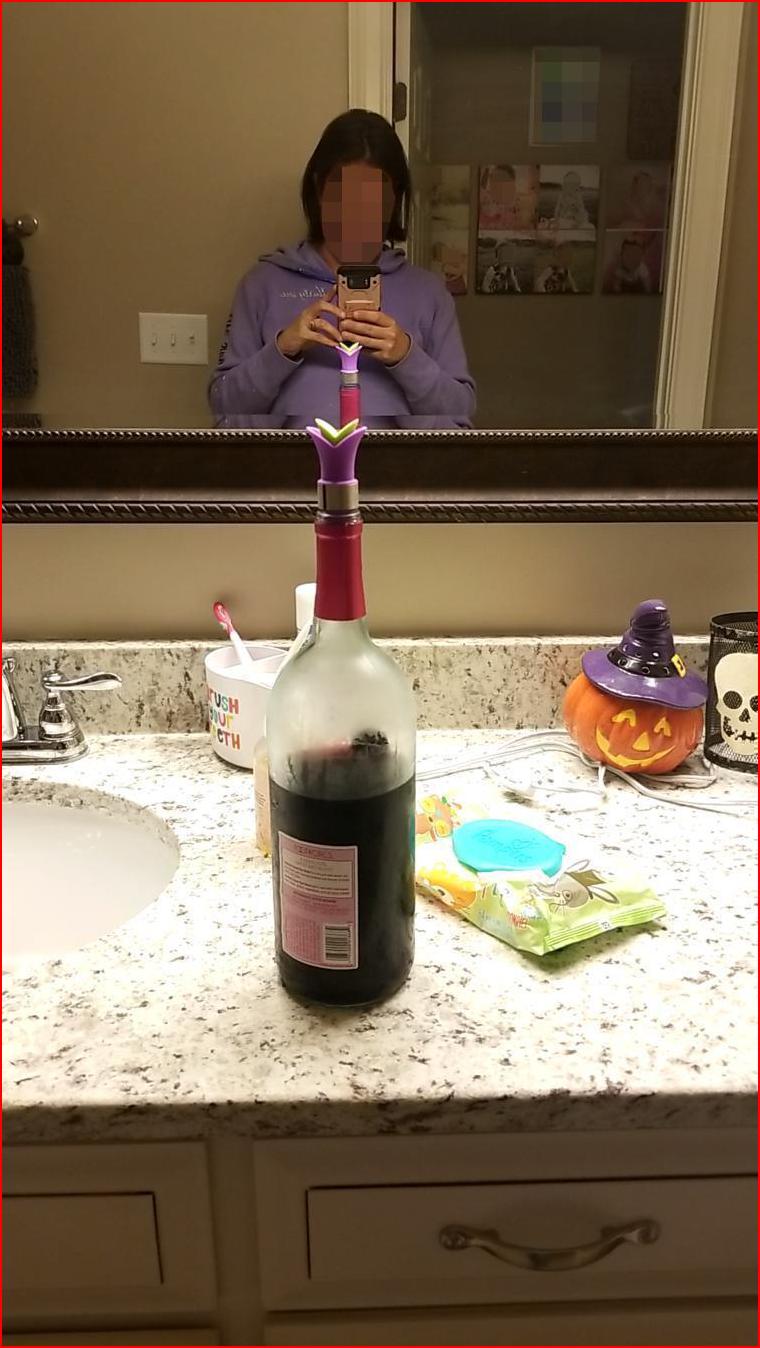}
\end{subfigure}
\begin{subfigure}[b]{\width\linewidth}
  \includegraphics[width=\linewidth, height=\height]{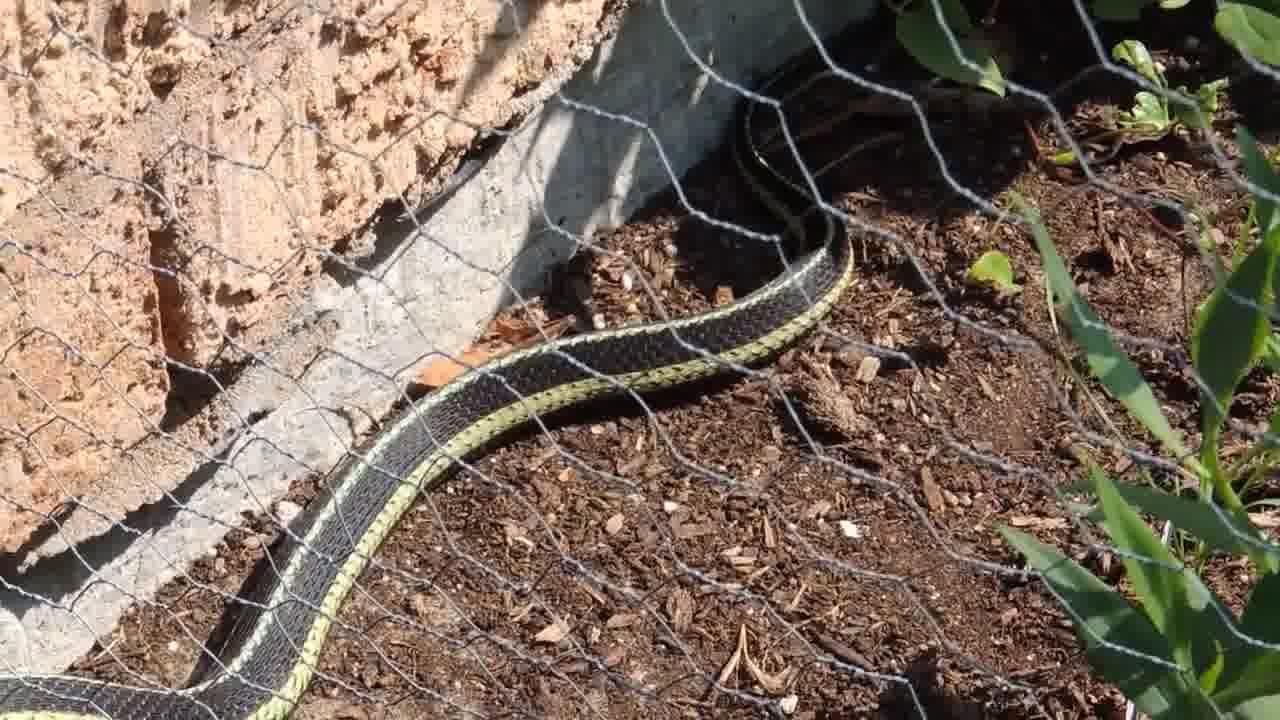}
\end{subfigure}
\hfill
\begin{subfigure}[b]{\width\linewidth}
  \includegraphics[width=\linewidth, height=\height]{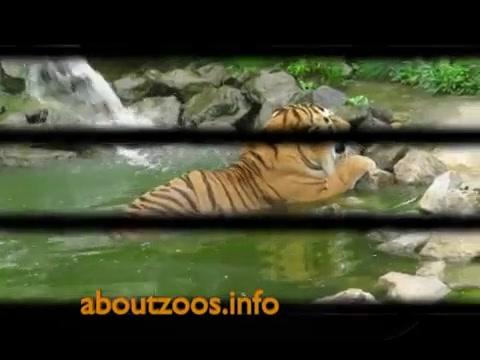}
\end{subfigure}
\hfill
\begin{subfigure}[b]{\width\linewidth}
  \includegraphics[width=\linewidth, height=\height]{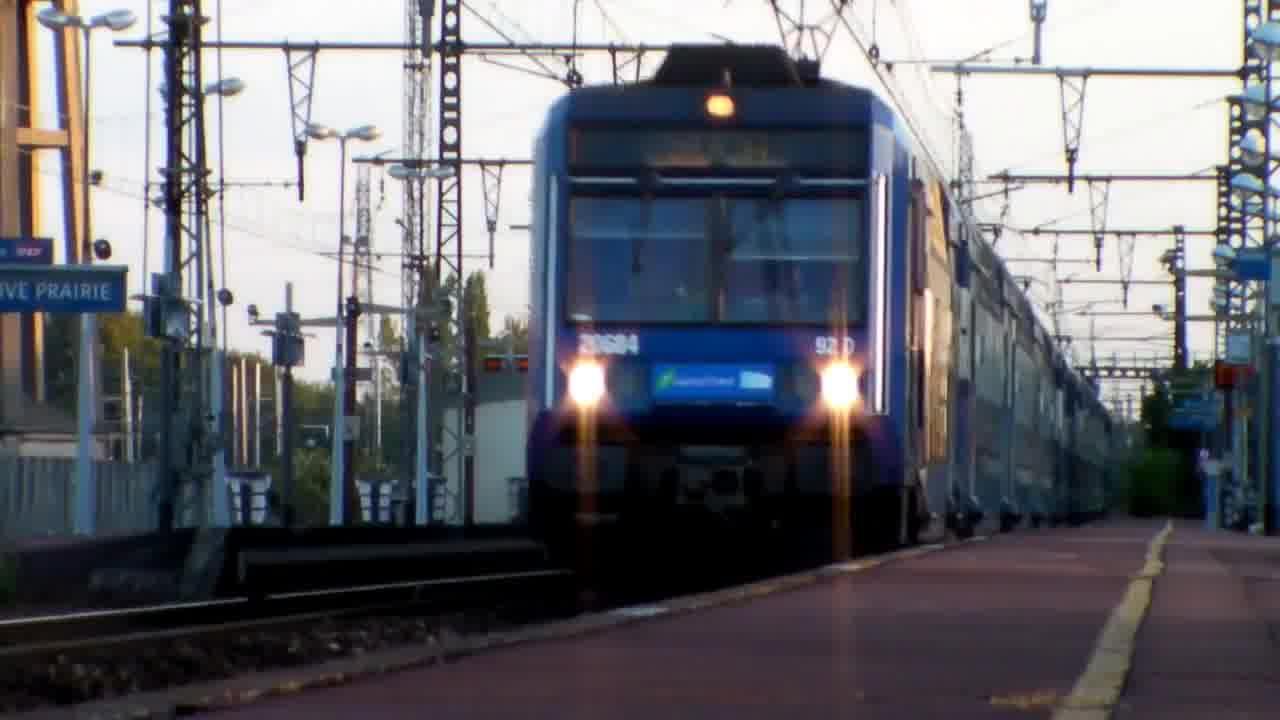}
\end{subfigure}
\hfill
\begin{subfigure}[b]{\width\linewidth}
  \includegraphics[width=\linewidth, height=\height]{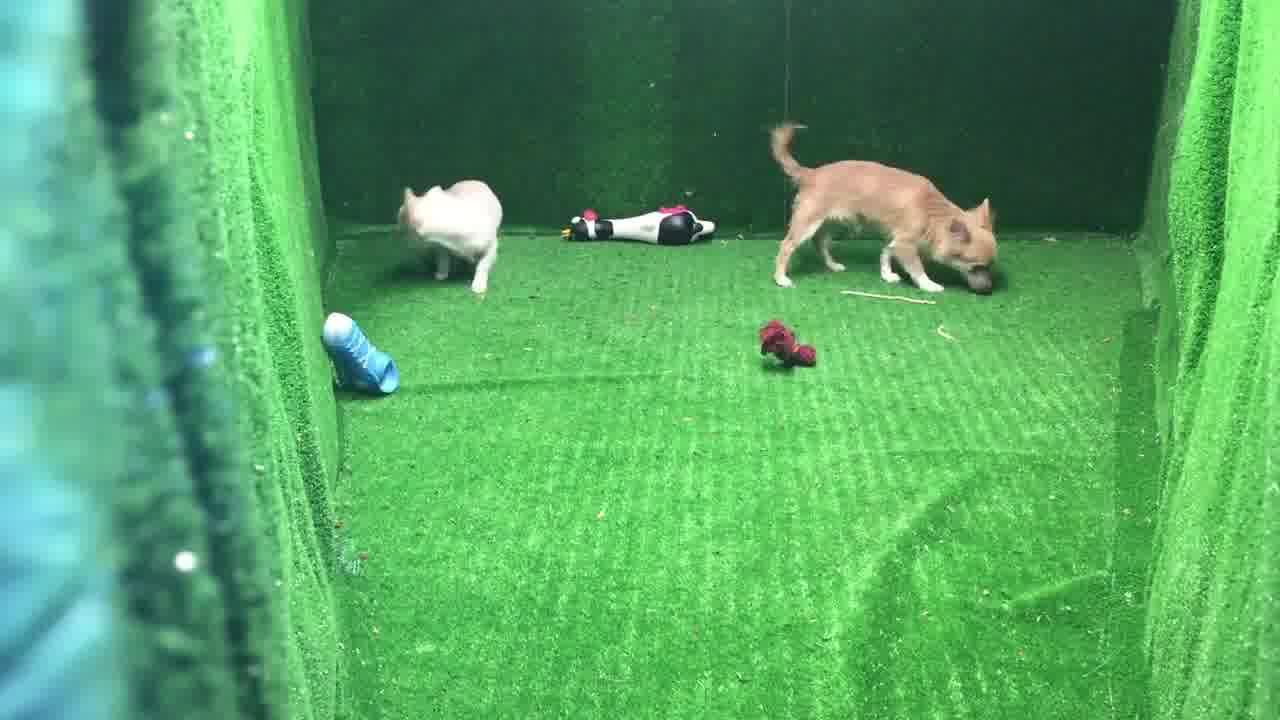}
\end{subfigure}
\begin{subfigure}[b]{\width\linewidth}
  \includegraphics[width=\linewidth, height=\height]{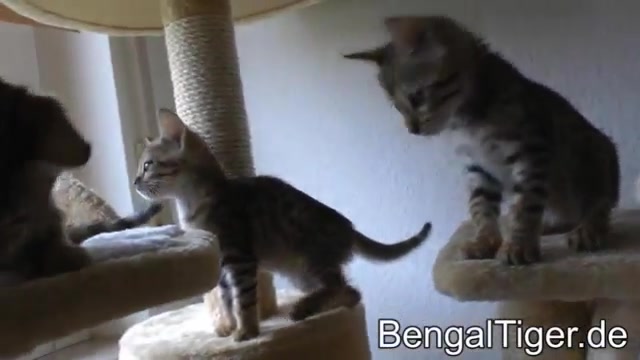}
\end{subfigure}
\hfill
\begin{subfigure}[b]{\width\linewidth}
  \includegraphics[width=\linewidth, height=\height]{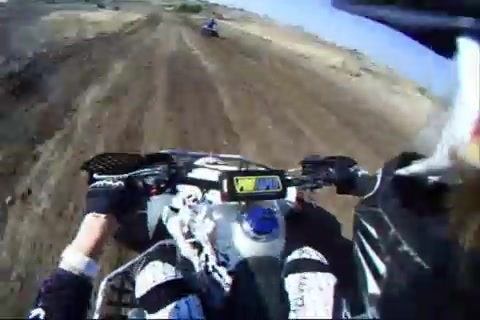}
\end{subfigure}
\hfill
\begin{subfigure}[b]{\width\linewidth}
  \includegraphics[width=\linewidth, height=\height]{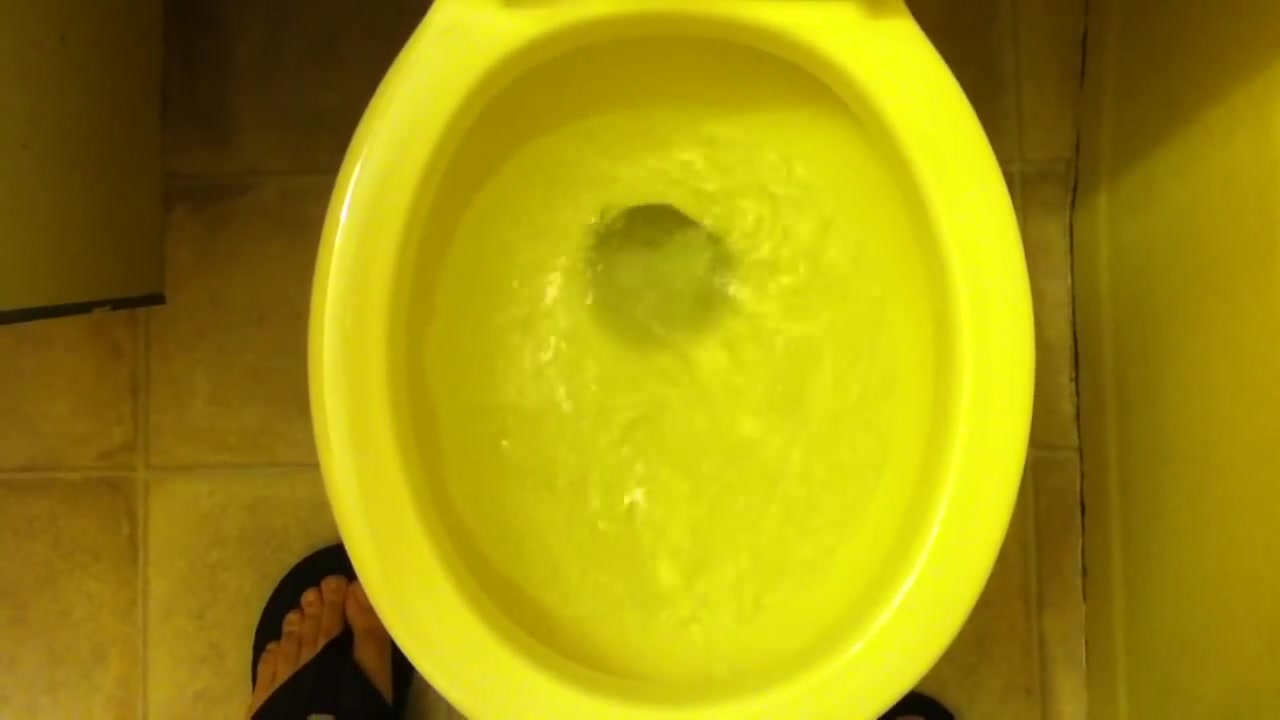}
\end{subfigure}
\hfill
\begin{subfigure}[b]{\width\linewidth}
  \includegraphics[width=\linewidth, height=\height]{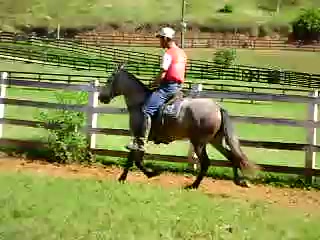}
\end{subfigure}
\caption{Dataset shifts.
Examples from \imagenetnew{} (first row), \objectnet{} (second row), \imagenetvidrobust{} (third row), and \ytbbrobust{} (fourth row).}
\end{figure}

\begin{figure}[ht!]
\begin{subfigure}[b]{\width\linewidth}
  \includegraphics[width=\linewidth, height=\height]{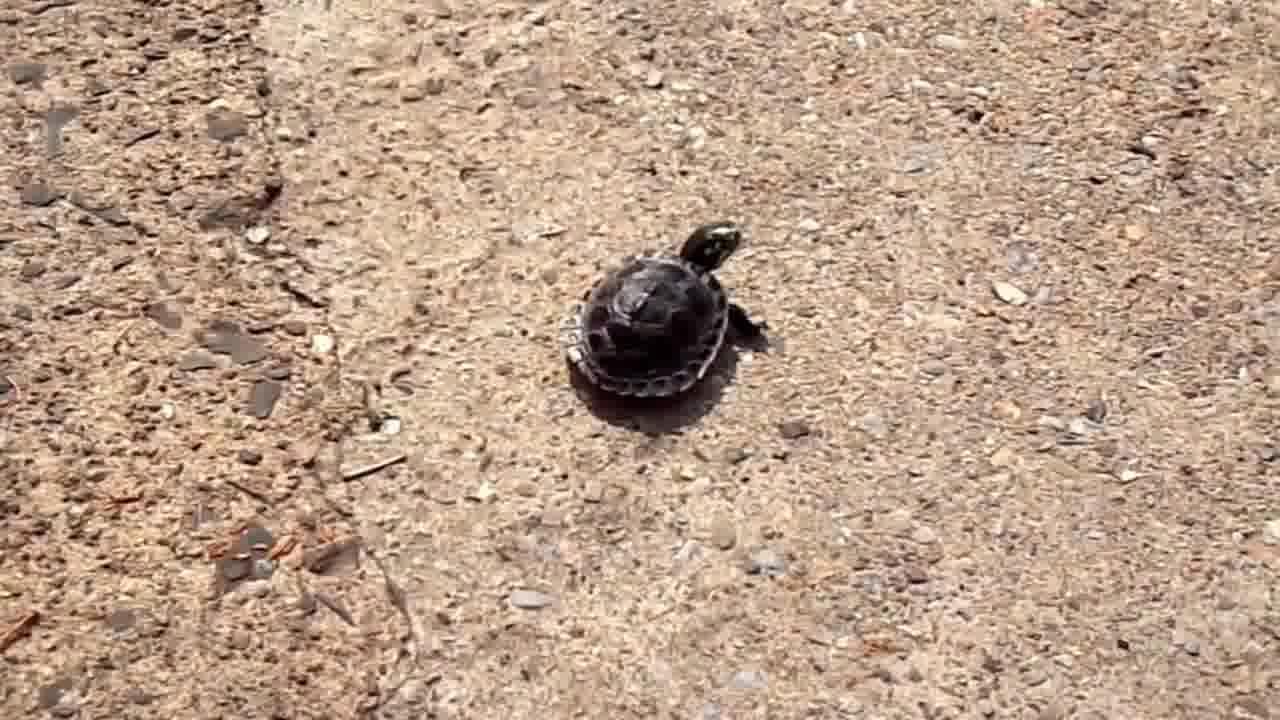}
\end{subfigure}
\hfill
\begin{subfigure}[b]{\width\linewidth}
  \includegraphics[width=\linewidth, height=\height]{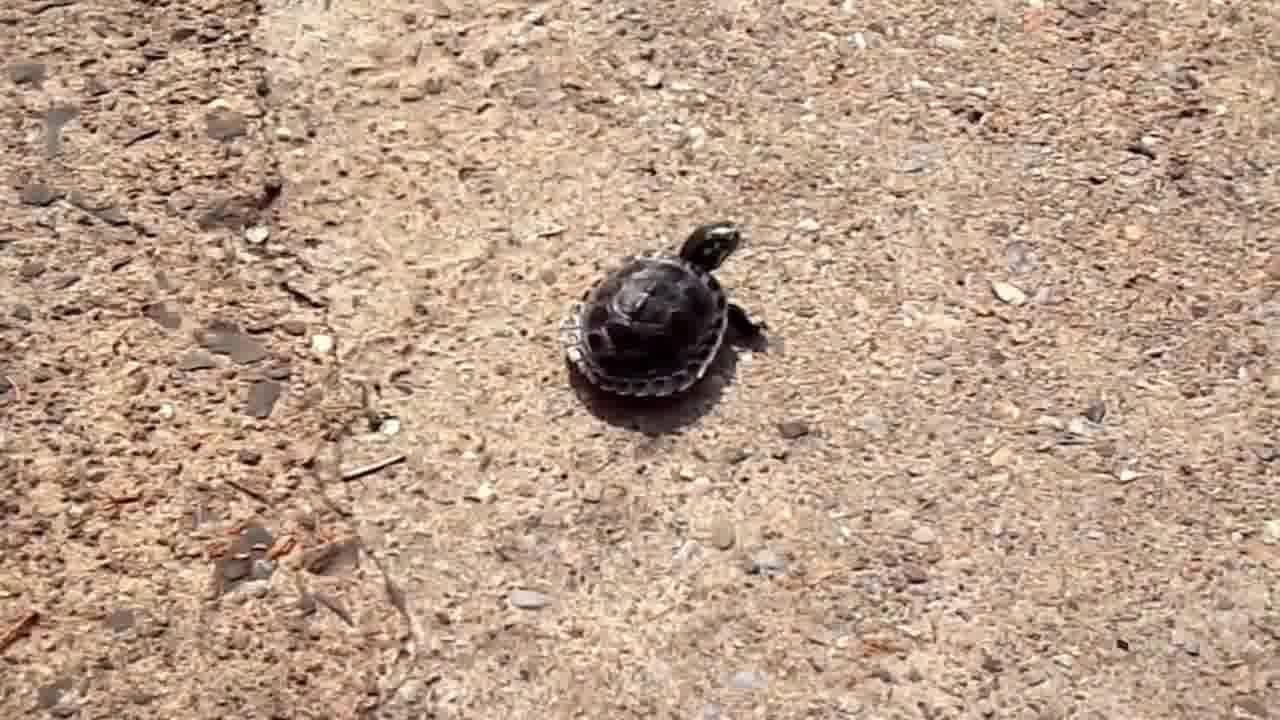}
\end{subfigure}
\hfill
\begin{subfigure}[b]{\width\linewidth}
  \includegraphics[width=\linewidth, height=\height]{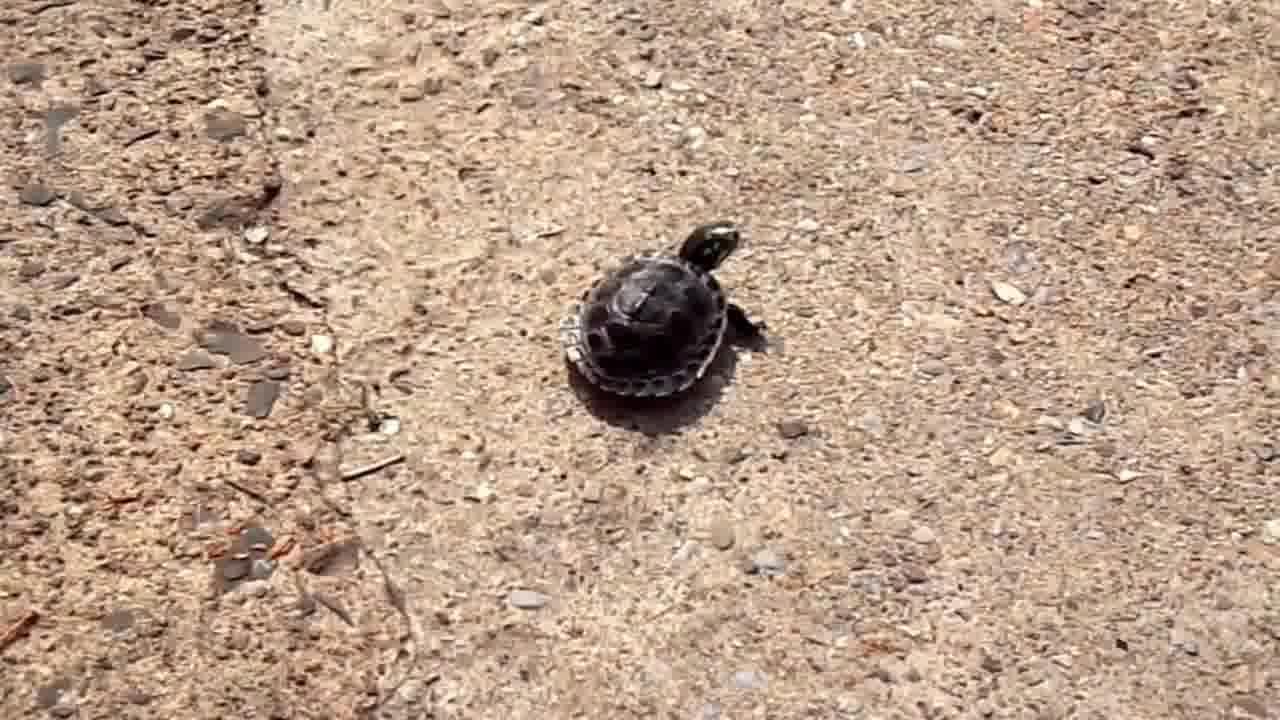}
\end{subfigure}
\hfill
\begin{subfigure}[b]{\width\linewidth}
  \includegraphics[width=\linewidth, height=\height]{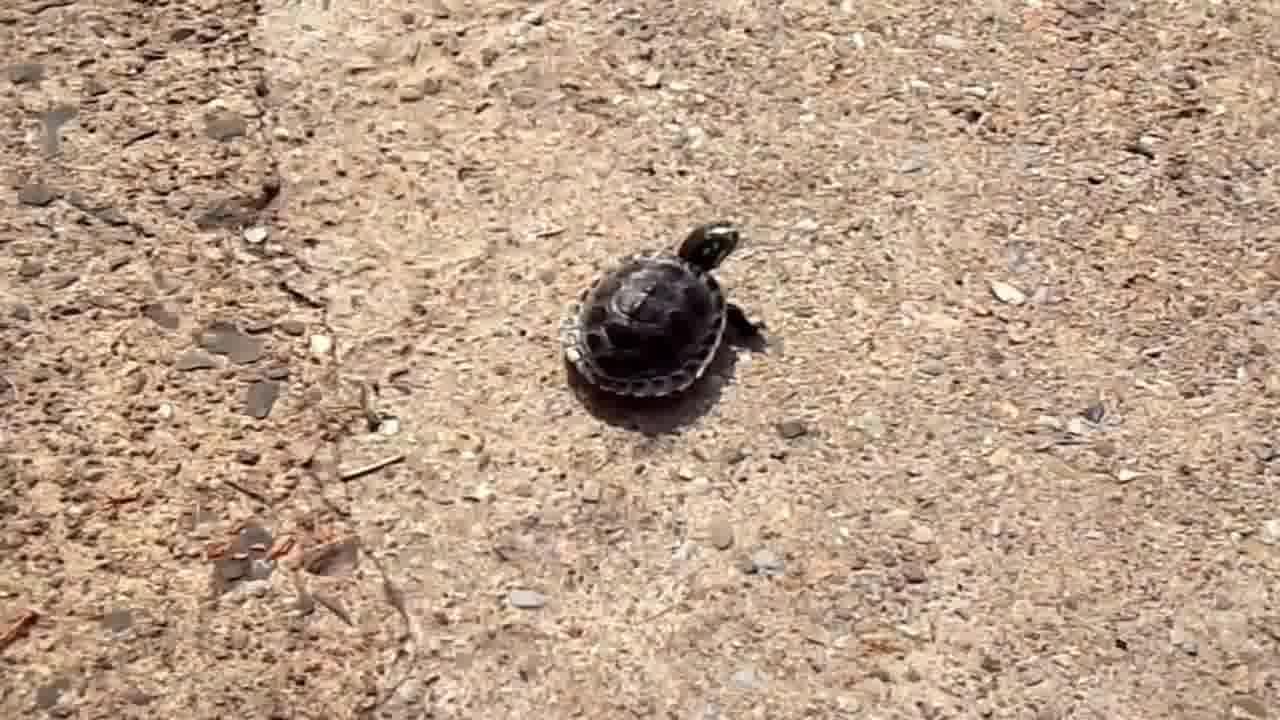}
\end{subfigure}
\begin{subfigure}[b]{\width\linewidth}
  \includegraphics[width=\linewidth, height=\height]{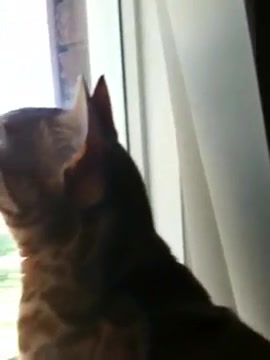}
\end{subfigure}
\hfill
\begin{subfigure}[b]{\width\linewidth}
  \includegraphics[width=\linewidth, height=\height]{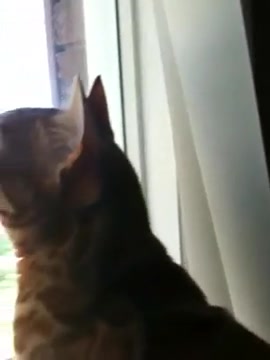}
\end{subfigure}
\hfill
\begin{subfigure}[b]{\width\linewidth}
  \includegraphics[width=\linewidth, height=\height]{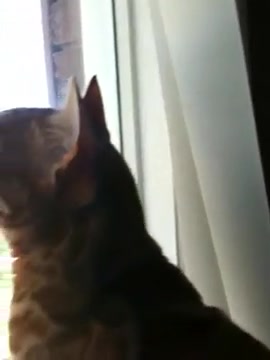}
\end{subfigure}
\hfill
\begin{subfigure}[b]{\width\linewidth}
  \includegraphics[width=\linewidth, height=\height]{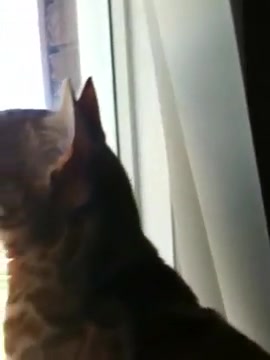}
\end{subfigure}
\caption{Consistency shifts. 
Sequences of video frames from \imagenetvidrobust{} (top) and YTBB-Robust (bottom).}
\end{figure}

\begin{figure}[ht!]
\begin{subfigure}[b]{\width\linewidth}
  \includegraphics[width=\linewidth, height=\height]{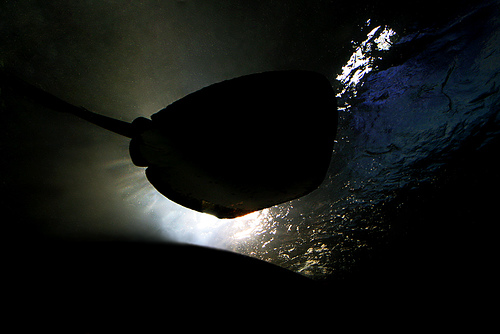}
\end{subfigure}
\hfill
\begin{subfigure}[b]{\width\linewidth}
  \includegraphics[width=\linewidth, height=\height]{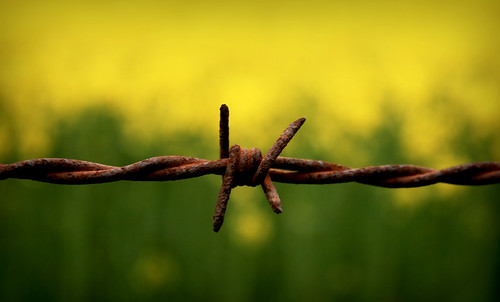}
\end{subfigure}
\hfill
\begin{subfigure}[b]{\width\linewidth}
  \includegraphics[width=\linewidth, height=\height]{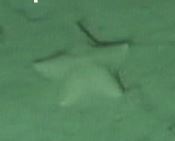}
\end{subfigure}
\hfill
\begin{subfigure}[b]{\width\linewidth}
  \includegraphics[width=\linewidth, height=\height]{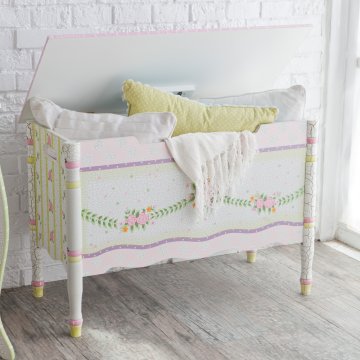}
\end{subfigure}
\caption{Adversarial shifts.
Examples from \imageneta{}.}
\end{figure}

\subsection{Synthetic distribution shift images}

\renewcommand\width{0.19}

\begin{figure}[h!]
\begin{subfigure}[b]{\width\textwidth}
  \includegraphics[width=\linewidth]{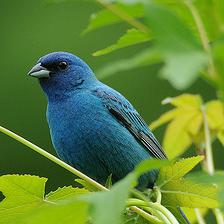} 
  \caption{original}
\end{subfigure}
\begin{subfigure}[b]{\width\textwidth}
  \includegraphics[width=\linewidth]{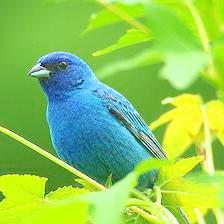} 
  \caption{brightness}
\end{subfigure}
\begin{subfigure}[b]{\width\textwidth}
\includegraphics[width=\linewidth]{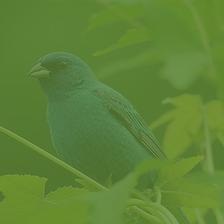} 
  \caption{contrast}
\end{subfigure}
\begin{subfigure}[b]{\width\textwidth}
\includegraphics[width=\linewidth]{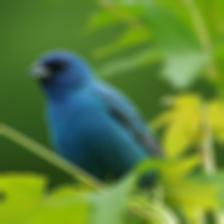}
  \caption{defocus blur}
\end{subfigure}
\begin{subfigure}[b]{\width\textwidth}
\includegraphics[width=\linewidth]{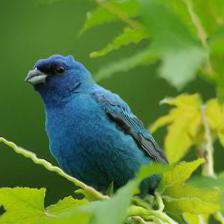}
  \caption{elastic transform}
\end{subfigure}
\begin{subfigure}[b]{\width\textwidth}
\includegraphics[width = \linewidth]{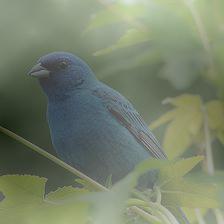} 
  \caption{fog}
\end{subfigure}
\begin{subfigure}[b]{\width\textwidth}
\includegraphics[width = \linewidth]{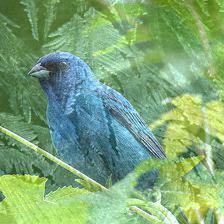} 
  \caption{frost}
\end{subfigure}
\begin{subfigure}[b]{\width\textwidth}
\includegraphics[width = \linewidth]{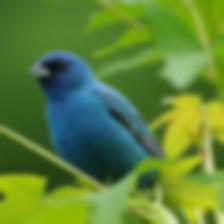} 
  \caption{gaussian blur}
\end{subfigure}
\begin{subfigure}[b]{\width\textwidth}
\includegraphics[width = \linewidth]{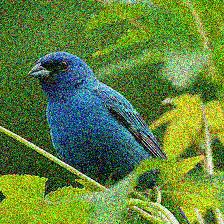}
  \caption{gaussian noise}
\end{subfigure}
\begin{subfigure}[b]{\width\textwidth}
\includegraphics[width = \linewidth]{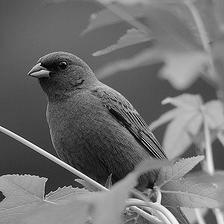} 
  \caption{greyscale}
\end{subfigure}
\begin{subfigure}[b]{\width\textwidth}
\includegraphics[width = \linewidth]{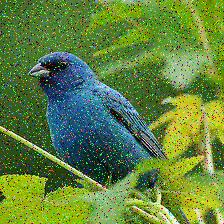} 
  \caption{impulse noise}
\end{subfigure}
\begin{subfigure}[b]{\width\textwidth}
\includegraphics[width = \linewidth]{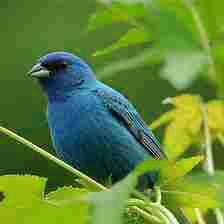}
  \caption{jpeg compression}
\end{subfigure}
\begin{subfigure}[b]{\width\textwidth}
\includegraphics[width = \linewidth]{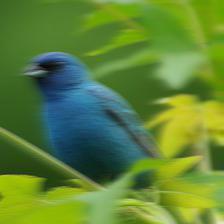}
  \caption{motion blur}
\end{subfigure}
\begin{subfigure}[b]{\width\textwidth}
\includegraphics[width = \linewidth]{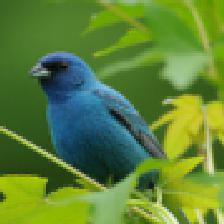}
  \caption{pixelate}
\end{subfigure}
\begin{subfigure}[b]{\width\textwidth}
\includegraphics[width = \linewidth]{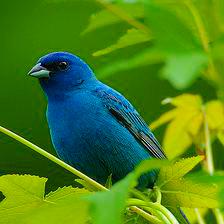} 
  \caption{saturate}
\end{subfigure}
\begin{subfigure}[b]{\width\textwidth}
\includegraphics[width = \linewidth]{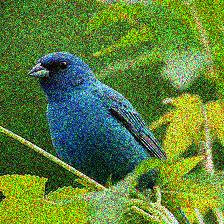}
  \caption{shot noise}
\end{subfigure}
\begin{subfigure}[b]{\width\textwidth}
\includegraphics[width = \linewidth]{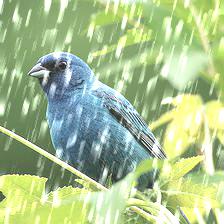}
  \caption{snow}
\end{subfigure}
\begin{subfigure}[b]{\width\textwidth}
\includegraphics[width = \linewidth]{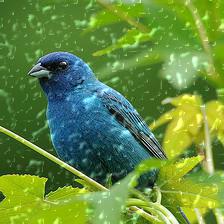} 
  \caption{spatter}
\end{subfigure}
\begin{subfigure}[b]{\width\textwidth}
\includegraphics[width = \linewidth]{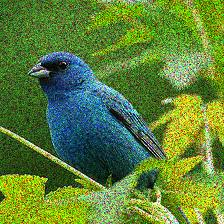} 
  \caption{speckle noise}
\end{subfigure}
\begin{subfigure}[b]{\width\textwidth}
\includegraphics[width = \linewidth]{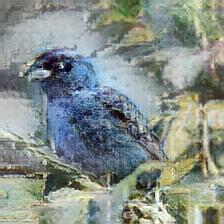}
  \caption{stylized imagenet}
\end{subfigure}
\begin{subfigure}[b]{\width\textwidth}
\includegraphics[width = \linewidth]{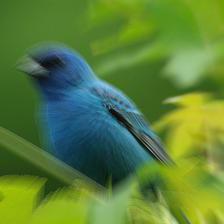}
  \caption{zoom blur}
\end{subfigure}
\begin{subfigure}[b]{\width\textwidth}
\includegraphics[width = \linewidth]{images/bird.jpeg}
  \caption{$\ell_p$-attack}
\end{subfigure}
\caption{
Sample demonstration of the synthetic distribution shifts in our testbed.
Note: This list is not complete. See \Cref{app:eval-settings} for a complete list.
}
\end{figure}

\clearpage

\section{Additional discussion of related work}
\label{app:related-work}

\subsection{\imagenetr{} \& \imagenetsketch{}}
\label{app:imagenet-r}

Recently, \citet{hendrycks2020faces} studied robustness of classifiers to a new dataset that measure distribution shift, \imagenetr{}, along with a new data augmentation method, DeepAugment. 
The authors make a number of comparisons in relation to an earlier version of this manuscript \cite{taori2019robustness}. 
In order to provide more clarity, we integrate the \imagenetr{} dataset and the DeepAugment models into our testbed in this paper.

\begin{figure*}[ht!]
    \centering
  \begin{subfigure}[t]{0.48\textwidth}
    \includegraphics[width=\linewidth]{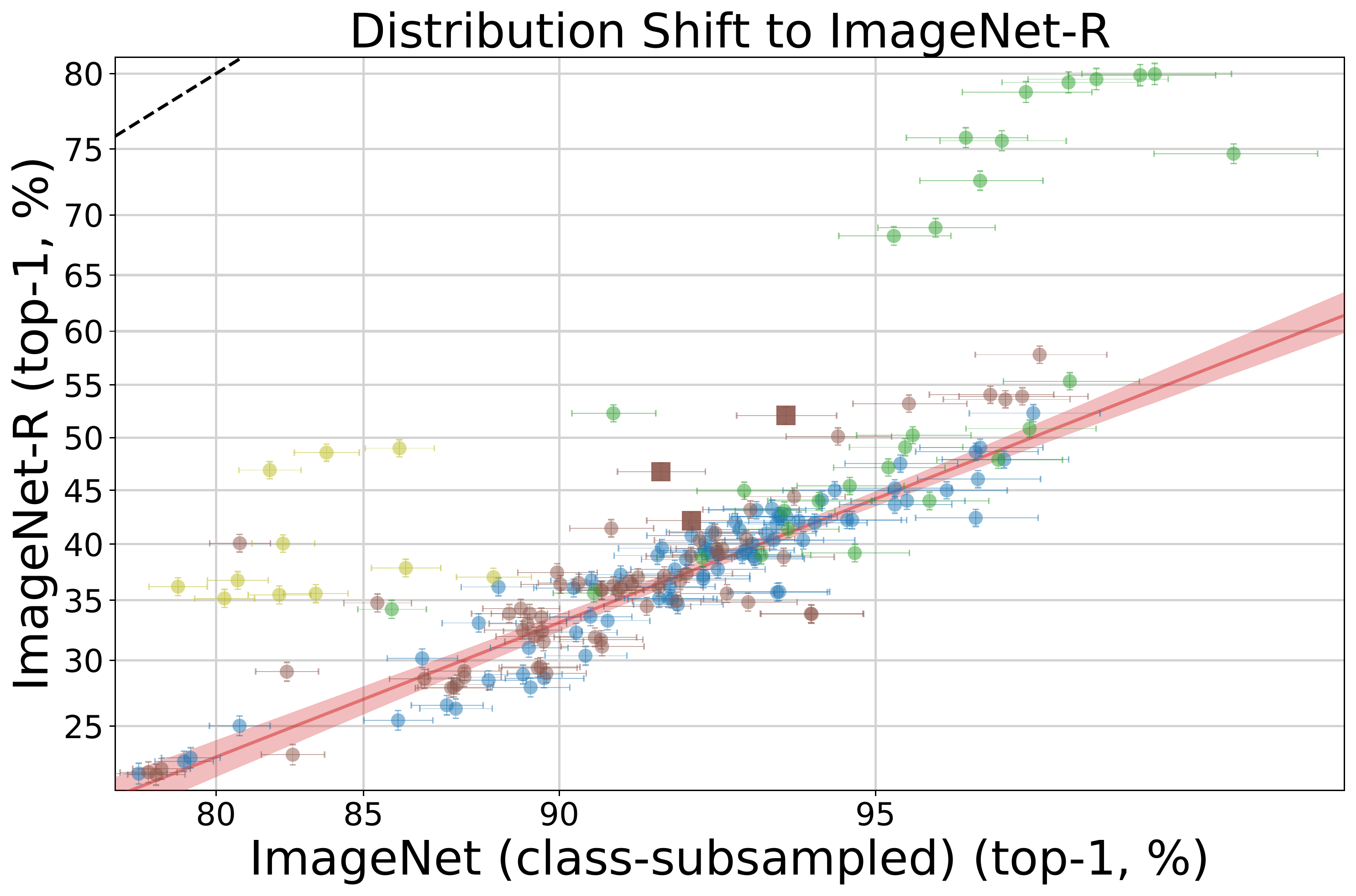}
  \end{subfigure}
  \hfill
  \begin{subfigure}[t]{0.48\textwidth}
    \includegraphics[width=\linewidth]{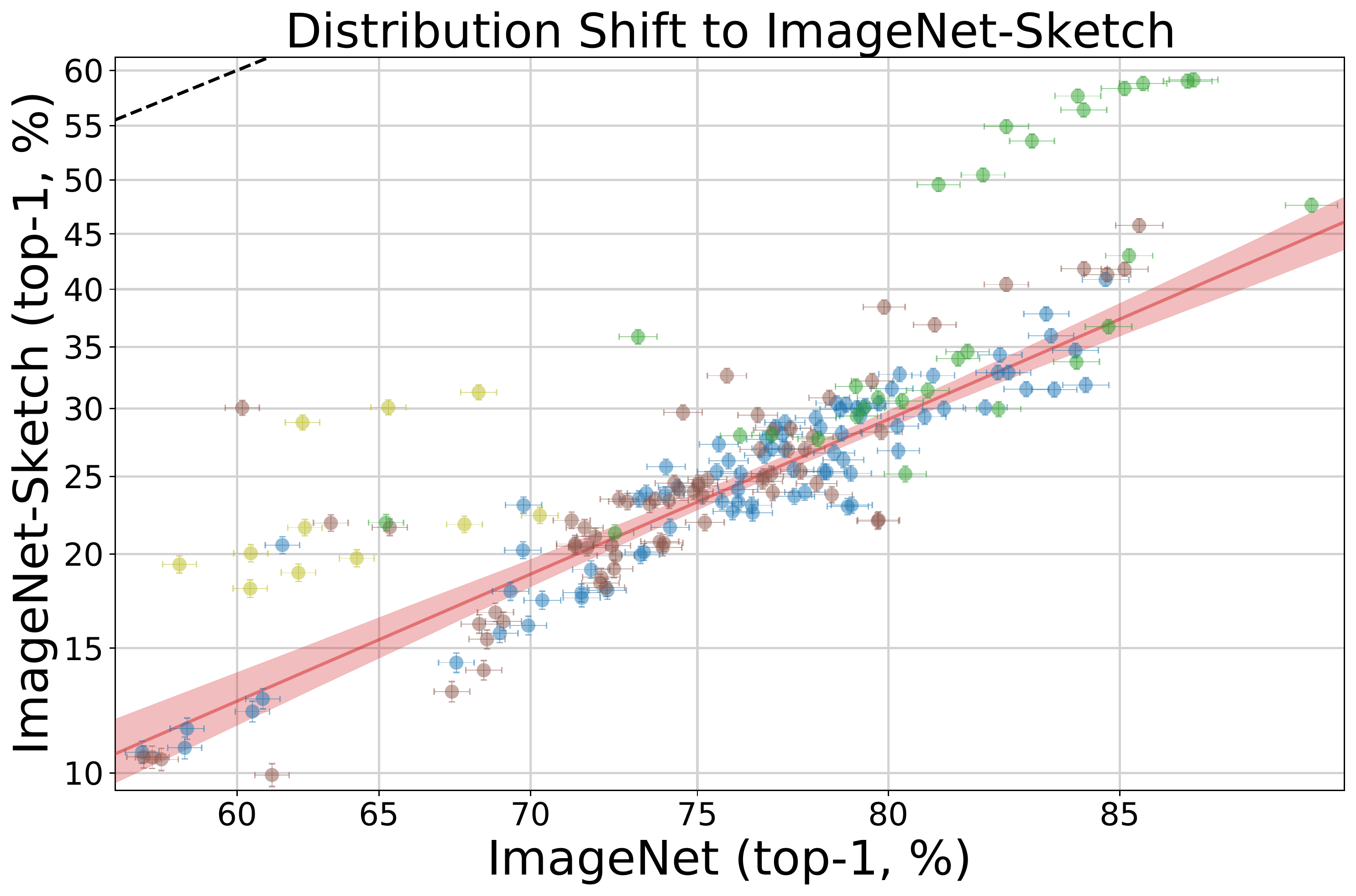}
  \end{subfigure}
  \begin{subfigure}[t]{\textwidth}
    \centering \includegraphics[width=0.9\textwidth]{figs/consistency_legend}
  \end{subfigure}
  \caption{
    Model accuracies on two datasets: \imagenetr{} (left), and \imagenetsketch{} (right).
    Both datasets create a distribution shift by selectively sampling images of renditions or sketches, respectively.
    Evaluations on these distribution shifts are similar to each other and follow the high-level trends of the other natural dataset distribution shifts in our testbed, with models trained on extra data providing the most robustness (though the effect is not uniform).
    On the left plot, DeepAugment models are highlighted in dark brown squares, 
    and ImageNet classes were subsampled to match the class distribution of \imagenetr{};
    the \imagenetsketch{} class distribution already matches ImageNet.
    Confidence intervals, axis scaling, and the linear fit are computed similarly to Figure \ref{fig:dataset-shifts}.
    }
  \label{fig:imagenet-r}
\end{figure*}

\textbf{ImageNet-R.} 
In \Cref{fig:imagenet-r}, we plot model accuracies on \imagenetr (endition) \cite{hendrycks2020faces} and a similar dataset of sketches, \imagenetsketch{} \cite{wang2019learning}. 
We find that a few of the models trained on more data substantially outperform the rest.
The top-right green cluster on both plots consists of several ResNet and ResNeXt models trained on 1 billion Instagram images \citep{wslimageseccv2018,yalniz2019billionscale,touvron2019FixRes} and EfficientNet-L2 (NoisyStudent) trained on the \jft{} dataset of 300 million images \citep{xie2019selftraining}.

However, as with the other dataset shifts, not all models trained on more data follow this trend.
Several ResNet models trained on either the YFCC 100 Million images dataset \cite{yalniz2019billionscale} or the full ImageNet 21k-class dataset \cite{alex2019big} have close to zero or negative effective robustness.

When interpreting the results of models trained on more data, a caveat is that the extra training data may contain renditions that do not occur in ImageNet.
To clarify this point, we have reached out to the authors of \cite{wslimageseccv2018} to obtain more information about the Instagram dataset.
We will update our paper when sufficient data becomes available to estimate the relative frequency of renditions in the Instagram dataset.
In the meantime, we note that the performance of the Instagram-trained models gives an answer to a question between the following two extremes:
(i) How much performance on \imagenetr{} do current models gain from a large, uncurated set of social media images that contains renditions?
(ii) How much robustness to \imagenetr{} do current models gain from a large, uncurated set of social media images that contains little to no renditions?

Interestingly, we also find that a number of $\ell_p$-adversarially robust models provide substantial effective and relative robustness on both datasets.
The top-left cluster of three yellow points on both plots are feature denoising models trained by \citet{xie2019feature}.
These results suggest that adversarial robustness and denoising blocks can be viable approaches for distribution shift comprised of renditions or sketches.

\begin{figure*}[t!]
    \centering
  \begin{subfigure}[t]{0.48\textwidth}
    \includegraphics[width=\linewidth]{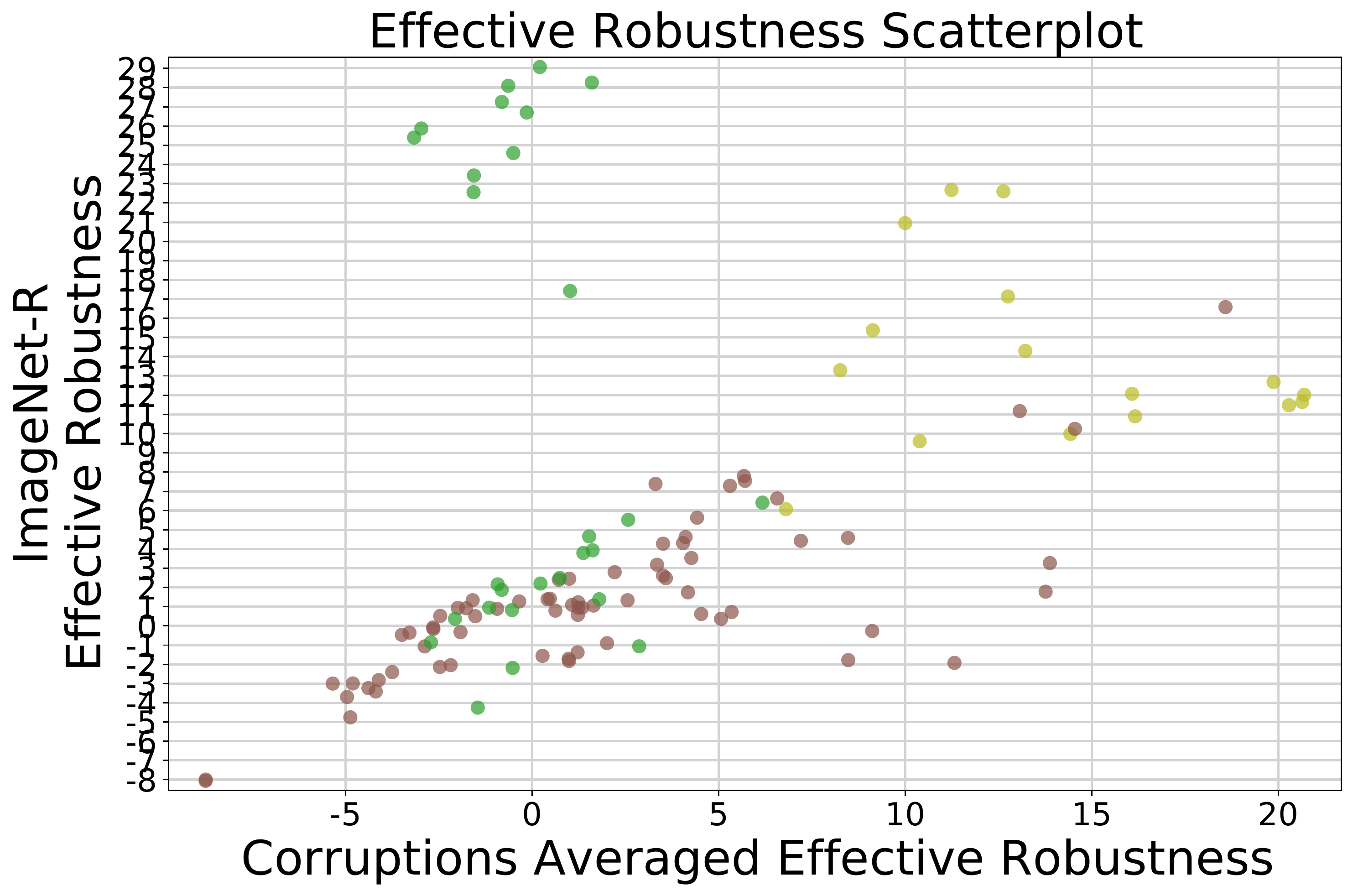}
  \end{subfigure}
  \hfill
  \begin{subfigure}[t]{0.48\textwidth}
    \includegraphics[width=\linewidth]{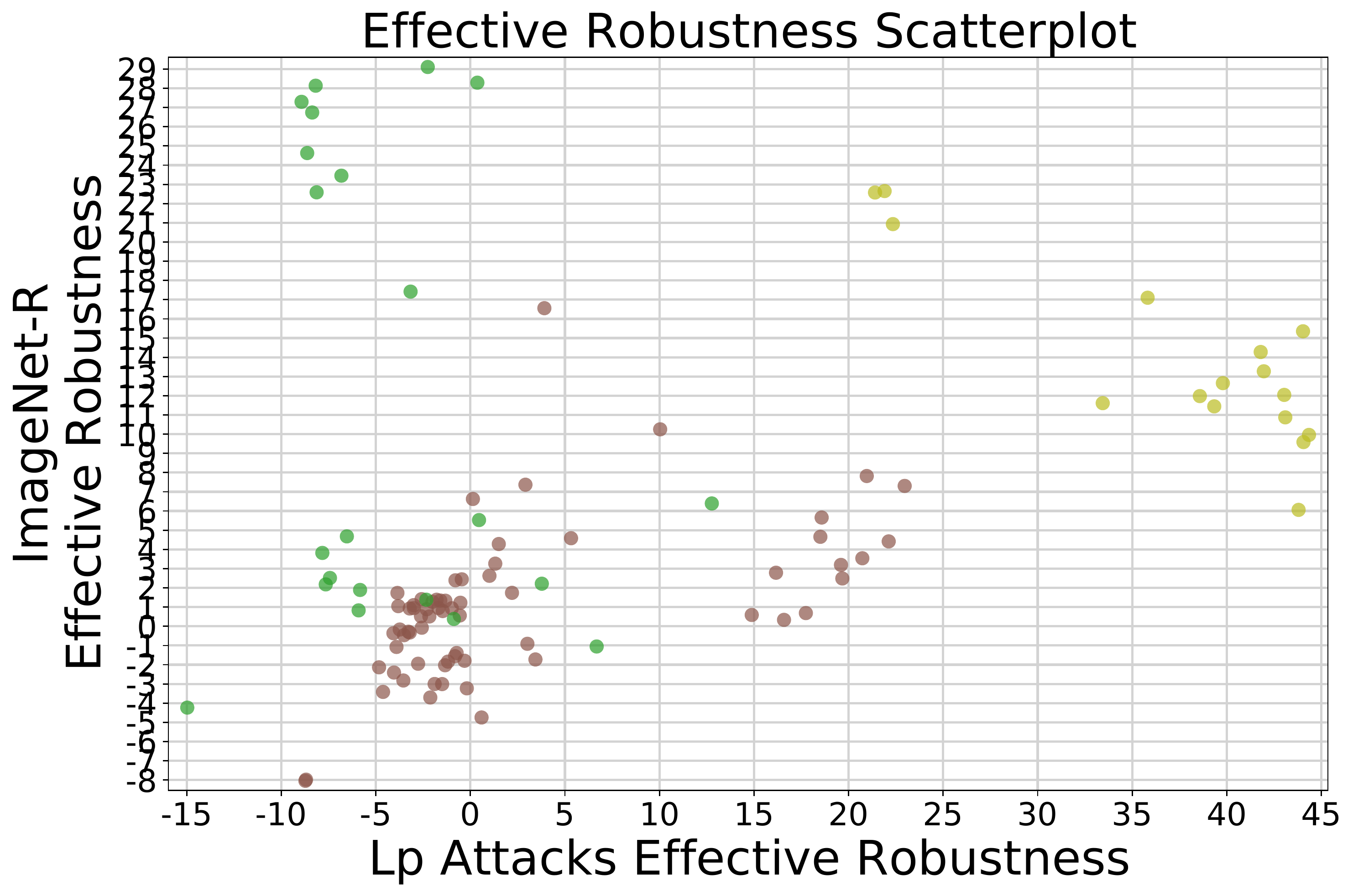}
  \end{subfigure}
  \begin{subfigure}[t]{\textwidth}
    \centering \includegraphics[width=0.9\textwidth]{appendix/eff_robust_legend2}
  \end{subfigure}
  \caption{
  We compare the effective robustness of models with their accuracy drop due to corruptions (left) and adversarial attacks (right). 
  The effective robustness is computed with respect to the linear fit on \imagenetr{}. 
  The measures are weakly correlated, indicating that improved robustness to corruptions or adversarial attacks does not in general improve effective robustness under \imagenetr{}.
    }
  \label{fig:imagenet-r-eff-robust}
\end{figure*}

A natural follow-up question is whether synthetic robustness is correlated with \imagenetr{} robustness. 
Similar to \Cref{app:eff-robust-syn-nat-robustness}, in \Cref{fig:imagenet-r-eff-robust} we compare effective robustness on synthetic distribution shifts against effective robustness on \imagenetr{}.
The scatter plots are weakly correlated (the Pearson correlation coefficients are $r = 0.35, 0.30$), indicating that improved robustness to corruptions or adversarial attacks in general does not improve effective robustness on \imagenetr{}. 
However, there does appear to be a strong trend for the brown points in the left plot.
Indeed, the correlation coefficient computed for only the ``other robustness intervention'' models is $r = 0.76$, 
suggesting that for this category of models, image corruptions robustness is well correlated with \imagenetr{} robustness.

\textbf{DeepAugment.}
\citet{hendrycks2020faces} also introduce a new data augmentation method, DeepAugment, which generates data augmentations by distorting the weights and activations of an image-to-image translation network.
As seen in \Cref{fig:imagenet-r}, the DeepAugment+Augmix models, the top two dark brown squares on the left plot, have higher effective robustness on \imagenetr{} than most other models ($\rho = 11.2\%$ for ResNeXt101 and $\rho = 10.2\%$ for ResNet50).

As mentioned above, some of the models trained on the large Instagram and JFT-300M datasets outperform all other approaches on \imagenetr{} including DeepAugment, but it is unclear how many images of renditions these datasets contain.
Among the other models trained only on ImageNet, the comparison between $\ell_p$-adversarial robustness and DeepAugment is nuanced.
The $\ell_p$-robust model of \cite{xie2019feature} has higher effective robustness but reduces standard ImageNet accuracy.
The highest accuracy on \imagenetr{} is also achieved by a model with an $\ell_p$-based robustness intervention (an AdvProp model \cite{xie2019adversarial}), but the model is derived from EfficientNet \cite{tan2019efficientnet} which achieves higher standard accuracy than the wide ResNeXt model \cite{xie2016aggregated} used in DeepAugment.
An interesting question for future work is how and why $\ell_p$-robustness helps on \imagenetr, e.g., by training a ResNeXt model with AdvProp.

On the anchor frames of \imagenetvidrobust{} and \ytbbrobust, DeepAgument provides effective robustness comparable to models trained on multiple synthetic perturbations (e.g., a combination of Gaussian noise, contrast, motion blur, and JPEG compression).
On \imagenetnew{} and \objectnet{}, DeepAugment does not provide effective robustness.

For ImageNet-C corruptions \cite{hendrycks2019benchmarking}, the combination of DeepAugment and AugMix offers substantial robustness.
Excluding $\ell_p$-adversarially robust models in the low accuracy regime,
the two models with highest effective robustness to ImageNet-C corruptions are DeepAugment- and Augmix- trained ResNet50 ($\rho = 14.2$\%) and ResNeXt101 ($\rho = 13.2$\%).
To put this in context, a ResNet50 trained directly on four of the ImageNet-C corruptions (Gaussian noise, contrast, motion blur, and JPEG compression) achieves an effective robustness of $\rho = 13.3$\%.

\subsection{Video robustness}
\label{app:video-robustness}

In the context of video robustness, \citet{gu2019using} have previously measured the performance of image classifiers on video sequences from the YouTube-BoundingBoxes(YT-BB) dataset \cite{real2017youtube}.
They find that video robustness is strongly correlated with accuracy on color corruptions such as brightness, hue, and saturation, with correlation coefficients $r$ near $0.95$.
There are two potential reasons our testbed finds these measures to be only weakly correlated with robustness on \ytbbrobust{} and \imagenetvidrobust{} ($r$ ranging from $0.1$ to $0.5$ - full table in \Cref{app:correlation}):

\textbf{Standard accuracy as a confounder.}
The analysis in \cite{gu2019using} does not account for standard accuracy as a confounder. The authors consider video robustness as accuracy within k frames of an anchor frame given that the anchor frame was correctly classified. While this definition does somewhat account for models with higher standard accuracies, it is natural to expect that models with higher standard accuracy are still more likely to predict the neighboring frames correctly given that the anchor frame was predicted correctly. Thus, standard accuracy will be correlated with video robustness. Moreover, our testbed reveals that standard accuracy is also correlated with corruption accuracy, and hence corruption accuracy will be correlated with video robustness as well. 

Additionally, it is worth noting that correlation does not mean that robustness to color corruptions \textit{cause} robustness on videos. For instance, our testbed contains a model trained on saturation as data augmentation. While this model is highly robust to saturation (exhibiting only a 1\% drop from standard accuracy to accuracy under saturations, compared to a baseline model exhibiting a 12\% drop), it is no more video robust than a baseline without the saturation training (the saturation-trained model still experiences an 18\% video robustness drop, compared to a baseline model exhibiting a 19\% drop). This example further shows the need for our measure of effective robustness as it explicitly corrects for the confounding effect of standard accuracy.

\textbf{Differences in data preparation.} 
\citet{gu2019using} split the full YT-BB dataset into training, validation, and testing splits, and evaluate \imagenet{} models on sequences from the test split.
In contrast, \citet{shankar2019systematic} derive datasets from ImageNet-Vid and Youtube-BB through a rigorous cleaning process: inspecting and annotating each sequence with human experts to check that subjects appear in frame throughout the sequence, match the correct class, and are not very blurry. This cleaning process indicates the derived datasets (\imagenetvidrobust{} and \ytbbrobust{}) are better calibrated to measuring classification performance.

%% file: appendix/model_list.tex
\begin{enumerate}
\item \model{BiT-M-R50x1-ILSVRC2012} \citep{alex2019big}. Trained with more data model. \footnotesize{\url{https://github.com/google-research/big_transfer}} 
\item \model{BiT-M-R50x3-ILSVRC2012} \citep{alex2019big}. Trained with more data model. \footnotesize{\url{https://github.com/google-research/big_transfer}} 
\item \model{BiT-M-R101x1-ILSVRC2012} \citep{alex2019big}. Trained with more data model. \footnotesize{\url{https://github.com/google-research/big_transfer}} 
\item \model{BiT-M-R101x3-ILSVRC2012} \citep{alex2019big}. Trained with more data model. \footnotesize{\url{https://github.com/google-research/big_transfer}} 
\item \model{BiT-M-R152x4-ILSVRC2012} \citep{alex2019big}. Trained with more data model. \footnotesize{\url{https://github.com/google-research/big_transfer}} 
\item \model{FixPNASNet} \citep{touvron2019FixRes}. Standard training model. \footnotesize{\url{https://github.com/facebookresearch/FixRes}} 
\item \model{FixResNeXt101\_32x48d} \citep{touvron2019FixRes}. Trained with more data model. \footnotesize{\url{https://github.com/facebookresearch/FixRes}} 
\item \model{FixResNeXt101\_32x48d\_v2} \citep{touvron2019FixRes}. Trained with more data model. \footnotesize{\url{https://github.com/facebookresearch/FixRes}} 
\item \model{FixResNet50} \citep{touvron2019FixRes}. Standard training model. \footnotesize{\url{https://github.com/facebookresearch/FixRes}} 
\item \model{FixResNet50CutMix} \citep{touvron2019FixRes}. Robustness intervention model. \footnotesize{\url{https://github.com/facebookresearch/FixRes}} 
\item \model{FixResNet50CutMix\_v2} \citep{touvron2019FixRes}. Robustness intervention model. \footnotesize{\url{https://github.com/facebookresearch/FixRes}} 
\item \model{FixResNet50\_no\_adaptation} \citep{touvron2019FixRes}. Standard training model. \footnotesize{\url{https://github.com/facebookresearch/FixRes}} 
\item \model{FixResNet50\_v2} \citep{touvron2019FixRes}. Standard training model. \footnotesize{\url{https://github.com/facebookresearch/FixRes}} 
\item \model{alexnet} \citep{alexnet}. Standard training model. \footnotesize{\url{https://github.com/Cadene/pretrained-models.pytorch}} 
\item \model{alexnet\_lpf2} \citep{zhang2019making}. Robustness intervention model. \footnotesize{\url{https://github.com/adobe/antialiased-cnns}} 
\item \model{alexnet\_lpf3} \citep{zhang2019making}. Robustness intervention model. \footnotesize{\url{https://github.com/adobe/antialiased-cnns}} 
\item \model{alexnet\_lpf5} \citep{zhang2019making}. Robustness intervention model. \footnotesize{\url{https://github.com/adobe/antialiased-cnns}} 
\item \model{bninception} \citep{ioffe2015batch}. Standard training model. \footnotesize{\url{https://github.com/Cadene/pretrained-models.pytorch}} 
\item \model{bninception-imagenet21k} \citep{ioffe2015batch}. Trained with more data model. \footnotesize{\url{https://github.com/dmlc/mxnet-model-gallery/blob/master/imagenet-21k-inception.md}} 
\item \model{cafferesnet101} \citep{resnet}. Standard training model. \footnotesize{\url{https://github.com/Cadene/pretrained-models.pytorch}} 
\item \model{densenet121} \citep{huang2016densely}. Standard training model. \footnotesize{\url{https://github.com/Cadene/pretrained-models.pytorch}} 
\item \model{densenet121\_lpf2} \citep{zhang2019making}. Robustness intervention model. \footnotesize{\url{https://github.com/adobe/antialiased-cnns}} 
\item \model{densenet121\_lpf3} \citep{zhang2019making}. Robustness intervention model. \footnotesize{\url{https://github.com/adobe/antialiased-cnns}} 
\item \model{densenet121\_lpf5} \citep{zhang2019making}. Robustness intervention model. \footnotesize{\url{https://github.com/adobe/antialiased-cnns}} 
\item \model{densenet161} \citep{huang2016densely}. Standard training model. \footnotesize{\url{https://github.com/Cadene/pretrained-models.pytorch}} 
\item \model{densenet169} \citep{huang2016densely}. Standard training model. \footnotesize{\url{https://github.com/Cadene/pretrained-models.pytorch}} 
\item \model{densenet201} \citep{huang2016densely}. Standard training model. \footnotesize{\url{https://github.com/Cadene/pretrained-models.pytorch}} 
\item \model{dpn107} \citep{chen2017dual}. Trained with more data model. \footnotesize{\url{https://github.com/Cadene/pretrained-models.pytorch}} 
\item \model{dpn131} \citep{chen2017dual}. Standard training model. \footnotesize{\url{https://github.com/Cadene/pretrained-models.pytorch}} 
\item \model{dpn68} \citep{chen2017dual}. Standard training model. \footnotesize{\url{https://github.com/Cadene/pretrained-models.pytorch}} 
\item \model{dpn68b} \citep{chen2017dual}. Trained with more data model. \footnotesize{\url{https://github.com/Cadene/pretrained-models.pytorch}} 
\item \model{dpn92} \citep{chen2017dual}. Trained with more data model. \footnotesize{\url{https://github.com/Cadene/pretrained-models.pytorch}} 
\item \model{dpn98} \citep{chen2017dual}. Standard training model. \footnotesize{\url{https://github.com/Cadene/pretrained-models.pytorch}} 
\item \model{efficientnet-b0} \citep{tan2019efficientnet}. Standard training model. \footnotesize{\url{https://github.com/tensorflow/tpu/tree/master/models/official/efficientnet}} 
\item \model{efficientnet-b0-advprop-autoaug} \citep{xie2019adversarial}. Robustness intervention model. \footnotesize{\url{https://github.com/tensorflow/tpu/tree/master/models/official/efficientnet}} 
\item \model{efficientnet-b0-autoaug} \citep{cubuk2018autoaugment}. Standard training model. \footnotesize{\url{https://github.com/tensorflow/tpu/tree/master/models/official/efficientnet}} 
\item \model{efficientnet-b1} \citep{tan2019efficientnet}. Standard training model. \footnotesize{\url{https://github.com/tensorflow/tpu/tree/master/models/official/efficientnet}} 
\item \model{efficientnet-b1-advprop-autoaug} \citep{xie2019adversarial}. Robustness intervention model. \footnotesize{\url{https://github.com/tensorflow/tpu/tree/master/models/official/efficientnet}} 
\item \model{efficientnet-b1-autoaug} \citep{cubuk2018autoaugment}. Standard training model. \footnotesize{\url{https://github.com/tensorflow/tpu/tree/master/models/official/efficientnet}} 
\item \model{efficientnet-b2} \citep{tan2019efficientnet}. Standard training model. \footnotesize{\url{https://github.com/tensorflow/tpu/tree/master/models/official/efficientnet}} 
\item \model{efficientnet-b2-advprop-autoaug} \citep{xie2019adversarial}. Robustness intervention model. \footnotesize{\url{https://github.com/tensorflow/tpu/tree/master/models/official/efficientnet}} 
\item \model{efficientnet-b2-autoaug} \citep{cubuk2018autoaugment}. Standard training model. \footnotesize{\url{https://github.com/tensorflow/tpu/tree/master/models/official/efficientnet}} 
\item \model{efficientnet-b3} \citep{tan2019efficientnet}. Standard training model. \footnotesize{\url{https://github.com/tensorflow/tpu/tree/master/models/official/efficientnet}} 
\item \model{efficientnet-b3-advprop-autoaug} \citep{xie2019adversarial}. Robustness intervention model. \footnotesize{\url{https://github.com/tensorflow/tpu/tree/master/models/official/efficientnet}} 
\item \model{efficientnet-b3-autoaug} \citep{cubuk2018autoaugment}. Standard training model. \footnotesize{\url{https://github.com/tensorflow/tpu/tree/master/models/official/efficientnet}} 
\item \model{efficientnet-b4} \citep{tan2019efficientnet}. Standard training model. \footnotesize{\url{https://github.com/tensorflow/tpu/tree/master/models/official/efficientnet}} 
\item \model{efficientnet-b4-advprop-autoaug} \citep{xie2019adversarial}. Robustness intervention model. \footnotesize{\url{https://github.com/tensorflow/tpu/tree/master/models/official/efficientnet}} 
\item \model{efficientnet-b4-autoaug} \citep{cubuk2018autoaugment}. Standard training model. \footnotesize{\url{https://github.com/tensorflow/tpu/tree/master/models/official/efficientnet}} 
\item \model{efficientnet-b5} \citep{tan2019efficientnet}. Standard training model. \footnotesize{\url{https://github.com/tensorflow/tpu/tree/master/models/official/efficientnet}} 
\item \model{efficientnet-b5-advprop-autoaug} \citep{xie2019adversarial}. Robustness intervention model. \footnotesize{\url{https://github.com/tensorflow/tpu/tree/master/models/official/efficientnet}} 
\item \model{efficientnet-b5-autoaug} \citep{cubuk2018autoaugment}. Standard training model. \footnotesize{\url{https://github.com/tensorflow/tpu/tree/master/models/official/efficientnet}} 
\item \model{efficientnet-b5-randaug} \citep{cubuk2019randaugment}. Standard training model. \footnotesize{\url{https://github.com/tensorflow/tpu/tree/master/models/official/efficientnet}} 
\item \model{efficientnet-b6-advprop-autoaug} \citep{xie2019adversarial}. Robustness intervention model. \footnotesize{\url{https://github.com/tensorflow/tpu/tree/master/models/official/efficientnet}} 
\item \model{efficientnet-b6-autoaug} \citep{cubuk2018autoaugment}. Standard training model. \footnotesize{\url{https://github.com/tensorflow/tpu/tree/master/models/official/efficientnet}} 
\item \model{efficientnet-b7-advprop-autoaug} \citep{xie2019adversarial}. Robustness intervention model. \footnotesize{\url{https://github.com/tensorflow/tpu/tree/master/models/official/efficientnet}} 
\item \model{efficientnet-b7-autoaug} \citep{cubuk2018autoaugment}. Standard training model. \footnotesize{\url{https://github.com/tensorflow/tpu/tree/master/models/official/efficientnet}} 
\item \model{efficientnet-b7-randaug} \citep{cubuk2019randaugment}. Standard training model. \footnotesize{\url{https://github.com/tensorflow/tpu/tree/master/models/official/efficientnet}} 
\item \model{efficientnet-b8-advprop-autoaug} \citep{xie2019adversarial}. Robustness intervention model. \footnotesize{\url{https://github.com/tensorflow/tpu/tree/master/models/official/efficientnet}} 
\item \model{efficientnet-l2-noisystudent} \citep{xie2019selftraining}. Trained with more data model. \footnotesize{\url{https://github.com/rwightman/pytorch-image-models}} 
\item \model{facebook\_adv\_trained\_resnet152\_baseline} \citep{xie2019feature}. Robustness intervention model. \footnotesize{\url{https://github.com/facebookresearch/ImageNet-Adversarial-Training}} 
\item \model{facebook\_adv\_trained\_resnet152\_denoise} \citep{xie2019feature}. Robustness intervention model. \footnotesize{\url{https://github.com/facebookresearch/ImageNet-Adversarial-Training}} 
\item \model{facebook\_adv\_trained\_resnext101\_denoiseAll} \citep{xie2019feature}. Robustness intervention model. \footnotesize{\url{https://github.com/facebookresearch/ImageNet-Adversarial-Training}} 
\item \model{fbresnet152} \citep{resnet}. Standard training model. \footnotesize{\url{https://github.com/Cadene/pretrained-models.pytorch}} 
\item \model{google\_resnet101\_jft-300M} \citep{sun2017revisiting}. Trained with more data model. \footnotesize{\url{}} 
\item \model{googlenet/inceptionv1} \citep{inceptionv1}. Standard training model. \footnotesize{\url{https://github.com/pytorch/vision/tree/master/torchvision/models}} 
\item \model{inceptionresnetv2} \citep{resnet}. Standard training model. \footnotesize{\url{https://github.com/Cadene/pretrained-models.pytorch}} 
\item \model{inceptionv3} \citep{szegedy2015rethinking}. Standard training model. \footnotesize{\url{https://github.com/Cadene/pretrained-models.pytorch}} 
\item \model{inceptionv4} \citep{szegedy2016inceptionv4}. Standard training model. \footnotesize{\url{https://github.com/Cadene/pretrained-models.pytorch}} 
\item \model{instagram-resnext101\_32x16d} \citep{wslimageseccv2018}. Trained with more data model. \footnotesize{\url{https://github.com/facebookresearch/WSL-Images}} 
\item \model{instagram-resnext101\_32x32d} \citep{wslimageseccv2018}. Trained with more data model. \footnotesize{\url{https://github.com/facebookresearch/WSL-Images}} 
\item \model{instagram-resnext101\_32x48d} \citep{wslimageseccv2018}. Trained with more data model. \footnotesize{\url{https://github.com/facebookresearch/WSL-Images}} 
\item \model{instagram-resnext101\_32x8d} \citep{wslimageseccv2018}. Trained with more data model. \footnotesize{\url{https://github.com/facebookresearch/WSL-Images}} 
\item \model{mnasnet0\_5} \citep{tan2018mnasnet}. Standard training model. \footnotesize{\url{https://github.com/pytorch/vision/tree/master/torchvision/models}} 
\item \model{mnasnet1\_0} \citep{tan2018mnasnet}. Standard training model. \footnotesize{\url{https://github.com/pytorch/vision/tree/master/torchvision/models}} 
\item \model{mobilenet\_v2} \citep{s2018mobilenetv2}. Standard training model. \footnotesize{\url{https://github.com/pytorch/vision/tree/master/torchvision/models}} 
\item \model{mobilenet\_v2\_lpf2} \citep{zhang2019making}. Robustness intervention model. \footnotesize{\url{https://github.com/adobe/antialiased-cnns}} 
\item \model{mobilenet\_v2\_lpf3} \citep{zhang2019making}. Robustness intervention model. \footnotesize{\url{https://github.com/adobe/antialiased-cnns}} 
\item \model{mobilenet\_v2\_lpf5} \citep{zhang2019making}. Robustness intervention model. \footnotesize{\url{https://github.com/adobe/antialiased-cnns}} 
\item \model{nasnetalarge} \citep{zoph2017learning}. Standard training model. \footnotesize{\url{https://github.com/Cadene/pretrained-models.pytorch}} 
\item \model{nasnetamobile} \citep{zoph2017learning}. Standard training model. \footnotesize{\url{https://github.com/Cadene/pretrained-models.pytorch}} 
\item \model{pnasnet5large} \citep{liu2017progressive}. Standard training model. \footnotesize{\url{https://github.com/Cadene/pretrained-models.pytorch}} 
\item \model{polynet} \citep{zhang2016polynet}. Standard training model. \footnotesize{\url{https://github.com/Cadene/pretrained-models.pytorch}} 
\item \model{resnet101} \citep{resnet}. Standard training model. \footnotesize{\url{https://github.com/Cadene/pretrained-models.pytorch}} 
\item \model{resnet101-tencent-ml-images} \citep{Wu_2019}. Trained with more data model. \footnotesize{\url{https://github.com/Tencent/tencent-ml-images}} 
\item \model{resnet101\_cutmix} \citep{yun2019cutmix}. Robustness intervention model. \footnotesize{\url{https://github.com/clovaai/CutMix-PyTorch}} 
\item \model{resnet101\_lpf2} \citep{zhang2019making}. Robustness intervention model. \footnotesize{\url{https://github.com/adobe/antialiased-cnns}} 
\item \model{resnet101\_lpf3} \citep{zhang2019making}. Robustness intervention model. \footnotesize{\url{https://github.com/adobe/antialiased-cnns}} 
\item \model{resnet101\_lpf5} \citep{zhang2019making}. Robustness intervention model. \footnotesize{\url{https://github.com/adobe/antialiased-cnns}} 
\item \model{resnet152} \citep{resnet}. Standard training model. \footnotesize{\url{https://github.com/Cadene/pretrained-models.pytorch}} 
\item \model{resnet152-imagenet11k} \citep{imagenet11kmodel}. Trained with more data model. \footnotesize{\url{https://github.com/tornadomeet/ResNet}} 
\item \model{resnet18} \citep{resnet}. Standard training model. \footnotesize{\url{https://github.com/Cadene/pretrained-models.pytorch}} 
\item \model{resnet18-rotation-nocrop\_40} \citep{engstrom2019exploring}. Robustness intervention model. \footnotesize{\url{https://github.com/MadryLab/spatial-pytorch}} 
\item \model{resnet18-rotation-random\_30} \citep{engstrom2019exploring}. Robustness intervention model. \footnotesize{\url{https://github.com/MadryLab/spatial-pytorch}} 
\item \model{resnet18-rotation-random\_40} \citep{engstrom2019exploring}. Robustness intervention model. \footnotesize{\url{https://github.com/MadryLab/spatial-pytorch}} 
\item \model{resnet18-rotation-standard\_40} \citep{engstrom2019exploring}. Robustness intervention model. \footnotesize{\url{https://github.com/MadryLab/spatial-pytorch}} 
\item \model{resnet18-rotation-worst10\_30} \citep{engstrom2019exploring}. Robustness intervention model. \footnotesize{\url{https://github.com/MadryLab/spatial-pytorch}} 
\item \model{resnet18-rotation-worst10\_40} \citep{engstrom2019exploring}. Robustness intervention model. \footnotesize{\url{https://github.com/MadryLab/spatial-pytorch}} 
\item \model{resnet18\_lpf2} \citep{zhang2019making}. Robustness intervention model. \footnotesize{\url{https://github.com/adobe/antialiased-cnns}} 
\item \model{resnet18\_lpf3} \citep{zhang2019making}. Robustness intervention model. \footnotesize{\url{https://github.com/adobe/antialiased-cnns}} 
\item \model{resnet18\_lpf5} \citep{zhang2019making}. Robustness intervention model. \footnotesize{\url{https://github.com/adobe/antialiased-cnns}} 
\item \model{resnet18\_ssl} \citep{yalniz2019billionscale}. Trained with more data model. \footnotesize{\url{https://github.com/facebookresearch/semi-supervised-ImageNet1K-models}} 
\item \model{resnet18\_swsl} \citep{yalniz2019billionscale}. Trained with more data model. \footnotesize{\url{https://github.com/facebookresearch/semi-supervised-ImageNet1K-models}} 
\item \model{resnet34} \citep{resnet}. Standard training model. \footnotesize{\url{https://github.com/Cadene/pretrained-models.pytorch}} 
\item \model{resnet34\_lpf2} \citep{zhang2019making}. Robustness intervention model. \footnotesize{\url{https://github.com/adobe/antialiased-cnns}} 
\item \model{resnet34\_lpf3} \citep{zhang2019making}. Robustness intervention model. \footnotesize{\url{https://github.com/adobe/antialiased-cnns}} 
\item \model{resnet34\_lpf5} \citep{zhang2019making}. Robustness intervention model. \footnotesize{\url{https://github.com/adobe/antialiased-cnns}} 
\item \model{resnet50} \citep{resnet}. Standard training model. \footnotesize{\url{https://github.com/Cadene/pretrained-models.pytorch}} 
\item \model{resnet50-randomized\_smoothing\_noise\_0.00} \citep{cohen2019certified}. Standard training model. \footnotesize{\url{https://github.com/locuslab/smoothing}} 
\item \model{resnet50-randomized\_smoothing\_noise\_0.25} \citep{cohen2019certified}. Robustness intervention model. \footnotesize{\url{https://github.com/locuslab/smoothing}} 
\item \model{resnet50-randomized\_smoothing\_noise\_0.50} \citep{cohen2019certified}. Robustness intervention model. \footnotesize{\url{https://github.com/locuslab/smoothing}} 
\item \model{resnet50-randomized\_smoothing\_noise\_1.00} \citep{cohen2019certified}. Robustness intervention model. \footnotesize{\url{https://github.com/locuslab/smoothing}} 
\item \model{resnet50-smoothing\_adversarial\_DNN\_2steps\_eps\_512\_noise\_0.25} \citep{salman2019provably}. Robustness intervention model. \footnotesize{\url{https://github.com/Hadisalman/smoothing-adversarial}} 
\item \model{resnet50-smoothing\_adversarial\_DNN\_2steps\_eps\_512\_noise\_0.50} \citep{salman2019provably}. Robustness intervention model. \footnotesize{\url{https://github.com/Hadisalman/smoothing-adversarial}} 
\item \model{resnet50-smoothing\_adversarial\_DNN\_2steps\_eps\_512\_noise\_1.00} \citep{salman2019provably}. Robustness intervention model. \footnotesize{\url{https://github.com/Hadisalman/smoothing-adversarial}} 
\item \model{resnet50-smoothing\_adversarial\_PGD\_1step\_eps\_512\_noise\_0.25} \citep{salman2019provably}. Robustness intervention model. \footnotesize{\url{https://github.com/Hadisalman/smoothing-adversarial}} 
\item \model{resnet50-smoothing\_adversarial\_PGD\_1step\_eps\_512\_noise\_0.50} \citep{salman2019provably}. Robustness intervention model. \footnotesize{\url{https://github.com/Hadisalman/smoothing-adversarial}} 
\item \model{resnet50-smoothing\_adversarial\_PGD\_1step\_eps\_512\_noise\_1.00} \citep{salman2019provably}. Robustness intervention model. \footnotesize{\url{https://github.com/Hadisalman/smoothing-adversarial}} 
\item \model{resnet50-vtab} \citep{zhai2019visual}. Standard training model. \footnotesize{\url{https://tfhub.dev/s?publisher=vtab}} 
\item \model{resnet50-vtab-exemplar} \citep{zhai2019visual}. Standard training model. \footnotesize{\url{https://tfhub.dev/s?publisher=vtab}} 
\item \model{resnet50-vtab-rotation} \citep{zhai2019visual}. Standard training model. \footnotesize{\url{https://tfhub.dev/s?publisher=vtab}} 
\item \model{resnet50-vtab-semi-exemplar} \citep{zhai2019visual}. Standard training model. \footnotesize{\url{https://tfhub.dev/s?publisher=vtab}} 
\item \model{resnet50-vtab-semi-rotation} \citep{zhai2019visual}. Standard training model. \footnotesize{\url{https://tfhub.dev/s?publisher=vtab}} 
\item \model{resnet50\_adv-train-free} \citep{shafahi2019adversarial}. Robustness intervention model. \footnotesize{\url{https://github.com/mahyarnajibi/FreeAdversarialTraining}} 
\item \model{resnet50\_augmix} \citep{hendrycks2019augmix}. Robustness intervention model. \footnotesize{\url{https://github.com/google-research/augmix}} 
\item \model{resnet50\_aws\_baseline}. Standard training model.  
\item \model{resnet50\_cutmix} \citep{yun2019cutmix}. Robustness intervention model. \footnotesize{\url{https://github.com/clovaai/CutMix-PyTorch}} 
\item \model{resnet50\_cutout} \citep{devries2017improved}. Robustness intervention model. \footnotesize{\url{https://github.com/clovaai/CutMix-PyTorch}} 
\item \model{resnet50\_deepaugment} \citep{hendrycks2020faces}. Robustness intervention model. \footnotesize{\url{https://github.com/hendrycks/imagenet-r}} 
\item \model{resnet50\_deepaugment\_augmix} \citep{hendrycks2020faces}. Robustness intervention model. \footnotesize{\url{https://github.com/hendrycks/imagenet-r}} 
\item \model{resnet50\_feature\_cutmix} \citep{yun2019cutmix}. Robustness intervention model. \footnotesize{\url{https://github.com/clovaai/CutMix-PyTorch}} 
\item \model{resnet50\_imagenet\_100percent\_batch64\_original\_images}. Standard training model.  
\item \model{resnet50\_imagenet\_subsample\_125\_classes\_batch64\_original\_images}. Standard training model.  
\item \model{resnet50\_imagenet\_subsample\_1\_of\_16\_batch64\_original\_images}. Standard training model.  
\item \model{resnet50\_imagenet\_subsample\_1\_of\_2\_batch64\_original\_images}. Standard training model.  
\item \model{resnet50\_imagenet\_subsample\_1\_of\_32\_batch64\_original\_images}. Standard training model.  
\item \model{resnet50\_imagenet\_subsample\_1\_of\_4\_batch64\_original\_images}. Standard training model.  
\item \model{resnet50\_imagenet\_subsample\_1\_of\_8\_batch64\_original\_images}. Standard training model.  
\item \model{resnet50\_imagenet\_subsample\_250\_classes\_batch64\_original\_images}. Standard training model.  
\item \model{resnet50\_imagenet\_subsample\_500\_classes\_batch64\_original\_images}. Standard training model.  
\item \model{resnet50\_l2\_eps3\_robust} \citep{robustness}. Robustness intervention model. \footnotesize{\url{https://github.com/MadryLab/robustness}} 
\item \model{resnet50\_linf\_eps4\_robust} \citep{robustness}. Robustness intervention model. \footnotesize{\url{https://github.com/MadryLab/robustness}} 
\item \model{resnet50\_linf\_eps8\_robust} \citep{robustness}. Robustness intervention model. \footnotesize{\url{https://github.com/MadryLab/robustness}} 
\item \model{resnet50\_lpf2} \citep{zhang2019making}. Robustness intervention model. \footnotesize{\url{https://github.com/adobe/antialiased-cnns}} 
\item \model{resnet50\_lpf3} \citep{zhang2019making}. Robustness intervention model. \footnotesize{\url{https://github.com/adobe/antialiased-cnns}} 
\item \model{resnet50\_lpf5} \citep{zhang2019making}. Robustness intervention model. \footnotesize{\url{https://github.com/adobe/antialiased-cnns}} 
\item \model{resnet50\_mixup} \citep{zhang2017mixup}. Robustness intervention model. \footnotesize{\url{https://github.com/clovaai/CutMix-PyTorch}} 
\item \model{resnet50\_ssl} \citep{yalniz2019billionscale}. Trained with more data model. \footnotesize{\url{https://github.com/facebookresearch/semi-supervised-ImageNet1K-models}} 
\item \model{resnet50\_swsl} \citep{yalniz2019billionscale}. Trained with more data model. \footnotesize{\url{https://github.com/facebookresearch/semi-supervised-ImageNet1K-models}} 
\item \model{resnet50\_trained\_on\_SIN} \citep{geirhos2019imagenettrained}. Robustness intervention model. \footnotesize{\url{https://github.com/rgeirhos/texture-vs-shape}} 
\item \model{resnet50\_trained\_on\_SIN\_and\_IN} \citep{geirhos2019imagenettrained}. Robustness intervention model. \footnotesize{\url{https://github.com/rgeirhos/texture-vs-shape}} 
\item \model{resnet50\_trained\_on\_SIN\_and\_IN\_then\_finetuned\_on\_IN} \citep{geirhos2019imagenettrained}. Robustness intervention model. \footnotesize{\url{https://github.com/rgeirhos/texture-vs-shape}} 
\item \model{resnet50\_with\_brightness\_aws}. Robustness intervention model.  
\item \model{resnet50\_with\_contrast\_aws}. Robustness intervention model.  
\item \model{resnet50\_with\_defocus\_blur\_aws}. Robustness intervention model.  
\item \model{resnet50\_with\_fog\_aws}. Robustness intervention model.  
\item \model{resnet50\_with\_frost\_aws}. Robustness intervention model.  
\item \model{resnet50\_with\_gaussian\_noise\_aws}. Robustness intervention model.  
\item \model{resnet50\_with\_gaussian\_noise\_contrast\_motion\_blur\_jpeg\_compression\_aws}. Robustness intervention model.  
\item \model{resnet50\_with\_greyscale\_aws}. Robustness intervention model.  
\item \model{resnet50\_with\_jpeg\_compression\_aws}. Robustness intervention model.  
\item \model{resnet50\_with\_motion\_blur\_aws}. Robustness intervention model.  
\item \model{resnet50\_with\_pixelate\_aws}. Robustness intervention model.  
\item \model{resnet50\_with\_saturate\_aws}. Robustness intervention model.  
\item \model{resnet50\_with\_spatter\_aws}. Robustness intervention model.  
\item \model{resnet50\_with\_zoom\_blur\_aws}. Robustness intervention model.  
\item \model{resnext101\_32x16d\_ssl} \citep{yalniz2019billionscale}. Trained with more data model. \footnotesize{\url{https://github.com/facebookresearch/semi-supervised-ImageNet1K-models}} 
\item \model{resnext101\_32x4d} \citep{xie2016aggregated}. Standard training model. \footnotesize{\url{https://github.com/Cadene/pretrained-models.pytorch}} 
\item \model{resnext101\_32x4d\_ssl} \citep{yalniz2019billionscale}. Trained with more data model. \footnotesize{\url{https://github.com/facebookresearch/semi-supervised-ImageNet1K-models}} 
\item \model{resnext101\_32x4d\_swsl} \citep{yalniz2019billionscale}. Trained with more data model. \footnotesize{\url{https://github.com/facebookresearch/semi-supervised-ImageNet1K-models}} 
\item \model{resnext101\_32x8d} \citep{xie2016aggregated}. Standard training model. \footnotesize{\url{https://github.com/pytorch/vision/tree/master/torchvision/models}} 
\item \model{resnext101\_32x8d\_deepaugment\_augmix} \citep{hendrycks2020faces}. Robustness intervention model. \footnotesize{\url{https://github.com/hendrycks/imagenet-r}} 
\item \model{resnext101\_32x8d\_ssl} \citep{yalniz2019billionscale}. Trained with more data model. \footnotesize{\url{https://github.com/facebookresearch/semi-supervised-ImageNet1K-models}} 
\item \model{resnext101\_32x8d\_swsl} \citep{yalniz2019billionscale}. Trained with more data model. \footnotesize{\url{https://github.com/facebookresearch/semi-supervised-ImageNet1K-models}} 
\item \model{resnext101\_64x4d} \citep{xie2016aggregated}. Standard training model. \footnotesize{\url{https://github.com/Cadene/pretrained-models.pytorch}} 
\item \model{resnext50\_32x4d} \citep{xie2016aggregated}. Standard training model. \footnotesize{\url{https://github.com/pytorch/vision/tree/master/torchvision/models}} 
\item \model{resnext50\_32x4d\_ssl} \citep{yalniz2019billionscale}. Trained with more data model. \footnotesize{\url{https://github.com/facebookresearch/semi-supervised-ImageNet1K-models}} 
\item \model{resnext50\_32x4d\_swsl} \citep{yalniz2019billionscale}. Trained with more data model. \footnotesize{\url{https://github.com/facebookresearch/semi-supervised-ImageNet1K-models}} 
\item \model{se\_resnet101} \citep{hu2017squeezeandexcitation}. Standard training model. \footnotesize{\url{https://github.com/Cadene/pretrained-models.pytorch}} 
\item \model{se\_resnet152} \citep{hu2017squeezeandexcitation}. Standard training model. \footnotesize{\url{https://github.com/Cadene/pretrained-models.pytorch}} 
\item \model{se\_resnet50} \citep{hu2017squeezeandexcitation}. Standard training model. \footnotesize{\url{https://github.com/Cadene/pretrained-models.pytorch}} 
\item \model{se\_resnext101\_32x4d} \citep{hu2017squeezeandexcitation}. Standard training model. \footnotesize{\url{https://github.com/Cadene/pretrained-models.pytorch}} 
\item \model{se\_resnext50\_32x4d} \citep{hu2017squeezeandexcitation}. Standard training model. \footnotesize{\url{https://github.com/Cadene/pretrained-models.pytorch}} 
\item \model{senet154} \citep{hu2017squeezeandexcitation}. Standard training model. \footnotesize{\url{https://github.com/Cadene/pretrained-models.pytorch}} 
\item \model{shufflenet\_v2\_x0\_5} \citep{ma2018shufflenet}. Standard training model. \footnotesize{\url{https://github.com/pytorch/vision/tree/master/torchvision/models}} 
\item \model{shufflenet\_v2\_x1\_0} \citep{ma2018shufflenet}. Standard training model. \footnotesize{\url{https://github.com/pytorch/vision/tree/master/torchvision/models}} 
\item \model{squeezenet1\_0} \citep{i2016squeezenet}. Standard training model. \footnotesize{\url{https://github.com/Cadene/pretrained-models.pytorch}} 
\item \model{squeezenet1\_1} \citep{i2016squeezenet}. Standard training model. \footnotesize{\url{https://github.com/Cadene/pretrained-models.pytorch}} 
\item \model{vgg11} \citep{vgg}. Standard training model. \footnotesize{\url{https://github.com/Cadene/pretrained-models.pytorch}} 
\item \model{vgg11\_bn} \citep{vgg}. Standard training model. \footnotesize{\url{https://github.com/Cadene/pretrained-models.pytorch}} 
\item \model{vgg13} \citep{vgg}. Standard training model. \footnotesize{\url{https://github.com/Cadene/pretrained-models.pytorch}} 
\item \model{vgg13\_bn} \citep{vgg}. Standard training model. \footnotesize{\url{https://github.com/Cadene/pretrained-models.pytorch}} 
\item \model{vgg16} \citep{vgg}. Standard training model. \footnotesize{\url{https://github.com/Cadene/pretrained-models.pytorch}} 
\item \model{vgg16\_bn} \citep{vgg}. Standard training model. \footnotesize{\url{https://github.com/Cadene/pretrained-models.pytorch}} 
\item \model{vgg16\_bn\_lpf2} \citep{zhang2019making}. Robustness intervention model. \footnotesize{\url{https://github.com/adobe/antialiased-cnns}} 
\item \model{vgg16\_bn\_lpf3} \citep{zhang2019making}. Robustness intervention model. \footnotesize{\url{https://github.com/adobe/antialiased-cnns}} 
\item \model{vgg16\_bn\_lpf5} \citep{zhang2019making}. Robustness intervention model. \footnotesize{\url{https://github.com/adobe/antialiased-cnns}} 
\item \model{vgg16\_lpf2} \citep{zhang2019making}. Robustness intervention model. \footnotesize{\url{https://github.com/adobe/antialiased-cnns}} 
\item \model{vgg16\_lpf3} \citep{zhang2019making}. Robustness intervention model. \footnotesize{\url{https://github.com/adobe/antialiased-cnns}} 
\item \model{vgg16\_lpf5} \citep{zhang2019making}. Robustness intervention model. \footnotesize{\url{https://github.com/adobe/antialiased-cnns}} 
\item \model{vgg19} \citep{vgg}. Standard training model. \footnotesize{\url{https://github.com/Cadene/pretrained-models.pytorch}} 
\item \model{vgg19\_bn} \citep{vgg}. Standard training model. \footnotesize{\url{https://github.com/Cadene/pretrained-models.pytorch}} 
\item \model{wide\_resnet101\_2} \citep{zagoruyko2016wide}. Standard training model. \footnotesize{\url{https://github.com/pytorch/vision/tree/master/torchvision/models}} 
\item \model{wide\_resnet50\_2} \citep{zagoruyko2016wide}. Standard training model. \footnotesize{\url{https://github.com/pytorch/vision/tree/master/torchvision/models}} 
\item \model{xception} \citep{chollet2016xception}. Standard training model. \footnotesize{\url{https://github.com/Cadene/pretrained-models.pytorch}} 

\end{enumerate}

%% file: appendix/model_table.tex
\foo{efficientnet-l2-noisystudent} & 88.32 & 1.11 &  &  \\ 
\foo{FixResNeXt101\_32x48d\_v2} & 86.36 & 0.97 & 65.65 &  \\ 
\foo{FixResNeXt101\_32x48d} & 86.26 & 0.95 & 65.56 &  \\ 
\foo{instagram-resnext101\_32x48d} & 85.44 & 1.26 & 65.53 & 24.1 \\ 
\foo{efficientnet-b8-advprop-autoaug} & 85.37 & 0.51 & 71.85 &  \\ 
\foo{BiT-M-R152x4-ILSVRC2012} & 85.18 & -0.31 & 67.26 &  \\ 
\foo{efficientnet-b7-advprop-autoaug} & 85.09 & 0.66 & 68.92 &  \\ 
\foo{instagram-resnext101\_32x32d} & 85.09 & 1.54 & 64.77 & 24.4 \\ 
\foo{BiT-M-R101x3-ILSVRC2012} & 84.78 & -1.35 & 63.44 &  \\ 
\foo{efficientnet-b6-advprop-autoaug} & 84.76 & 0.75 & 68.65 & 50.67 \\ 
\foo{efficientnet-b7-randaug} & 84.73 & 0.11 & 69.12 &  \\ 
\foo{efficientnet-b7-autoaug} & 84.33 & 0.32 & 62.77 &  \\ 
\foo{efficientnet-b5-advprop-autoaug} & 84.3 & 0.51 & 67.76 & 50.17 \\ 
\foo{resnext101\_32x8d\_swsl} & 84.29 & 1.19 & 63.17 & 23.22 \\ 
\foo{instagram-resnext101\_32x16d} & 84.18 & 1.51 & 63.22 & 29.19 \\ 
\foo{BiT-M-R50x3-ILSVRC2012} & 84.15 & -0.76 & 60.23 &  \\ 
\foo{efficientnet-b6-autoaug} & 84.13 & 0.14 & 63.42 & 34.29 \\ 
\foo{FixPNASNet} & 83.7 & -0.0 & 61.35 & 22.8 \\ 
\foo{efficientnet-b5-autoaug} & 83.63 & 0.25 & 62.3 & 32.43 \\ 
\foo{efficientnet-b5-randaug} & 83.53 & 0.08 & 63.35 & 34.41 \\ 
\foo{resnext101\_32x4d\_swsl} & 83.23 & 1.41 & 60.09 & 21.73 \\ 
\foo{efficientnet-b5} & 83.11 & 0.17 & 60.28 & 35.18 \\ 
\foo{pnasnet5large} & 82.74 & 0.21 & 61.76 & 29.46 \\ 
\foo{instagram-resnext101\_32x8d} & 82.69 & 1.59 & 60.81 & 30.13 \\ 
\foo{efficientnet-b4-advprop-autoaug} & 82.69 & 0.42 & 64.88 & 50.72 \\ 
\foo{efficientnet-b4-autoaug} & 82.55 & 0.17 & 59.59 & 34.24 \\ 
\foo{BiT-M-R101x1-ILSVRC2012} & 82.52 & -0.42 & 58.28 &  \\ 
\foo{nasnetalarge} & 82.51 & 0.48 & 61.74 & 36.99 \\ 
\foo{efficientnet-b4} & 82.23 & -0.64 & 57.2 & 37.06 \\ 
\foo{resnext50\_32x4d\_swsl} & 82.18 & 1.26 & 56.38 & 21.09 \\ 
\foo{resnext101\_32x16d\_ssl} & 81.84 & 0.3 & 58.63 & 22.34 \\ 
\foo{resnext101\_32x8d\_ssl} & 81.63 & 0.73 & 57.96 & 20.82 \\ 
\foo{senet154} & 81.3 & -0.07 & 54.11 & 30.65 \\ 
\foo{resnet50\_swsl} & 81.18 & 1.35 & 53.95 & 21.39 \\ 
\foo{efficientnet-b3-advprop-autoaug} & 81.09 & 0.29 & 60.6 & 51.09 \\ 
\foo{efficientnet-b3-autoaug} & 81.05 & 0.17 & 55.5 & 31.76 \\ 
\foo{resnext101\_32x4d\_ssl} & 80.93 & 0.48 & 55.65 & 20.54 \\ 
\foo{polynet} & 80.86 & 0.36 & 54.02 & 23.05 \\ 
\foo{BiT-M-R50x1-ILSVRC2012} & 80.4 & -0.63 & 52.21 & 12.5 \\ 
\foo{resnext50\_32x4d\_ssl} & 80.33 & 0.44 & 52.57 & 19.75 \\ 
\foo{inceptionresnetv2} & 80.27 & 0.32 & 56.85 & 34.85 \\ 
\foo{se\_resnext101\_32x4d} & 80.24 & 0.47 & 52.26 & 28.77 \\ 
\foo{efficientnet-b3} & 80.21 & -0.48 & 53.31 & 34.22 \\ 
\foo{inceptionv4} & 80.08 & 0.5 & 55.52 & 28.02 \\ 
\foo{resnext101\_32x8d\_deepaugment\_augmix} & 79.9 & 0.25 & 65.56 &  \\ 
\foo{resnet101\_cutmix} & 79.83 & -0.39 & 50.15 & 25.6 \\ 
\foo{efficientnet-b2-autoaug} & 79.78 & 0.17 & 53.5 & 30.93 \\ 
\foo{FixResNet50CutMix\_v2} & 79.76 & -1.21 & 43.44 & 18.19 \\ 
\foo{dpn107} & 79.75 & -0.47 & 52.37 & 30.64 \\ 
\foo{FixResNet50CutMix} & 79.74 & -1.22 & 43.39 & 18.14 \\ 
\foo{efficientnet-b2-advprop-autoaug} & 79.6 & -0.25 & 55.17 & 46.33 \\ 
\foo{dpn131} & 79.43 & -0.2 & 52.06 & 30.38 \\ 
\foo{dpn92} & 79.4 & -0.65 & 49.29 & 25.69 \\ 
\foo{resnext101\_32x8d} & 79.31 & -0.34 & 49.68 & 25.38 \\ 
\foo{resnet50\_ssl} & 79.23 & 0.52 & 50.15 & 20.57 \\ 
\foo{dpn98} & 79.22 & 0.08 & 51.82 & 30.14 \\ 
\foo{google\_resnet101\_jft-300M} & 79.2 & -0.23 & 53.49 & 26.84 \\ 
\foo{FixResNet50\_v2} & 79.1 & -0.62 & 43.31 & 15.38 \\ 
\foo{se\_resnext50\_32x4d} & 79.08 & 0.27 & 50.65 & 24.74 \\ 
\foo{FixResNet50} & 79.0 & -0.67 & 43.25 & 15.3 \\ 
\foo{resnext101\_64x4d} & 78.96 & -0.2 & 52.06 & 23.57 \\ 
\foo{efficientnet-b2} & 78.89 & -0.39 & 50.05 & 33.88 \\ 
\foo{wide\_resnet101\_2} & 78.85 & -0.87 & 48.2 & 25.24 \\ 
\foo{xception} & 78.82 & 0.06 & 51.7 & 26.32 \\ 
\foo{efficientnet-b1-autoaug} & 78.72 & -0.07 & 51.19 & 30.69 \\ 
\foo{se\_resnet152} & 78.66 & 0.45 & 50.94 & 28.42 \\ 
\foo{resnet50\_cutmix} & 78.6 & -1.1 & 44.7 & 26.46 \\ 
\foo{efficientnet-b1-advprop-autoaug} & 78.54 & -0.23 & 53.7 & 46.54 \\ 
\foo{wide\_resnet50\_2} & 78.47 & -0.61 & 46.23 & 26.13 \\ 
\foo{se\_resnet101} & 78.4 & 0.43 & 50.12 & 28.2 \\ 
\foo{resnet152} & 78.31 & 0.27 & 47.81 & 22.48 \\ 
\foo{resnet101-tencent-ml-images} & 78.25 & 0.04 & 47.77 &  \\ 
\foo{resnet50\_feature\_cutmix} & 78.21 & -0.42 & 44.33 & 25.36 \\ 
\foo{resnext101\_32x4d} & 78.19 & -0.13 & 50.96 & 22.38 \\ 
\foo{resnet101\_lpf3} & 78.12 & -0.27 & 46.52 & 22.48 \\ 
\foo{efficientnet-b1} & 77.91 & -0.24 & 47.07 & 31.33 \\ 
\foo{resnet101\_lpf5} & 77.91 & 0.1 & 46.54 & 23.13 \\ 
\foo{resnet101\_lpf2} & 77.8 & 0.3 & 46.06 & 22.01 \\ 
\foo{se\_resnet50} & 77.64 & 0.08 & 48.11 & 27.55 \\ 
\foo{resnext50\_32x4d} & 77.62 & 0.1 & 45.56 & 22.52 \\ 
\foo{resnet50\_augmix} & 77.54 & -0.53 & 50.78 & 26.01 \\ 
\foo{resnet50\_mixup} & 77.47 & -0.54 & 48.2 & 21.95 \\ 
\foo{fbresnet152} & 77.39 & 0.02 & 49.98 & 23.4 \\ 
\foo{resnet101} & 77.37 & 0.01 & 46.06 & 21.85 \\ 
\foo{inceptionv3} & 77.32 & 0.29 & 49.83 & 25.72 \\ 
\foo{densenet161} & 77.14 & 0.13 & 49.36 & 22.22 \\ 
\foo{efficientnet-b0-advprop-autoaug} & 77.08 & 0.21 & 49.9 & 44.31 \\ 
\foo{resnet50\_cutout} & 77.07 & -0.65 & 43.81 & 19.8 \\ 
\foo{FixResNet50\_no\_adaptation} & 77.04 & -0.02 & 44.68 & 20.61 \\ 
\foo{dpn68b} & 77.03 & -0.28 & 45.67 & 18.7 \\ 
\foo{resnet50\_lpf5} & 77.03 & -0.53 & 43.54 & 22.03 \\ 
\foo{densenet201} & 76.9 & -0.12 & 47.63 & 23.95 \\ 
\foo{efficientnet-b0-autoaug} & 76.84 & -0.39 & 45.27 & 30.66 \\ 
\foo{resnet50\_lpf3} & 76.82 & -0.12 & 43.3 & 21.83 \\ 
\foo{resnet50\_lpf2} & 76.79 & -0.25 & 42.22 & 20.91 \\ 
\foo{resnet50\_trained\_on\_SIN\_and\_IN\_then\_finetuned\_on\_IN} & 76.72 & -0.04 & 43.96 & 22.78 \\ 
\foo{resnet50\_deepaugment} & 76.66 & 0.73 & 53.91 & 29.65 \\ 
\foo{efficientnet-b0} & 76.53 & -0.79 & 43.84 & 31.05 \\ 
\foo{resnet50-vtab-rotation} & 76.5 & -0.49 & 41.93 &  \\ 
\foo{cafferesnet101} & 76.2 & 0.08 & 44.83 & 25.54 \\ 
\foo{resnet152-imagenet11k} & 76.18 & 2.09 & 47.33 & 30.64 \\ 
\foo{resnet50\_aws\_baseline} & 76.14 & -0.36 & 42.13 & 21.46 \\ 
\foo{resnet50} & 76.13 & -0.77 & 41.59 & 21.24 \\ 
\foo{resnet50\_imagenet\_100percent\_batch64\_original\_images} & 75.98 & -0.56 & 41.61 & 21.89 \\ 
\foo{dpn68} & 75.87 & -0.56 & 45.46 & 17.71 \\ 
\foo{resnet50\_deepaugment\_augmix} & 75.82 & -0.08 & 58.29 & 33.73 \\ 
\foo{resnet50-randomized\_smoothing\_noise\_0.00} & 75.69 & 0.31 & 41.75 & 21.32 \\ 
\foo{densenet169} & 75.6 & 0.19 & 46.67 & 21.79 \\ 
\foo{resnet50-vtab} & 75.54 & 0.22 & 43.61 &  \\ 
\foo{resnet50\_with\_brightness\_aws} & 75.28 & -0.28 & 43.9 & 22.78 \\ 
\foo{resnet50\_with\_spatter\_aws} & 75.21 & -0.29 & 42.81 & 22.45 \\ 
\foo{densenet121\_lpf3} & 75.14 & -0.35 & 40.48 & 20.01 \\ 
\foo{densenet121\_lpf5} & 75.03 & 0.13 & 41.84 & 21.13 \\ 
\foo{densenet121\_lpf2} & 75.03 & 0.41 & 41.24 & 20.82 \\ 
\foo{resnet50\_with\_saturate\_aws} & 74.89 & -0.27 & 42.4 & 20.46 \\ 
\foo{resnet50\_trained\_on\_SIN\_and\_IN} & 74.59 & 0.55 & 47.91 & 22.96 \\ 
\foo{resnet34\_lpf2} & 74.48 & 0.15 & 41.54 & 20.96 \\ 
\foo{densenet121} & 74.43 & 0.13 & 43.54 & 20.01 \\ 
\foo{resnet34\_lpf3} & 74.34 & 0.25 & 42.22 & 20.97 \\ 
\foo{vgg19\_bn} & 74.22 & 0.18 & 37.94 & 16.51 \\ 
\foo{resnet34\_lpf5} & 74.19 & 0.46 & 41.22 & 21.09 \\ 
\foo{resnet50-vtab-exemplar} & 74.1 & 0.3 & 44.73 &  \\ 
\foo{nasnetamobile} & 74.08 & -0.29 & 44.78 & 22.89 \\ 
\foo{vgg16\_bn\_lpf5} & 74.04 & -0.4 & 36.19 & 18.91 \\ 
\foo{vgg16\_bn\_lpf2} & 74.01 & 0.13 & 36.06 & 17.81 \\ 
\foo{vgg16\_bn\_lpf3} & 73.92 & 0.5 & 36.33 & 18.33 \\ 
\foo{resnet50\_with\_frost\_aws} & 73.78 & 0.29 & 42.39 & 20.96 \\ 
\foo{resnet50\_with\_jpeg\_compression\_aws} & 73.63 & -0.21 & 41.76 & 38.34 \\ 
\foo{bninception} & 73.52 & 1.0 & 40.59 & 21.27 \\ 
\foo{mnasnet1\_0} & 73.46 & -0.47 & 36.42 & 18.78 \\ 
\foo{vgg16\_bn} & 73.36 & -0.09 & 35.69 & 16.19 \\ 
\foo{resnet34} & 73.31 & 0.12 & 40.48 & 21.23 \\ 
\foo{resnet18\_swsl} & 73.29 & 1.74 & 39.95 & 18.79 \\ 
\foo{resnet50\_with\_gaussian\_noise\_aws} & 72.97 & 0.21 & 45.56 & 43.88 \\ 
\foo{resnet50\_with\_gaussian\_noise\_contrast\_motion\_blur\_jpeg\_compression\_aws} & 72.72 & 0.05 & 51.8 & 22.91 \\ 
\foo{mobilenet\_v2\_lpf2} & 72.62 & -0.56 & 34.46 & 17.46 \\ 
\foo{resnet18\_ssl} & 72.6 & 1.24 & 39.51 & 19.19 \\ 
\foo{mobilenet\_v2\_lpf3} & 72.57 & -0.23 & 34.78 & 17.6 \\ 
\foo{mobilenet\_v2\_lpf5} & 72.51 & -0.1 & 34.9 & 17.73 \\ 
\foo{vgg19} & 72.38 & -0.01 & 32.43 & 20.65 \\ 
\foo{vgg16\_lpf5} & 72.33 & 0.15 & 31.89 & 19.86 \\ 
\foo{vgg16\_lpf3} & 72.19 & -0.19 & 32.18 & 19.37 \\ 
\foo{vgg16\_lpf2} & 72.16 & -0.2 & 31.98 & 19.13 \\ 
\foo{resnet50\_with\_contrast\_aws} & 72.0 & -0.42 & 40.85 & 17.29 \\ 
\foo{mobilenet\_v2} & 71.88 & -0.13 & 33.96 & 17.49 \\ 
\foo{resnet50\_with\_fog\_aws} & 71.76 & -0.83 & 37.9 & 17.19 \\ 
\foo{resnet18\_lpf3} & 71.68 & -0.43 & 36.84 & 20.17 \\ 
\foo{vgg16} & 71.59 & -0.33 & 31.3 & 20.14 \\ 
\foo{vgg13\_bn} & 71.59 & 0.01 & 31.76 & 15.16 \\ 
\foo{resnet18\_lpf2} & 71.39 & -0.09 & 36.88 & 19.8 \\ 
\foo{resnet18\_lpf5} & 71.39 & -0.51 & 36.86 & 20.22 \\ 
\foo{resnet18-rotation-standard\_40} & 71.28 & -0.05 & 36.46 & 20.26 \\ 
\foo{vgg11\_bn} & 70.37 & -0.1 & 31.7 & 18.05 \\ 
\foo{resnet50-randomized\_smoothing\_noise\_0.25} & 70.29 & 0.28 & 40.66 & 63.94 \\ 
\foo{vgg13} & 69.93 & -0.27 & 28.53 & 19.32 \\ 
\foo{googlenet/inceptionv1} & 69.78 & 1.01 & 38.84 & 21.85 \\ 
\foo{resnet18} & 69.76 & 0.46 & 35.01 & 19.51 \\ 
\foo{shufflenet\_v2\_x1\_0} & 69.36 & -0.48 & 30.87 & 16.66 \\ 
\foo{resnet18-rotation-worst10\_30} & 69.13 & 0.72 & 34.06 & 22.51 \\ 
\foo{vgg11} & 69.02 & -0.33 & 28.61 & 22.38 \\ 
\foo{resnet18-rotation-random\_30} & 68.88 & 0.19 & 32.88 & 18.63 \\ 
\foo{resnet18-rotation-worst10\_40} & 68.6 & -0.05 & 32.24 & 22.65 \\ 
\foo{resnet50\_with\_pixelate\_aws} & 68.5 & 1.17 & 39.58 & 18.85 \\ 
\foo{resnet18-rotation-random\_40} & 68.35 & 0.73 & 31.87 & 17.89 \\ 
\foo{facebook\_adv\_trained\_resnext101\_denoiseAll} & 68.33 & -0.11 & 40.86 & 41.42 \\ 
\foo{resnet50-smoothing\_adversarial\_DNN\_2steps\_eps\_512\_noise\_0.25} & 67.87 & -0.31 & 40.57 & 62.89 \\ 
\foo{mnasnet0\_5} & 67.6 & -0.37 & 27.9 & 17.39 \\ 
\foo{resnet50\_with\_motion\_blur\_aws} & 67.46 & 1.49 & 38.71 & 15.34 \\ 
\foo{resnet18-rotation-nocrop\_40} & 65.37 & 1.23 & 30.1 & 20.5 \\ 
\foo{facebook\_adv\_trained\_resnet152\_denoise} & 65.32 & 0.38 & 37.97 & 39.48 \\ 
\foo{bninception-imagenet21k} & 65.24 & 1.78 & 32.8 & 30.3 \\ 
\foo{resnet50-randomized\_smoothing\_noise\_0.50} & 64.24 & 0.04 & 39.8 & 61.41 \\ 
\foo{resnet50\_with\_greyscale\_aws} & 63.33 & 0.49 & 28.33 & 18.16 \\ 
\foo{resnet50\_linf\_eps4\_robust} & 62.42 & 0.53 & 32.37 & 60.3 \\ 
\foo{facebook\_adv\_trained\_resnet152\_baseline} & 62.34 & 0.58 & 35.77 & 37.63 \\ 
\foo{resnet50-smoothing\_adversarial\_DNN\_2steps\_eps\_512\_noise\_0.50} & 62.19 & -0.04 & 39.14 & 59.26 \\ 
\foo{resnet50-vtab-semi-exemplar} & 61.62 & 0.98 & 33.85 &  \\ 
\foo{resnet50\_with\_zoom\_blur\_aws} & 61.25 & 1.22 & 33.27 & 13.01 \\ 
\foo{resnet50-vtab-semi-rotation} & 60.92 & 0.94 & 26.38 &  \\ 
\foo{shufflenet\_v2\_x0\_5} & 60.55 & -0.27 & 23.58 & 16.08 \\ 
\foo{resnet50\_adv-train-free} & 60.49 & -0.03 & 29.41 & 57.42 \\ 
\foo{resnet50-smoothing\_adversarial\_PGD\_1step\_eps\_512\_noise\_0.25} & 60.47 & -0.45 & 37.21 & 58.49 \\ 
\foo{resnet50\_trained\_on\_SIN} & 60.18 & 1.4 & 39.42 & 19.25 \\ 
\foo{squeezenet1\_1} & 58.18 & 0.12 & 20.18 & 16.08 \\ 
\foo{squeezenet1\_0} & 58.09 & -0.26 & 20.17 & 18.06 \\ 
\foo{resnet50\_l2\_eps3\_robust} & 57.9 & 0.33 & 31.83 & 56.25 \\ 
\foo{alexnet\_lpf2} & 57.23 & -0.38 & 22.54 & 29.09 \\ 
\foo{alexnet\_lpf3} & 56.89 & -0.41 & 22.77 & 30.67 \\ 
\foo{alexnet\_lpf5} & 56.58 & -0.41 & 22.77 & 31.71 \\ 
\foo{alexnet} & 56.52 & -0.28 & 21.55 & 24.08 \\ 
\foo{resnet50-smoothing\_adversarial\_PGD\_1step\_eps\_512\_noise\_0.50} & 54.66 & -0.31 & 35.7 & 53.09 \\ 
\foo{resnet50-randomized\_smoothing\_noise\_1.00} & 53.12 & 0.12 & 34.93 & 52.11 \\ 
\foo{resnet50-smoothing\_adversarial\_DNN\_2steps\_eps\_512\_noise\_1.00} & 51.87 & 0.23 & 34.43 & 50.95 \\ 
\foo{resnet50\_linf\_eps8\_robust} & 47.91 & 1.35 & 23.93 & 46.97 \\ 
\foo{resnet50-smoothing\_adversarial\_PGD\_1step\_eps\_512\_noise\_1.00} & 44.28 & 0.2 & 29.84 & 43.57 \\ 
\foo{resnet50\_with\_defocus\_blur\_aws} & 31.9 & 1.3 & 18.18 & 9.29 \\ 

%% file: appendix/eff_robust_corr_table.tex
avg\_corruptions & 0.25 & 0.06 & 0.6 & 0.5 & 0.65 & 0.52 & 0.02 \\ 
avg\_pgd & -0.04 & -0.19 & 0.3 & 0.35 & 0.84 & 0.7 & -0.12 \\ 
brightness\_in-memory & 0.34 & 0.11 & 0.32 & 0.3 & 0.29 & 0.23 & 0.13 \\ 
brightness\_on-disk & 0.56 & 0.48 & 0.56 & 0.39 & 0.22 & 0.15 & 0.16 \\ 
contrast\_in-memory & 0.15 & 0.07 & 0.14 & 0.04 & -0.61 & -0.5 & 0.14 \\ 
contrast\_on-disk & 0.26 & 0.28 & 0.17 & 0.05 & -0.61 & -0.54 & 0.15 \\ 
defocus\_blur\_in-memory & 0.27 & -0.04 & 0.66 & 0.56 & 0.43 & 0.27 & -0.05 \\ 
defocus\_blur\_on-disk & 0.39 & 0.39 & 0.65 & 0.49 & 0.28 & 0.17 & 0.12 \\ 
elastic\_transform\_in-memory & 0.14 & -0.12 & 0.49 & 0.42 & 0.75 & 0.63 & -0.15 \\ 
elastic\_transform\_on-disk & 0.3 & 0.21 & 0.57 & 0.41 & 0.65 & 0.58 & 0.01 \\ 
fog\_in-memory & 0.14 & 0.07 & -0.04 & -0.07 & -0.59 & -0.56 & 0.02 \\ 
fog\_on-disk & 0.28 & 0.31 & 0.04 & -0.03 & -0.64 & -0.6 & 0.04 \\ 
frost\_in-memory & 0.15 & -0.12 & 0.42 & 0.44 & 0.54 & 0.42 & -0.02 \\ 
frost\_on-disk & 0.32 & 0.15 & 0.53 & 0.45 & 0.44 & 0.36 & 0.08 \\ 
gaussian\_blur\_in-memory & 0.27 & -0.07 & 0.67 & 0.57 & 0.47 & 0.33 & -0.05 \\ 
gaussian\_blur\_on-disk & 0.41 & 0.4 & 0.65 & 0.48 & 0.26 & 0.16 & 0.13 \\ 
gaussian\_noise\_in-memory & -0.01 & -0.13 & 0.41 & 0.38 & 0.68 & 0.51 & -0.04 \\ 
gaussian\_noise\_on-disk & 0.08 & 0.0 & 0.4 & 0.34 & 0.71 & 0.62 & 0.07 \\ 
glass\_blur\_on-disk & 0.24 & 0.17 & 0.56 & 0.45 & 0.61 & 0.53 & -0.0 \\ 
greyscale & 0.3 & 0.17 & 0.11 & 0.29 & 0.09 & -0.06 & 0.04 \\ 
impulse\_noise\_in-memory & -0.06 & -0.1 & 0.35 & 0.34 & 0.65 & 0.45 & -0.05 \\ 
impulse\_noise\_on-disk & 0.04 & 0.0 & 0.34 & 0.31 & 0.72 & 0.6 & 0.03 \\ 
jpeg\_compression\_in-memory & 0.04 & -0.11 & 0.43 & 0.41 & 0.8 & 0.62 & -0.01 \\ 
jpeg\_compression\_on-disk & 0.09 & 0.01 & 0.44 & 0.4 & 0.8 & 0.65 & 0.03 \\ 
motion\_blur\_in-memory & 0.2 & -0.02 & 0.51 & 0.43 & 0.56 & 0.41 & -0.08 \\ 
motion\_blur\_on-disk & 0.32 & 0.25 & 0.58 & 0.43 & 0.39 & 0.31 & 0.07 \\ 
pgd.l2.eps0.1 & -0.03 & -0.01 & 0.18 & 0.25 & 0.64 & 0.44 & -0.33 \\ 
pgd.l2.eps0.5 & -0.05 & -0.22 & 0.31 & 0.34 & 0.71 & 0.63 & -0.11 \\ 
pgd.linf.eps0.5 & -0.05 & -0.23 & 0.28 & 0.33 & 0.84 & 0.7 & -0.13 \\ 
pgd.linf.eps2 & 0.01 & -0.18 & 0.3 & 0.31 & 0.76 & 0.69 & 0.05 \\ 
pixelate\_in-memory & 0.27 & 0.03 & 0.61 & 0.48 & 0.66 & 0.53 & 0.05 \\ 
pixelate\_on-disk & 0.31 & 0.16 & 0.62 & 0.46 & 0.63 & 0.54 & 0.12 \\ 
saturate\_in-memory & 0.35 & 0.08 & 0.4 & 0.43 & 0.38 & 0.27 & 0.12 \\ 
saturate\_on-disk & 0.55 & 0.43 & 0.46 & 0.41 & 0.26 & 0.16 & 0.13 \\ 
shot\_noise\_in-memory & -0.01 & -0.14 & 0.41 & 0.39 & 0.69 & 0.51 & -0.05 \\ 
shot\_noise\_on-disk & 0.07 & -0.01 & 0.4 & 0.35 & 0.71 & 0.62 & 0.06 \\ 
snow\_in-memory & 0.26 & 0.02 & 0.39 & 0.35 & 0.61 & 0.55 & 0.04 \\ 
snow\_on-disk & 0.33 & 0.14 & 0.5 & 0.43 & 0.6 & 0.51 & 0.04 \\ 
spatter\_in-memory & 0.09 & -0.06 & 0.36 & 0.36 & 0.8 & 0.66 & -0.08 \\ 
spatter\_on-disk & 0.26 & 0.08 & 0.5 & 0.43 & 0.75 & 0.63 & -0.04 \\ 
speckle\_noise\_in-memory & 0.0 & -0.13 & 0.43 & 0.39 & 0.71 & 0.55 & -0.04 \\ 
speckle\_noise\_on-disk & 0.08 & -0.02 & 0.42 & 0.36 & 0.74 & 0.66 & 0.02 \\ 
stylized\_imagenet & 0.32 & 0.24 & 0.31 & 0.3 & 0.44 & 0.31 & -0.02 \\ 
zoom\_blur\_in-memory & 0.21 & 0.23 & 0.45 & 0.35 & 0.45 & 0.36 & -0.01 \\ 
zoom\_blur\_on-disk & 0.26 & 0.21 & 0.55 & 0.41 & 0.49 & 0.39 & 0.0 \\ 